\pdfoutput=1

\documentclass[11pt]{article}

\usepackage[]{emnlp2022}

\usepackage{times}
\usepackage{latexsym}

\usepackage[T1]{fontenc}

\usepackage[utf8]{inputenc}

\usepackage{microtype}
\usepackage{xspace}

\usepackage{subfig}
\usepackage{booktabs}
\usepackage{CJKutf8}

\usepackage{graphicx}
\usepackage{booktabs}
\usepackage{xcolor}
\usepackage{xspace}
\usepackage{graphicx}
\usepackage{subfig}
\usepackage{multirow}
\usepackage{array}
\usepackage{verbatim}
\usepackage{mdwlist}
\usepackage{epigraph}
\usepackage{float}

\title{Affective Idiosyncratic Responses to Music}

\newcommand{\chinese}[1]{\begin{CJK*}{UTF8}{gbsn}#1\end{CJK*}}
\newcommand{\myparagraph}[1]{\noindent \textbf{#1}}

\author{
Sky CH-Wang$^{\circ}$\quad
Evan Li$^{\circ}$\quad
Oliver Li$^{\circ}$\quad 
Smaranda Muresan$^{\circ\bullet}$\quad
Zhou Yu$^{\circ}$\quad \\
$^{\circ}$Department of Computer Science, Columbia University \\
$^{\bullet}$Data Science Institute, Columbia University \\
\texttt{skywang@cs.columbia.edu} \\
\texttt{\{el3078, al4143, smara, zy2461\}@columbia.edu}
}

\begin{document}
\maketitle
\begin{abstract}
Affective responses to music are highly personal. Despite consensus that idiosyncratic factors play a key role in regulating how listeners emotionally respond to music, precisely measuring the marginal effects of these variables has proved challenging. To address this gap, we develop computational methods to measure affective responses to music from over 403M listener comments on a Chinese social music platform. Building on studies from music psychology in systematic and quasi-causal analyses, we test for musical, lyrical, contextual, demographic, and mental health effects that drive listener affective responses. Finally, motivated by the social phenomenon known as \chinese{网抑云} (wǎng-yì-yún), we identify influencing factors of platform user self-disclosures, the social support they receive, and notable differences in discloser user activity.

\end{abstract}

\section{Introduction}

Music can evoke powerful emotions in listeners \cite{meyer1956emotion}. However, our emotional reactions to it are not universal---affective responses to music are highly \textit{personal}. Just as you may wonder why your friend is sobbing to a song that you only feel ambivalent about, a listener's emotional response to music not only varies with inherent audio or lyrical features \cite{hevner1935expression, webster2005emotional, van2011emotional}, %
but also with other factors such as a listener's demographics, mental health conditions, and surrounding environment %
\cite{krugman1943affective, robazza1994emotional, gregory1996cross, juslin2008emotional, saarikallio2013affective, garrido2018music}. As a result of this idiosyncrasy, it has been extremely difficult to precisely measure the marginal effects of these variables on a 
listener's affective response \cite{yang2007music, beveridge2018popular}. This difficulty is further compounded especially when examining how a collection of these factors influence individual affective reactions in combination \cite{gomez2021music}. 

Music psychology has long focused on identifying the relationships between human affect and music, both in those that are perceived and those that are felt. Perceiving and feeling emotions in music, while highly related, are not identical \cite{hunter2010music}. Examining the latter has proved challenging, as in addition to insufficient scale for finding significance, measuring felt emotions in participatory studies often interferes with the experience itself \cite{gabrielsson2010strong}. While recent computational studies have attempted 
citizen science approaches for %
annotation \cite{gutierrez2021emotion}, reliability remains an issue; annotator confusion persists between the concepts of perceived and induced emotions \cite{juslin2019musical}. Our work expands this line of research by examining affective responses to music in a natural setting: an online social music platform. %

We test for differences in affective responses to music by computationally measuring expressed emotions from a massive study of over 403M listener comments on one of China's largest social music platforms, Netease Cloud Music. Our paper offers the following three contributions. First, we reveal several nuances in listener affective responses against a host of musical, lyrical, and contextual factors, showing evidence of emotional contagion. %
Second, in a multi-modal quasi-causal analysis, we show that listeners of different genders and ages vary in their reactions to musical stimuli and identify specific features driving demographic effects on affective responses. Third, motivated by the social phenomenon known as
\chinese{网抑云},\footnote{\chinese{网抑云}, pronounced as wǎng-yì-yún, is a pun on the platform name \chinese{网易云} that refers to the outpour of emotional and personal comments on the social music platform especially late at night and under sad songs. \chinese{抑}, here, is the first character of the word for depression---\chinese{抑郁}.}
we systematically study self-disclosures of mental health disorders on the platform, identifying driving factors of this behavior, the social support they receive, and differences in discloser user activity.

\section{Data}

Our work is drawn from one of the largest music streaming services in China, Netease Cloud Music, and focuses on Chinese-language user content.

\myparagraph{Netease Cloud Music.} \chinese{网易云音乐} (wǎng-yì-yún-yīn-yuè) has over 185 million monthly active users \cite{dredge_2022}. Unlike mainstream music streaming services in the United States such as Spotify and Apple Music, Netease Cloud Music is a \textit{social music platform} \cite{zhou2018homophily, wang2020exploring}. Here, among other unique features,
each song, album, and playlist have comment sections that serve as discussion boards, where users can post top-level comments as well as reply to or up-vote existing ones. 
These platform interactions serve as a natural setting on listener responses, where users are able to post freely\footnote{See our \nameref{sec:limitations} section for a discussion on platform moderation and censorship.} in the comment sections of what they are currently listening to.
Users are required to create an account to access most of the platform's features; when doing so, users optionally input personal demographic information like age, gender, and location, which they can then choose to display as public or private.

\myparagraph{Dataset Collection.} %
To collect a representative sample of public platform commenting activity, we adapt traditional snowball sampling \cite{atkinson2001accessing} across multiple random seeds to build an exhaustive list of user, song, album, and playlist entity ids on the platform. We then uniformly sample from the set of entities that have at least one public comment posted. Data was collected from all public content ranging from the platform's inception, 2013, to 2022, totaling over 455K albums, 2.87M songs, 1.36M playlists, 29.9M users, and 403M comments. A detailed breakdown of our data and a view of the interaction interface of the platform are shown in Appendix Section \ref{sec:datastats}. The study and data collection were approved by an Institutional Review Board as exempt. %

\section{Measuring Affective Response}

We measure affective responses to music as expressed in comments posted under their comment sections. Since not all comments are indicative of a user's emotional response, %
we %
sample a %
subset of user content and examine both the experiencer of the emotion and its expressed stimulus, before conducting our analysis.

\begin{table}[t]
    \centering
    \begin{tabular}{l}
    \toprule
        
        \chinese{我只想和你一个人做那些浪漫到极致的事} \\ %
        \textit{Translation}: I just want to do the most\\
        romantic things with you alone \\ [1.2ex] 
        
        \chinese{果然不该来的。混蛋老爸，气死我了！} \\ %
        \textit{Translation}: I shouldn't have come. \\
        Asshole dad, pisses me off! \\ [1.2ex]
        
        \chinese{太棒了好听太治愈了我莫名有点想哭} \\ %
        \textit{Translation}: It's so good, it's so healing,\\ 
        I feel like crying for some reason \\
        
    \bottomrule
    \end{tabular}
    \caption{Example top-level comments indicating an affective response.}
    \label{tab:examplecomments}
\end{table}

\myparagraph{Emotion Experiencer.} Two annotators first manually annotated 1000 comments selected uniformly at random to identify the experiencer of the emotion expressed in top-level comments. Top-level comments were chosen to limit dyadic interaction effects and are used to measure affective responses later on. With an initial Cohen's $\kappa$ of 0.80 and with disagreements resolved via discussion, similar to \cite{mohammad-etal-2014-semantic}, we find that the experiencer of the emotion expressed in the comment is often the commenter themselves (99.1\%); we thus maintain this assumption in our later experiments. Selected examples and annotation guidelines are shown in Table \ref{tab:examplecomments} and Appendix Section \ref{sec:annotationguidelines}, respectively.

\myparagraph{Affective Stimulus.} Next, annotators were tasked with identifying %
what caused the emotional response in the comment itself. %
Annotators labeled for comments containing emotions that could explicitly be said to \textit{not} originate from music---under the BRECVEMA framework of music-induced emotions \cite{JUSLIN2013235}, emotions are evoked in listeners via a combination of mechanisms related to aesthetic appreciation, entrainment, visual imagination, and emotional associations with past experiences, among other factors. A listener's emotional state also has an effect on their music choice; for example, listeners often use music for mood regulation, or as a coping mechanism \cite{stewart2019music, schafer2020music}.
Here, we make no explicit causal assumptions of music choice and seek only to measure comment affective responses. 
With an initial Cohen's $\kappa$ of 0.76 and with disagreements resolved via discussion, we only find a few instances (3.3\%) where affective stimuli may be explicitly attributed elsewhere. 
There are a few patterns among these irrelevant comments: namely, that 
they primarily 
relate to album images, quotations, and easily identifiable spam messages, i.e. ``\chinese{沙发}'' (meaning ``first comment''). Aiming for high precision, %
we create simple regular expressions and redundancy filtering to increase the relevance of comments with affective content, achieving a precision of 98.8\% on a held-out test set of the same size. Specific annotation guidelines and filtering methods are shown in Appendix Sections \ref{sec:annotationguidelines} and \ref{sec:commentpreprocessing}, respectively.

\myparagraph{Measuring Affective Response.} We characterize emotions across a 2-dimensional plane of valence and arousal following the Russell model of emotions \cite{russell1980circumplex}, representing the degree of positivity and emotion intensity, respectively. Specifically, we employ a lexicon-based approach to measure valence and arousal in music comments, using one of the largest crowd-sourced datasets for the Chinese language---Chinese EmoBank \cite{yu-etal-2016-building}---containing 5512 words annotated for their valence and arousal. In the following sections, these measures of expressed emotions in comments are what we define as listener affective response.

\section{Variations in Affective Response}

Computing comment-level valence and arousal by averaging across word-level scores,\footnote{To confirm that our findings were robust against high-volume sentiment terms, following a distribution analysis of comment valence and arousal scores, we drop the top three most frequent terms---\chinese{好听}, \chinese{好}, and \chinese{喜欢}, which roughly translate to "sounds good", "good", and "like", respectively---and recompute our experiments, obtaining similar results.} we analyze variations in listener affective responses to (1) musical and (2) lyrical features, (3) contextual factors, and (4) user demographic variables.

\begin{figure*}[!t]
    \centering
    \begin{tabular}{ccc}
    \subfloat[Tempo]{\includegraphics[width=0.30\textwidth]{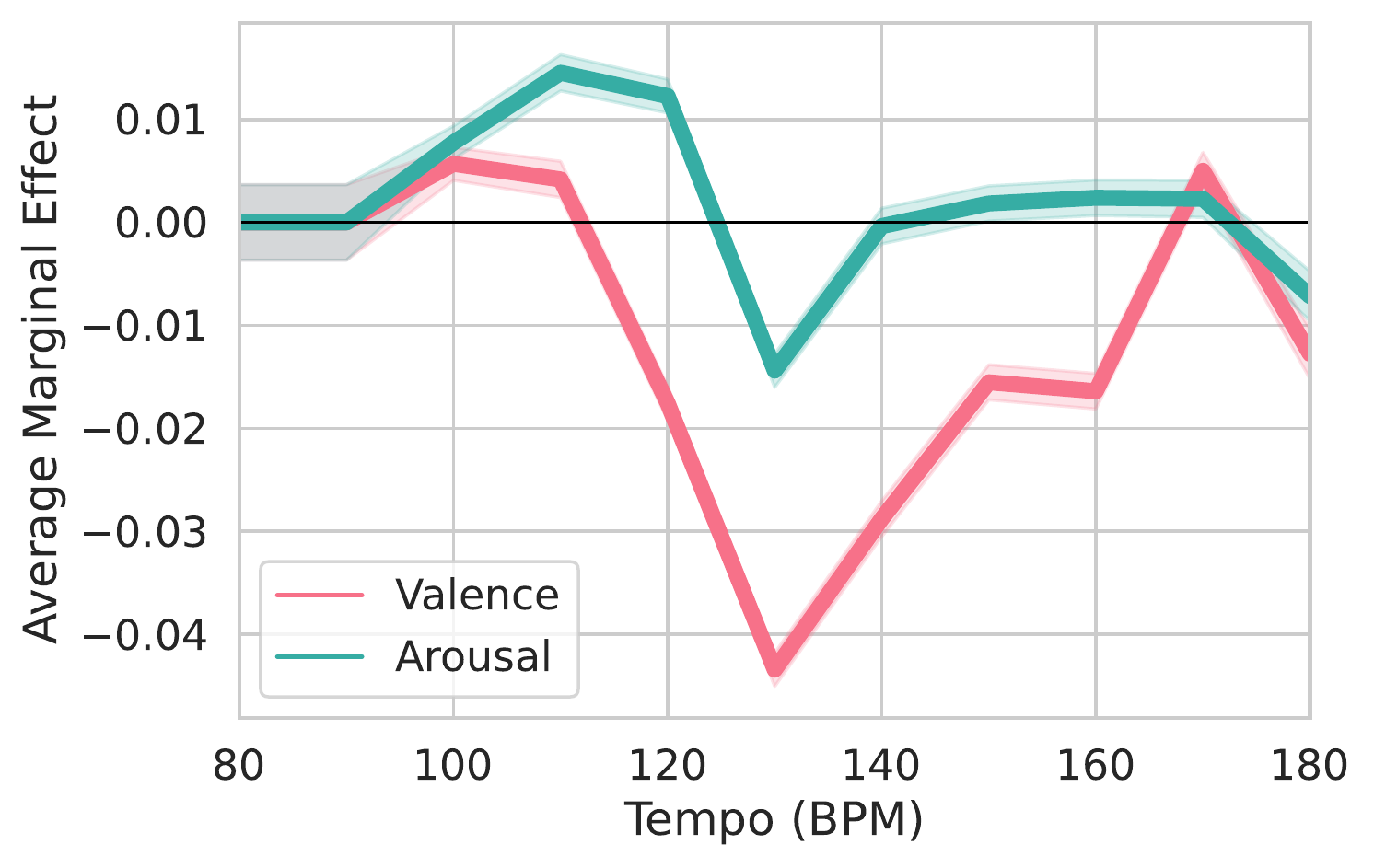}} & %
    \subfloat[Loudness]{\includegraphics[width=0.30\textwidth]{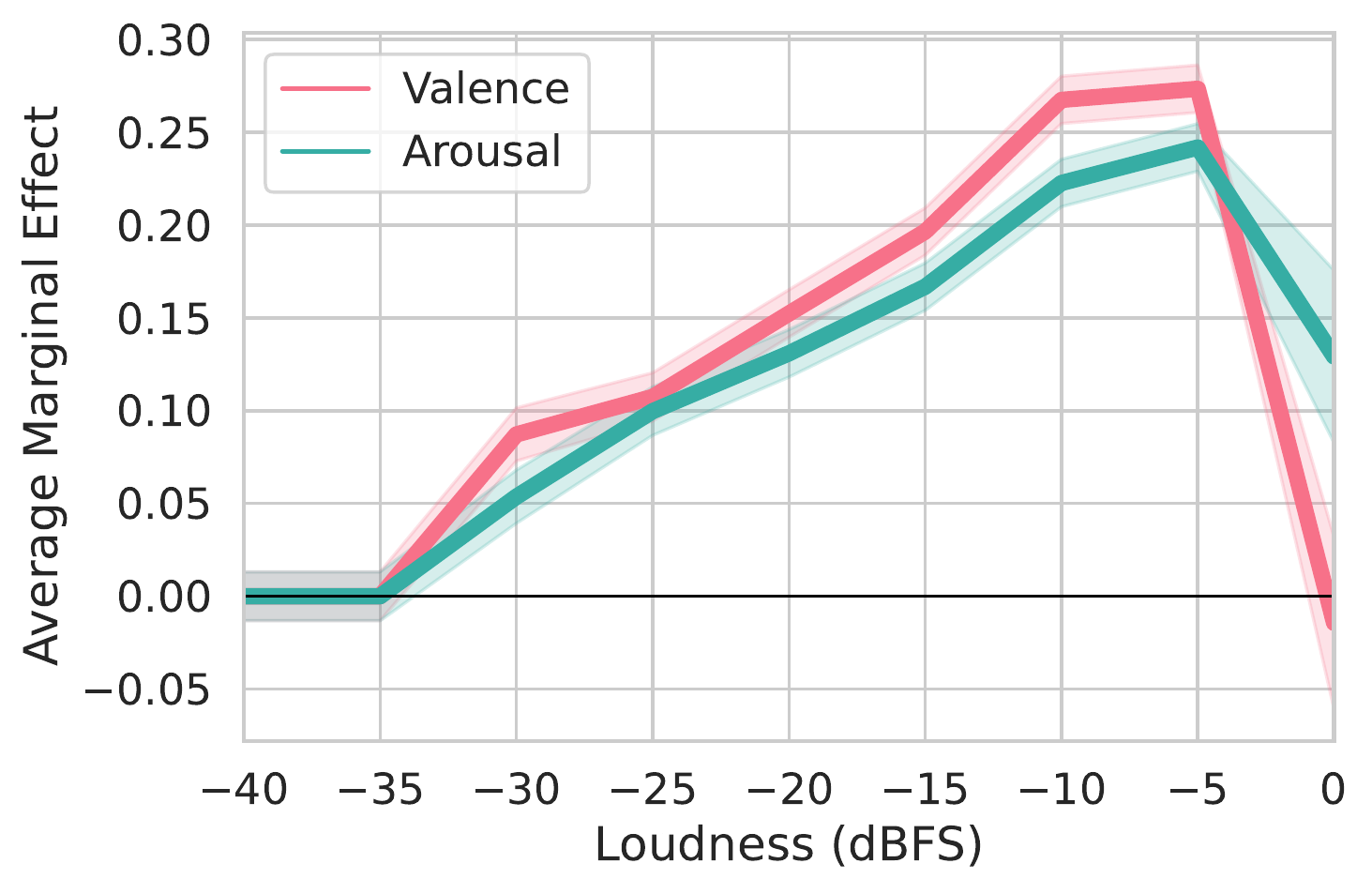}} & %
    \subfloat[Mode]{\includegraphics[width=0.30\textwidth]{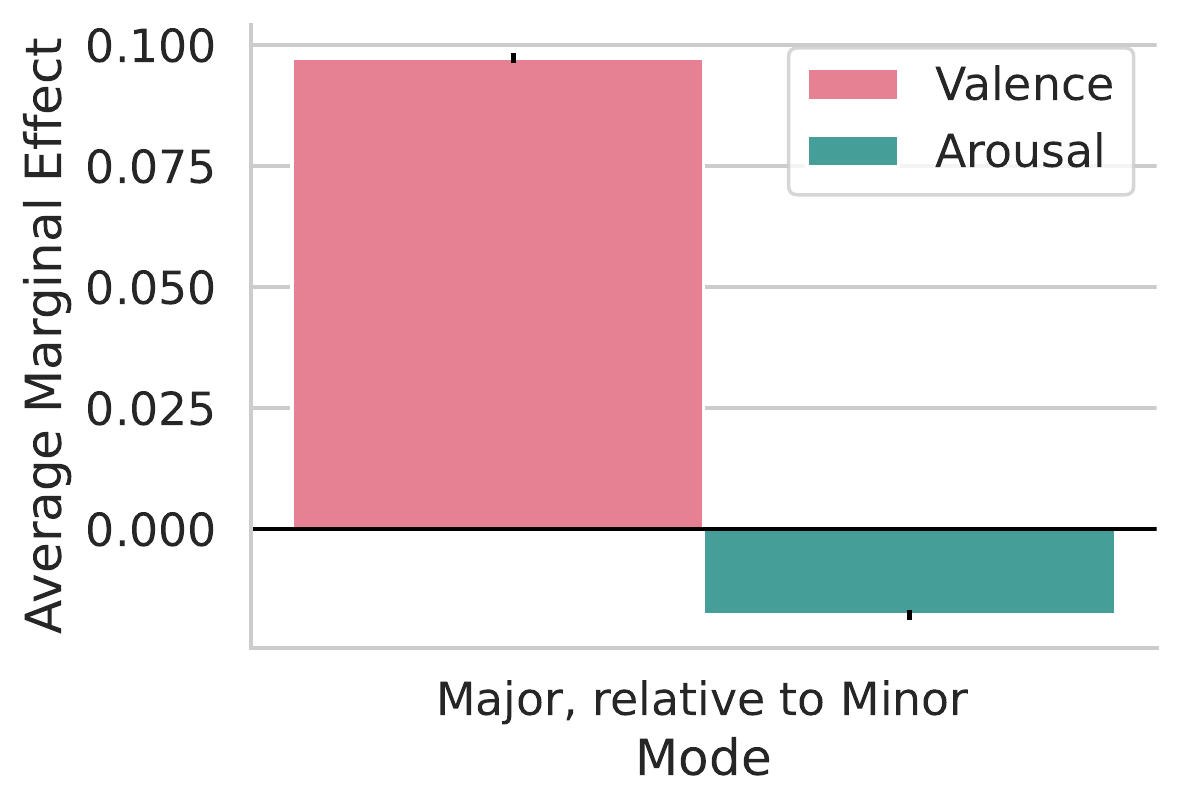}} \\ %
    \subfloat[Timbre-Brightness]{\includegraphics[width=0.30\textwidth]{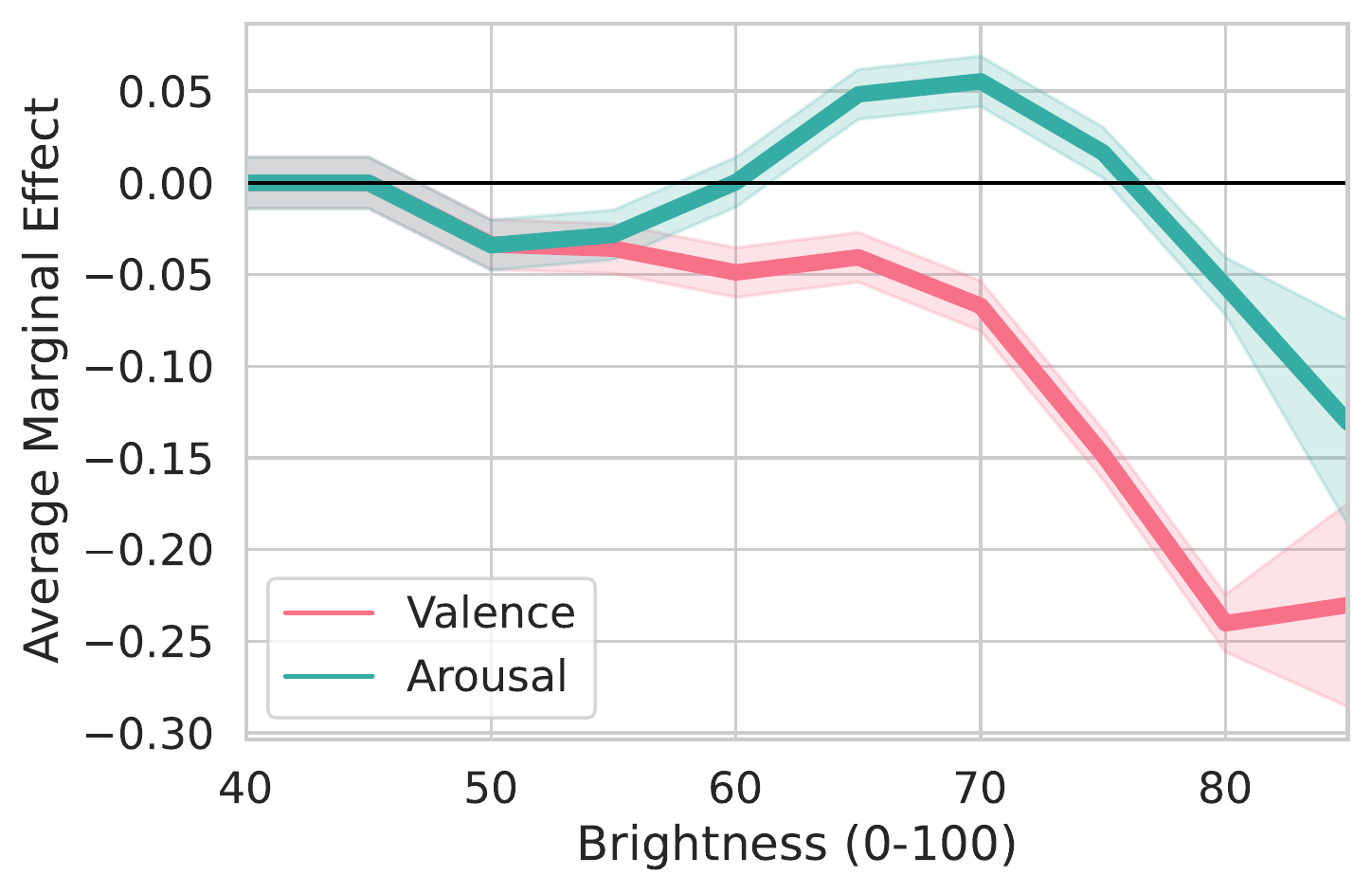}} & %
    \subfloat[Positive Emotion]{\includegraphics[width=0.30\textwidth]{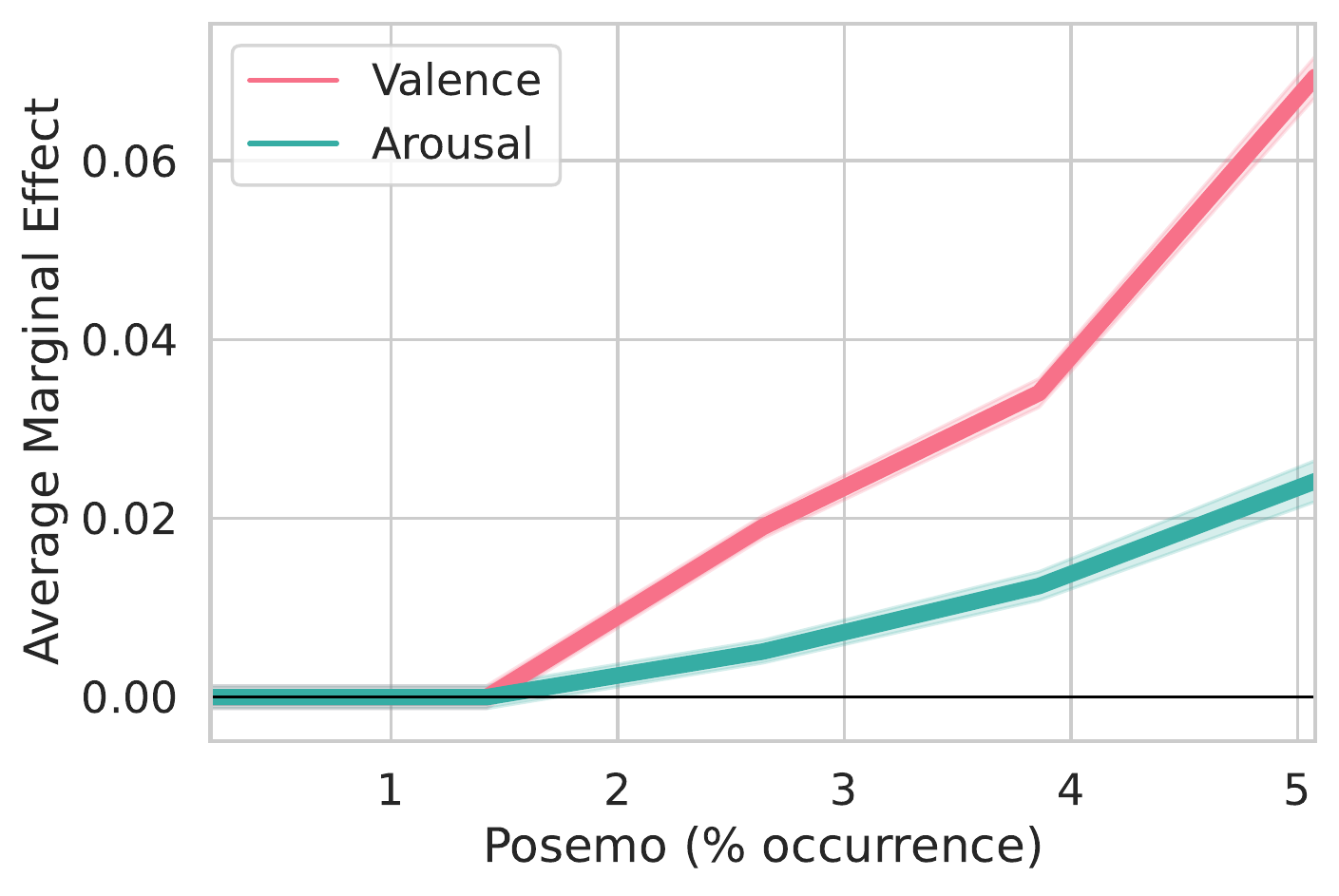}} & %
    \subfloat[1st Person Singular Pronouns]{\includegraphics[width=0.30\textwidth]{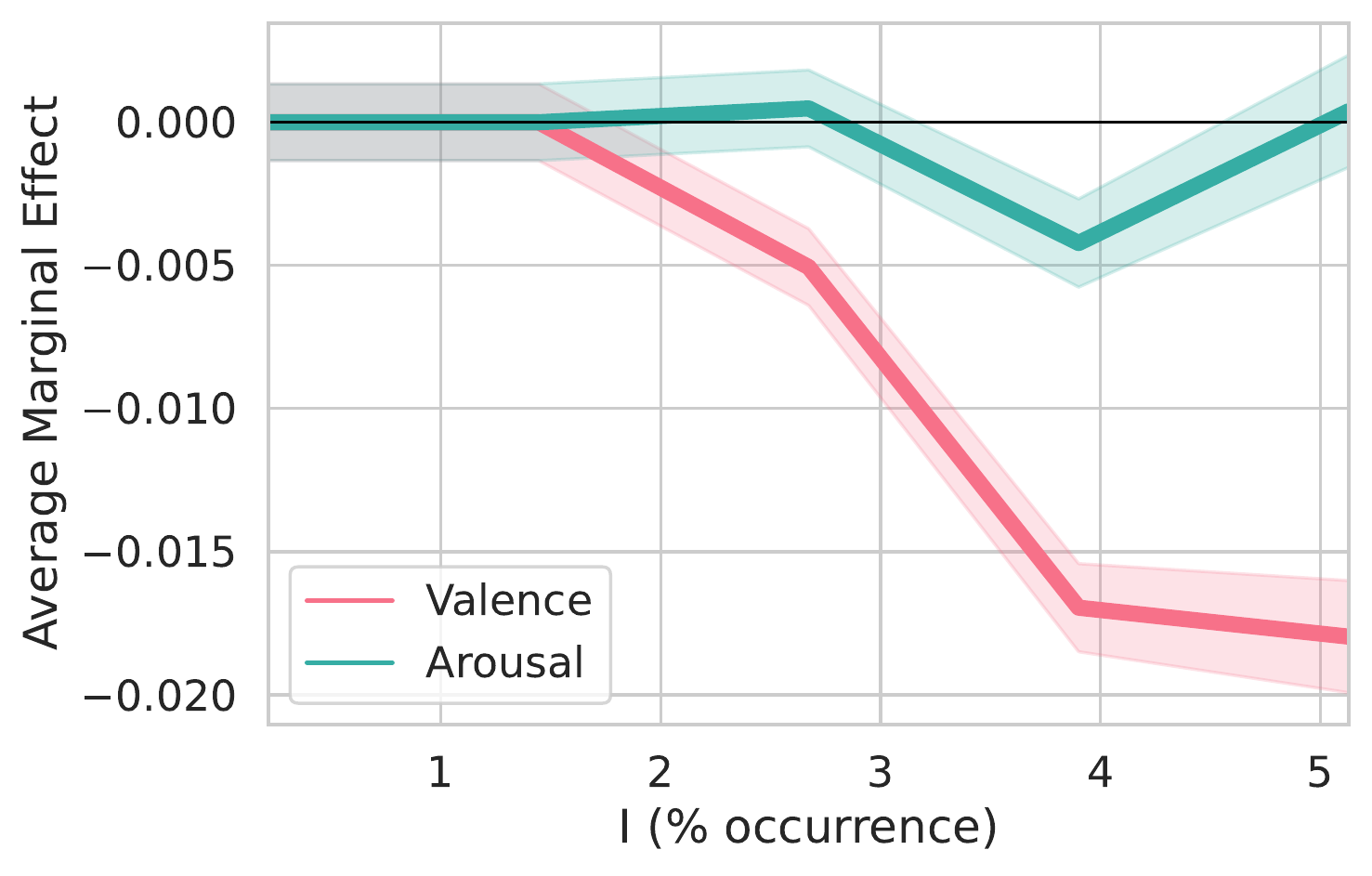}} %
    \end{tabular}
    \caption{
    Average marginal effects of musical and lyrical features on listener affective responses with other features and listener demographics as controls.
    Standard errors are shown;
    \textcolor{red}{valence} in \textcolor{red}{red}, \textcolor{blue}{arousal} in \textcolor{blue}{blue}.
    The complete set of figures for musical and lyrical features are shown in Appendix Section \ref{sec:expandedresults}, Figures \ref{fig:musicfeatures_expanded}-\ref{fig:lyricfeatures_expanded_4}.
    }
    \label{fig:musicfeatures}
\end{figure*}

\subsection{Musical and Lyrical Features}
\label{sec:musicalandlyricalvariations}

In prior work, while much emphasis is placed on identifying the causes of perceived emotions in music, less emphasis has been placed on emotional responses, which are highly influenced by extra-musical and contextual factors in listeners \cite{gomez2021music}.
Recent work has attempted to use physiological signals and self-reported emotions to measure emotional responses in listeners \cite{hu2018relationships}, though this has proved challenging partly due to a high degree of intercorrelations and confounds, causing the number of trials needed to measure such effects to be intractable relative to typical experiment scale \cite{eerola2013emotional}.
Using our data, we test for the marginal effects of musical and lyrical features on affective responses.

\myparagraph{Methods.} To understand the marginal contributions of these variables on affective responses, we fit separate multivariable linear regression models on response valence and arousal, including the features described below as regressors. As affective responses are highly idiosyncratic \cite{juslin2008emotional}, we further control for listener demographics, namely age, gender, and location. We then test for multicollinearity by computing the variance inflation factor (VIF) for each variable and iteratively remove collinear variables in our regression that have a VIF greater than 5. In our analyses, we stratify continuous variables (i.e. tempo) into fixed-length category indicator variables (i.e., 80-90 BPM, 90-100 BPM, and so on) and measure the average marginal effects (AME) on valence and arousal of each stratum, using the first of such categories as the reference group (i.e., the AME of 90-100 BPM, and so on, relative to 80-90 BPM).

\myparagraph{Musical Features.} We use \textit{librosa} \cite{mcfee2015librosa}, \textit{pydub} \cite{robert2018pydub}, and \textit{timbral\_models} of the AudioCommons project \cite{font2016audio} to derive song file musical features. We extract (1) \textbf{tempo} and (2) \textbf{tempo standard deviation}, both in beats per minute (BPM) \cite{ellis2007beat}; (3) \textbf{loudness}, measured as the average decibels relative to full scale (dBFS) value of the entire song; (4) \textbf{mode}, namely, major or minor, and (5) \textbf{key}, i.e. C\# minor \cite{krumhansl2001cognitive}; as well as eight additional timbral features, or the perceived sound qualities of a piece of music. They are, specifically,
(6) \textbf{depth}, related to the emphasis of low frequency content, 
(7) \textbf{brightness}, a measure that correlates both with the spectral centroid and the ratio of high frequencies to the sum of all energy of a sound, and 
(8) \textbf{warmth}, often created by low and midrange frequencies and associated with lower harmonics \cite{pearce5first};
(9) \textbf{roughness}, a sound's buzzing, harsh, and raspy quality \cite{vassilakis2007sra};
(10) \textbf{sharpness}, measuring high frequency content \cite{zwicker2013psychoacoustics};
(11) \textbf{hardness}, the amount of aggression \cite{pearce2019modelling};
(12) \textbf{reverberation}, a sound or echo's persistence after it is initially produced \cite{jan2012blind};
and (13) \textbf{boominess}, a sound's deep resonant quality as measured by the booming index \cite{hatano2000booming}.
Here, reverberation is classed as a binary variable, %
while all other timbral features are measured and clipped to values between 0 and 100 following their regression models' training data domain considerations.

\myparagraph{Psycholinguistic Lyrical Features.} Similar to \citet{mihalcea2012lyrics} while limiting our analysis to Chinese language songs, we extract coarse psycholinguistic lexical features of lyrics. Specifically, we preprocess lyrics with regular expressions to remove extraneous information as shown in Appendix Section \ref{sec:lyricpreprocessing} and use the Simplified Chinese version \cite{huang2012development} of the Linguistic Inquiry and Word Count (LIWC) lexicon \cite{pennebaker2015development} to create normalized counts of tokenized word semantic classes.

\myparagraph{Topic-wise Lyrical Features.} To capture thematic trends across song lyrics, we train a 20-topic LDA model on preprocessed song lyrics and manually label each topic with its prominent theme, i.e. Nationalism/China or Hometown/Childhood. Labeled lyrical topics and the top words associated with each are shown in Appendix Section \ref{sec:lyricpreprocessing}.

\myparagraph{Results.} Figure \ref{fig:musicfeatures} reveals five important trends in affective responses across a variety of musical and lyrical features. 

First, tempo exhibits a bimodal distribution relative to both valence and arousal; listeners are most intensely positive for tempos of around 110 BPM and 160 BPM, with the former eliciting greater arousal. Higher tempo variation also sees similar increases in affective responses, although tempo standard deviations of around 35-40 BPM produce the opposite effect, with arousal peaking earlier than valence. %
Our findings are consistent with prior work on listener self-ratings and measured physiological responses that have used coarse categorizations of tempo, i.e. ``fast'' tempo \cite{liu2018effects}, or the presence and absence of tempo variation \cite{kamenetsky1997effect}, as opposed to the continuous measures we use here.

Second, consistent with prior work \cite{schubert2004modeling, gomez2007relationships}, loudness generally produces a strong positive correlation with more intensely positive reactions; changes in loudness also see a greater change in AME than that of tempo. However, this trend is reversed for songs that are loudest (i.e. between -5 and 0 dBFS)---while unexplored in prior work within music psychology, this observation intuitively follows neural downregulation responses to excessively loud or unpleasant sounds \cite{hirano2006effect, koelsch2014brain}.

Third, consistent across all keys (Appendix Figure \ref{fig:musicfeatures_key}), major mode in songs has a greater valence and a lower arousal than minor mode. This observation extends prior work investigating the interaction effects between mode and affective responses \cite{van2011emotional} in a western tonal context, suggesting that associations of sadness and happiness by way of musical mode are also consistent in Chinese listeners.

Fourth, increases in most timbral characteristics see similar increases in the intensity and positivity of reactions up until a point of extremity, whereafter the opposite effect is observed. The only exceptions are roughness and warmth, in which both valence and arousal see monotonically decreasing and increasing trends, respectively. Our results for brightness specifically provide nuance into how, when exploring an expanded set of timbral characteristics and moving beyond only varying timbre through different instruments \cite{hailstone2009s, eerola2012timbre, wallmark2018embodied}, excess of a timbral feature can produce the opposite initial effect on affective responses.

Fifth, listener affective reactions mirror the psychological states expressed in lyrics. Changes in response valence and arousal closely match the proportion of LIWC category terms for affective processes. Greater use of positive emotion terms sees greater response positivity (r=0.93,p<0.05), while the opposite is true for negative emotion terms (r=-0.92,p<0.05), and both saw rises in response intensity with their increased use. Furthermore, increases in first-person pronouns also see decreases in valence (r=-0.94,p<0.05), mirroring work on the depressed psychological states reflected through their increased use \cite{pennebaker2011secret}. Interpreted together with our findings on musical features, these observations mirror emotional contagion \cite{JUSLIN2013235}, where the recognition of emotions expressed in music evokes similar emotions in listeners.

These findings, compared to prior work, %
highlight the importance of using finer-grained measurements on an extended set of features and controls to provide a more nuanced analysis of emotional responses to musical and lyrical stimuli. Expanded results with the full list of figures are shown in Appendix %
Figures \ref{fig:musicfeatures_expanded}-\ref{fig:lyricfeatures_expanded_4}.

\subsection{Contextual Factors}
\label{sec:contextual}

\begin{figure}[!t]
    \centering
    \raisebox{-0.5\height}{\includegraphics[width=0.23\textwidth]{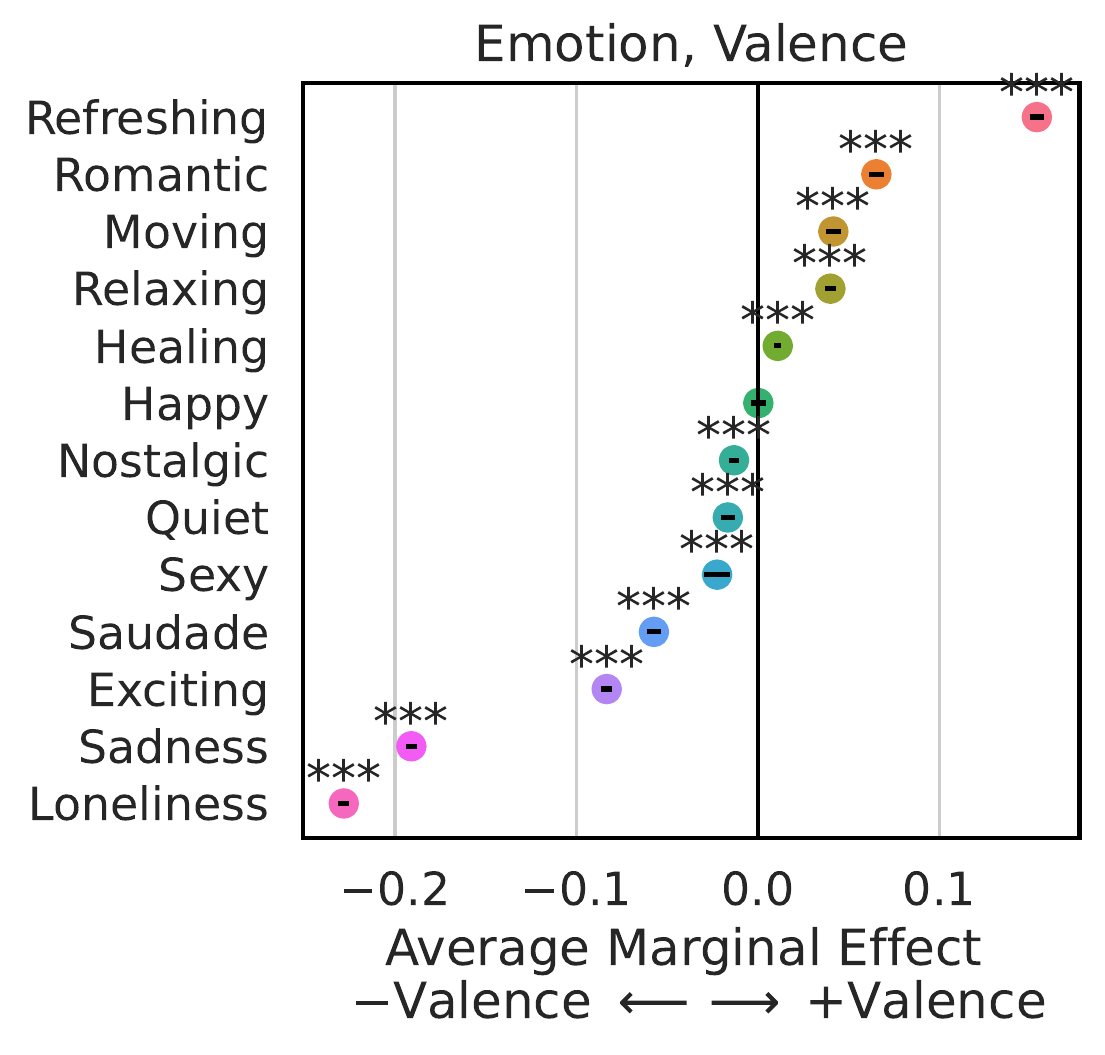}}
    \raisebox{-0.5\height}{\includegraphics[width=0.23\textwidth]{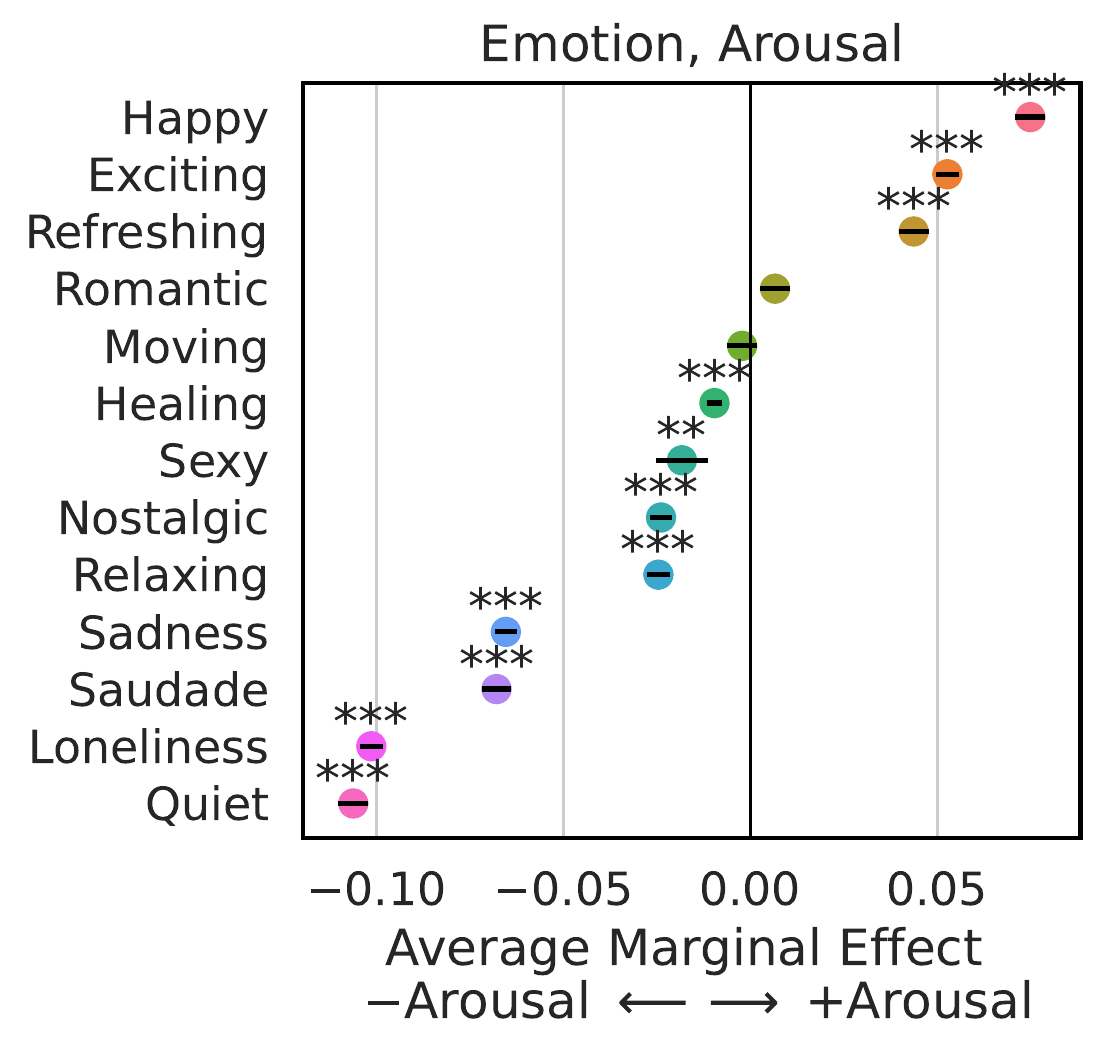}} \\
    \raisebox{-0.5\height}{\includegraphics[width=0.23\textwidth]{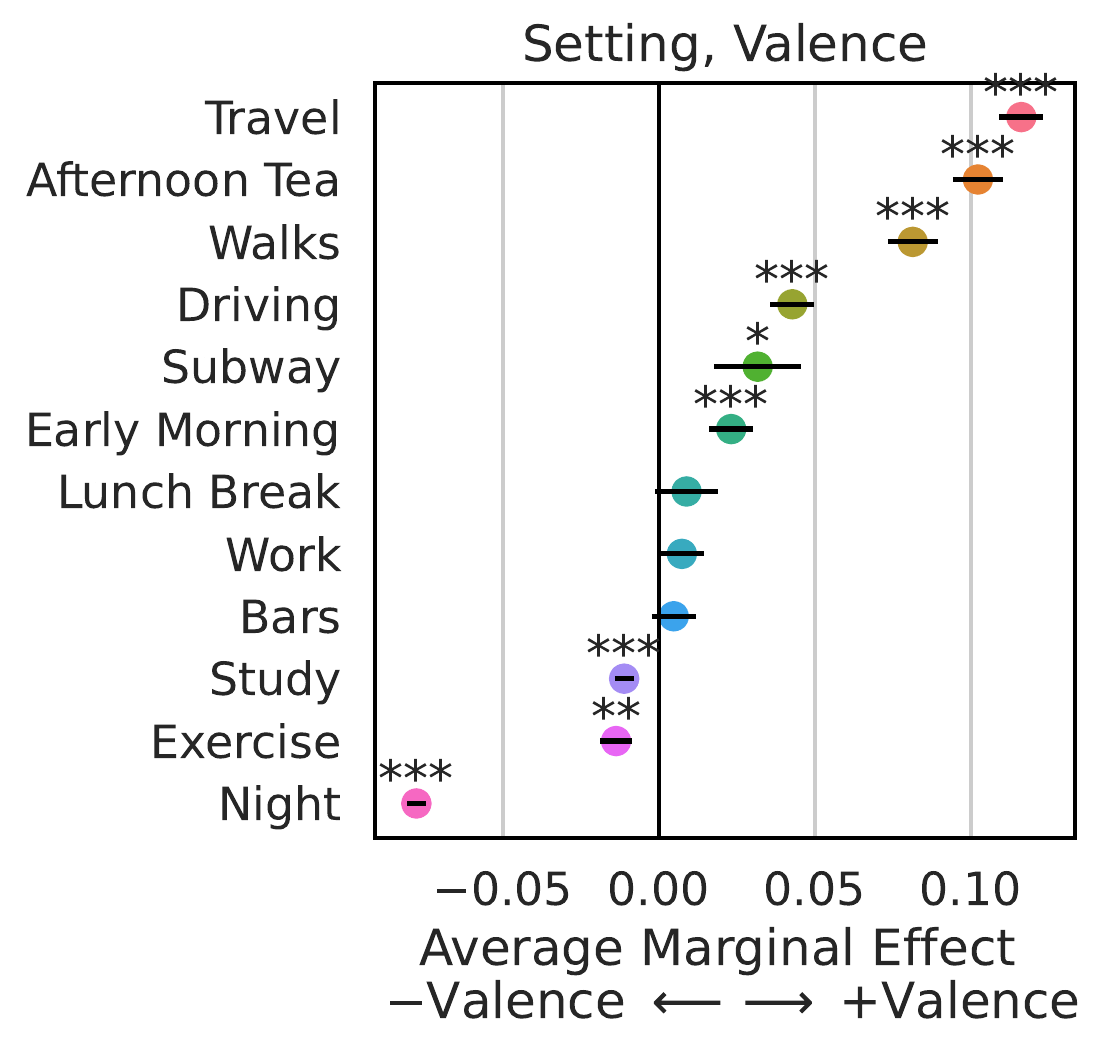}}
    \raisebox{-0.5\height}{\includegraphics[width=0.23\textwidth]{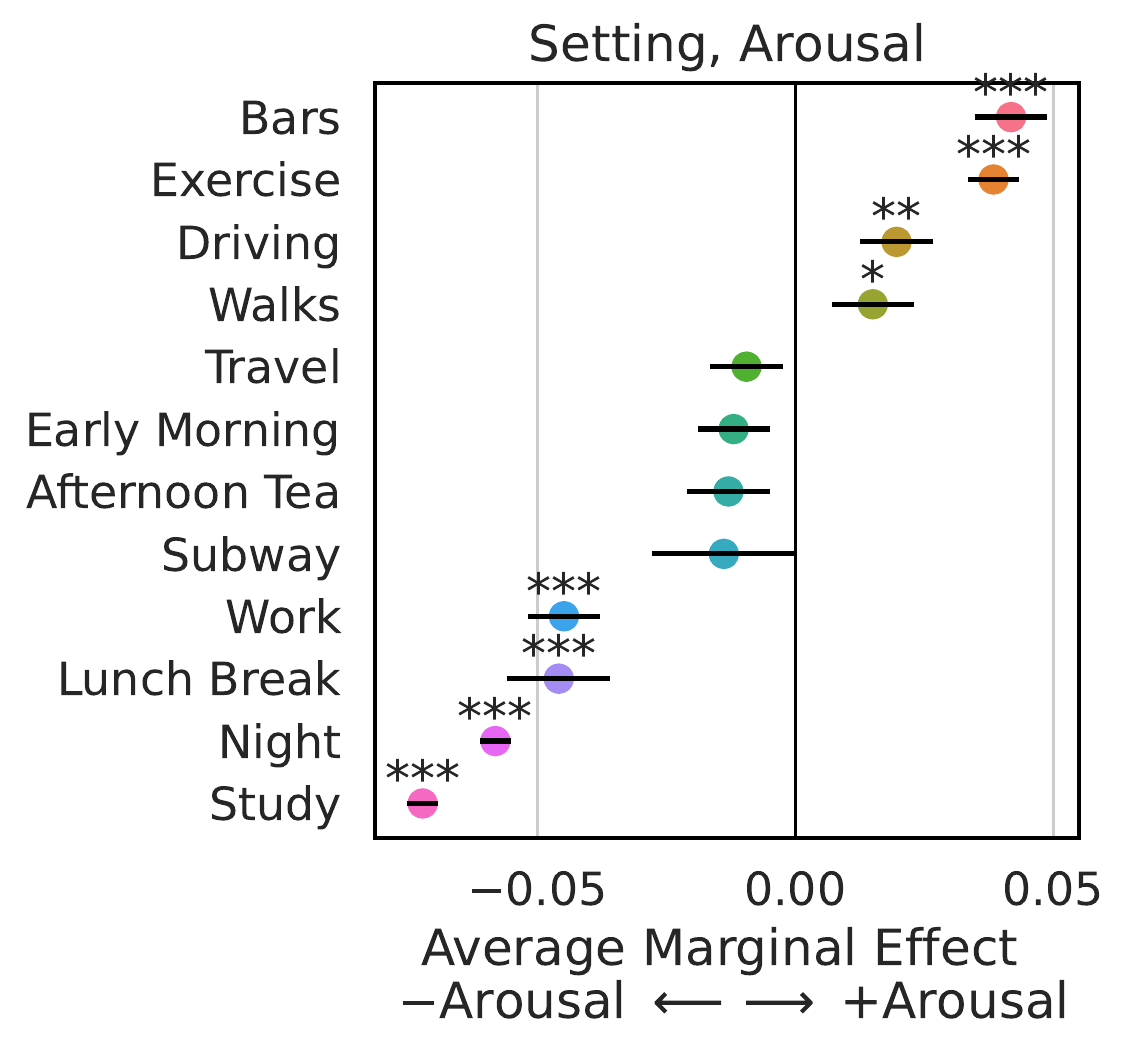}}
    \caption{Averaged marginal effects of contextual choices in emotion-tagged (top) and setting-tagged (bottom) playlists on listener affective responses; standard errors are shown.} 
    \label{fig:setting}
\end{figure}

Extramusical factors such as listening context (e.g., listening to music when grabbing coffee vs. when exercising) also influence the emotional effects of music \cite{sloboda2001emotions, greasley2011exploring, vuoskoski2015extramusical}. %
Prior work has primarily utilized experience sampling methods \cite{csikszentmihalyi1989optimal} to study musical experiences in everyday contexts---where participants are polled at random intervals during the day---though generalizations to the population at large have proved difficult with small sample sizes \cite{sloboda2001functions, juslin2008experience}. 
While we are unable to obtain information about the physical setting a user was in (i.e. that a user was exercising when listening to a song), here, using our data on playlists and treating the choices of users in listening to playlists of specific types as context, we tease out the marginal effects that these choices have on affective responses.

\myparagraph{Choice as Context.} We obtain context variables on 1.36M playlists through their tags, used by creators to label individual playlists. Tags consist of a set of physical setting (e.g.,  afternoon tea), emotional (e.g.,  nostalgic), and thematic (e.g., video game music) categories, in addition to language (e.g.,  Chinese) and stylistic (e.g.,  jazz) labels. As users primarily discover new playlists within the platform by browsing specific tags, we treat these tags as implicit signals of \textit{choice} with these listening contexts, aiming to identify those that may differ on the emotional responses produced—i.e. that a user chose to listen to an exercise tagged playlist rather than an afternoon tea tagged one—noting that we do not make explicit causal assumptions behind the factors that led to these user choices.

\begin{figure*}[!t]
    \centering
    \begin{tabular}{ccc}
    \subfloat[Tempo, Women/Men]{\includegraphics[width=0.30\textwidth]{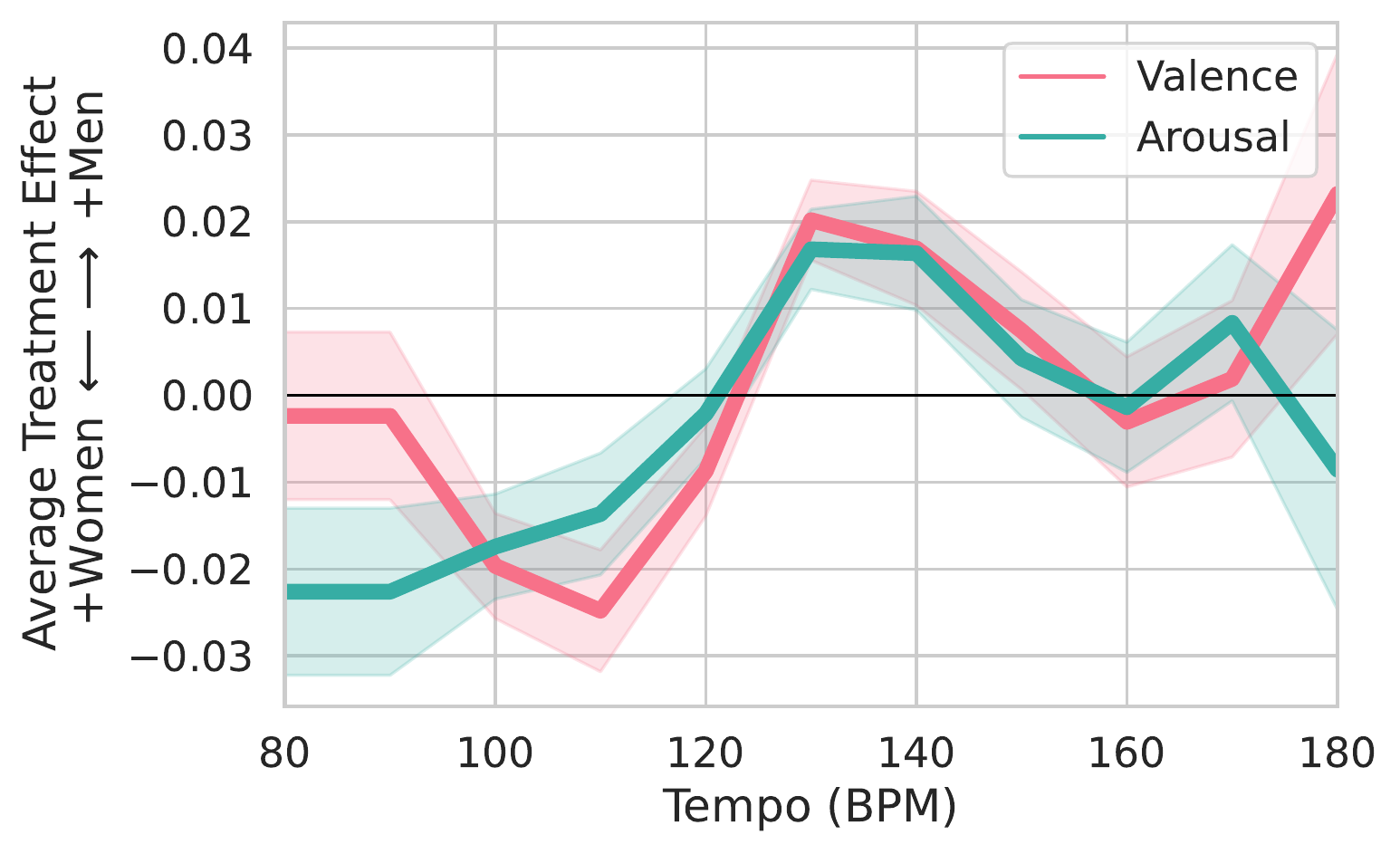}} & 
    \subfloat[Loudness, Women/Men]{\includegraphics[width=0.30\textwidth]{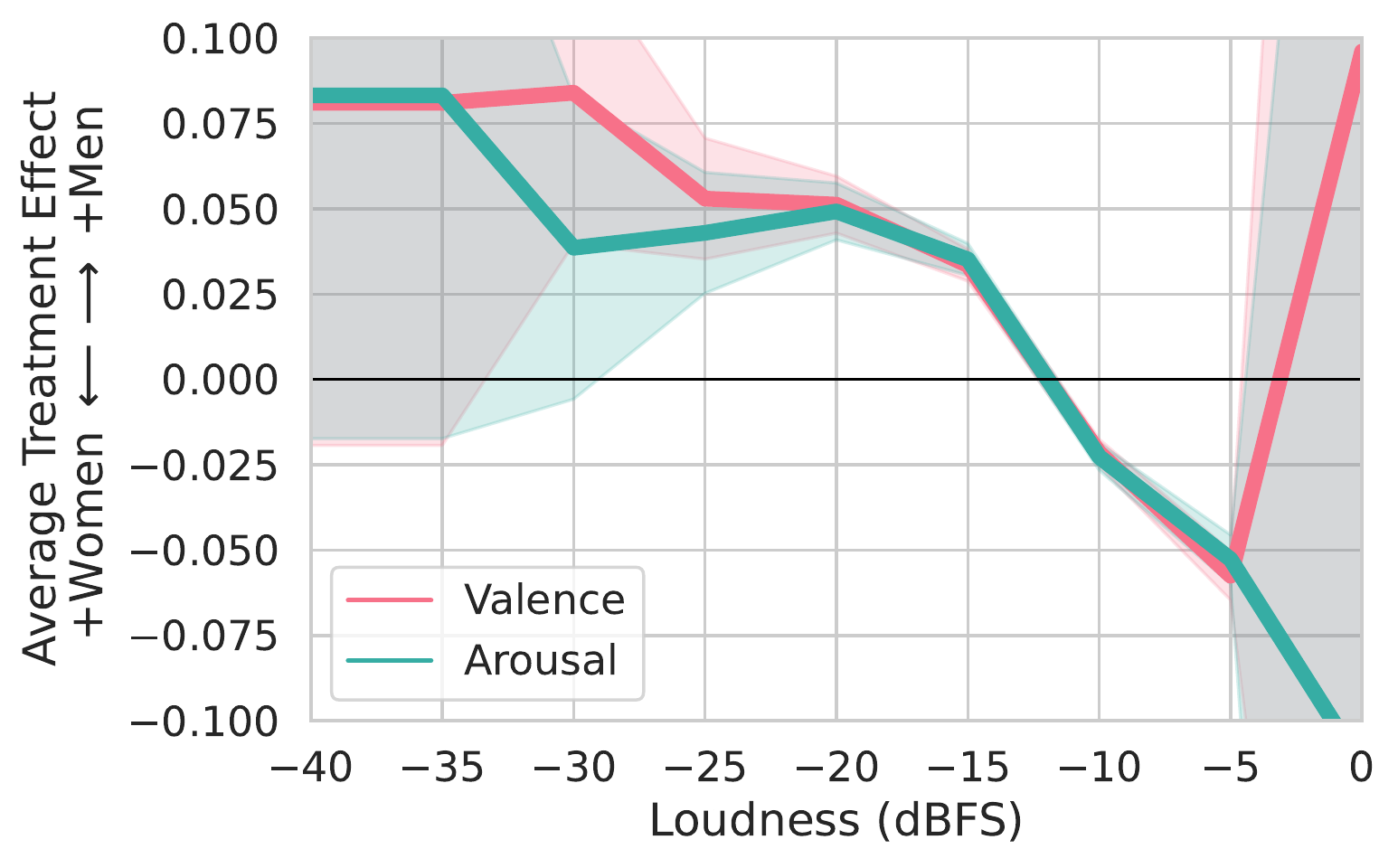}} &
    \subfloat[Mode, Women/Men]{\includegraphics[width=0.30\textwidth]{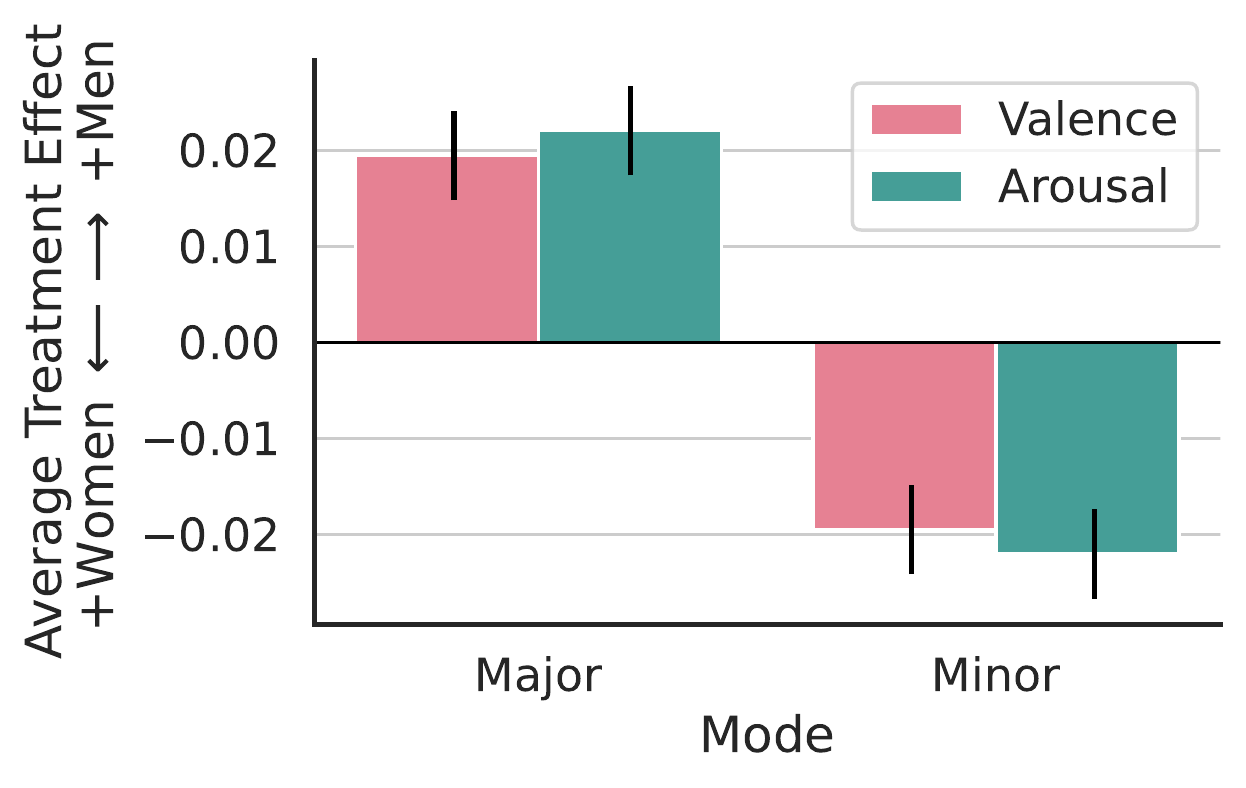}} \\
    \subfloat[Negative Emotion, Women/Men]{\includegraphics[width=0.30\textwidth]{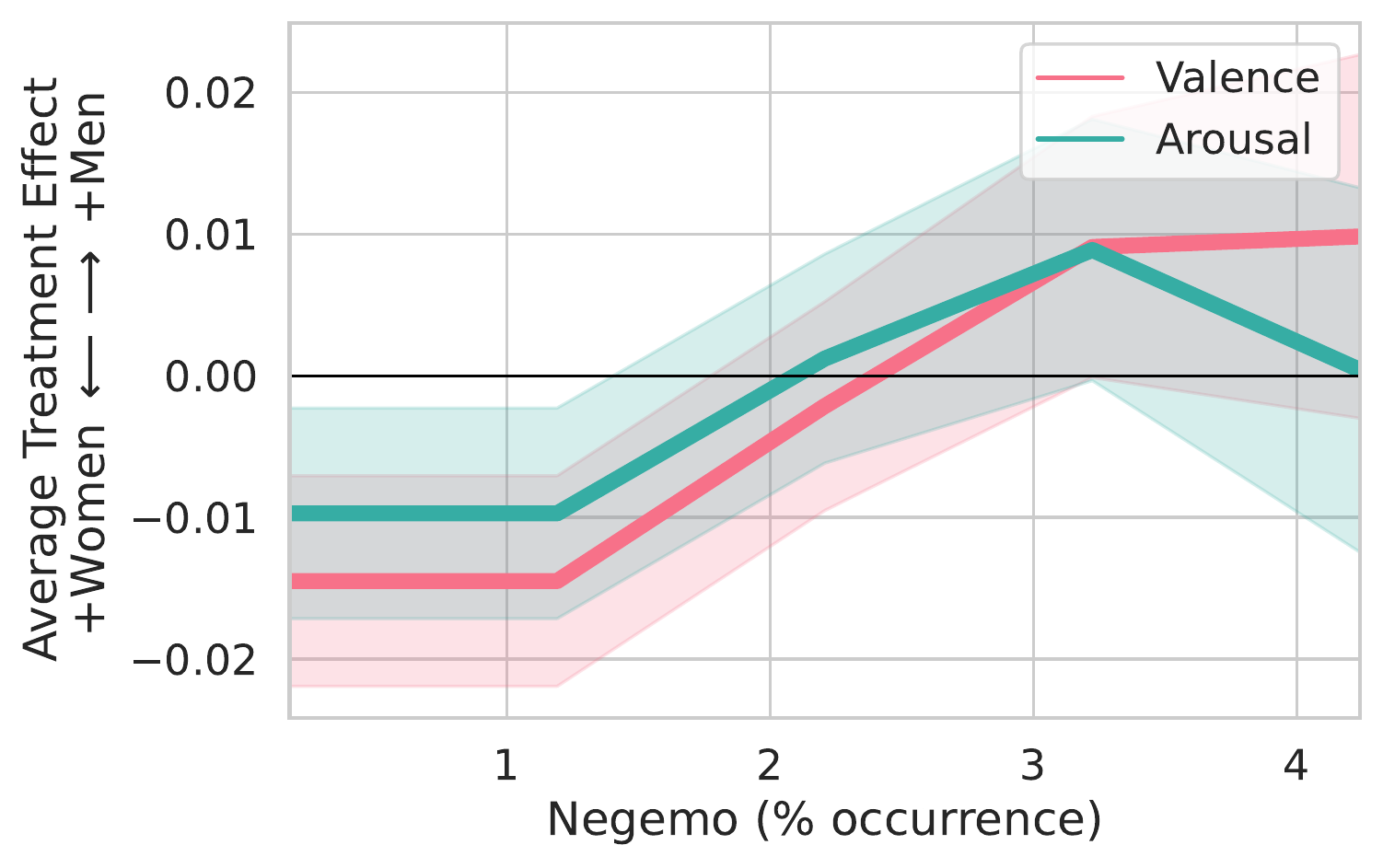}} &
    \subfloat[Hardness, b.i.t. 90s/b.i.t. 00s]{\includegraphics[width=0.30\textwidth]{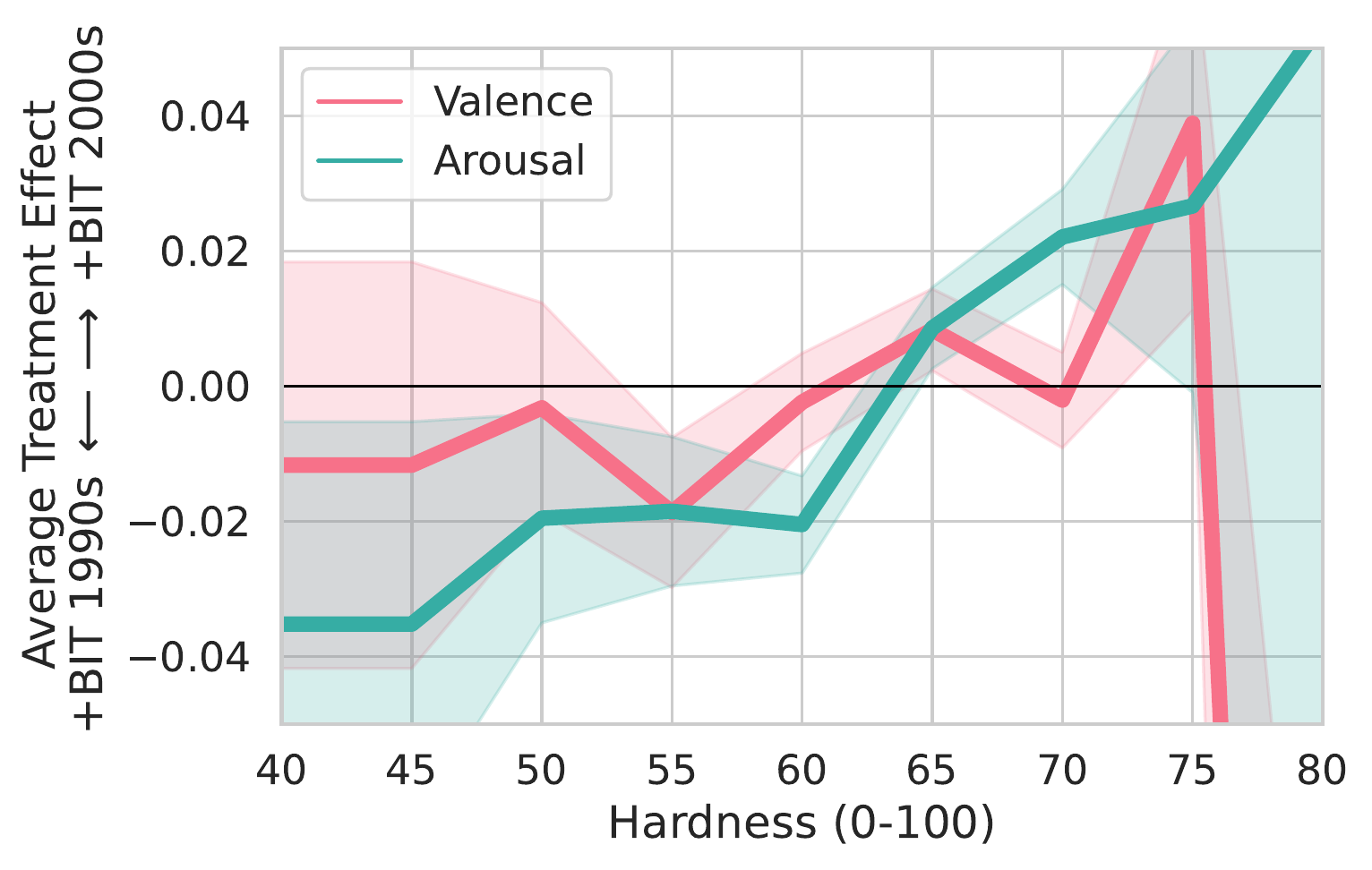}} & 
    \subfloat[Mode, b.i.t. 90s/b.i.t. 00s]{\includegraphics[width=0.30\textwidth]{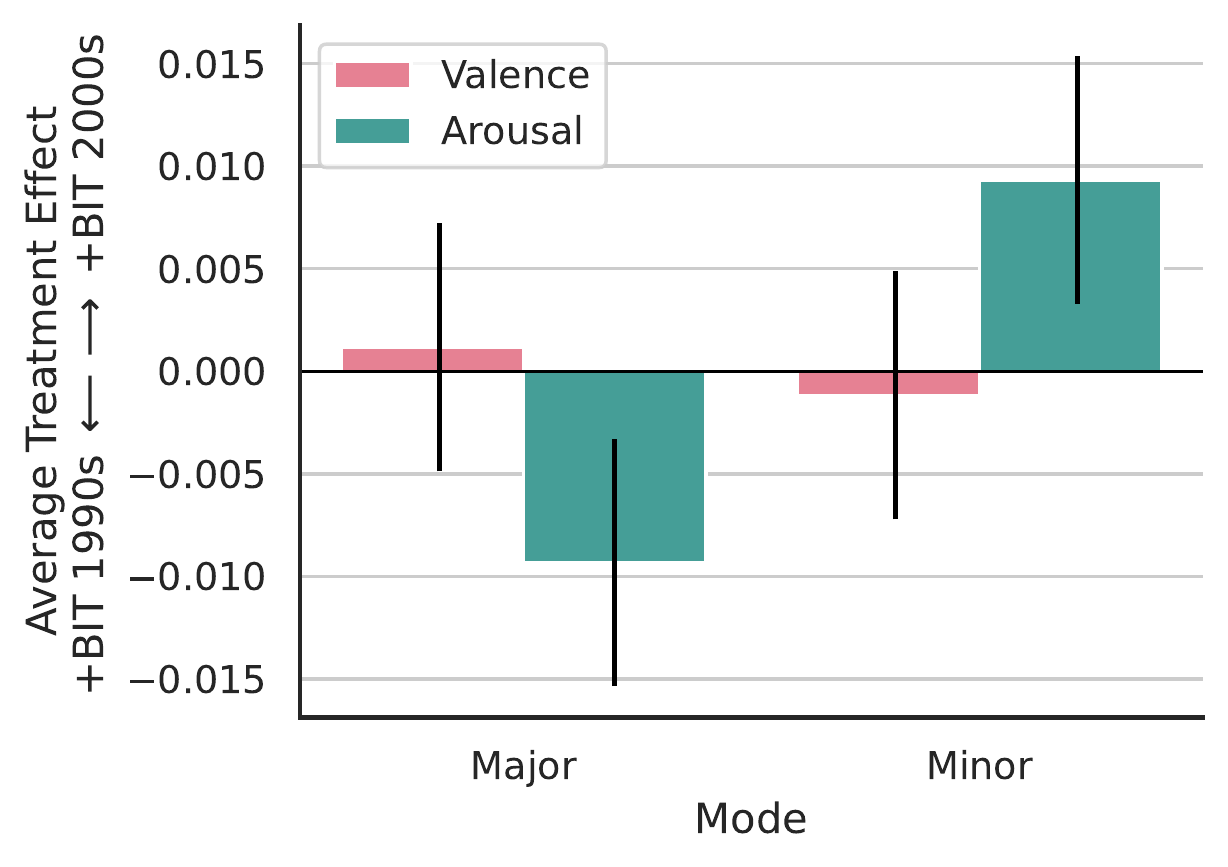}} \\ 
    \end{tabular}
    \caption{
    Relative average treatment effects of gender (women/men) and age-based (born in the (b.i.t.) 1990s/b.i.t. 2000s) demographic groups on listener response valence and arousal against musical and lyrical features. Standard errors are shown; \textcolor{red}{valence} in \textcolor{red}{red}, \textcolor{blue}{arousal} in \textcolor{blue}{blue}.
    The complete set of figures for musical and lyrical features are shown in Appendix Section \ref{sec:expandedresults}, Figures \ref{fig:demographics_expanded_women_men}-\ref{fig:demographics_expanded_1990_2000}.
    }
    \label{fig:demographics}
\end{figure*}

\myparagraph{Methods.} To identify the marginal effects of contextual choices on affective responses, we fit separate mixed-effect multivariable linear regression models on response valence and arousal, including tagged category indicator variables as features and control for listener demographics. To further control for differences between playlist songs, we include them as random effects; for computational tractability, we include only random effects for songs that are labeled with 10 or more unique tags.

\myparagraph{Results.} Cultivating affect is a driving reason behind why users create playlists \cite{denora2000music, siles2019genres}, and our results point to how playlists created by users are also generally successful at cultivating these affects among the general user population as well. As shown in Figure \ref{fig:setting}, playlists tagged by leisurely activity categories corresponded to the highest positivity in responses, and are consistent with prior work on stress levels in everyday situations \cite{vastfjall2012music}, while arousal trends mirror diurnal shifts in emotion and physical activity \cite{golder2011diurnal}. Expanded results for all tagged categories are shown in Appendix Section \ref{sec:expandedresults}, Figures \ref{fig:setting_overall_setting}-\ref{fig:setting_overall_language}.

\subsection{Demographic Variations}

Individuality is a driving factor in how listeners experience musically-evoked emotions \cite{yang2007music, juslin2008emotional, gomez2021music}. However, measuring \textit{how} individual differences affect emotional responses to music has proved challenging, with many researchers citing the insufficiency of typical experiment scale as a primary reason \cite{juslin2008experience, lundqvist2009emotional, cameron2013waiting}, especially in the presence of confounders. For listener demographics, prior work has seen conflicting observations of how demographic effects modulate affective responses against musical features. For example, some observe that age and gender modulate emotional responses against tempo, mode, volume, and pitch \cite{webster2005emotional, chen2020felt}, while others report the absence of such demographic effects or even contrasting observations \cite{robazza1994emotional, cameron2013waiting}. These contrary results might be due to variable experimental setups between studies, wherein the method of measurement will often interfere with the 
experience itself \cite{gabrielsson2010strong}. This raises the importance of studying emotional reactions in a natural setting when analyzing affective responses to music in everyday situations.
Here, we test for demographic differences in affective responses in relation to
song features using our data.

\myparagraph{Demographic Variables.} Our analysis focuses on two main demographic variables, namely listener gender\footnote{Platform-provided options for gender are limited to binary options (\chinese{男} man and \chinese{女} woman); see \nameref{sec:ethical}.} and age.\footnote{In the form of ''\chinese{XX后}'', e.g. \chinese{00后} refers to individuals that are born between 2000 and 2010.} We operate within the constraints of platform-provided choices in user registration for our variable categories and use only publicly displayed user data in our analysis.

\myparagraph{Methods.} To test for differences between pairs of demographic groups in their affective responses to musical and lyrical features, we formalize alternations between groups as \textit{treatments} and compute average treatment effects (ATE). In order to account for covariates and reduce bias due to confounding variables, we construct a multi-modal stratified propensity score matching (PSM) model as a quasi-causal analysis of demographic effects. Here, we formalize comments as \textit{subjects}; the propensity score, defined traditionally as the likelihood of being assigned to a treatment group based on observed characteristics of the subject \cite{rosenbaum1983central}, is thus a scaled estimate of the likelihood of a commenter being of a demographic group $g_i$ given a set of song features $f_i$, or $P(g_i|f_i)$. We estimate this probability---the propensity score---via logistic regression on a song's musical and lyrical features, and match data points within stratified deciles of this score to mitigate confounding bias \cite{rosenbaum1984reducing, paul2017feature}. Within these matched and stratified deciles, we fit separate linear regression models on response valence and arousal against specific song features, weighting and pooling stratum-specific estimated treatment effects to estimate the ATE \cite{imbens2004nonparametric} and its variance \cite{lunceford2004stratification}. Consistent with prior work in musical emotions \cite{kamenetsky1997effect} and in social psychology on how cultural constructions of gender may account for differences in emotional display \cite{bem1974measurement}, we observe that response valence and arousal by demographic groups differ in their distributions---for example, as shown in Appendix Section \ref{sec:demographicbaselines}, comments made by women are on average higher in both valence and arousal than those made by men. Therefore, we test specifically for standardized \textit{change} in affective responses across song features within demographic groups. Finally, as in Section \ref{sec:musicalandlyricalvariations}, we stratify continuous variables in our analyses into fixed-length categories and estimate the ATE of each stratum.

\myparagraph{Results.} Shown in Figure \ref{fig:demographics}, we find that listener age and gender both modulated affective responses to statistically significant degrees across a series of musical and lyrical features. Compared to men, women had more intensely positive affective reactions for songs that were louder (>-12 dBFS), of lower tempo (<120 BPM), of minimal tempo standard deviation (<5), of minor mode, and that had reverb; though gender differences often diminished (i.e. tempo >160BPM) or became statistically insignificant at feature extremities. Lyrically, women were affected more negatively with a greater proportion of negemo terms, while men were affected more positively for posemo terms, consistent with observed gendered responses in other mediums \cite{bradley2001emotion, fernandez2012physiological}. Compared to women, men had more intensely positive affective responses to darker, flatter, softer, smoother, and warmer timbral characteristics. Age effects were much less pronounced, with those born in the 2000s reacting more positively to harder and rougher timbral features as well as minor modes than those born in the 1990s.
Full results are shown in Appendix Section \ref{sec:expandedresults}, Figures \ref{fig:demographics_expanded_women_men} to \ref{fig:demographics_expanded_1990_2000}.

Affective experiences are central reasons for music consumption \cite{denora1999music}, though music choices often misalign with intended well-being outcomes \cite{stewart2019music}. We hope our work further facilitates more effective and more intentional music choices in daily consumption to achieve these desired results.
Under appraisal theory, affective responses are learned and conditioned through individual lived experiences rather than innate to certain biological factors \cite{brody2010gender}. These findings on demographic effects should then be interpreted to be \textit{products} of the social norms, values, and lived experiences \cite{de2016contesting} of those who may self-identify under the broad demographic groups in question; with the platform,
song, and comment board
being part of the context in which these emotions are deployed.

\section{Disclosures of Mental Health Disorders}
\label{sec:mentalhealthmain}

In the context of social media, given the frequent benefits of anonymity \cite{de2014mental} and social connectedness \cite{bazarova2014self}, self-disclosures of personal details can be a method to find social support, advice, and belonging \cite{ernala2018characterizing, yang2019channel}.
This phenomenon gives life to \chinese{"网抑云,"} which refers to the outpour of emotional and personal comments on the social music platform, especially late at night and under sad songs.
While known colloquially and in popular culture,\footnote{``\chinese{网抑云}'' was termed one of 2020's top ten internet buzzwords, \href{https://baike.baidu.com/item/2020\%E5\%8D\%81\%E5\%A4\%A7\%E7\%BD\%91\%E7\%BB\%9C\%E7\%83\%AD\%E8\%AF\%8D/54318377}{\chinese{https://baike.baidu.com/item/2020十大网络热词/}}} the mechanisms behind self-disclosure phenomena in the context of social music platforms are not well understood.
Motivated to better understand disclosures of mental health disorders in a musically-situated social environment, we frame them as affective responses \cite{ho2018psychological} and test for factors driving this behavior, the social support they receive, and differences in discloser user activity.
Addressing these unknowns will help us understand how users may use social music platforms for therapeutic purposes \cite{schriewer2016music} and guide us to better support vulnerable and at-risk individuals.

\begin{figure}[!t]
    \centering
    \raisebox{-0.5\height}{\includegraphics[width=0.40\textwidth]{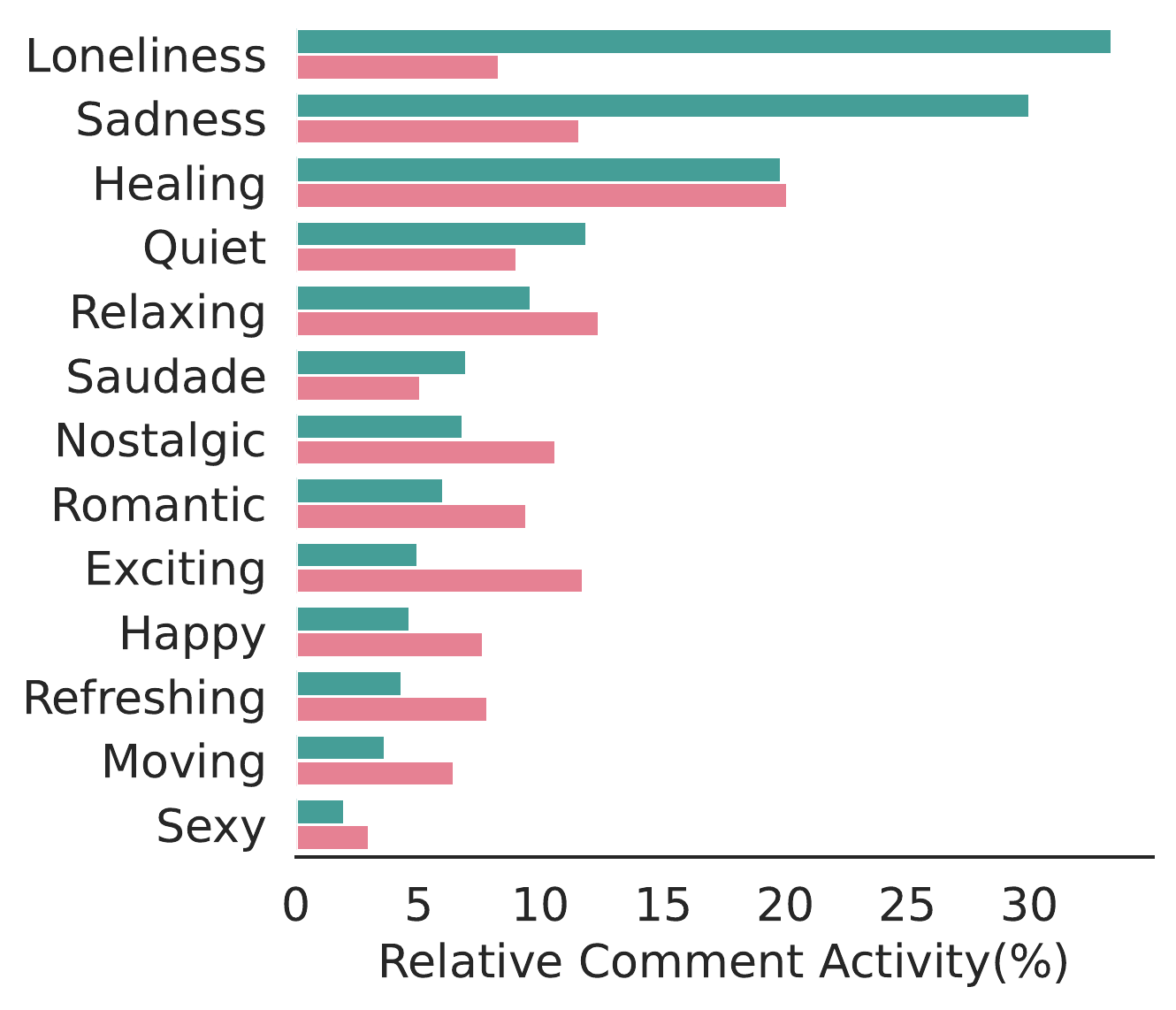}} \\
    \raisebox{-0.5\height}{\includegraphics[width=0.40\textwidth]{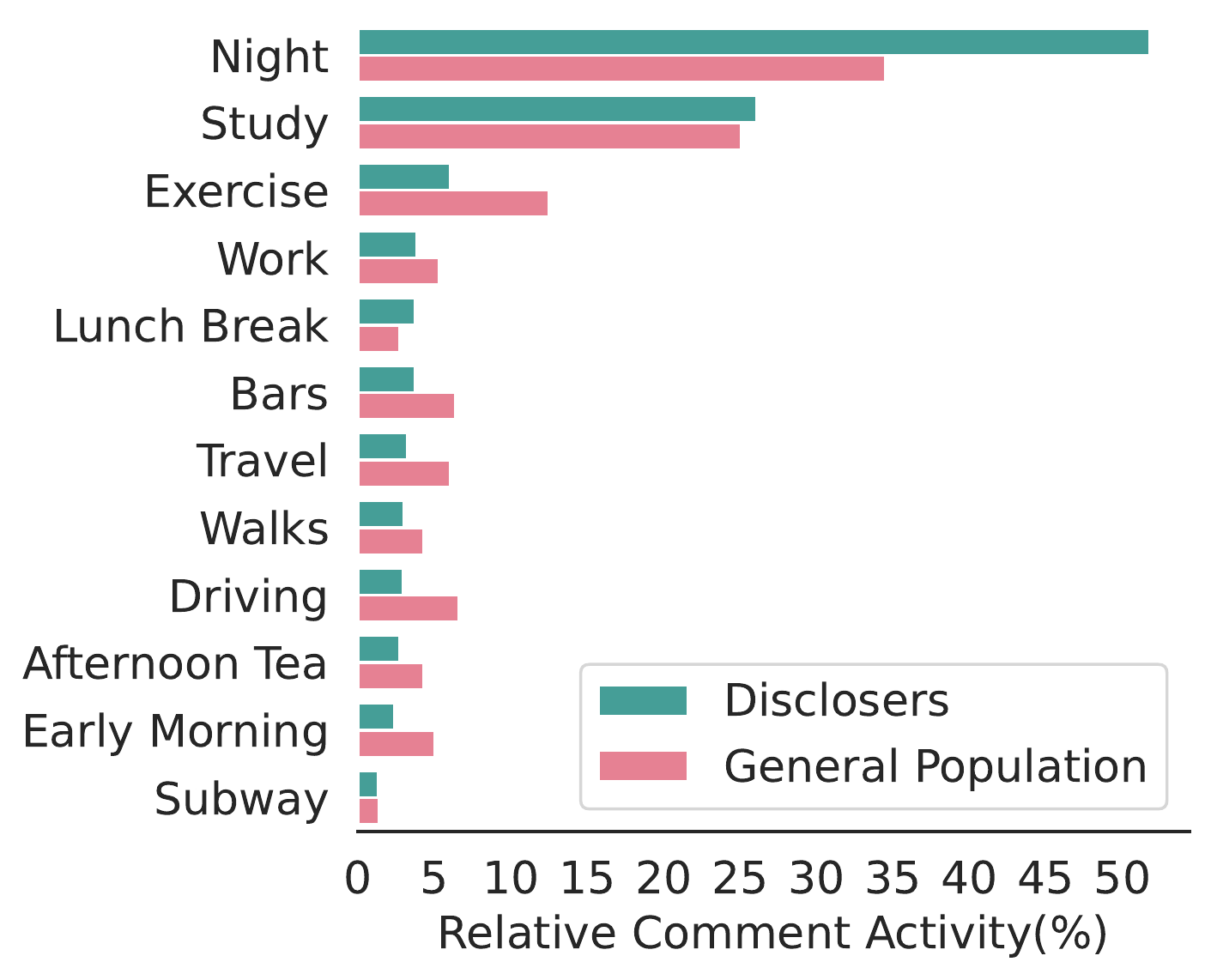}}
    \caption{Relative tagged playlist commenting activity between \textcolor{blue}{disclosers} and the set of \textcolor{red}{all users} on emotion (top) and setting (bottom) tagged playlists. Note that as each playlist may have up to three unique tags, relative tag percentages do not add up to 100\%. The complete set of figures for all playlist tag categories are shown in Appendix Section \ref{sec:mentalhealthdisclosures}, Figure \ref{fig:playlistengagement}.} 
    \label{fig:engagement}
\end{figure}

\myparagraph{Dataset Collection.} In the absence of clinically-aligned user data \cite{harrigian2020state}, we source disorder terms from the DSM-5-TR\footnote{
See \nameref{sec:limitations} for a discussion on method caveats and standards of diagnosis in clinical psychology.} \cite{american2013diagnostic} and utilize regular expressions to identify disclosures of self-reported statements of diagnosis \cite{coppersmith2014quantifying, coppersmith2015adhd, cohan2018smhd} for mental health disorders in music comments. 
Two Chinese native speakers
further manually filter for %
genuine statements of disclosure
(i.e., excluding jokes, quotes, and clearly disingenuous statements),
resulting in %
1133 users with self-reported mental health disorders.
We find that, out of all disclosers, most disclose depression (81.2\%), anxiety (19.9\%), and bipolar (18.5\%) disorders; additionally, most users (60.6\%) self-identify as women, consistent with constituent gender differences of affective disorders in national studies \cite{huang2019prevalence}.
Disclosers show greater platform usage (Kolmogorov-Smirnov on user levels, p<0.01),
insomnia-aligned diurnal user activity consistent with disorder symptoms \cite{taylor2005epidemiology, harvey2008sleep}, 
increased engagement with playlists of sadder natures, e.g., as shown in Figure \ref{fig:engagement}, \texttt{loneliness} (+302\%), \texttt{sadness} (+158\%), and \texttt{night} (+50.1\%),  
and decreased engagement with playlists of more active natures, e.g., \texttt{exercise} (-51.7\%),
compared to %
typical users. These observations mirror affective disorder activity trends \cite{cooney2013exercise} and suggest that people with affective disorders are more likely to use music reflective of negative emotions than positive emotions to manage feelings of sadness and depression \cite{stewart2019music}.
A detailed breakdown of our data, comorbidities, and our specific regular expressions are described in Appendix Section \ref{sec:mentalhealthdisclosures}.

\myparagraph{Affective Response.} Treating the act of self-disclosure as an affective response, we test for factors driving this behavior. Statements of self-disclosure are more likely to appear as top-level comments (78.1\%) than as replies (21.9\%); 
top-level disclosures are biased towards songs with features generally associated with sadness \cite{juslin2004expression} (p<0.01 for all features other than tempo, Kolmogorov-Smirnov), i.e. softer songs with minor modes, and towards playlists with tags of the same nature, while disclosures in reply occur to comments
that indicate emotional distress, %
or
that are themselves replies to existing comments made by the discloser (``\chinese{怎么了}'', meaning ``what's wrong''). 
Responses to the first are split in function, with disclosers either expressing their diagnosis in empathy for encouragement (``\chinese{...我一年前也确诊了，事情会好起来的}'', meaning ``...I was diagnosed a year ago, things will be better''), or to commiserate (``\chinese{我也确诊了，活着好难...}'', meaning ``I was diagnosed too, living is so hard...''), showing evidence of resonance \cite{miller2015resonance, rosa2019resonance} and high person-centered condolence \cite{high2012review}.

\myparagraph{Social Support.} Characterizing audience engagement around self-disclosure comments in their content, we identify supportive comments according to the four major classes of social support around health concerns---prescriptive, informational, instrumental, and emotional support---from established literature \cite{turner1983social, george1989social, de2014mental} and label the main type of support each comment falls under. We then fit logistic regression models on the dependent variable of receipt, aiming to identify \textit{where} users are more likely to receive a supportive comment in response to disclosure; including song features as independent variables and song popularity, comment length, user demographics, and %
comment LDA topic distributions as controls. 
We observe that emotional (52\%, e.g., ``\chinese{加油，事情一定会好起来的我保证}'', meaning ``good luck, everything will be better I promise'') and prescriptive support (31\%, e.g., ``\chinese{听一些令人振奋的歌曲吧}'', meaning  ``listen to heart raising songs'') largely exceeds informational (9\%, e.g., ``\chinese{...治疗可能会有帮助，两年治疗后我...}'', meaning ``...therapy could help, after two years of therapy I...'') and instrumental (8\%, e.g., ``\chinese{...你可以私聊我}'', meaning ``...you can private message me'') forms in response to disclosures. 
Several psycholinguistic lyrical features proved statistically significant (p<0.05) in predicting if a disclosure comment to a song would receive a supportive reply; the rate of terms in lyrics relating to social processes, specifically friend (+2.23) and ingest (+0.97), positively predict this prosocial behavior, and negative emotion terms (-2.04) do so negatively, mirroring negative correlations between sadness and prosocial tendencies \cite{ye2020effect}. For musical features, only reverberation did so positively (+0.90). 
While past work has studied the prosocial effects of music, most have only used a limited set of author-chosen songs \cite{greitemeyer2009effects, kennedy2013relationship} or crowd-sourced prosocial perceptions \cite{ruth2017heal}; here, we specifically identify what makes for prosocial songs and situate our study in the context of social support to mental health self-disclosures.
Taken together, these observations not only provide ample pointers for music therapists on musical and dyadic conversational means for more successful emotion-focused interventions \cite{jensen2001effects}
but also guide users on how to effectively find social support on the platform when needed \cite{de2017language}.

\section{Discussion and Future Work}

In this work, we sought to examine the driving factors behind variations in emotional reactions to music, via a large-scale computational study of a Chinese social music platform. Our analyses here reveal several nuances in how idiosyncratic variables elicit emotional responses, with a degree of precision that prior studies have often lacked thus far. In a case study of mental health self-disclosures in music comments, we characterized a type of discourse in the context of a popular social phenomenon, demonstrated the importance of posting location in determining the social support disclosures would receive, and revealed several factors driving the prosociality of music in this context. We see our present work situated in the broader context of studying emotionality in music and in the design of platforms to promote healthier interactions more centered on user well-being. Here, we highlight a few limitations and directions for future work; models, code, and anonymized data are made available at \href{https://github.com/skychwang/music-emotions}{https://github.com/skychwang/music-emotions}.

The music we listen to has a strong effect on our moods \cite{mccraty1998effects}. The integration of emotional response analysis into music recommendation systems could promote healthier recommendations \cite{konstan2012recommender, singh2020building} more cognizant of listener well-being outcomes. No one size fits all, and more sophisticated analyses could better capture more factors that explain emotional response variations towards creating more personalized music emotion recommendation systems. 

While our work measures the effects of demographic variables on emotional responses, there remains a bio-psycho-social question on identifying the causes behind \textit{why} this variation exists as it relates to song features. Lived experiences condition our emotions \cite{brody1997gender}; future work could aim, through significant theoretical and qualitative study, to better identify the relationships and causes behind these variation outcomes. 

Several open questions also remain as to whether \textit{risk} may be qualified in this context in relation to well-being. Specifically, it would be interesting to study how recommendation interactions may disproportionately affect those afflicted with mental health disorders, and how we may design platforms, in the context of well-being outcomes, under normative goals of equity and distributive justice \cite{rawls2001justice}.

\section*{Ethical Considerations}
\label{sec:ethical}

\myparagraph{Data Release.} For user comments, taking user privacy considerations into account, we release the set of comment ids used in our analyses---which researchers are able to use in conjunction with the Netease API to obtain original comment content---mirroring Twitter data release guidelines for academic research.

\myparagraph{Identity Affiliation.} In studying demographic effects, we examine only the aggregate behavior of users who make public their demographic self-identification \textit{choices} during registration under platform constraints. In particular, we note that platform choices for gender are limited only to binary options---\chinese{男} men and \chinese{女} women. These choices should not be interpreted to have taken into account gender fluidity considerations or the multidimensional spectrum of gender identities \cite{larson-2017-gender}. 

\section*{Limitations}
\label{sec:limitations}

\myparagraph{Measuring Affective Response.} In particular, we mirror the concerns by \citet{mohammad2020practical}; notably, that (1) emotion lexicons are limited in coverage and do not include all possible terms in a language, and that (2) as languages and, in particular, our perceptions of words in them are by nature entities of change that inherently possess socio-cultural variations, emotion scores for words are not immutable, neither longitudinally nor socio-culturally. As such, while we have attempted to mitigate for this limitation by (1) choosing the largest Chinese emotion lexicon annotated for words sourced from the domain of social media and (2) comparing our findings to that of previous smaller-scale in-person studies that use varying methods to measure emotion when possible---even as no ``gold standard'' measure of emotional response exists, physiological, behavioral, or otherwise \cite{mauss2009measures}---we encourage future work to further examine these phenomena in a greater variety of contexts. Further, our study does not make explicit causal claims around factors of music choice and user predisposition, i.e. what caused users to choose to listen to a specific song, or what their states of mind were prior to making this choice. While our work shows evidence of variations in affective responses correlated with musical, lyrical, demographic, and mental health factors, like the quasi-causal results estimating demographic effects on listener affective responses, we do not argue that these alone explain the entirety of the associated variations. In moving towards truly causal studies \cite{feder2021causal}, we encourage further direct participatory work to further examine these observations in larger, more controlled, and even cross-cultural contexts.

\myparagraph{Censorship and Moderation.} Users are able to report comments that violate platform rules,\footnote{\href{http://music.163.com/m/topic/13336053}{http://music.163.com/m/topic/13336053}} and active moderation of user content exists on the platform. As we use only public posts on the platform, it is thus important to interpret our findings in the context of internet censorship in China \cite{vuori2015lexicon}. In particular, as noted by previous studies on mental health postings in Chinese social media \cite{cui2022social}, comments that go against certain government objectives---such as the ``stability and unity for a harmonious society'' \cite{wang2012china}, which mental health-related postings may go against---are often censored \cite{paltemaa2020meta}. While pilot tests matching regular expressions on such phrases within platform comments still yielded significant quantities, the degree of censorship that these types of comments receive remains unclear.

\myparagraph{Statements of Diagnosis.} As we study users with self-reported statements of diagnosis, our method only potentially captures a sub-population of each disorder---those who choose to disclose a diagnosis on a public platform under the option of anonymity. While we have attempted to increase the precision of identifying individuals who are diagnosed with specific disorders through 
significant manual annotation, 
in the lack of clinically-aligned user data, we nonetheless are unable to verify if genuine-appearing disclosures of mental health disorder diagnoses are ultimately truthful. However, as noted by \cite{coppersmith2014quantifying}, given the stigmas often associated with mental illnesses, it seems unlikely that users would disclose that they were diagnosed with a condition they do not possess. Individuals who may be diagnosed with affective disorders undoubtedly also remain in the set of all users that we compare disclosers against and, as such, our results on platform user activity differences should only be interpreted in the context of discovering broad themes---not as ground truths of comparisons between those who are diagnosed and those who aren't. 
Finally, we also note concerns in clinical psychology on the heterogeneity of psychiatric diagnoses, which remains contentious in current literature. Notably, that standards of diagnosis all use different decision-making rules, that significant overlaps exist in symptoms between diagnoses,
and that they may instead mask the complex underlying causes of human distress with potentially scientifically meaningless labels
\cite{allsopp2019heterogeneity}.

\section*{Acknowledgements}

We thank David Jurgens, Michelle Cohn, Xiang Zhou, Maximillian Chen, Kexin Fan, and the anonymous reviewers for their helpful comments, thoughts, and discussions. This material is based upon work supported by the National Science Foundation Graduate Research Fellowship under Grant No. DGE-2036197. 

\bibliography{emnlp2022}

\begin{thebibliography}{120}
\expandafter\ifx\csname natexlab\endcsname\relax\def\natexlab#1{#1}\fi

\bibitem[{Allsopp et~al.(2019)Allsopp, Read, Corcoran, and
  Kinderman}]{allsopp2019heterogeneity}
Kate Allsopp, John Read, Rhiannon Corcoran, and Peter Kinderman. 2019.
\newblock Heterogeneity in psychiatric diagnostic classification.
\newblock \emph{Psychiatry research}, 279:15--22.

\bibitem[{{American Psychiatric Association}(2022)}]{american2013diagnostic}
{American Psychiatric Association}. 2022.
\newblock \emph{Diagnostic and statistical manual of mental disorders, fifth
  edition, text revision}.
\newblock American psychiatric association Washington, DC.

\bibitem[{Atkinson and Flint(2001)}]{atkinson2001accessing}
Rowland Atkinson and John Flint. 2001.
\newblock Accessing hidden and hard-to-reach populations: Snowball research
  strategies.
\newblock \emph{Social research update}, 33(1):1--4.

\bibitem[{Bareeqa et~al.(2021)Bareeqa, Ahmed, Samar, Yasin, Zehra, Monese, and
  Gouthro}]{bareeqa2021prevalence}
Syeda~Beenish Bareeqa, Syed~Ijlal Ahmed, Syeda~Sana Samar, Waqas Yasin, Sani
  Zehra, George~M Monese, and Robert~V Gouthro. 2021.
\newblock Prevalence of depression, anxiety and stress in china during covid-19
  pandemic: A systematic review with meta-analysis.
\newblock \emph{The International Journal of Psychiatry in Medicine},
  56(4):210--227.

\bibitem[{Bazarova and Choi(2014)}]{bazarova2014self}
Natalya~N Bazarova and Yoon~Hyung Choi. 2014.
\newblock Self-disclosure in social media: Extending the functional approach to
  disclosure motivations and characteristics on social network sites.
\newblock \emph{Journal of Communication}, 64(4):635--657.

\bibitem[{Bem(1974)}]{bem1974measurement}
Sandra~L Bem. 1974.
\newblock The measurement of psychological androgyny.
\newblock \emph{Journal of consulting and clinical psychology}, 42(2):155.

\bibitem[{Beveridge and Knox(2018)}]{beveridge2018popular}
Scott Beveridge and Don Knox. 2018.
\newblock Popular music and the role of vocal melody in perceived emotion.
\newblock \emph{Psychology of Music}, 46(3):411--423.

\bibitem[{Bradley et~al.(2001)Bradley, Codispoti, Sabatinelli, and
  Lang}]{bradley2001emotion}
Margaret~M Bradley, Maurizio Codispoti, Dean Sabatinelli, and Peter~J Lang.
  2001.
\newblock Emotion and motivation ii: sex differences in picture processing.
\newblock \emph{Emotion}, 1(3):300.

\bibitem[{Brody(1997)}]{brody1997gender}
Leslie~R Brody. 1997.
\newblock Gender and emotion: Beyond stereotypes.
\newblock \emph{Journal of Social issues}, 53(2):369--393.

\bibitem[{Brody and Hall(2010)}]{brody2010gender}
Leslie~R Brody and Judith~A Hall. 2010.
\newblock Gender, emotion, and socialization.
\newblock In \emph{Handbook of gender research in psychology}, pages 429--454.
  Springer.

\bibitem[{Cameron et~al.(2013)Cameron, Baker, and
  Peterson}]{cameron2013waiting}
Michaelle Cameron, Julie Baker, and Mark Peterson. 2013.
\newblock Waiting for service: The effects of music volume and gender.
\newblock \emph{Services Marketing Quarterly}, 34(4):257--273.

\bibitem[{Chen et~al.(2020)Chen, Huang, Hei, and Zeng}]{chen2020felt}
Xuqian Chen, Shengqiao Huang, Xueting Hei, and Hongyuan Zeng. 2020.
\newblock Felt emotion elicited by music: are sensitivities to various musical
  features different for young children and young adults?
\newblock \emph{The Spanish Journal of Psychology}, 23.

\bibitem[{Cohan et~al.(2018)Cohan, Desmet, Yates, Soldaini, MacAvaney, and
  Goharian}]{cohan2018smhd}
Arman Cohan, Bart Desmet, Andrew Yates, Luca Soldaini, Sean MacAvaney, and
  Nazli Goharian. 2018.
\newblock Smhd: a large-scale resource for exploring online language usage for
  multiple mental health conditions.
\newblock In \emph{27th International Conference on Computational Linguistics},
  pages 1485--1497. ACL.

\bibitem[{Cooney et~al.(2013)Cooney, Dwan, Greig, Lawlor, Rimer, Waugh,
  McMurdo, and Mead}]{cooney2013exercise}
Gary~M Cooney, Kerry Dwan, Carolyn~A Greig, Debbie~A Lawlor, Jane Rimer,
  Fiona~R Waugh, Marion McMurdo, and Gillian~E Mead. 2013.
\newblock Exercise for depression.
\newblock \emph{Cochrane database of systematic reviews}, (9).

\bibitem[{Coppersmith et~al.(2014)Coppersmith, Dredze, and
  Harman}]{coppersmith2014quantifying}
Glen Coppersmith, Mark Dredze, and Craig Harman. 2014.
\newblock Quantifying mental health signals in twitter.
\newblock In \emph{Proceedings of the workshop on computational linguistics and
  clinical psychology: From linguistic signal to clinical reality}, pages
  51--60.

\bibitem[{Coppersmith et~al.(2015)Coppersmith, Dredze, Harman, and
  Hollingshead}]{coppersmith2015adhd}
Glen Coppersmith, Mark Dredze, Craig Harman, and Kristy Hollingshead. 2015.
\newblock From adhd to sad: Analyzing the language of mental health on twitter
  through self-reported diagnoses.
\newblock In \emph{Proceedings of the 2nd workshop on computational linguistics
  and clinical psychology: from linguistic signal to clinical reality}, pages
  1--10.

\bibitem[{Csikszentmihalyi and LeFevre(1989)}]{csikszentmihalyi1989optimal}
Mihaly Csikszentmihalyi and Judith LeFevre. 1989.
\newblock Optimal experience in work and leisure.
\newblock \emph{Journal of personality and social psychology}, 56(5):815.

\bibitem[{Cui et~al.(2022)Cui, Zhang, Pang, Jaidka, Sherman, Jakhetiya, Ungar,
  and Guntuku}]{cui2022social}
Jesse Cui, Tingdan Zhang, Dandan Pang, Kokil Jaidka, Garrick Sherman, Vinit
  Jakhetiya, Lyle Ungar, and Sharath~Chandra Guntuku. 2022.
\newblock Social media reveals urban-rural differences in stress across china.
\newblock \emph{ICWSM}.

\bibitem[{de~Boise(2016)}]{de2016contesting}
Sam de~Boise. 2016.
\newblock Contesting ‘sex’and ‘gender’difference in emotions through
  music use in the uk.
\newblock \emph{Journal of Gender Studies}, 25(1):66--84.

\bibitem[{De~Choudhury and De(2014)}]{de2014mental}
Munmun De~Choudhury and Sushovan De. 2014.
\newblock Mental health discourse on reddit: Self-disclosure, social support,
  and anonymity.
\newblock In \emph{Eighth international AAAI conference on weblogs and social
  media}.

\bibitem[{De~Choudhury and Kiciman(2017)}]{de2017language}
Munmun De~Choudhury and Emre Kiciman. 2017.
\newblock The language of social support in social media and its effect on
  suicidal ideation risk.
\newblock In \emph{Proceedings of the International AAAI Conference on Web and
  Social Media}, volume~11, pages 32--41.

\bibitem[{DeNora(1999)}]{denora1999music}
Tia DeNora. 1999.
\newblock Music as a technology of the self.
\newblock \emph{Poetics}, 27(1):31--56.

\bibitem[{DeNora(2000)}]{denora2000music}
Tia DeNora. 2000.
\newblock \emph{Music in everyday life}.
\newblock Cambridge University Press.

\bibitem[{Dredge(2022)}]{dredge_2022}
Stuart Dredge. 2022.
\newblock Netease cloud music reveals its revenues grew 43\% in 2021.

\bibitem[{Eerola et~al.(2012)Eerola, Ferrer, and Alluri}]{eerola2012timbre}
Tuomas Eerola, Rafael Ferrer, and Vinoo Alluri. 2012.
\newblock Timbre and affect dimensions: Evidence from affect and similarity
  ratings and acoustic correlates of isolated instrument sounds.
\newblock \emph{Music Perception: An Interdisciplinary Journal}, 30(1):49--70.

\bibitem[{Eerola et~al.(2013)Eerola, Friberg, and Bresin}]{eerola2013emotional}
Tuomas Eerola, Anders Friberg, and Roberto Bresin. 2013.
\newblock Emotional expression in music: contribution, linearity, and
  additivity of primary musical cues.
\newblock \emph{Frontiers in psychology}, 4:487.

\bibitem[{Ellis(2007)}]{ellis2007beat}
Daniel~PW Ellis. 2007.
\newblock Beat tracking by dynamic programming.
\newblock \emph{Journal of New Music Research}, 36(1):51--60.

\bibitem[{Ernala et~al.(2018)Ernala, Labetoulle, Bane, Birnbaum, Rizvi, Kane,
  and De~Choudhury}]{ernala2018characterizing}
Sindhu~Kiranmai Ernala, Tristan Labetoulle, Fred Bane, Michael~L Birnbaum,
  Asra~F Rizvi, John~M Kane, and Munmun De~Choudhury. 2018.
\newblock Characterizing audience engagement and assessing its impact on social
  media disclosures of mental illnesses.
\newblock In \emph{Twelfth international AAAI conference on web and social
  media}.

\bibitem[{Feder et~al.(2021)Feder, Keith, Manzoor, Pryzant, Sridhar,
  Wood-Doughty, Eisenstein, Grimmer, Reichart, Roberts
  et~al.}]{feder2021causal}
Amir Feder, Katherine~A Keith, Emaad Manzoor, Reid Pryzant, Dhanya Sridhar,
  Zach Wood-Doughty, Jacob Eisenstein, Justin Grimmer, Roi Reichart, Margaret~E
  Roberts, et~al. 2021.
\newblock Causal inference in natural language processing: Estimation,
  prediction, interpretation and beyond.
\newblock \emph{arXiv preprint arXiv:2109.00725}.

\bibitem[{Fern{\'a}ndez et~al.(2012)Fern{\'a}ndez, Pascual, Soler, Elices,
  Portella, and Fern{\'a}ndez-Abascal}]{fernandez2012physiological}
Cristina Fern{\'a}ndez, Juan~C Pascual, Joaquim Soler, Matilde Elices, Maria~J
  Portella, and Enrique Fern{\'a}ndez-Abascal. 2012.
\newblock Physiological responses induced by emotion-eliciting films.
\newblock \emph{Applied psychophysiology and biofeedback}, 37(2):73--79.

\bibitem[{Font et~al.(2016)Font, Brookes, Fazekas, Guerber, La~Burthe, Plans,
  Plumbley, Shaashua, Wang, and Serra}]{font2016audio}
Frederic Font, Tim Brookes, George Fazekas, Martin Guerber, Amaury La~Burthe,
  David Plans, Mark~d. Plumbley, Meir Shaashua, Wenwu Wang, and Xavier Serra.
  2016.
\newblock audio commons: bringing creative commons audio content to the
  creative industries.
\newblock \emph{journal of the audio engineering society}.

\bibitem[{Gabrielsson and Lindstr{\"o}m(2010)}]{gabrielsson2010strong}
Alf Gabrielsson and Siv Lindstr{\"o}m. 2010.
\newblock Strong experiences with music.
\newblock \emph{Handbook of music and emotion: Theory, research, applications},
  pages 547--574.

\bibitem[{Garrido et~al.(2018)Garrido, Stevens, Chang, Dunne, and
  Perz}]{garrido2018music}
Sandra Garrido, Catherine~J Stevens, Esther Chang, Laura Dunne, and Janette
  Perz. 2018.
\newblock Music and dementia: individual differences in response to
  personalized playlists.
\newblock \emph{Journal of Alzheimer's Disease}, 64(3):933--941.

\bibitem[{George et~al.(1989)George, Blazer, Hughes, and
  Fowler}]{george1989social}
Linda~K George, Dan~G Blazer, Dana~C Hughes, and Nancy Fowler. 1989.
\newblock Social support and the outcome of major depression.
\newblock \emph{The British Journal of Psychiatry}, 154(4):478--485.

\bibitem[{Golder and Macy(2011)}]{golder2011diurnal}
Scott~A Golder and Michael~W Macy. 2011.
\newblock Diurnal and seasonal mood vary with work, sleep, and daylength across
  diverse cultures.
\newblock \emph{Science}, 333(6051):1878--1881.

\bibitem[{Gomez and Danuser(2007)}]{gomez2007relationships}
Patrick Gomez and Brigitta Danuser. 2007.
\newblock Relationships between musical structure and psychophysiological
  measures of emotion.
\newblock \emph{Emotion}, 7(2):377.

\bibitem[{G{\'o}mez-Ca{\~n}{\'o}n et~al.(2021)G{\'o}mez-Ca{\~n}{\'o}n, Cano,
  Eerola, Herrera, Hu, Yang, and G{\'o}mez}]{gomez2021music}
Juan~Sebasti{\'a}n G{\'o}mez-Ca{\~n}{\'o}n, Estefan{\'\i}a Cano, Tuomas Eerola,
  Perfecto Herrera, Xiao Hu, Yi-Hsuan Yang, and Emilia G{\'o}mez. 2021.
\newblock Music emotion recognition: Toward new, robust standards in
  personalized and context-sensitive applications.
\newblock \emph{IEEE Signal Processing Magazine}, 38(6):106--114.

\bibitem[{Greasley and Lamont(2011)}]{greasley2011exploring}
Alinka~E Greasley and Alexandra Lamont. 2011.
\newblock Exploring engagement with music in everyday life using experience
  sampling methodology.
\newblock \emph{Musicae Scientiae}, 15(1):45--71.

\bibitem[{Gregory and Varney(1996)}]{gregory1996cross}
Andrew~H Gregory and Nicholas Varney. 1996.
\newblock Cross-cultural comparisons in the affective response to music.
\newblock \emph{Psychology of Music}, 24(1):47--52.

\bibitem[{Greitemeyer(2009)}]{greitemeyer2009effects}
Tobias Greitemeyer. 2009.
\newblock Effects of songs with prosocial lyrics on prosocial behavior: Further
  evidence and a mediating mechanism.
\newblock \emph{Personality and Social Psychology Bulletin}, 35(11):1500--1511.

\bibitem[{Guti{\'e}rrez~P{\'a}ez et~al.(2021)Guti{\'e}rrez~P{\'a}ez,
  G{\'o}mez-Ca{\~n}{\'o}n, Porcaro, Santos, Hern{\'a}ndez-Leo, and
  G{\'o}mez}]{gutierrez2021emotion}
Nicol{\'a}s~Felipe Guti{\'e}rrez~P{\'a}ez, Juan~Sebasti{\'a}n
  G{\'o}mez-Ca{\~n}{\'o}n, Lorenzo Porcaro, Patricia Santos, Davinia
  Hern{\'a}ndez-Leo, and Emilia G{\'o}mez. 2021.
\newblock Emotion annotation of music: A citizen science approach.
\newblock In \emph{International Conference on Collaboration Technologies and
  Social Computing}, pages 51--66. Springer.

\bibitem[{Hailstone et~al.(2009)Hailstone, Omar, Henley, Frost, Kenward, and
  Warren}]{hailstone2009s}
Julia~C Hailstone, Rohani Omar, Susie~MD Henley, Chris Frost, Michael~G
  Kenward, and Jason~D Warren. 2009.
\newblock It's not what you play, it's how you play it: Timbre affects
  perception of emotion in music.
\newblock \emph{Quarterly journal of experimental psychology},
  62(11):2141--2155.

\bibitem[{Harrigian et~al.(2020)Harrigian, Aguirre, and
  Dredze}]{harrigian2020state}
Keith Harrigian, Carlos Aguirre, and Mark Dredze. 2020.
\newblock On the state of social media data for mental health research.
\newblock \emph{arXiv preprint arXiv:2011.05233}.

\bibitem[{Harvey(2008)}]{harvey2008sleep}
Allison~G Harvey. 2008.
\newblock Sleep and circadian rhythms in bipolar disorder: seeking synchrony,
  harmony, and regulation.
\newblock \emph{American journal of psychiatry}, 165(7):820--829.

\bibitem[{Hatano and Hashimoto(2000)}]{hatano2000booming}
Shigeko Hatano and Takeo Hashimoto. 2000.
\newblock Booming index as a measure for evaluating booming sensation.
\newblock In \emph{Proc. Inter-Noise}, 233, pages 1--6.

\bibitem[{Hevner(1935)}]{hevner1935expression}
Kate Hevner. 1935.
\newblock Expression in music: a discussion of experimental studies and
  theories.
\newblock \emph{Psychological review}, 42(2):186.

\bibitem[{High and Dillard(2012)}]{high2012review}
Andrew~C High and James~Price Dillard. 2012.
\newblock A review and meta-analysis of person-centered messages and social
  support outcomes.
\newblock \emph{Communication Studies}, 63(1):99--118.

\bibitem[{Hirano et~al.(2006)Hirano, Fujita, Watanabe, Niwa, Takahashi,
  Kanematsu, Ido, Tomida, and Onozuka}]{hirano2006effect}
Yoshiyuki Hirano, Masafumi Fujita, Kazuko Watanabe, Masami Niwa, Toru
  Takahashi, Masayuki Kanematsu, Yasushi Ido, Mihoko Tomida, and Minoru
  Onozuka. 2006.
\newblock Effect of unpleasant loud noise on hippocampal activities during
  picture encoding: an fmri study.
\newblock \emph{Brain and cognition}, 61(3):280--285.

\bibitem[{Ho et~al.(2018)Ho, Hancock, and Miner}]{ho2018psychological}
Annabell Ho, Jeff Hancock, and Adam~S Miner. 2018.
\newblock Psychological, relational, and emotional effects of self-disclosure
  after conversations with a chatbot.
\newblock \emph{Journal of Communication}, 68(4):712--733.

\bibitem[{Hu et~al.(2018)Hu, Li, and Ng}]{hu2018relationships}
Xiao Hu, Fanjie Li, and Tzi-Dong~Jeremy Ng. 2018.
\newblock On the relationships between music-induced emotion and physiological
  signals.
\newblock In \emph{ISMIR}, pages 362--369.

\bibitem[{Huang et~al.(2012)Huang, Chung, Hui, Lin, Seih, Lam, Chen, Bond, and
  Pennebaker}]{huang2012development}
Chin-Lan Huang, Cindy~K Chung, Natalie Hui, Yi-Cheng Lin, Yi-Tai Seih, Ben~CP
  Lam, Wei-Chuan Chen, Michael~H Bond, and James~W Pennebaker. 2012.
\newblock The development of the chinese linguistic inquiry and word count
  dictionary.
\newblock \emph{Chinese Journal of Psychology}.

\bibitem[{Huang et~al.(2019)Huang, Wang, Wang, Liu, Yu, Yan, Yu, Kou, Xu, Lu
  et~al.}]{huang2019prevalence}
Yueqin Huang, YU~Wang, Hong Wang, Zhaorui Liu, Xin Yu, Jie Yan, Yaqin Yu,
  Changgui Kou, Xiufeng Xu, Jin Lu, et~al. 2019.
\newblock Prevalence of mental disorders in china: a cross-sectional
  epidemiological study.
\newblock \emph{The Lancet Psychiatry}, 6(3):211--224.

\bibitem[{Hunter and Schellenberg(2010)}]{hunter2010music}
Patrick~G Hunter and E~Glenn Schellenberg. 2010.
\newblock Music and emotion.
\newblock In \emph{Music perception}, pages 129--164. Springer.

\bibitem[{Imbens(2004)}]{imbens2004nonparametric}
Guido~W Imbens. 2004.
\newblock Nonparametric estimation of average treatment effects under
  exogeneity: A review.
\newblock \emph{Review of Economics and statistics}, 86(1):4--29.

\bibitem[{Jan and Wang(2012)}]{jan2012blind}
Tariqullah Jan and Wenwu Wang. 2012.
\newblock Blind reverberation time estimation based on laplace distribution.
\newblock In \emph{2012 Proceedings of the 20th European Signal Processing
  Conference (EUSIPCO)}, pages 2050--2054. IEEE.

\bibitem[{Jensen(2001)}]{jensen2001effects}
Kaja~L Jensen. 2001.
\newblock The effects of selected classical music on self-disclosure.
\newblock \emph{Journal of music therapy}, 38(1):2--27.

\bibitem[{Juslin(2013)}]{JUSLIN2013235}
Patrik~N. Juslin. 2013.
\newblock From everyday emotions to aesthetic emotions: Towards a unified
  theory of musical emotions.
\newblock \emph{Physics of Life Reviews}, 10(3):235--266.

\bibitem[{Juslin(2019)}]{juslin2019musical}
Patrik~N Juslin. 2019.
\newblock \emph{Musical emotions explained: Unlocking the secrets of musical
  affect}.
\newblock Oxford University Press, USA.

\bibitem[{Juslin and Laukka(2004)}]{juslin2004expression}
Patrik~N Juslin and Petri Laukka. 2004.
\newblock Expression, perception, and induction of musical emotions: A review
  and a questionnaire study of everyday listening.
\newblock \emph{Journal of new music research}, 33(3):217--238.

\bibitem[{Juslin et~al.(2008)Juslin, Liljestr{\"o}m, V{\"a}stfj{\"a}ll,
  Barradas, and Silva}]{juslin2008experience}
Patrik~N Juslin, Simon Liljestr{\"o}m, Daniel V{\"a}stfj{\"a}ll, Gon{\c{c}}alo
  Barradas, and Ana Silva. 2008.
\newblock An experience sampling study of emotional reactions to music:
  listener, music, and situation.
\newblock \emph{Emotion}, 8(5):668.

\bibitem[{Juslin and V{\"a}stfj{\"a}ll(2008)}]{juslin2008emotional}
Patrik~N Juslin and Daniel V{\"a}stfj{\"a}ll. 2008.
\newblock Emotional responses to music: The need to consider underlying
  mechanisms.
\newblock \emph{Behavioral and brain sciences}, 31(5):559--575.

\bibitem[{Kamenetsky et~al.(1997)Kamenetsky, Hill, and
  Trehub}]{kamenetsky1997effect}
Stuart~B Kamenetsky, David~S Hill, and Sandra~E Trehub. 1997.
\newblock Effect of tempo and dynamics on the perception of emotion in music.
\newblock \emph{Psychology of Music}, 25(2):149--160.

\bibitem[{Kennedy(2013)}]{kennedy2013relationship}
Patrick Kennedy. 2013.
\newblock The relationship between prosocial music and helping behaviour and
  its mediators: An irish college sample.
\newblock \emph{Journal of European Psychology Students}, 4(1).

\bibitem[{Koelsch(2014)}]{koelsch2014brain}
Stefan Koelsch. 2014.
\newblock Brain correlates of music-evoked emotions.
\newblock \emph{Nature Reviews Neuroscience}, 15(3):170--180.

\bibitem[{Konstan and Riedl(2012)}]{konstan2012recommender}
Joseph~A Konstan and John Riedl. 2012.
\newblock Recommender systems: from algorithms to user experience.
\newblock \emph{User modeling and user-adapted interaction}, 22(1):101--123.

\bibitem[{Krugman(1943)}]{krugman1943affective}
Herbert~E Krugman. 1943.
\newblock Affective response to music as a function of familiarity.
\newblock \emph{The Journal of Abnormal and Social Psychology}, 38(3):388.

\bibitem[{Krumhansl(2001)}]{krumhansl2001cognitive}
Carol~L Krumhansl. 2001.
\newblock \emph{Cognitive foundations of musical pitch}, volume~17.
\newblock Oxford University Press.

\bibitem[{Larson(2017)}]{larson-2017-gender}
Brian Larson. 2017.
\newblock Gender as a variable in natural-language processing: Ethical
  considerations.
\newblock In \emph{Proceedings of the First {ACL} Workshop on Ethics in Natural
  Language Processing}, pages 1--11, Valencia, Spain. Association for
  Computational Linguistics.

\bibitem[{Liu et~al.(2018)Liu, Liu, Wei, Li, Yuan, Wu, Wang, and
  Zhao}]{liu2018effects}
Ying Liu, Guangyuan Liu, Dongtao Wei, Qiang Li, Guangjie Yuan, Shifu Wu,
  Gaoyuan Wang, and Xingcong Zhao. 2018.
\newblock Effects of musical tempo on musicians’ and non-musicians’
  emotional experience when listening to music.
\newblock \emph{Frontiers in Psychology}, page 2118.

\bibitem[{Lunceford and Davidian(2004)}]{lunceford2004stratification}
Jared~K Lunceford and Marie Davidian. 2004.
\newblock Stratification and weighting via the propensity score in estimation
  of causal treatment effects: a comparative study.
\newblock \emph{Statistics in medicine}, 23(19):2937--2960.

\bibitem[{Lundqvist et~al.(2009)Lundqvist, Carlsson, Hilmersson, and
  Juslin}]{lundqvist2009emotional}
Lars-Olov Lundqvist, Fredrik Carlsson, Per Hilmersson, and Patrik~N Juslin.
  2009.
\newblock Emotional responses to music: Experience, expression, and physiology.
\newblock \emph{Psychology of music}, 37(1):61--90.

\bibitem[{Mauss and Robinson(2009)}]{mauss2009measures}
Iris~B Mauss and Michael~D Robinson. 2009.
\newblock Measures of emotion: A review.
\newblock \emph{Cognition and emotion}, 23(2):209--237.

\bibitem[{McCraty et~al.(1998)McCraty, Barrios-Choplin, Atkinson, and
  Tomasino}]{mccraty1998effects}
Rollin McCraty, Bob Barrios-Choplin, Michael Atkinson, and Dana Tomasino. 1998.
\newblock The effects of different types of music on mood, tension, and mental
  clarity.
\newblock \emph{Alternative therapies in health and medicine}, 4(1):75--84.

\bibitem[{McFee et~al.(2015)McFee, Raffel, Liang, Ellis, McVicar, Battenberg,
  and Nieto}]{mcfee2015librosa}
Brian McFee, Colin Raffel, Dawen Liang, Daniel~P Ellis, Matt McVicar, Eric
  Battenberg, and Oriol Nieto. 2015.
\newblock librosa: Audio and music signal analysis in python.
\newblock In \emph{Proceedings of the 14th python in science conference},
  volume~8, pages 18--25. Citeseer.

\bibitem[{Meyer(1956)}]{meyer1956emotion}
Leonard~B Meyer. 1956.
\newblock \emph{Emotion and meaning in music}.
\newblock University of chicago Press.

\bibitem[{Mihalcea and Strapparava(2012)}]{mihalcea2012lyrics}
Rada Mihalcea and Carlo Strapparava. 2012.
\newblock Lyrics, music, and emotions.
\newblock In \emph{Proceedings of the 2012 Joint Conference on Empirical
  Methods in Natural Language Processing and Computational Natural Language
  Learning}, pages 590--599.

\bibitem[{Miller(2015)}]{miller2015resonance}
Vincent Miller. 2015.
\newblock Resonance as a social phenomenon.
\newblock \emph{Sociological Research Online}, 20(2):58--70.

\bibitem[{Mohammad et~al.(2014)Mohammad, Zhu, and
  Martin}]{mohammad-etal-2014-semantic}
Saif Mohammad, Xiaodan Zhu, and Joel Martin. 2014.
\newblock Semantic role labeling of emotions in tweets.
\newblock In \emph{Proceedings of the 5th Workshop on Computational Approaches
  to Subjectivity, Sentiment and Social Media Analysis}, pages 32--41,
  Baltimore, Maryland. Association for Computational Linguistics.

\bibitem[{Mohammad(2020)}]{mohammad2020practical}
Saif~M Mohammad. 2020.
\newblock Practical and ethical considerations in the effective use of emotion
  and sentiment lexicons.
\newblock \emph{arXiv preprint arXiv:2011.03492}.

\bibitem[{{National Bureau of Statistics of China}(2021)}]{national2021main}
{National Bureau of Statistics of China}. 2021.
\newblock Main data of the seventh national population census.

\bibitem[{Paltemaa et~al.(2020)Paltemaa, Vuori, Mattlin, and
  Katajisto}]{paltemaa2020meta}
Lauri Paltemaa, Juha~A Vuori, Mikael Mattlin, and Jouko Katajisto. 2020.
\newblock Meta-information censorship and the creation of the chinanet bubble.
\newblock \emph{Information, Communication \& Society}, 23(14):2064--2080.

\bibitem[{Paul(2017)}]{paul2017feature}
Michael Paul. 2017.
\newblock Feature selection as causal inference: Experiments with text
  classification.
\newblock In \emph{Proceedings of the 21st Conference on Computational Natural
  Language Learning (CoNLL 2017)}, pages 163--172.

\bibitem[{Pearce et~al.(2017)Pearce, Brookes, and Mason}]{pearce5first}
Andy Pearce, Tim Brookes, and Russell Mason. 2017.
\newblock First prototype of timbral characterisation tool for semantically
  annotating non-musical.
\newblock \emph{Audio Commons project deliverable D}, 5.

\bibitem[{Pearce et~al.(2019)Pearce, Brookes, and Mason}]{pearce2019modelling}
Andy Pearce, Tim Brookes, and Russell Mason. 2019.
\newblock Modelling timbral hardness.
\newblock \emph{Applied Sciences}, 9(3):466.

\bibitem[{Pennebaker(2011)}]{pennebaker2011secret}
James~W Pennebaker. 2011.
\newblock The secret life of pronouns.
\newblock \emph{New Scientist}, 211(2828):42--45.

\bibitem[{Pennebaker et~al.(2015)Pennebaker, Boyd, Jordan, and
  Blackburn}]{pennebaker2015development}
James~W Pennebaker, Ryan~L Boyd, Kayla Jordan, and Kate Blackburn. 2015.
\newblock The development and psychometric properties of liwc2015.
\newblock Technical report.

\bibitem[{Rawls(2001)}]{rawls2001justice}
John Rawls. 2001.
\newblock \emph{Justice as fairness: A restatement}.
\newblock Harvard University Press.

\bibitem[{Robazza et~al.(1994)Robazza, Macaluso, and
  D'Urso}]{robazza1994emotional}
Claudio Robazza, Cristina Macaluso, and Valentina D'Urso. 1994.
\newblock Emotional reactions to music by gender, age, and expertise.
\newblock \emph{Perceptual and Motor skills}, 79(2):939--944.

\bibitem[{Robert et~al.(2018)Robert, Webbie et~al.}]{robert2018pydub}
James Robert, Marc Webbie, et~al. 2018.
\newblock \href {http://pydub.com/} {Pydub}.

\bibitem[{Rosa(2019)}]{rosa2019resonance}
Hartmut Rosa. 2019.
\newblock \emph{Resonance: A sociology of our relationship to the world}.
\newblock John Wiley \& Sons.

\bibitem[{Rosenbaum and Rubin(1983)}]{rosenbaum1983central}
Paul~R Rosenbaum and Donald~B Rubin. 1983.
\newblock The central role of the propensity score in observational studies for
  causal effects.
\newblock \emph{Biometrika}, 70(1):41--55.

\bibitem[{Rosenbaum and Rubin(1984)}]{rosenbaum1984reducing}
Paul~R Rosenbaum and Donald~B Rubin. 1984.
\newblock Reducing bias in observational studies using subclassification on the
  propensity score.
\newblock \emph{Journal of the American statistical Association},
  79(387):516--524.

\bibitem[{Russell(1980)}]{russell1980circumplex}
James~A Russell. 1980.
\newblock A circumplex model of affect.
\newblock \emph{Journal of personality and social psychology}, 39(6):1161.

\bibitem[{Ruth(2017)}]{ruth2017heal}
Nicolas Ruth. 2017.
\newblock “heal the world”: A field experiment on the effects of music with
  prosocial lyrics on prosocial behavior.
\newblock \emph{Psychology of Music}, 45(2):298--304.

\bibitem[{Saarikallio et~al.(2013)Saarikallio, Nieminen, and
  Brattico}]{saarikallio2013affective}
Suvi Saarikallio, Sirke Nieminen, and Elvira Brattico. 2013.
\newblock Affective reactions to musical stimuli reflect emotional use of music
  in everyday life.
\newblock \emph{Musicae Scientiae}, 17(1):27--39.

\bibitem[{Sch{\"a}fer et~al.(2020)Sch{\"a}fer, Saarikallio, and
  Eerola}]{schafer2020music}
Katharina Sch{\"a}fer, Suvi Saarikallio, and Tuomas Eerola. 2020.
\newblock Music may reduce loneliness and act as social surrogate for a friend:
  evidence from an experimental listening study.
\newblock \emph{Music \& Science}, 3:2059204320935709.

\bibitem[{Schriewer and Bulaj(2016)}]{schriewer2016music}
Karl Schriewer and Grzegorz Bulaj. 2016.
\newblock Music streaming services as adjunct therapies for depression,
  anxiety, and bipolar symptoms: convergence of digital technologies, mobile
  apps, emotions, and global mental health.
\newblock \emph{Frontiers in public health}, 4:217.

\bibitem[{Schubert(2004)}]{schubert2004modeling}
Emery Schubert. 2004.
\newblock Modeling perceived emotion with continuous musical features.
\newblock \emph{Music perception}, 21(4):561--585.

\bibitem[{Siles et~al.(2019)Siles, Segura-Castillo, Sancho, and
  Sol{\'\i}s-Quesada}]{siles2019genres}
Ignacio Siles, Andr{\'e}s Segura-Castillo, M{\'o}nica Sancho, and Ricardo
  Sol{\'\i}s-Quesada. 2019.
\newblock Genres as social affect: Cultivating moods and emotions through
  playlists on spotify.
\newblock \emph{Social Media+ Society}, 5(2):2056305119847514.

\bibitem[{Singh et~al.(2020)Singh, Halpern, Thain, Christakopoulou, Chi, Chen,
  and Beutel}]{singh2020building}
Ashudeep Singh, Yoni Halpern, Nithum Thain, Konstantina Christakopoulou, E~Chi,
  Jilin Chen, and Alex Beutel. 2020.
\newblock Building healthy recommendation sequences for everyone: A safe
  reinforcement learning approach.
\newblock In \emph{Proceedings of the FAccTRec Workshop, Online}, pages 26--27.

\bibitem[{Sloboda et~al.(2001)Sloboda, O'Neill, and
  Ivaldi}]{sloboda2001functions}
John~A Sloboda, Susan~A O'Neill, and Antonia Ivaldi. 2001.
\newblock Functions of music in everyday life: An exploratory study using the
  experience sampling method.
\newblock \emph{Musicae scientiae}, 5(1):9--32.

\bibitem[{Sloboda and O’neill(2001)}]{sloboda2001emotions}
John~A Sloboda and Susan~A O’neill. 2001.
\newblock Emotions in everyday listening to music.
\newblock \emph{Music and emotion: Theory and research}, 8:415--429.

\bibitem[{Stewart et~al.(2019)Stewart, Garrido, Hense, and
  McFerran}]{stewart2019music}
Joanna Stewart, Sandra Garrido, Cherry Hense, and Katrina McFerran. 2019.
\newblock Music use for mood regulation: Self-awareness and conscious listening
  choices in young people with tendencies to depression.
\newblock \emph{Frontiers in psychology}, 10:1199.

\bibitem[{Taylor et~al.(2005)Taylor, Lichstein, Durrence, Reidel, and
  Bush}]{taylor2005epidemiology}
Daniel~J Taylor, Kenneth~L Lichstein, H~Heith Durrence, Brant~W Reidel, and
  Andrew~J Bush. 2005.
\newblock Epidemiology of insomnia, depression, and anxiety.
\newblock \emph{Sleep}, 28(11):1457--1464.

\bibitem[{Turner et~al.(1983)Turner, Frankel, and Levin}]{turner1983social}
R~Jay Turner, B~Gail Frankel, and Deborah~M Levin. 1983.
\newblock Social support: Conceptualization, measurement, and implications for
  mental health.
\newblock \emph{Research in community \& mental health}.

\bibitem[{Van~der Zwaag et~al.(2011)Van~der Zwaag, Westerink, and Van~den
  Broek}]{van2011emotional}
Marjolein~D Van~der Zwaag, Joyce~HDM Westerink, and Egon~L Van~den Broek. 2011.
\newblock Emotional and psychophysiological responses to tempo, mode, and
  percussiveness.
\newblock \emph{Musicae Scientiae}, 15(2):250--269.

\bibitem[{Vassilakis and Fitz(2007)}]{vassilakis2007sra}
Pantelis~N Vassilakis and K~Fitz. 2007.
\newblock Sra: A web-based research tool for spectral and roughness analysis of
  sound signals.
\newblock In \emph{Proceedings of the 4th Sound and Music Computing (SMC)
  Conference}, pages 319--325.

\bibitem[{V{\"a}stfj{\"a}ll et~al.(2012)V{\"a}stfj{\"a}ll, Juslin, and
  Hartig}]{vastfjall2012music}
Daniel V{\"a}stfj{\"a}ll, Patrik~N Juslin, and Terry Hartig. 2012.
\newblock Music, subjective wellbeing, and health: The role of everyday
  emotions.
\newblock \emph{Music, health, and wellbeing}, pages 405--423.

\bibitem[{Vuori and Paltemaa(2015)}]{vuori2015lexicon}
Juha~Antero Vuori and Lauri Paltemaa. 2015.
\newblock The lexicon of fear: Chinese internet control practice in sina weibo
  microblog censorship.
\newblock \emph{Surveillance \& society}, 13(3/4):400--421.

\bibitem[{Vuoskoski and Eerola(2015)}]{vuoskoski2015extramusical}
Jonna~K Vuoskoski and Tuomas Eerola. 2015.
\newblock Extramusical information contributes to emotions induced by music.
\newblock \emph{Psychology of Music}, 43(2):262--274.

\bibitem[{Wallmark et~al.(2018)Wallmark, Iacoboni, Deblieck, and
  Kendall}]{wallmark2018embodied}
Zachary Wallmark, Marco Iacoboni, Choi Deblieck, and Roger~A Kendall. 2018.
\newblock Embodied listening and timbre: Perceptual, acoustical, and neural
  correlates.
\newblock \emph{Music Perception: An Interdisciplinary Journal},
  35(3):332--363.

\bibitem[{Wang and Fu(2020)}]{wang2020exploring}
Han Wang and RongRong Fu. 2020.
\newblock Exploring user experience of music social mode-take netease cloud
  music as an example.
\newblock In \emph{International Conference on Applied Human Factors and
  Ergonomics}, pages 993--999. Springer.

\bibitem[{Wang(2012)}]{wang2012china}
Shaojung~Sharon Wang. 2012.
\newblock China’s internet lexicon: Symbolic meaning and commoditization of
  grass mud horse in the harmonious society.
\newblock \emph{First Monday}.

\bibitem[{Webster and Weir(2005)}]{webster2005emotional}
Gregory~D Webster and Catherine~G Weir. 2005.
\newblock Emotional responses to music: Interactive effects of mode, texture,
  and tempo.
\newblock \emph{Motivation and Emotion}, 29(1):19--39.

\bibitem[{Yang et~al.(2019)Yang, Yao, Seering, and Kraut}]{yang2019channel}
Diyi Yang, Zheng Yao, Joseph Seering, and Robert Kraut. 2019.
\newblock The channel matters: Self-disclosure, reciprocity and social support
  in online cancer support groups.
\newblock In \emph{Proceedings of the 2019 chi conference on human factors in
  computing systems}, pages 1--15.

\bibitem[{Yang et~al.(2007)Yang, Su, Lin, and Chen}]{yang2007music}
Yi-Hsuan Yang, Ya-Fan Su, Yu-Ching Lin, and Homer~H Chen. 2007.
\newblock Music emotion recognition: The role of individuality.
\newblock In \emph{Proceedings of the international workshop on Human-centered
  multimedia}, pages 13--22.

\bibitem[{Ye et~al.(2020)Ye, Long, Liu, and Xu}]{ye2020effect}
Yingying Ye, Tingting Long, Cuizhen Liu, and Dan Xu. 2020.
\newblock The effect of emotion on prosocial tendency: the moderating effect of
  epidemic severity under the outbreak of covid-19.
\newblock \emph{Frontiers in psychology}, 11:588701.

\bibitem[{Yu et~al.(2016)Yu, Lee, Hao, Wang, He, Hu, Lai, and
  Zhang}]{yu-etal-2016-building}
Liang-Chih Yu, Lung-Hao Lee, Shuai Hao, Jin Wang, Yunchao He, Jun Hu, K.~Robert
  Lai, and Xuejie Zhang. 2016.
\newblock Building {C}hinese affective resources in valence-arousal dimensions.
\newblock In \emph{Proceedings of the 2016 Conference of the North {A}merican
  Chapter of the Association for Computational Linguistics: Human Language
  Technologies}, pages 540--545, San Diego, California. Association for
  Computational Linguistics.

\bibitem[{Zhou et~al.(2018)Zhou, Xu, and Zhao}]{zhou2018homophily}
Zhenkun Zhou, Ke~Xu, and Jichang Zhao. 2018.
\newblock Homophily of music listening in online social networks of china.
\newblock \emph{Social Networks}, 55:160--169.

\bibitem[{Zwicker and Fastl(2013)}]{zwicker2013psychoacoustics}
Eberhard Zwicker and Hugo Fastl. 2013.
\newblock \emph{Psychoacoustics: Facts and models}, volume~22.
\newblock Springer Science \& Business Media.

\end{thebibliography}
\bibliographystyle{acl_natbib}

\newpage
\clearpage

\newpage

\appendix

\section*{Overview of Appendix}

We provide, as supplementary material, additional information about our dataset, annotation guidelines, preprocessing details, and expanded results across all experiments.

\section{Data}
\label{sec:datastats}

This section describes summary statistics of our data, as well as a view of the platform's user interface.

\subsection{Platform Interface}

\begin{figure*}[!t]
    \centering
    \begin{tabular}{ccc}
    {\includegraphics[height=6cm]{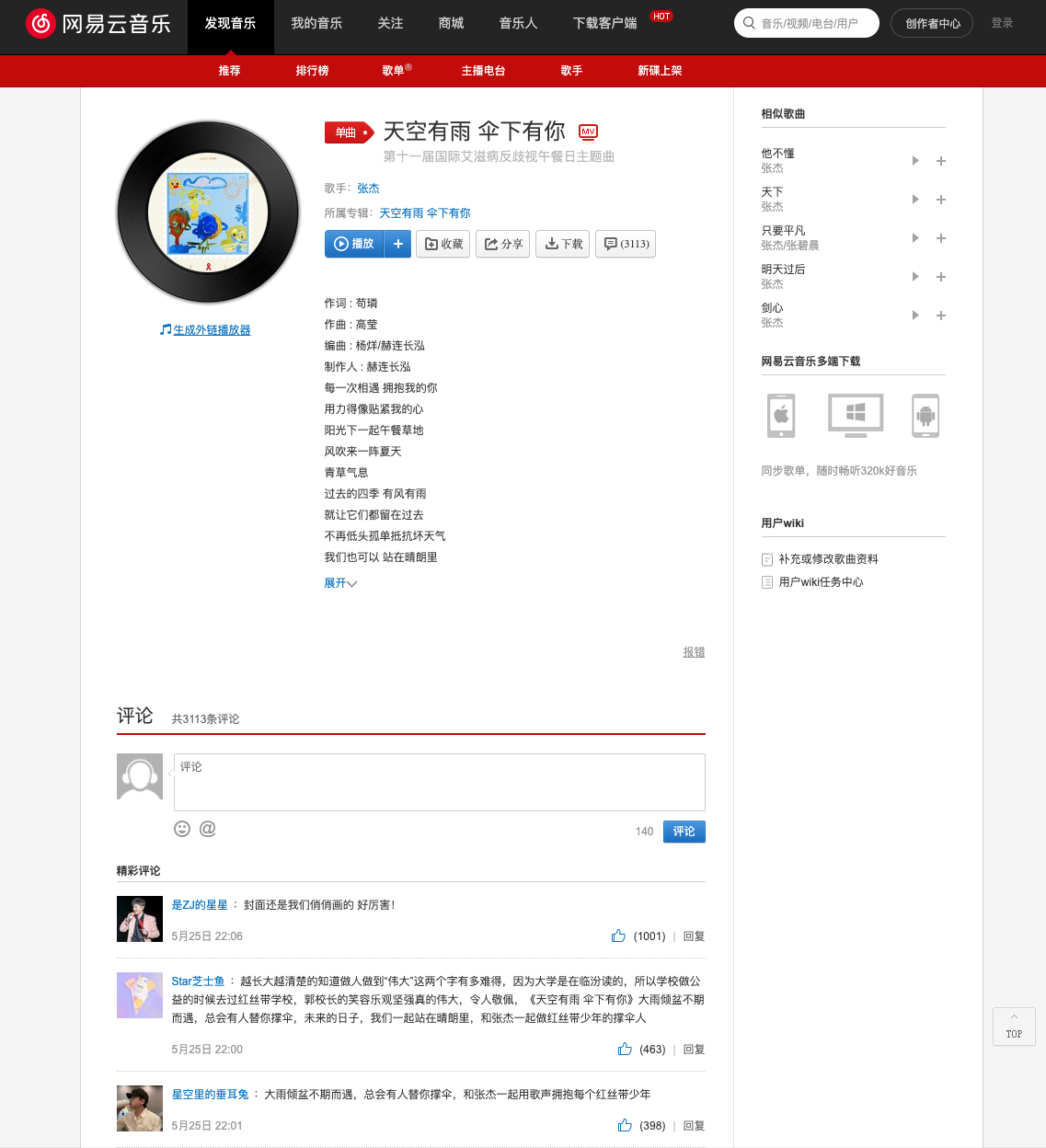}} &
    {\includegraphics[height=6cm]{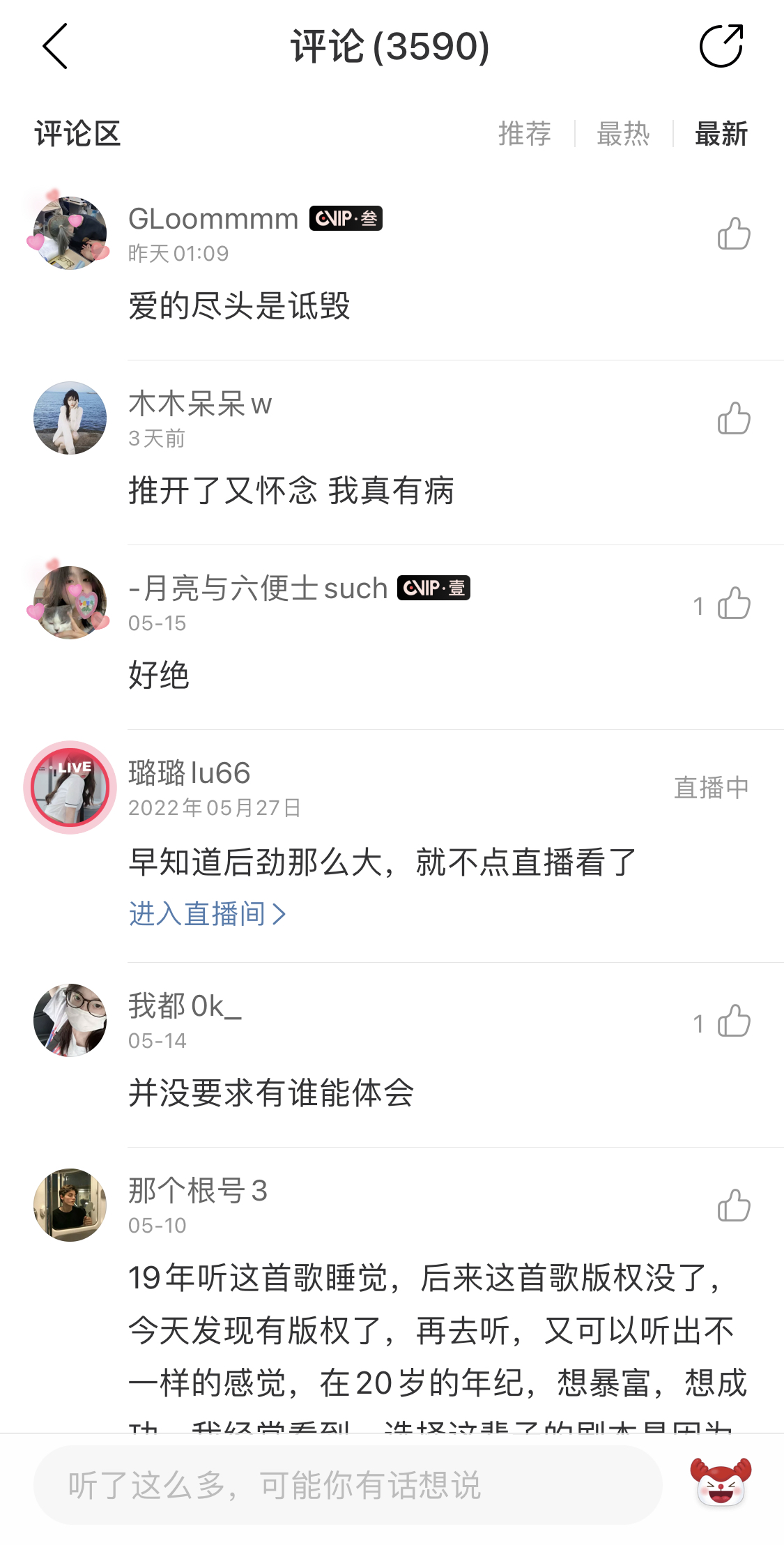}}
    \end{tabular}
    \caption{
    Screenshots of the platform's in-browser web page interface (left), showing the description, lyrics, and comment board of a song, and iOS in-app interface (right), showing the comment board of a song.
    }
    \label{fig:interfaces}
\end{figure*}

Users are able to interact with the platform through their browsers, native OS applications, and phone apps. Screenshots of a song's interface are shown in Figure \ref{fig:interfaces}, as is a view of the iOS application's commenting page for a song.

\subsection{Users}

\begin{figure*}[!t]
    \centering
    \begin{tabular}{ccc}
    {\includegraphics[width=0.40\textwidth]{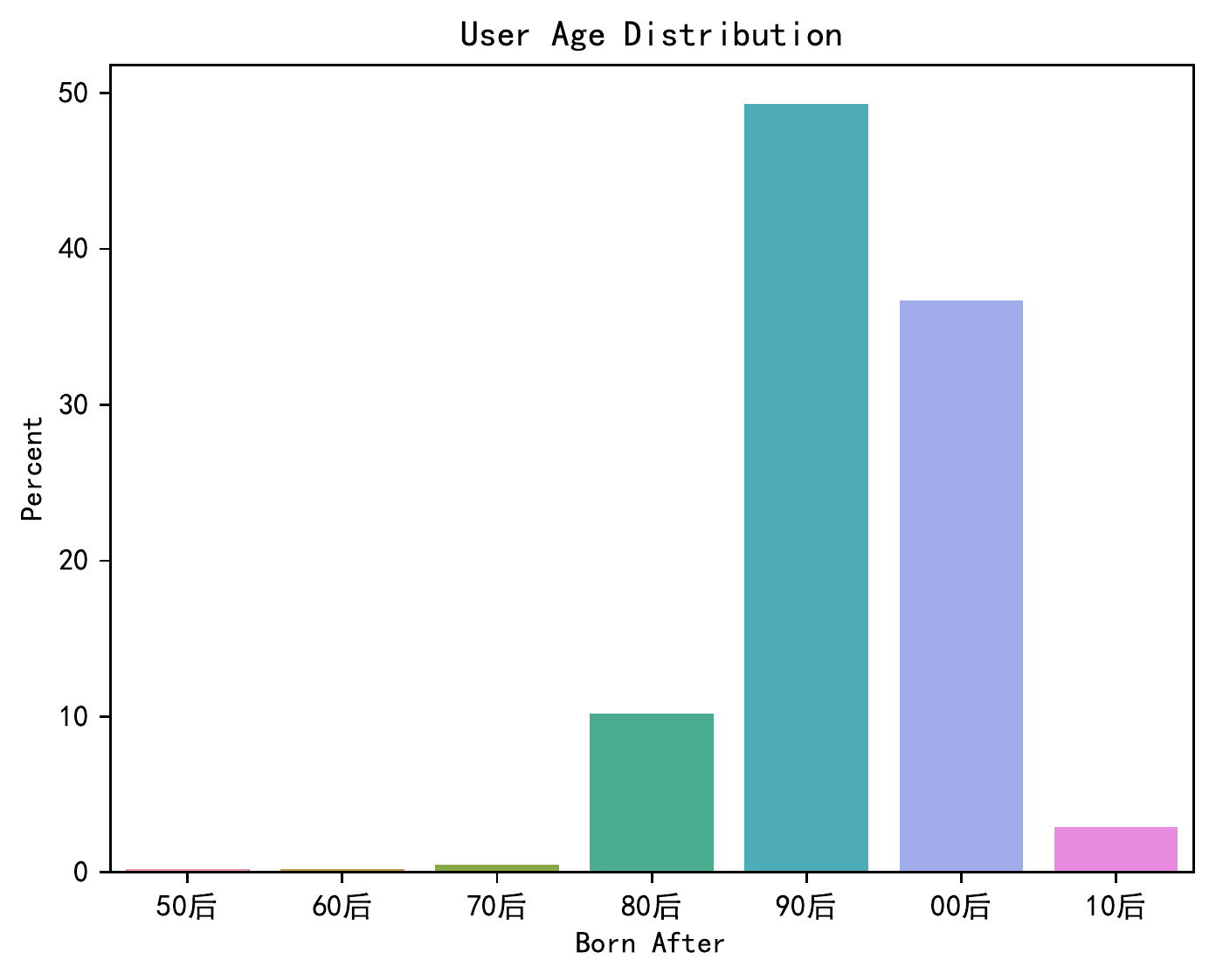}} &
    {\includegraphics[width=0.40\textwidth]{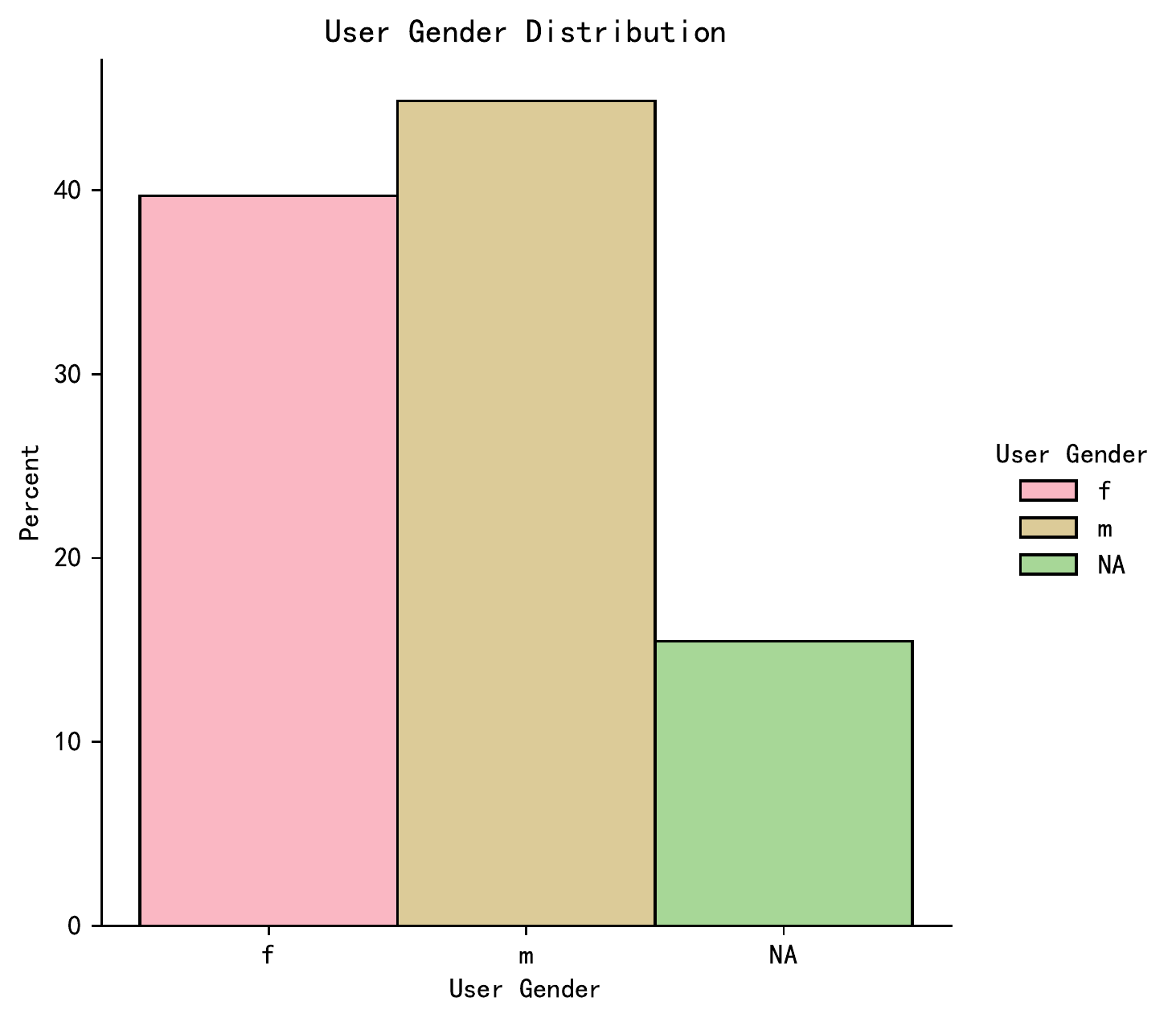}}
    \end{tabular}
    \caption{
    User age distributions (left) according to the decade of birth (i.e. 00\chinese{后} for those born in the 2000s), and user gender distributions (right) across platform-available choices for gender. Here, NA implies that the user omitted to input gender information during registration.
    }
    \label{fig:user-age-gender}
\end{figure*}

User age, gender, and region distributions (Figure \ref{fig:user-age-gender}) show that the majority of users are young men that hail from major metropolitan areas.
The top 20 regions that users hail from are, in descending order, Beijing (4.71\%), Guangzhou (4.22\%), Shanghai (3.80\%), Chengdu (3.47\%), Shenzhen (2.58\%), Chongqing (2.56\%), Nanjing (2.51\%), Wuhan (2.43\%), Hangzhou (2.21\%), Changsha (1.95\%), Xian, (1.91\%), Overseas-Other (1.77\%), Zhengzhou (1.54\%), Hefei (1.30\%), Tianjin (1.28\%), Suzhou (1.22\%), Kunming (1.20\%), Urumqi (1.19\%), Jinan (1.03\%), Fuzhou (0.99\%), Qingdao (0.91\%), Nanning (0.89\%), Nanchang (0.88\%), Shenyang (0.83\%), Harbin (0.83\%), Foshan (0.77\%), Dongguan (0.74\%), Guiyang (0.74\%), Shijiazhuang (0.73\%), and Ningbo (0.68\%).
General trends for user gender and region taken in the context of user ages mirror population trends indicated by the Chinese Census \cite{national2021main}.

\subsection{Songs}

Song comment and comment token distributions are shown in Figure \ref{fig:song-comments}; lyric preprocessing and topic modeling details are in Appendix Section \ref{sec:lyricpreprocessing}.

\subsection{Playlists}

\begin{figure*}[!t]
    \centering
    \begin{tabular}{ccc}
    {\includegraphics[width=0.4\textwidth]{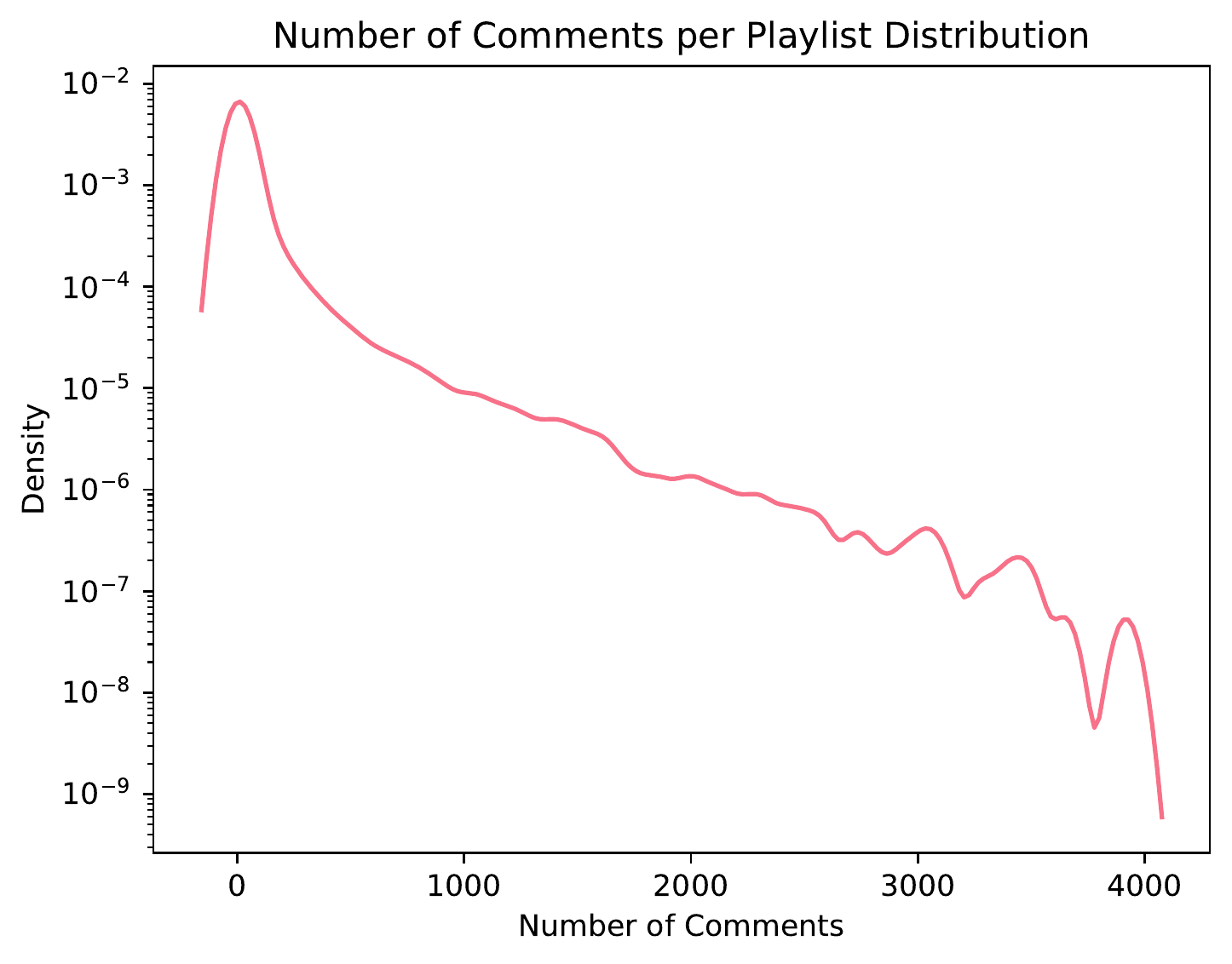}} &
    {\includegraphics[width=0.4\textwidth]{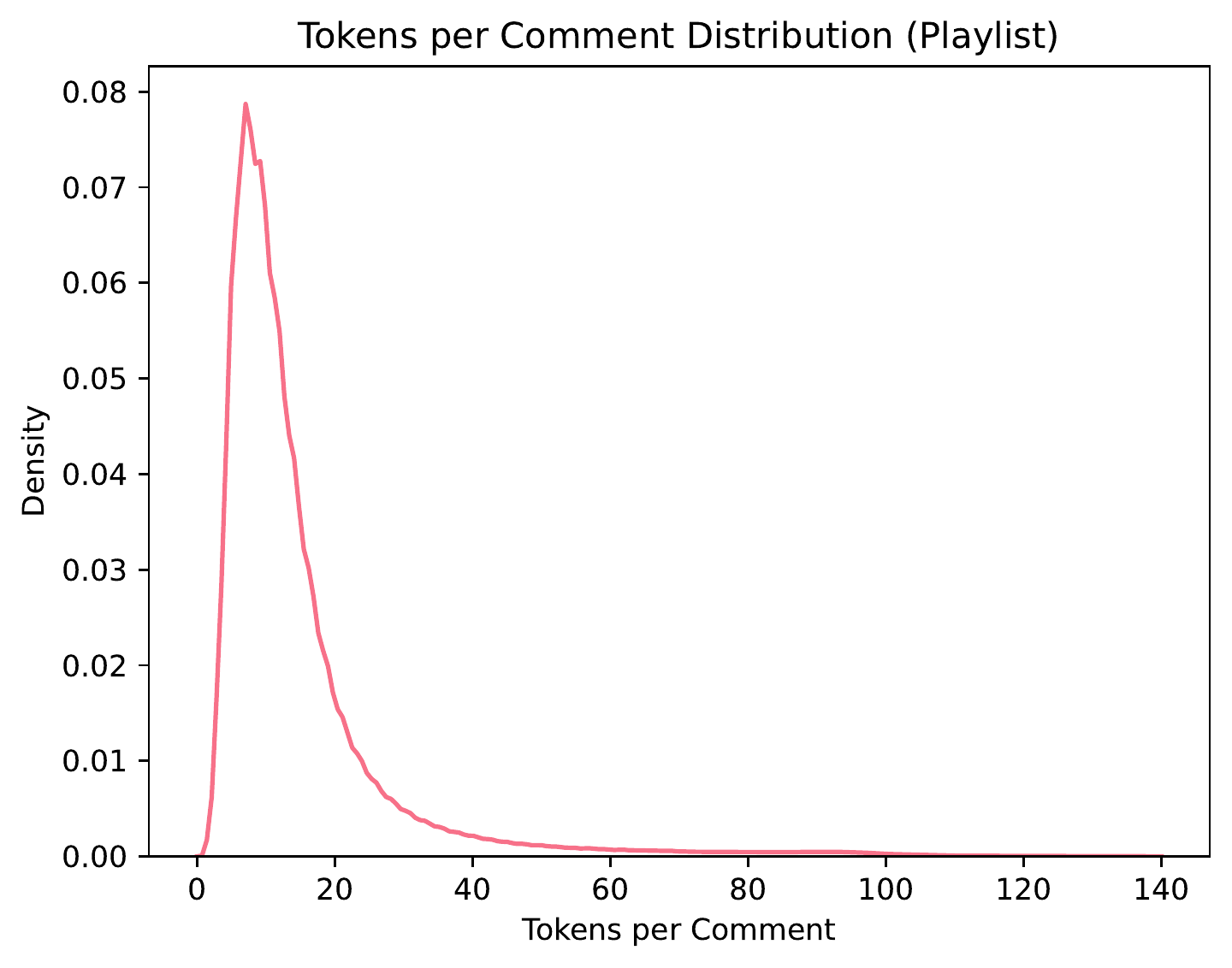}}
    \end{tabular}
    \caption{
    Comment (left) and comment token (right) distributions across all playlists with at least one comment.
    }
    \label{fig:playlist-comments}
\end{figure*}

Playlist comment distributions and comment token distributions are shown in Figure \ref{fig:playlist-comments}. The top 20 most popular tags used for playlists are, in descending order,
Language-Western (21.5\%), Style-Pop (19.2\%), Language-Chinese (15.1\%), Style-Electronic (13.4\%), Emotion-Healing (7.66\%), Theme-ACG (7.56\%), Setting-Night (7.03\%), Style-Soft (6.56\%), Theme-Games (6.18\%), Theme-Movies (6.10\%), Emotion-Relaxing (5.87\%), Emotion-Exciting (5.33\%), Style-Rock (5.32\%), Style-Rap (5.08\%), Emotion-Nostalgic (4.91\%), Emotion-Quiet (4.65\%), Setting-Study (4.50\%), Theme-Classics (4.11\%), and Emotion-Sadness (4.09\%).
Note here that as playlists may each contain at most three tags, summing percentages for all tags exceeds 100\%.

\subsection{Albums}

\begin{figure*}[!t]
    \centering
    \begin{tabular}{ccc}
    {\includegraphics[width=0.3\textwidth]{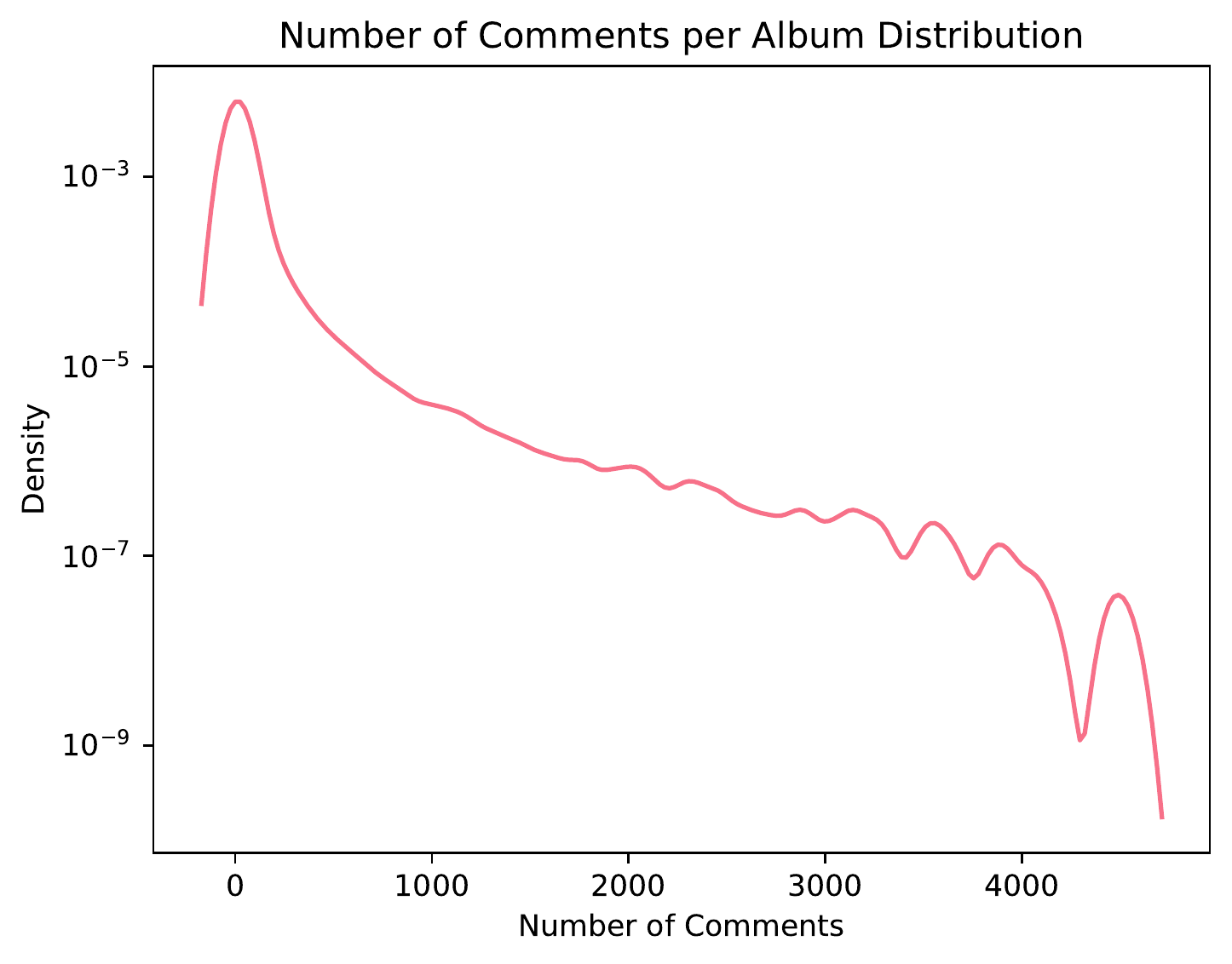}} &
    {\includegraphics[width=0.3\textwidth]{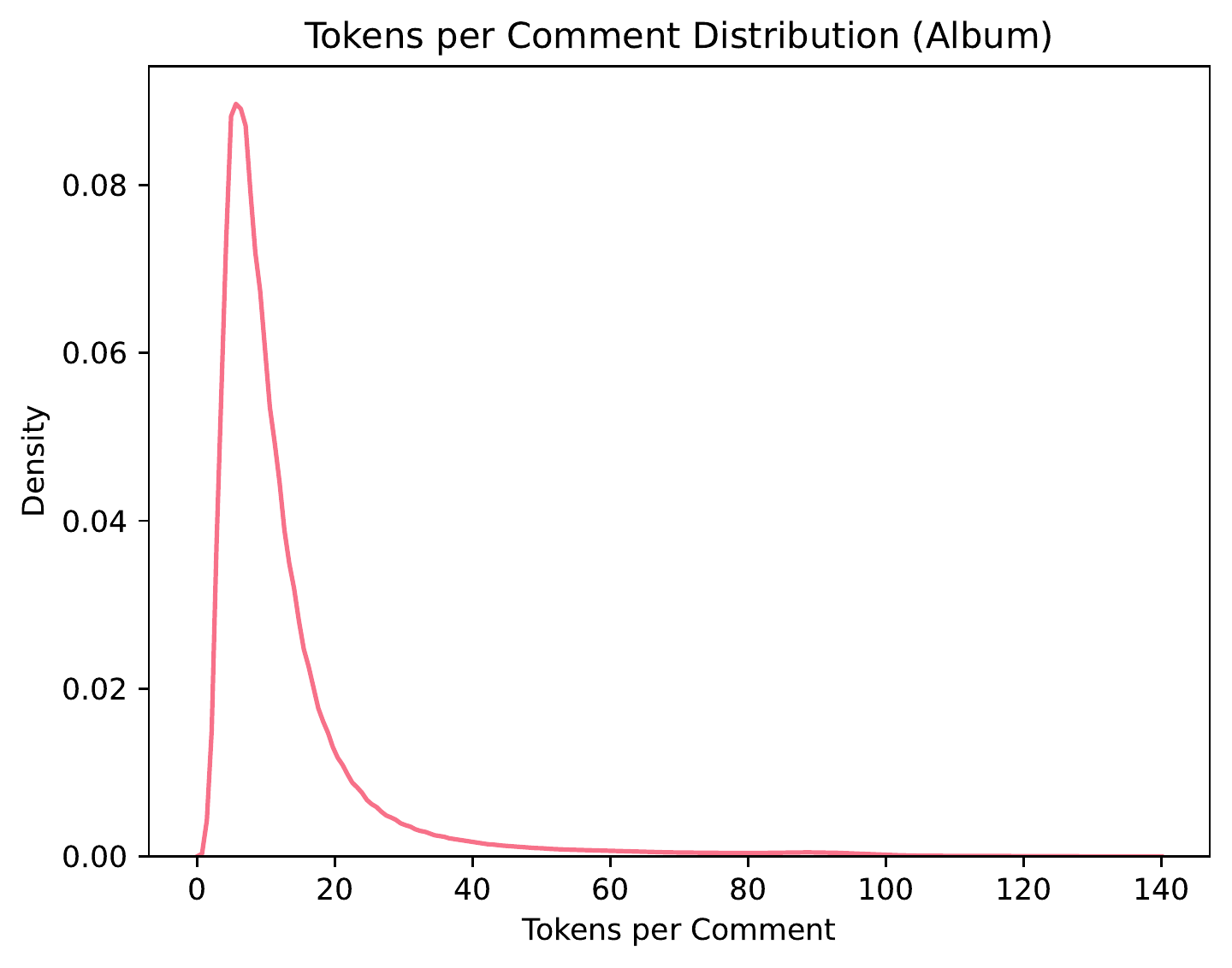}} &
     {\includegraphics[width=0.3\textwidth]{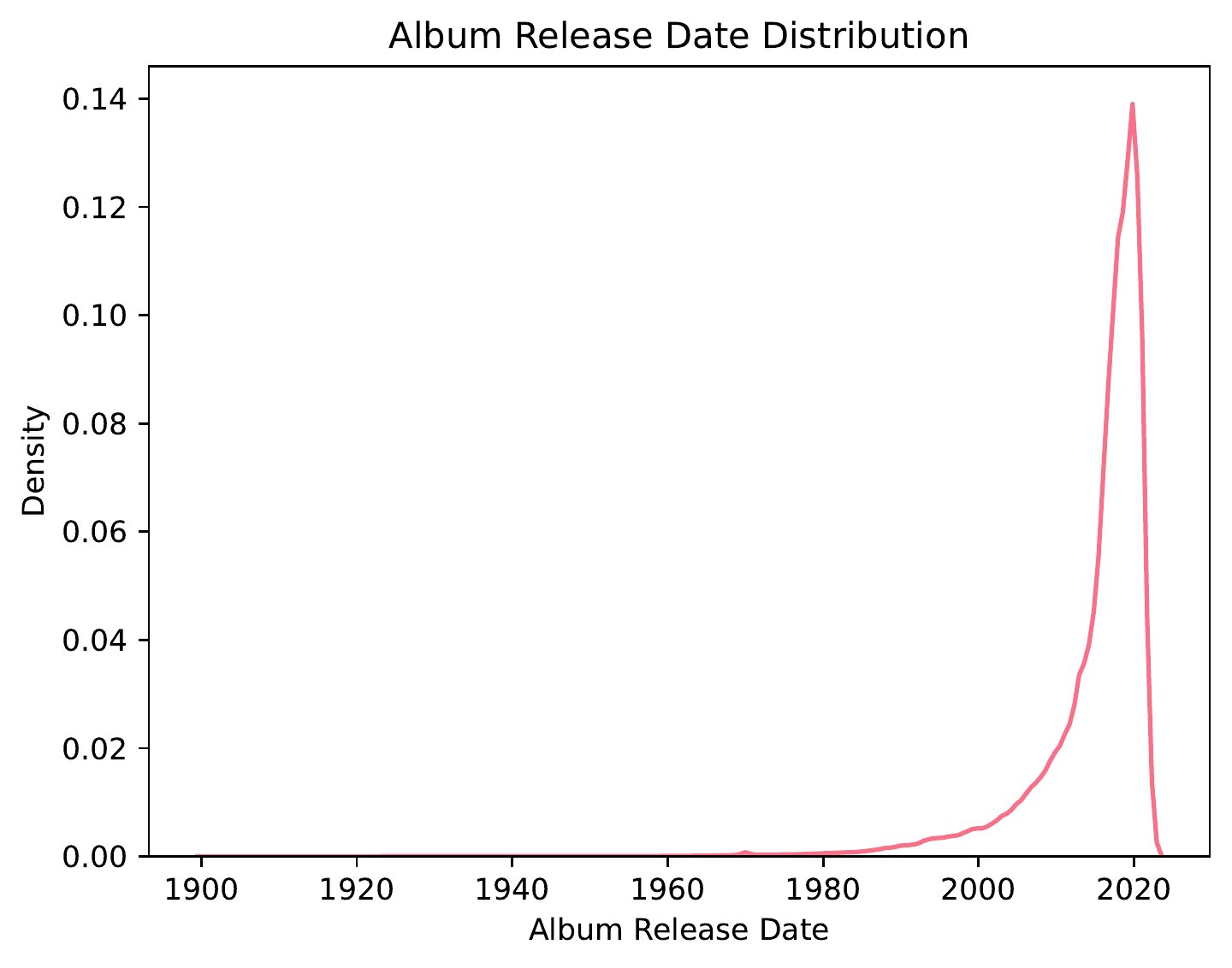}}
    \end{tabular}
    \caption{
    Comment (left) and comment token (middle) distributions across all albums with at least one comment, as well as album release date distributions (right).
    }
    \label{fig:album-comments}
\end{figure*}

Album comment, comment token, and release date distributions are shown in Figure \ref{fig:album-comments}. Songs with at least one comment show exponential bias towards recently released music. 

\subsection{Artists}

\begin{figure*}[!t]
    \centering
    \begin{tabular}{ccc}
    {\includegraphics[width=0.4\textwidth]{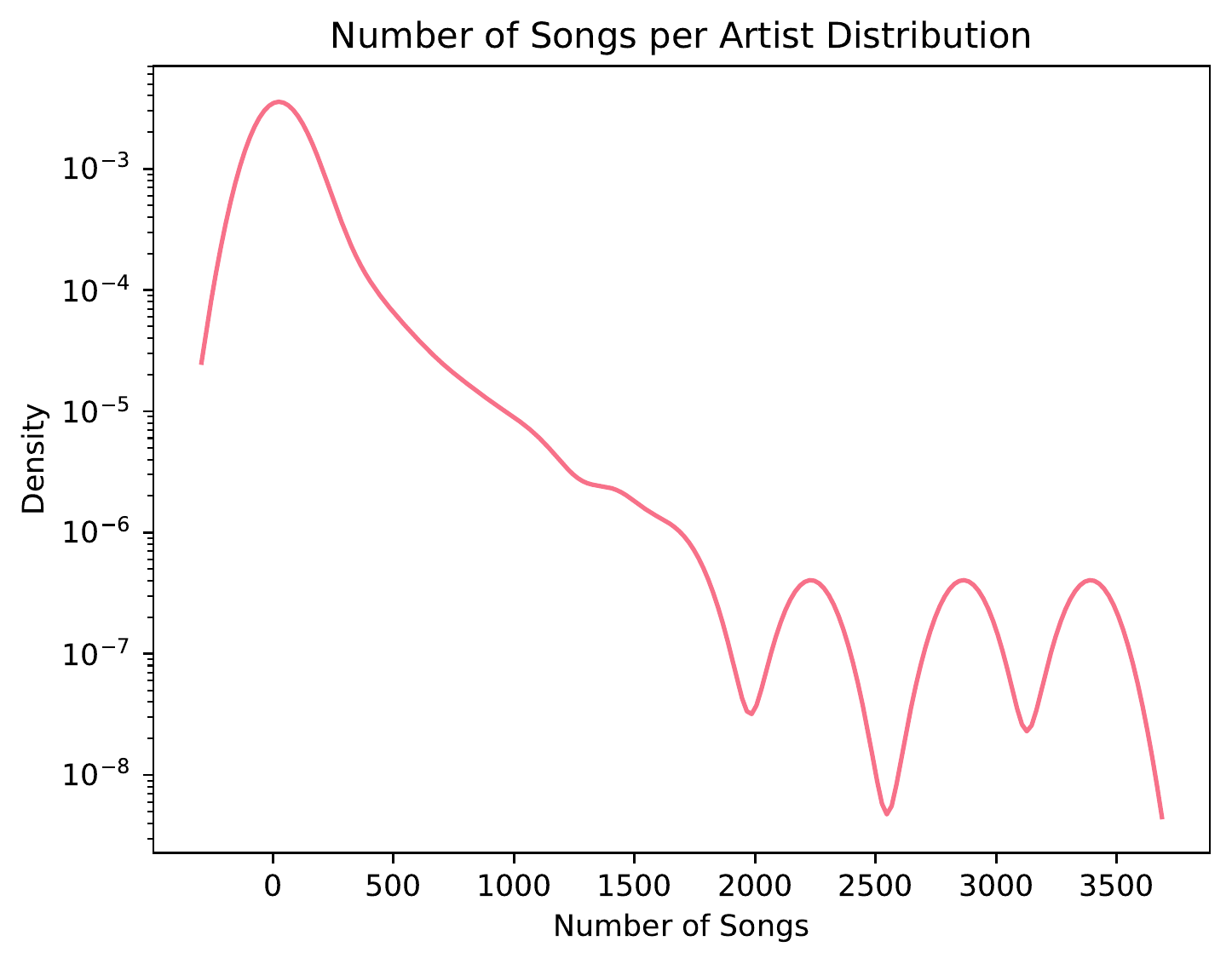}} &
    {\includegraphics[width=0.4\textwidth]{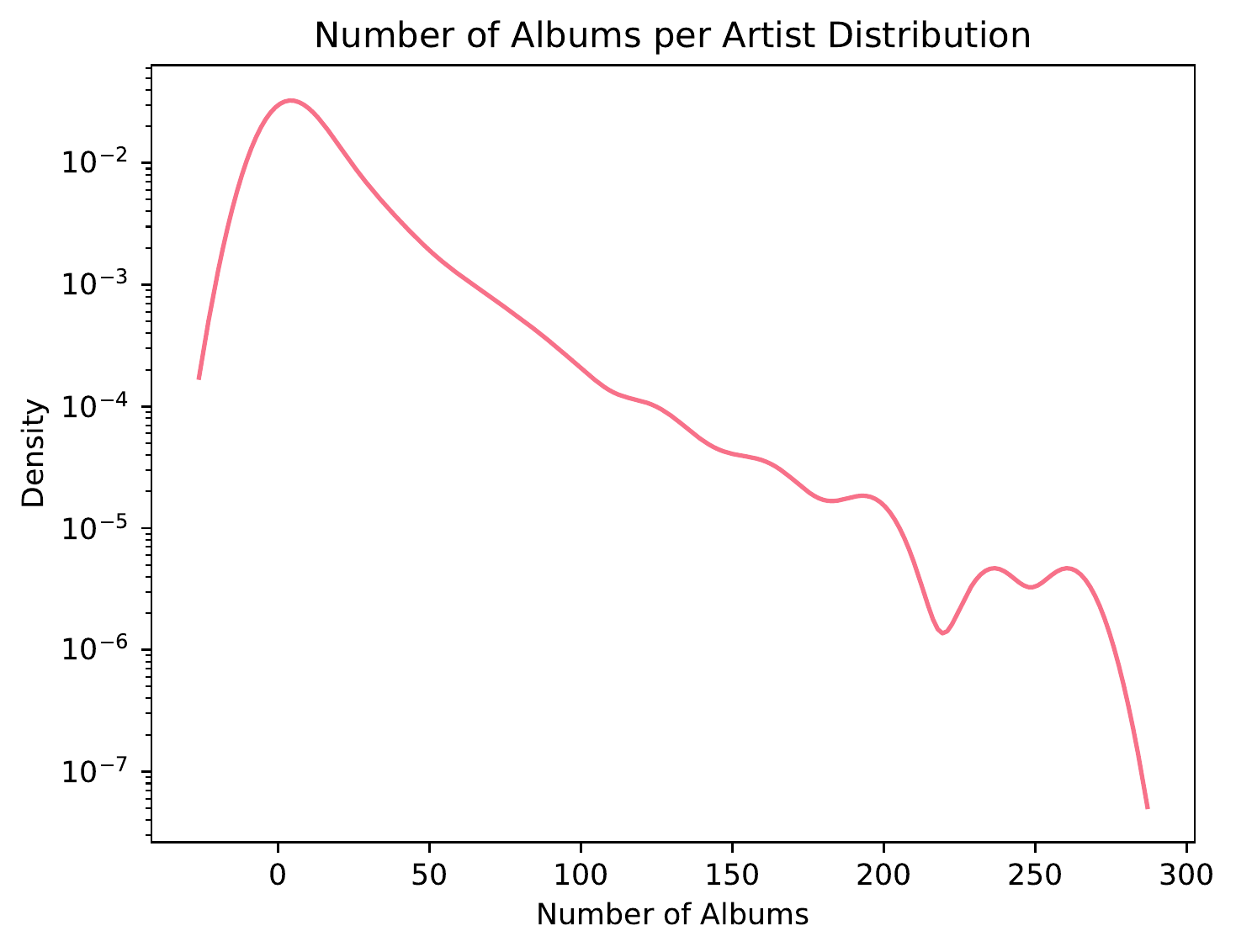}}
    \end{tabular}
    \caption{
    Song (left) and album (right) distributions per artist across all artists. Platform-listed artists with the highest amount of songs and albums are generic compilations of multiple artists, e.g. ``\chinese{华语群星}'' (``Chinese stars'').
    }
    \label{fig:artist}
\end{figure*}

A distribution of the number of albums and songs per artist is shown in Figure \ref{fig:artist}.

\subsection{Demographic Baselines}
\label{sec:demographicbaselines}

Users of different demographic groups have varying comment valence and arousal means and standard deviations. These statistics, stratified by demographic groups on gender and age, are shown in Table \ref{tab:dembaselines}.

\begin{table}
    \centering
    \begin{tabular}{ l|rr } 
     \toprule
     Group & Valence (m./std.) & Arousal (m./std.) \\
     \midrule
     Women & 5.93/1.33 & 5.23/1.09 \\ 
     Men & 5.72/1.38 & 5.14/1.13 \\ 
     \midrule
     10\chinese{后} & 5.76/1.37 & 5.17/1.12 \\ 
     05\chinese{后} & 5.79/1.40 & 5.21/1.10 \\ 
     00\chinese{后} & 5.85/1.38 & 5.22/1.10 \\ 
     95\chinese{后} & 5.80/1.36 & 5.17/1.11 \\ 
     90\chinese{后} & 5.76/1.34 & 5.12/1.11 \\ 
     85\chinese{后} & 5.74/1.36 & 5.13/1.12 \\ 
     80\chinese{后} & 5.79/1.32 & 5.10/1.11 \\ 
     75\chinese{后} & 5.75/1.32 & 5.08/1.10 \\ 
     70\chinese{后} & 5.81/1.31 & 5.09/1.09 \\ 
     65\chinese{后} & 5.81/1.33 & 5.13/1.11 \\ 
     60\chinese{后} & 5.91/1.33 & 5.15/1.10 \\ 
     55\chinese{后} & 5.81/1.36 & 5.18/1.11 \\ 
     50\chinese{后} & 5.77/1.40 & 5.19/1.11 \\ 
     \bottomrule
    \end{tabular}
    \caption{
    Comment valence and arousal mean (m.) and standard deviations (std.) for demographic groups on gender and age.
    }
    \label{tab:dembaselines}
\end{table}

\section{Emotion Annotation Guidelines}
\label{sec:annotationguidelines}

This section describes the annotation guidelines used by annotators in our pilot studies to determine, in top-level comments, (1) the emotion experiencer, or who was the primary experiencer of the emotions expressed in comments, and (2) the affective stimulus of the emotion expressed in the comment. Annotators consisted of two Chinese native speakers and were asked to annotate a set of 1000 randomly selected comments on the platform.

Annotators were first tasked with familiarizing themselves with the BRECVEMA framework of musically evoked emotions \cite{JUSLIN2013235} before being presented with the following questionnaires for annotation:

\vspace{\baselineskip}
\noindent \textbf{Question 1: The Emotion Experiencer}\\
\textbf{Comment:} \chinese{真特么的带感这曲子！}\\
Q. Who was the primary experiencer of the emotion expressed in the comment?
\begin{itemize}
\itemsep 0em
\item The commenter themselves.
\item Someone other than the commenter themselves.
\item This comment possesses no emotional content.
\end{itemize}

\noindent \textbf{Question 2: The Affective Stimulus}\\
\textbf{Comment:} \chinese{真特么的带感这曲子！}\\
Q. What was the primary affective stimulus of the emotion expressed in the comment?
\begin{itemize}
\itemsep 0em
\item The song, album, or playlist.
\item Something other than the song, album, or playlist.
\item This comment possesses no emotional content.
\end{itemize}

As stated, annotators were asked to resolve initial annotation disagreements through discussion in order to come up with a set of annotations that both agreed on.

\section{Lyric Topics and Preprocessing}
\label{sec:lyricpreprocessing}

This section describes our lyrical preprocessing methods and 20-topic LDA model results on song lyrics.

\myparagraph{Preprocessing.} We first identify instrumental music by matching lyric data on the substring \chinese{纯音乐}, used by the platform to denote songs of this category. For non-instrumental pieces, we filter out lines with song metadata (e.g. composers) by removing lines that match the following regex:

\texttt{\chinese{:|：|《|》|produced by|vocals by|recorded by|edited by|mixed by|mastered by| - | － }}

As repeated lyrics are denoted with overlapping time stamps, e.g.

\texttt{\chinese{[1:00.00][2:00.00] 雨淋湿了天空}}

indicates that the line \chinese{雨淋湿了天空} is repeated at minutes 1 and 2), we further unfurl and reorder lines by timestamp, duplicating lines when necessary. Further tokenization details are shown in  \nameref{sec:commentpreprocessing}.

\myparagraph{Topic Modeling.} We train a 20-topic LDA model on preprocessed song lyrics and manually label each lyric with its prominent theme. While some degree of variation exists for listener affective responses across songs of each topic, these topic distributions are primarily used as lyrical content controls in our regression models. Labeled topics and their top words are shown in Table \ref{tab:lyrictopics}.

\begin{table*}[h]
    \centering
    \begin{tabular}{ l|l } 
     \toprule
     Topic Theme & Top Tokens \\
     \midrule
     
     0. Romance/Sentiment & \chinese{爱, 未, 今天, 便, 似, 一生, 令, 心, 没, 心中, 里, 太, 愿, 仍然, 想, 没法,} \\
     \chinese{(爱情/感性/伤感)} & \chinese{一起, 讲, 吻, 快乐} \\
     
     1. Youth/Hope/Warmth & \chinese{梦想, 希望, 地方, 梦, 世界, 身旁, 远方, 青春, 路, 模样, 时光, 方向,} \\
     \chinese{(青春/希望/阳光)} & \chinese{未来, 流浪, 勇敢, 阳光, 带, 温暖, 生命, 心中} \\
     
     2. Transcendental & \chinese{人间, 江湖, 天地, 皆, 天下, 少年, 山河, 剑, 生死, 笑, 问, 间, 世间, 道,} \\
     \chinese{(人生/社会/超俗)} & \chinese{万里, 便, 江山, 英雄, 合, 此生} \\
     
     3. Hometown/Childhood & \chinese{花, 家, 牵挂, 长大, 噢, 回家, 记得, 回答, 说话, 画, 天涯, 挣扎, 走, 呐,} \\
     \chinese{(故乡/童年)} & \chinese{害怕, 变化, 落下, 傻, 年华, 故事} \\
     
     4. Friendship/Hedonism & \chinese{吃, 不要, 没, 兄弟, 快, 音乐, 钱, 没有, 起来, 新, 听, 带, 买, 今天, 走,} \\
     \chinese{(享受/欲望/世俗)} & \chinese{站, 玩, 现在, 喝, 说唱} \\
     
     5. Love/Lust & \chinese{喔, 女, 男, 合, 阮, 喝, 一杯, 一半, 讲, 耶, 人生, 爱, 伊, 酒, 甲, 拢,} \\
     \chinese{(恋爱/情欲)} & \chinese{咿呀, 啊啊啊, 心爱, 搁} \\
     
     6. Memories/Regret & \chinese{没有, 想, 没, 不会, 知道, 现在, 不想, 已经, 生活, 里, 太, 真的, 想要,} \\
     \chinese{(从前/失望)} & \chinese{时间, 总是, 听, 曾经, 其实, 不能, 一直} \\
     
     7. Nature/Spring & \chinese{唱, 美丽, 姑娘, 飞, 月亮, 草原, 歌, 吹, 故乡, 春天, 开, 歌声, 轻轻,} \\
     \chinese{(阳光/故乡/自然/草原)} & \chinese{歌唱, 亲爱, 太阳, 一片, 花儿, 远方, 阳光} \\
     
     8. Breakups/Sadness & \chinese{走, 没有, 爱, 手, 寂寞, 温柔, 快乐, 不要, 懂, 回头, 以后, 梦, 朋友,} \\
     \chinese{(分手/情感/失恋)} & \chinese{难过, 自由, 不会, 最后, 记得, 沉默, 拥有} \\
     
     9. Nostalgia/Melancholy & \chinese{相思, 一曲, 醉, 听, 落, 岁月, 梦, 红尘, 明月, 桃花, 笑, 人间, 花, 叹,} \\
     \chinese{(桑感/忧愁/思念/思乡)} & \chinese{不见, 故人, 春风, 似, 间, 清风, 见} \\
     
     10. Heartbreak/Loneliness & \chinese{爱, 心, 爱情, 眼泪, 哭, 太, 寂寞, 不要, 泪, 越, 女人, 心碎, 恨, 伤, 美} \\
     \chinese{(爱情/失恋/孤独/伤心)} & \chinese{幸福, 想, 错, 后悔, 不会} \\
     
     11. Wistful/Sentimental & \chinese{梦, 一生, 情, 心, 愿, 心中, 今生, 难, 梦里, 往事, 雨, 问, 岁月, 泪,} \\
     \chinese{(思念/孤独)} & \chinese{匆匆, 人生, 如今, 相逢, 相思, 风雨} \\
     
     12. Family/Longing & \chinese{妈妈, 唔, 哒, 想, 喵, 好想你, 爸爸, 妳, 宝贝, 话, 嘅, 滴, 摇, 快, 系, 滴答,} \\
     \chinese{(家庭/思念)} & \chinese{讲, 一只, 唿, 笑} \\
     
     13. Celestial/Awe & \chinese{里, 风, 天空, 故事, 听, 记忆, 温柔, 城市, 雨, 回忆, 梦, 黑夜, 时光,} \\
     \chinese{(孤独/渺小)} & \chinese{相遇, 声音, 风景, 夜空, 梦境, 流星} \\
     
     14. Love & \chinese{爱, 想, 爱情, 心, 忘记, 没有, 离开, 永远, 等待, 明白, 回忆, 不会,} \\
     \chinese{(爱情)} & \chinese{未来, 不要, 我爱你, 相信, 一起, 不能, 愿意, 一次} \\
     
     15. Countryside/Family & \chinese{呦, 哥哥, 嗨, 里, 妹妹, 哥, 走, 转, 白, 长, 嘞, 熘, 红, 山, 开, 水, 见, 笑,} \\
     \chinese{(乡村/山水)} & \chinese{送} \\
     
     16. Blossoming/Joy & \chinese{想, 一起, 喜欢, 陪, 爱, 知道, 想要, 笑, 世界, 微笑, 拥抱, 感觉, 慢慢,} \\
     \chinese{(爱情/幸福)} & \chinese{眼睛, 听, 心, 心里, 我要, 带, 幸福} \\
     
     17. Nationalism/China & \chinese{中国, 恭喜, 新, 菩萨, 熘, 南无, 祖国, 祝, 北京, 人民, 英雄, 来来来,} \\
     \chinese{(爱国)} & \chinese{新年, 吼, 平安, 东方, 历史, 阿弥陀佛, 祝福, 菩提} \\
     
     18. Time/Nihilism & \chinese{一天, 时间, 再见, 身边, 永远, 脸, 世界, 改变, 思念, 从前, 想念,} \\
     \chinese{(时间/转瞬即逝)} & \chinese{明天, 远, 出现, 看见, 回忆, 昨天, 一点, 一遍, 一年} \\
     
     19. Being/Existential & \chinese{世界, 无法, 灵魂, 现实, 需要, 不断, 成为, 黑暗, 继续, 命运, 生命,} \\
     \chinese{(存在/生命的意义)} & \chinese{身体, 内心, 像是, 保持, 自我, 有人, 每个, 孤独, 自由} \\
     
    \bottomrule
    \end{tabular}
    \caption{
    Manually labeled lyrical topics and their top tokens, as captured from a 20-topic LDA model trained on preprocessed song lyrics.
    }
    \label{tab:lyrictopics}
\end{table*}

\section{Text Preprocessing}
\label{sec:commentpreprocessing}

This section describes our text preprocessing pipeline for all text data on the platform, namely (1) lyrics and (2) listener comments. 

\myparagraph{Preprocessing.} We analyze only Chinese language content, using Google's Compact Language Detector v3 (\texttt{gcld3}\footnote{https://github.com/google/cld3}) to detect text language and keep only Chinese language texts. We then convert all traditional Chinese characters to their simplified forms using \texttt{hanziconv}\footnote{https://pythonhosted.org/hanziconv/} to ensure consistency in our experiments---i.e. when calculating LIWC scores, for which we use the simplified Chinese version \cite{huang2012development}---and finally tokenize with \texttt{jieba}.\footnote{https://github.com/fxsjy/jieba}

\myparagraph{Filtering for Affective Content.} Following annotations of listener comments, we filter out all comments that match the following regular expressions (Table \ref{tab:filters}), aiming to increase the precision of comments in our analysis that indicate an affective response. The following filters generally match with easily identifiable spam messages, i.e. ``first comment'', album images, and quotations.

\begin{table}[H]
    \centering
    \begin{tabular}{c} 
    \toprule
    \chinese{沙发,第一,第二, 第三, 第四, 第五, 第六,} \\
    \chinese{第七, 第八, 第九, 第十, 第1, 第2, 第3, 第4,} \\
    \chinese{第5, 第6, 第7, 第8, 第9, 一楼, 留名, 封面,} \\
    \chinese{没人, 来晚了, 板凳, 求, 前排, 识曲, 后排,} \\
    \chinese{一条, 好少, 不火, 助攻, 作者, 评论, 人呢,} \\
    \chinese{来了, “*”, "*", ‘*’, '*',《*》, <*>, ：, :, 9+} \\
    \bottomrule
    \end{tabular}
    \caption{
    Regular expressions used to filter out irrelevant listener comments that do not indicate an affective response.
    }
    \label{tab:filters}
\end{table}

\section{Expanded Results for Variations in Affective Response}
\label{sec:expandedresults}

\begin{figure*}[!t]
    \centering
    \begin{tabular}{ccc}
    {\includegraphics[width=0.30\textwidth]{plots/AME/mp3_features_tempo.pdf}} & %
    {\includegraphics[width=0.30\textwidth]{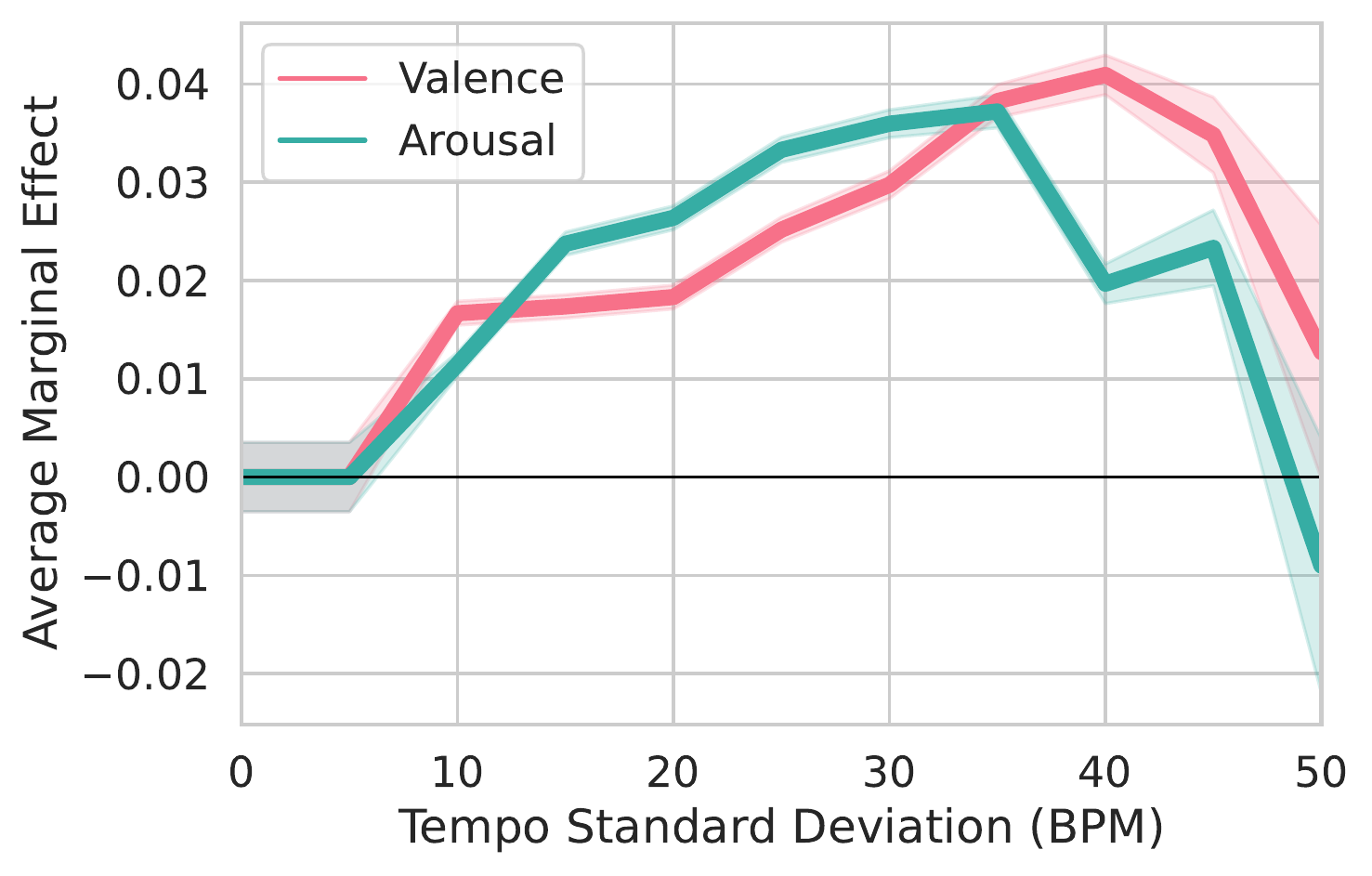}} & %
    {\includegraphics[width=0.30\textwidth]{plots/AME/mp3_features_loudness.pdf}} \\ %
    {\includegraphics[width=0.30\textwidth]{plots/AME/mp3_features_timbre_brightness.pdf}} & %
    {\includegraphics[width=0.30\textwidth]{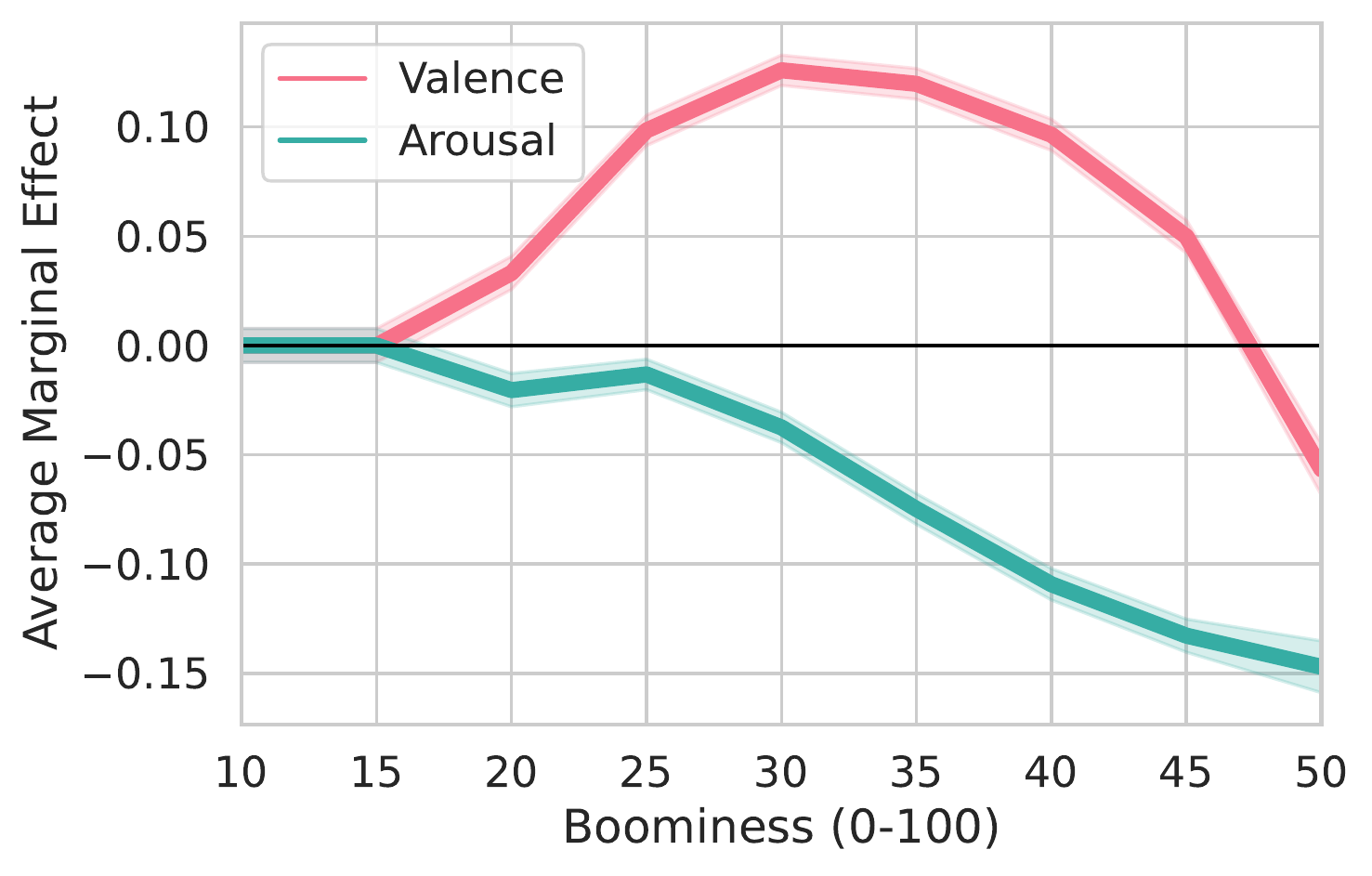}} & %
    {\includegraphics[width=0.30\textwidth]{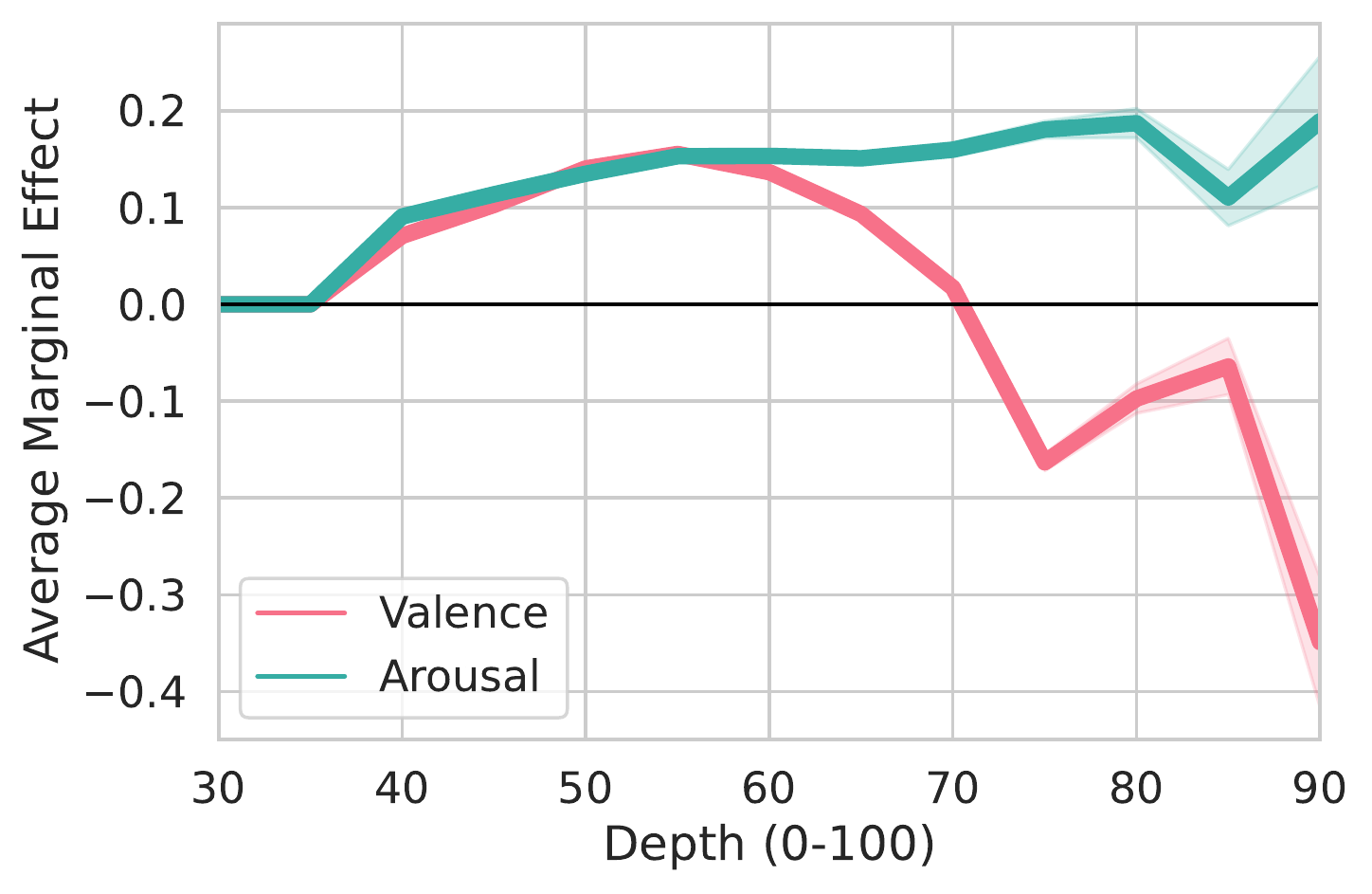}} \\ %
    {\includegraphics[width=0.30\textwidth]{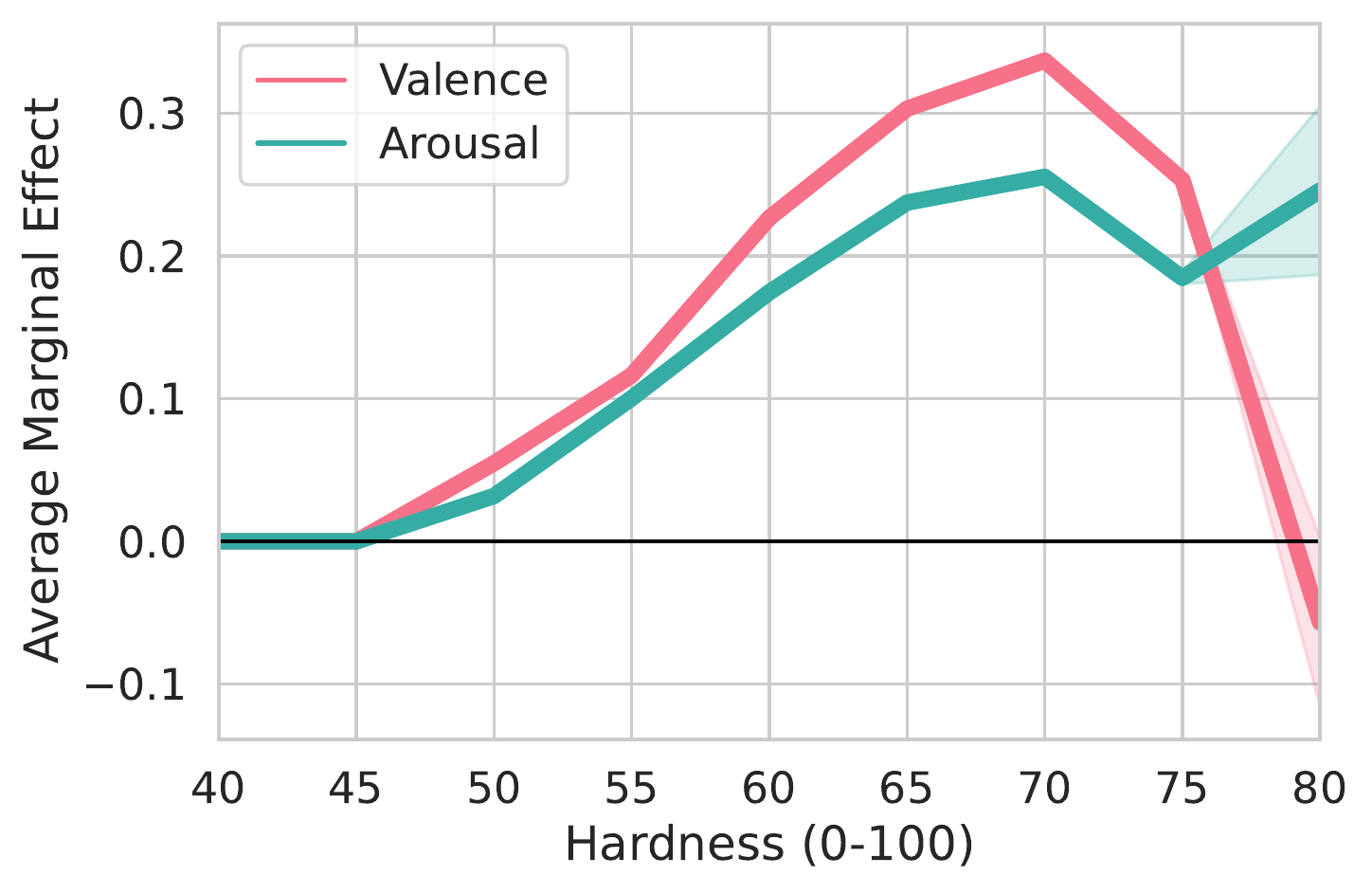}} & %
    {\includegraphics[width=0.30\textwidth]{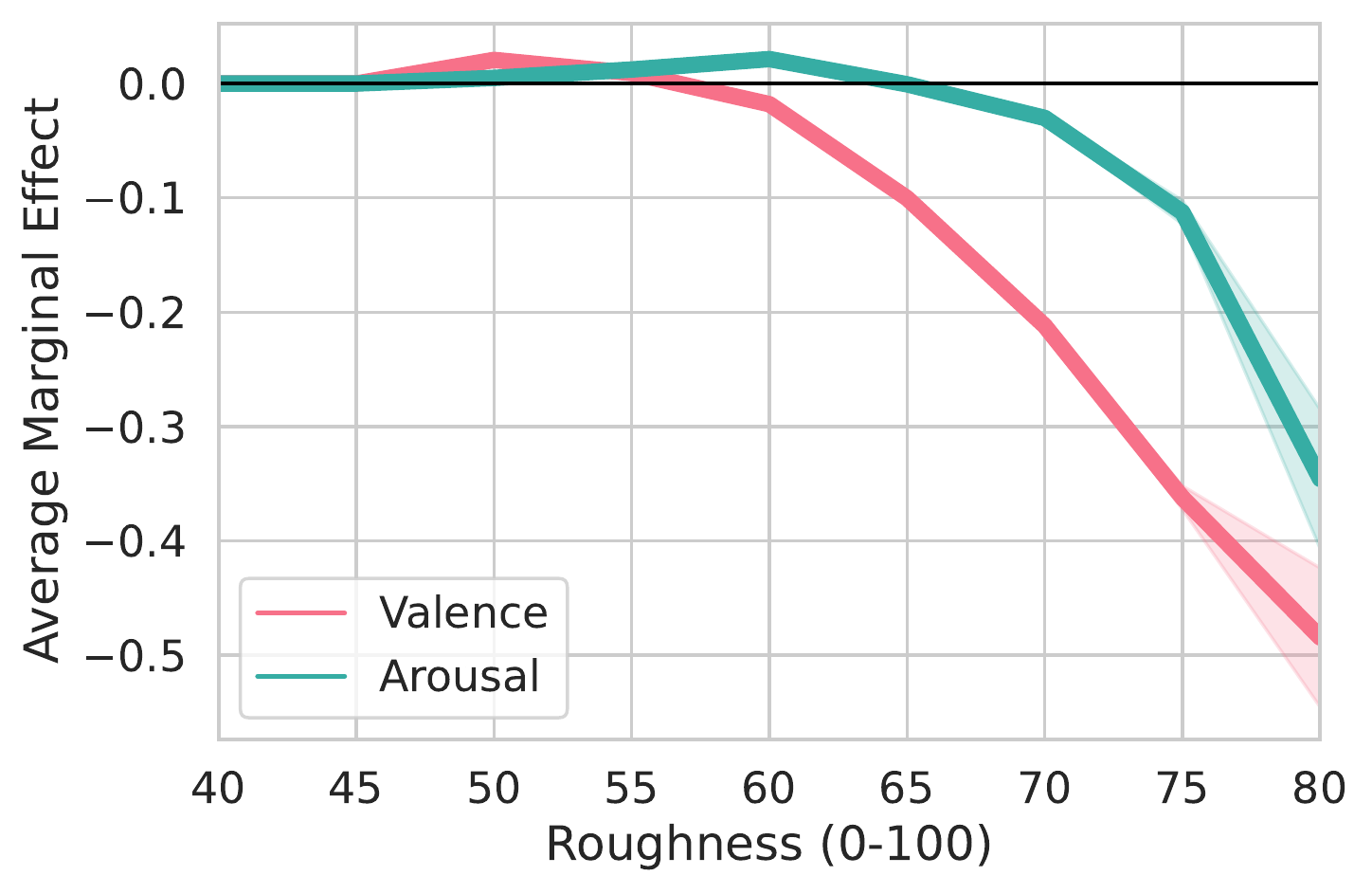}} & %
    {\includegraphics[width=0.30\textwidth]{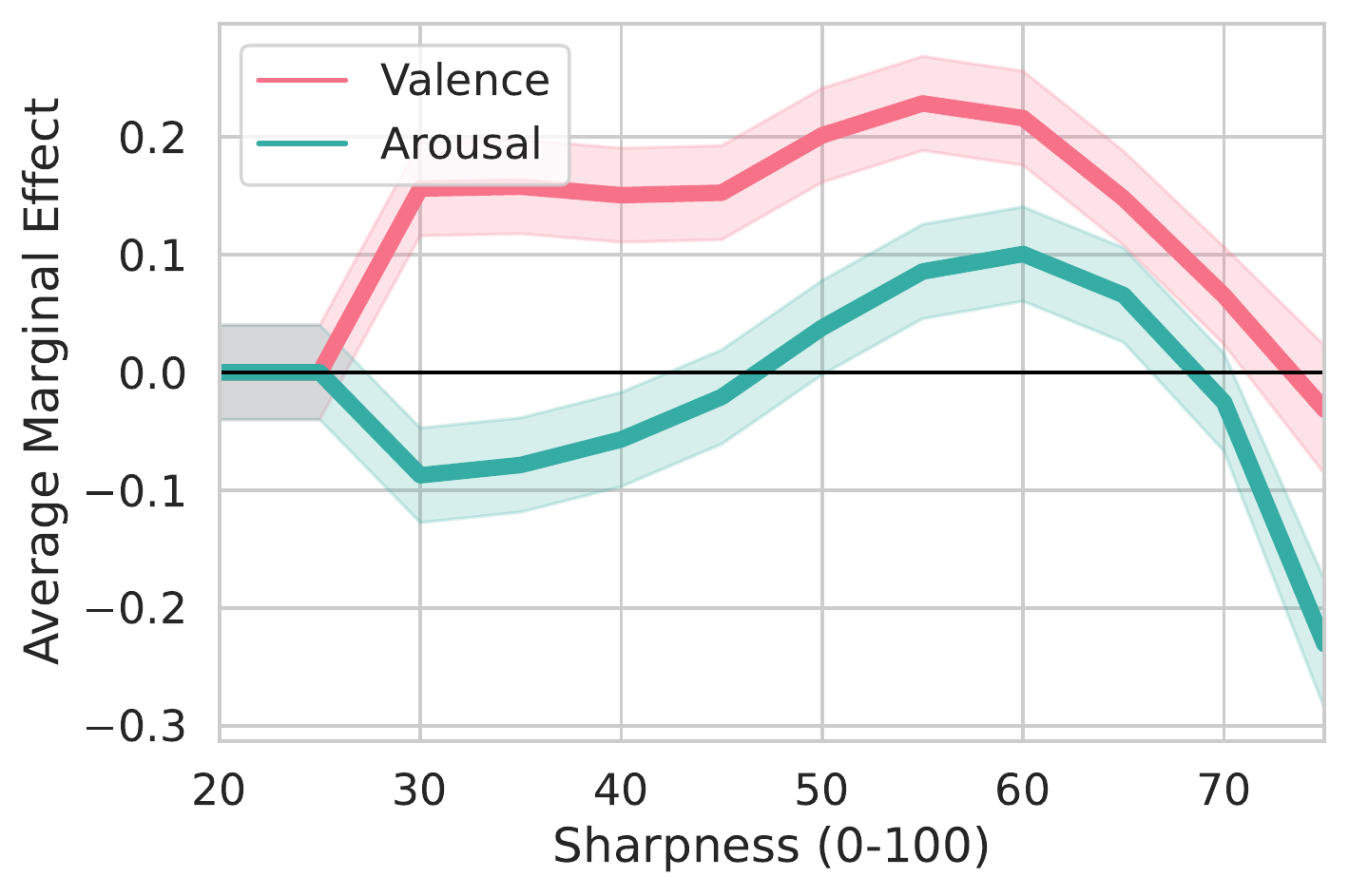}} \\ %
    {\includegraphics[width=0.30\textwidth]{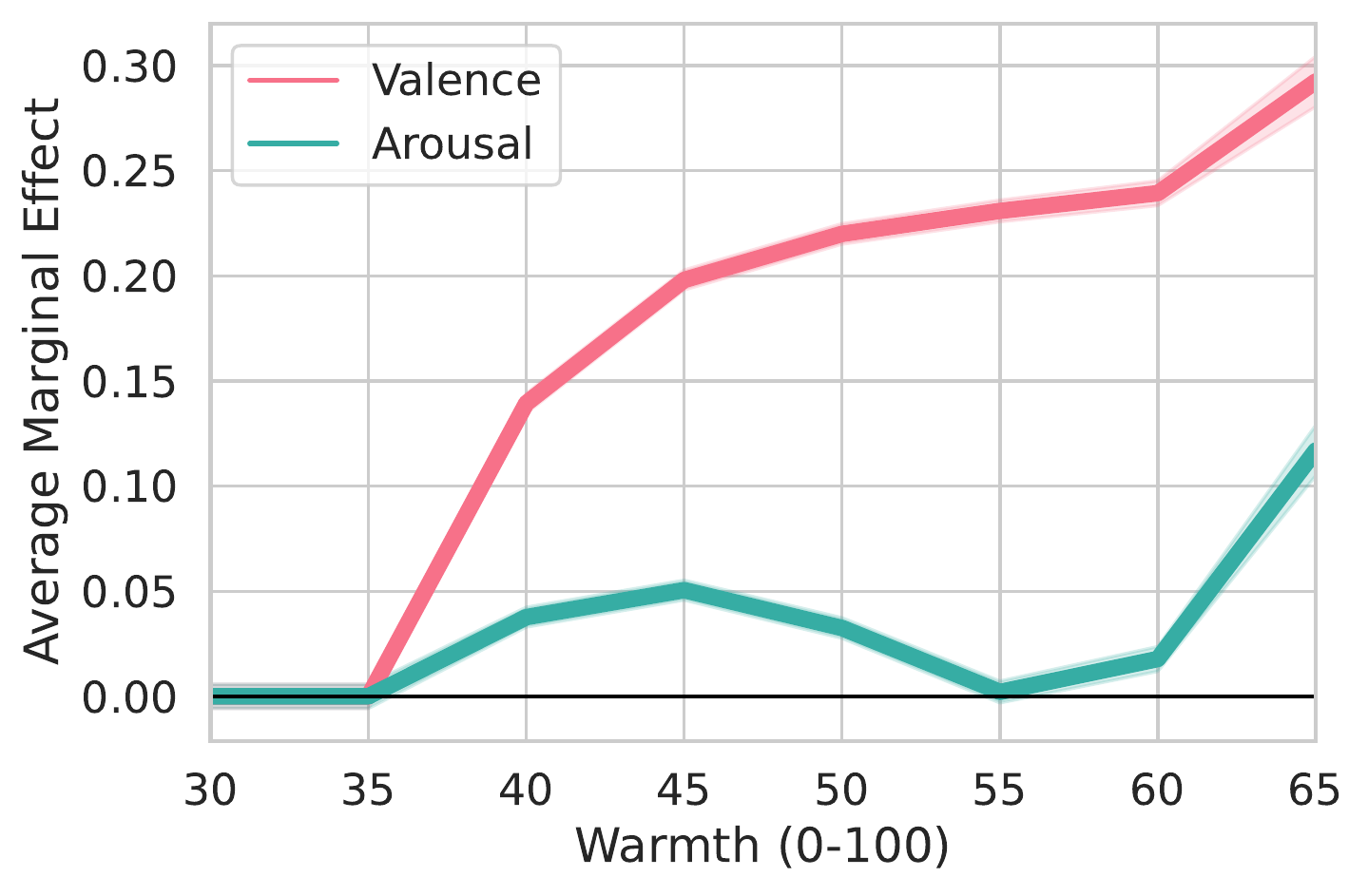}} & %
    {\includegraphics[width=0.30\textwidth]{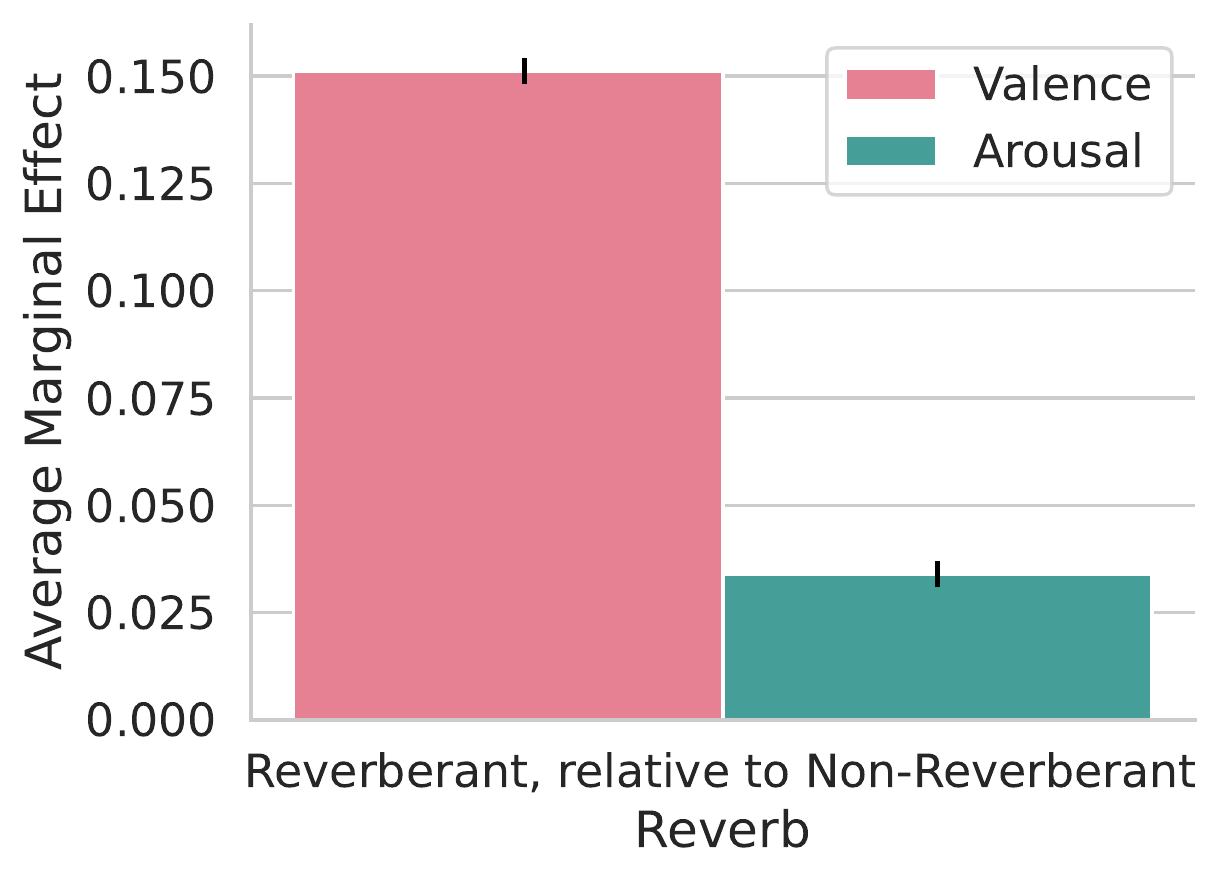}} & %
    {\includegraphics[width=0.30\textwidth]{plots/AME/mp3_features_mode.pdf}} \\ %
    \end{tabular}
    \caption{
    Average marginal effects of \textbf{musical features} on listener affective responses, controlling for lyrical features and listener demographics.
    Standard errors are shown;
    \textcolor{red}{valence} in \textcolor{red}{red}, \textcolor{blue}{arousal} in \textcolor{blue}{blue}.
    }
    \label{fig:musicfeatures_expanded}
\end{figure*}

\begin{figure*}[!t]
    \centering
    \begin{tabular}{ccc}
    {\includegraphics[width=0.90\textwidth]{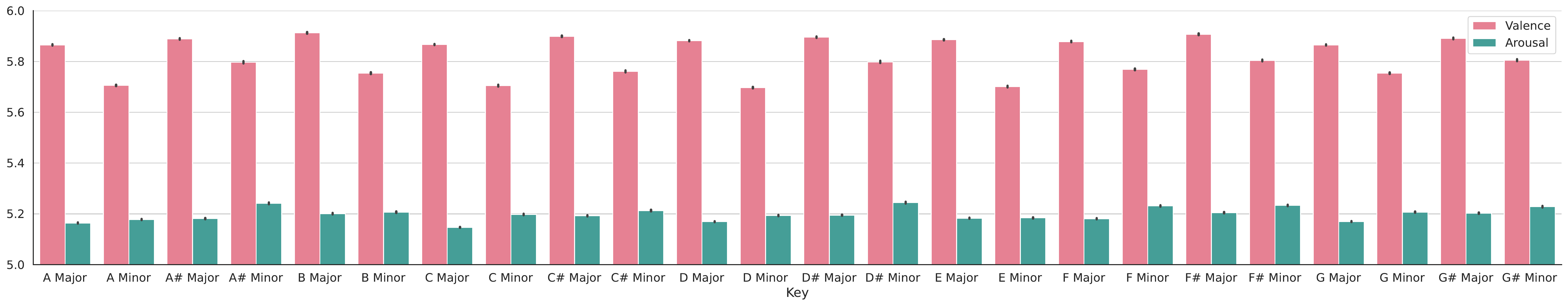}} & %
    \end{tabular}
    \caption{
    Raw valence and arousal scores for variations in listener affective responses with respect to \textbf{key}. Across all keys, valence response to major mode keys is consistently higher than that of their corresponding minor mode key, while the opposite relationship exists for arousal response.
    Standard errors are shown;
    \textcolor{red}{valence} in \textcolor{red}{red}, \textcolor{blue}{arousal} in \textcolor{blue}{blue}.
    }
    \label{fig:musicfeatures_key}
\end{figure*}

\begin{figure*}[!t]
    \centering
    \begin{tabular}{ccc}
    {\includegraphics[width=0.30\textwidth]{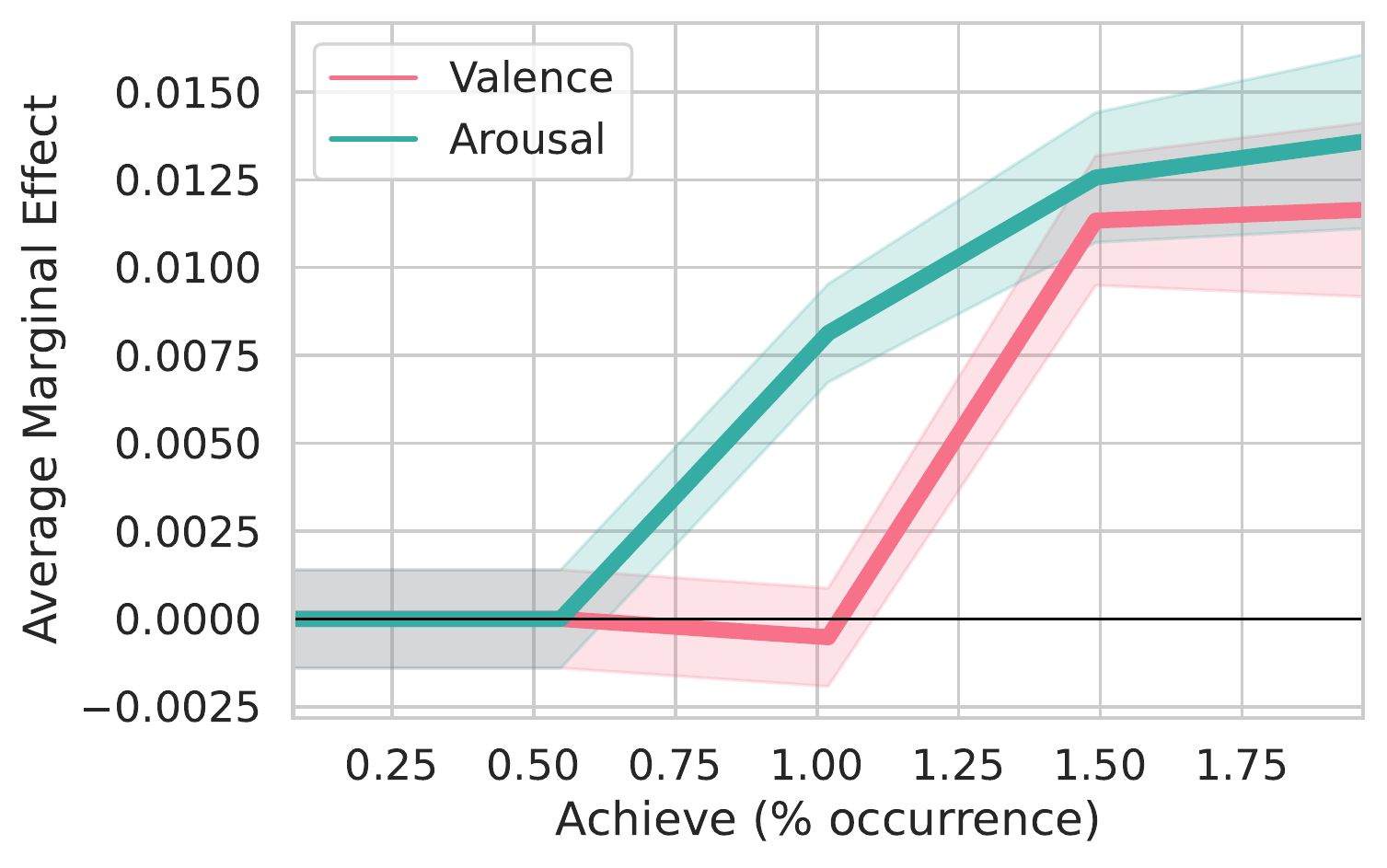}} &
    {\includegraphics[width=0.30\textwidth]{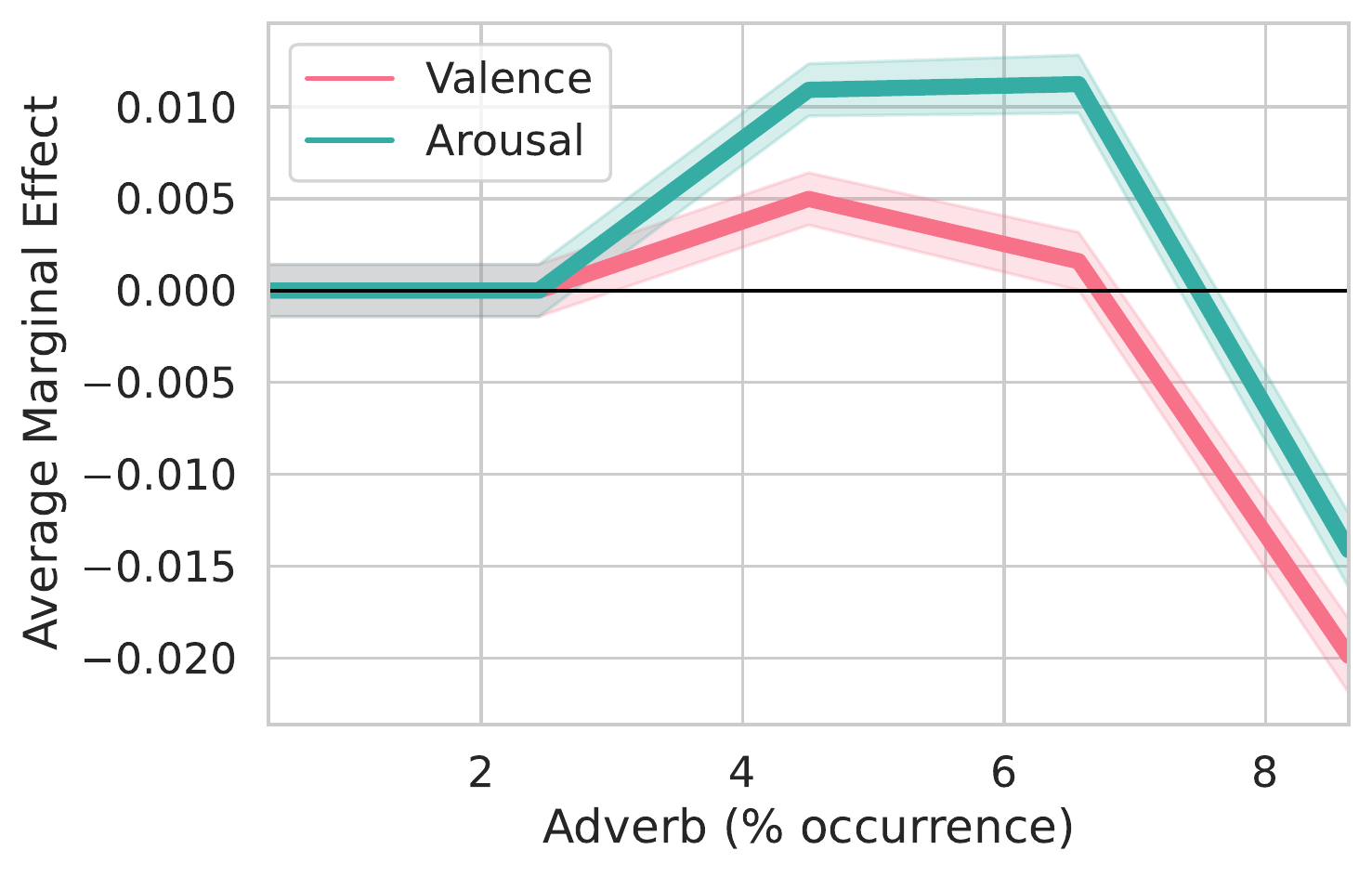}} & 
    {\includegraphics[width=0.30\textwidth]{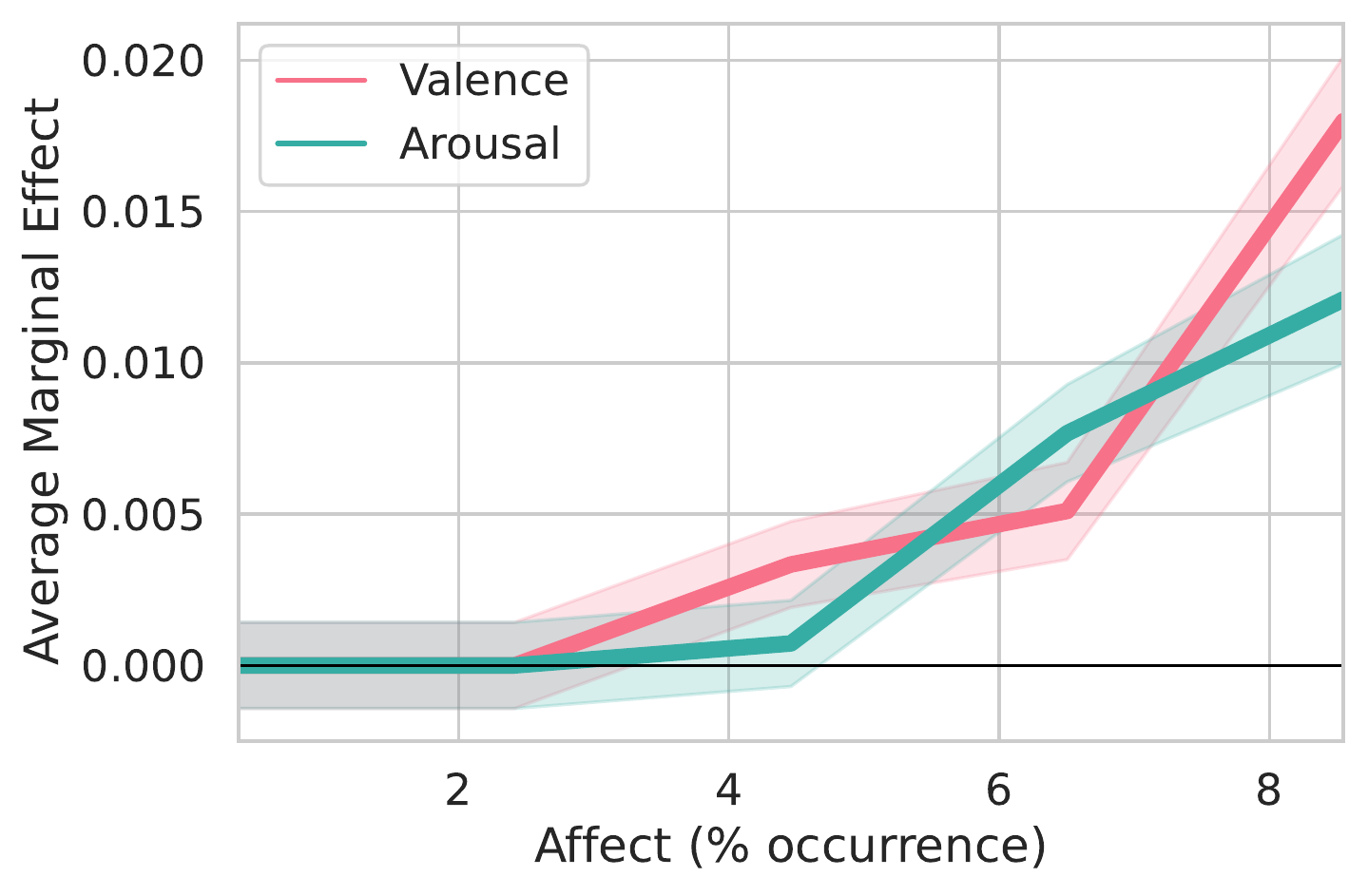}} \\
    {\includegraphics[width=0.30\textwidth]{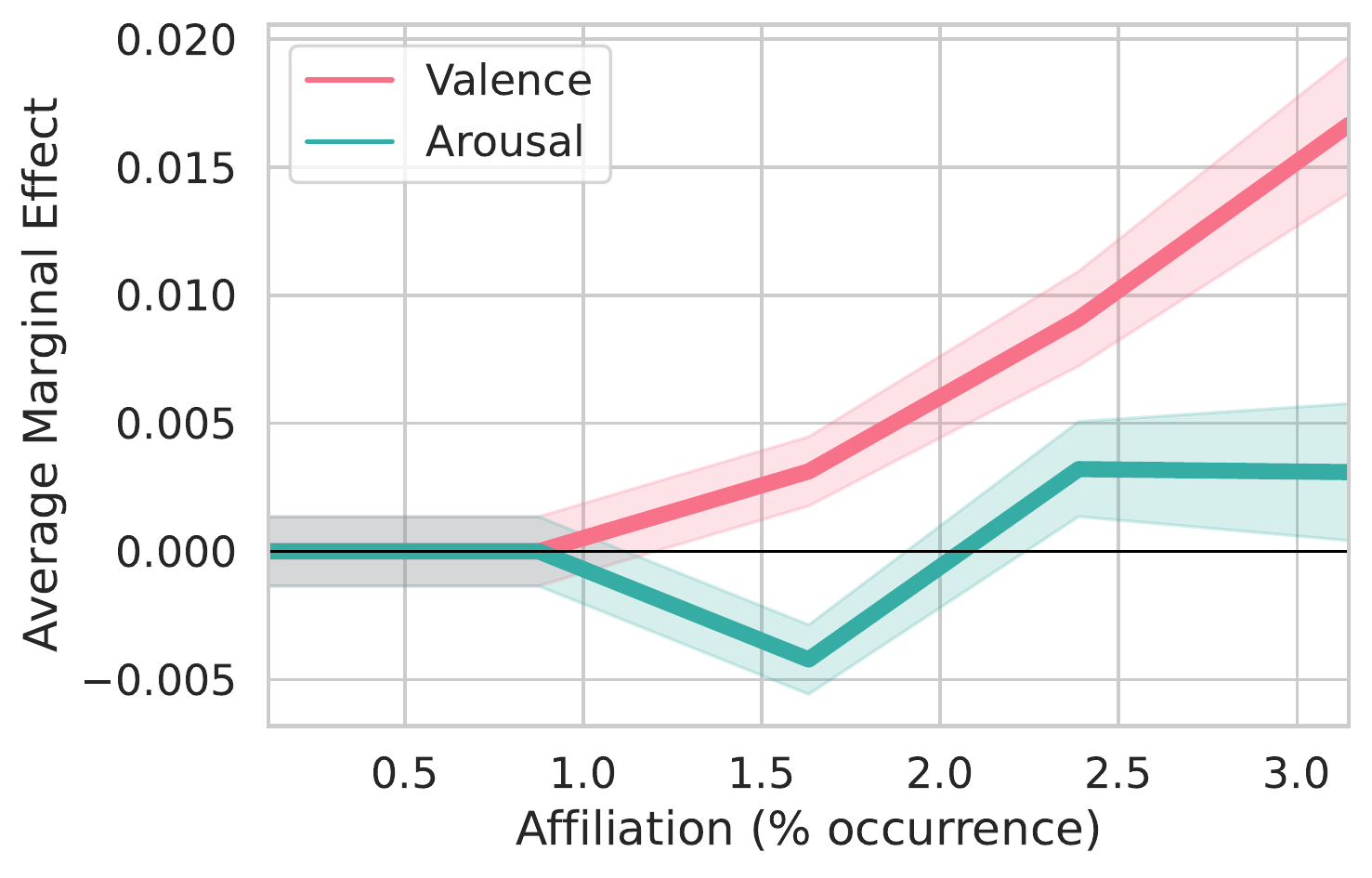}} & 
    {\includegraphics[width=0.30\textwidth]{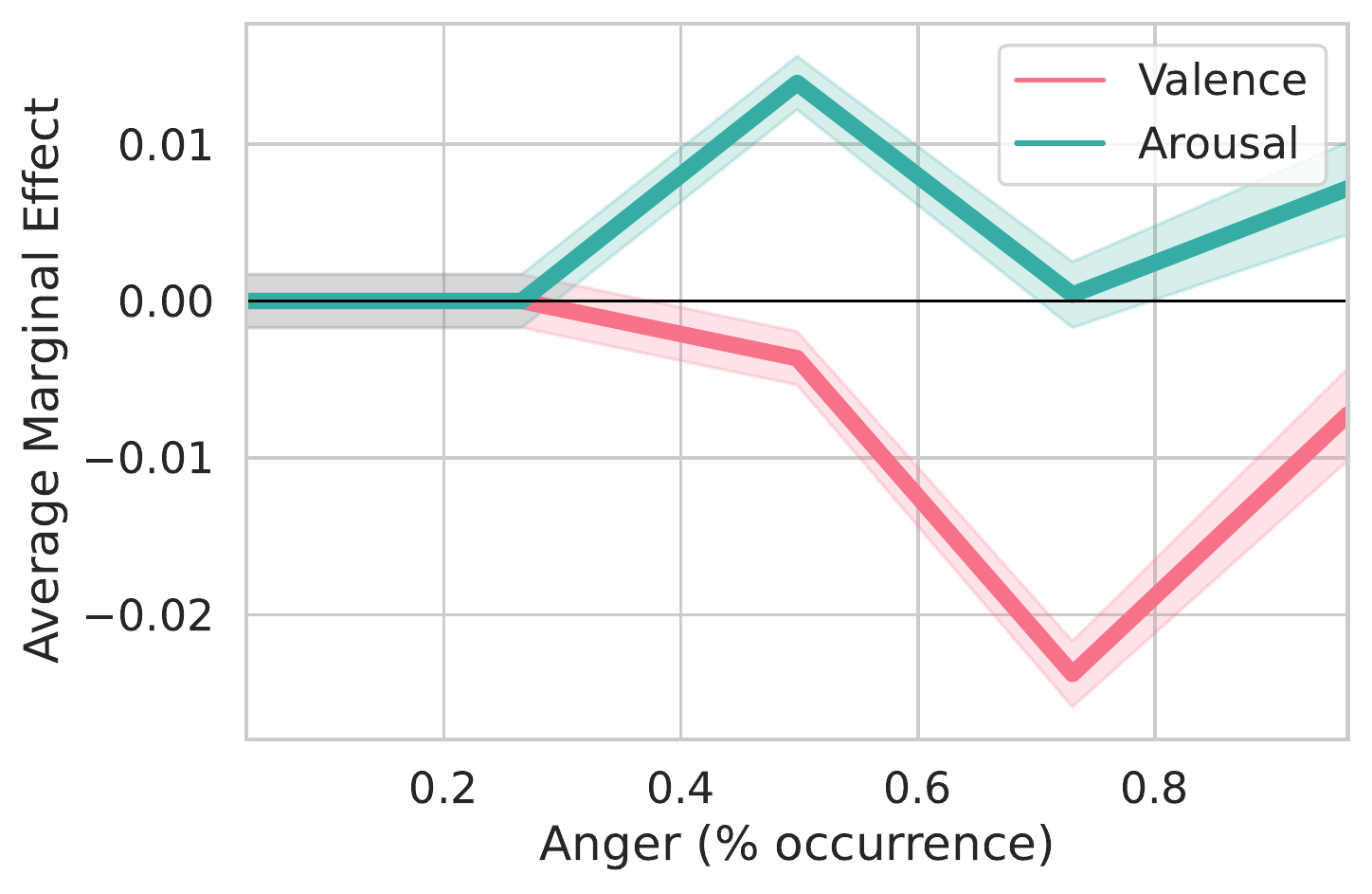}} & 
    {\includegraphics[width=0.30\textwidth]{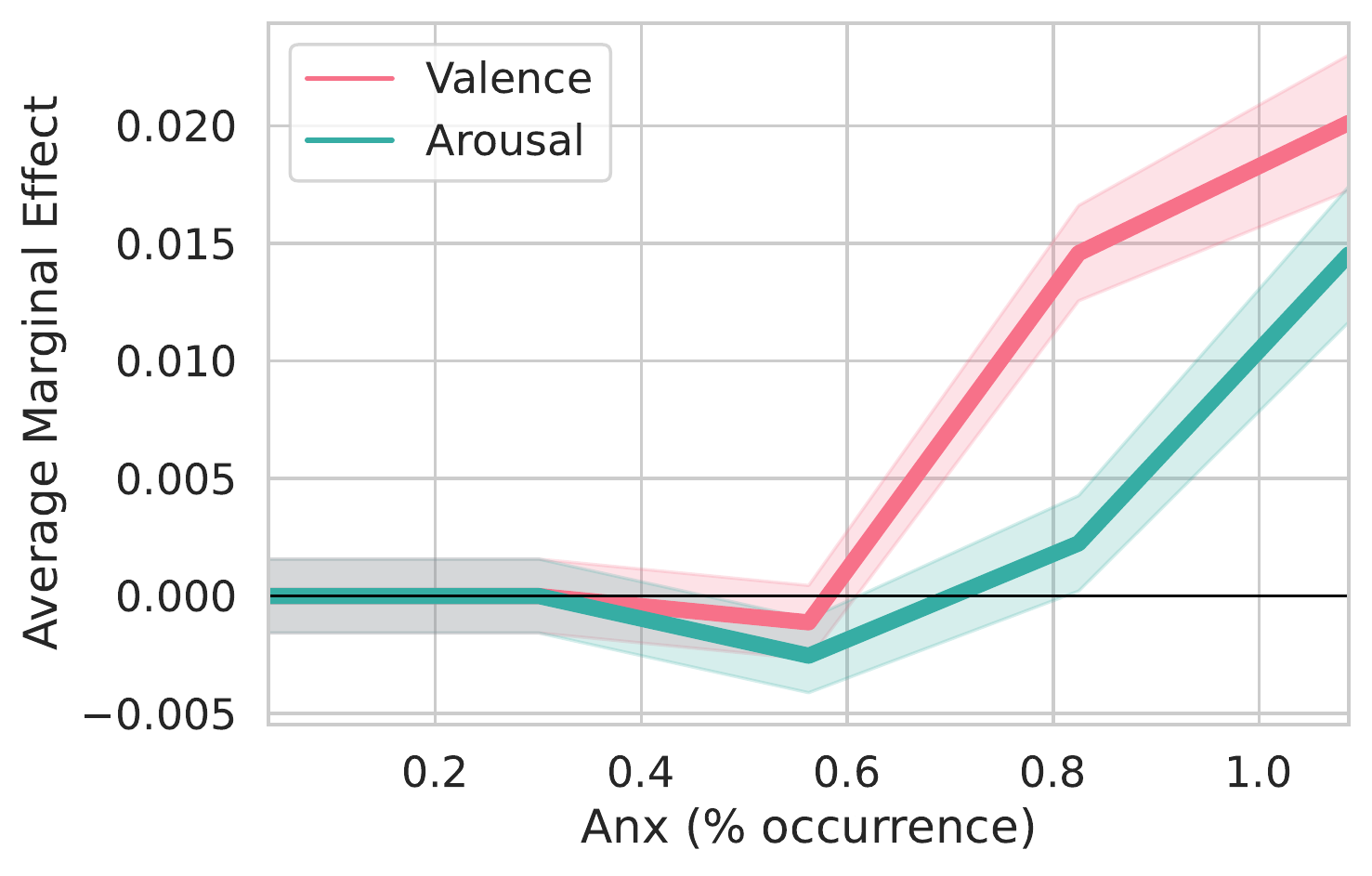}} \\ 
    {\includegraphics[width=0.30\textwidth]{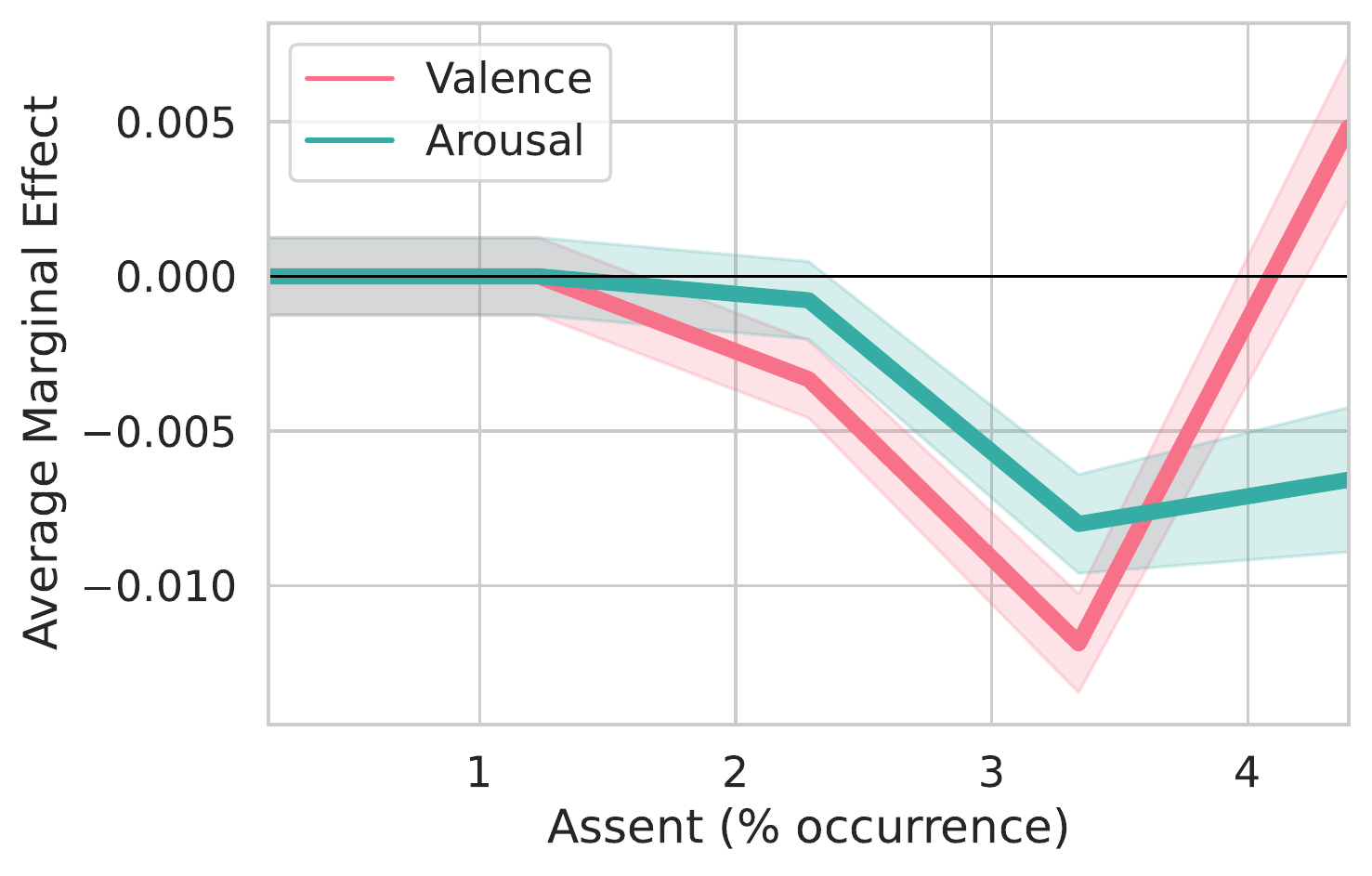}} & 
    {\includegraphics[width=0.30\textwidth]{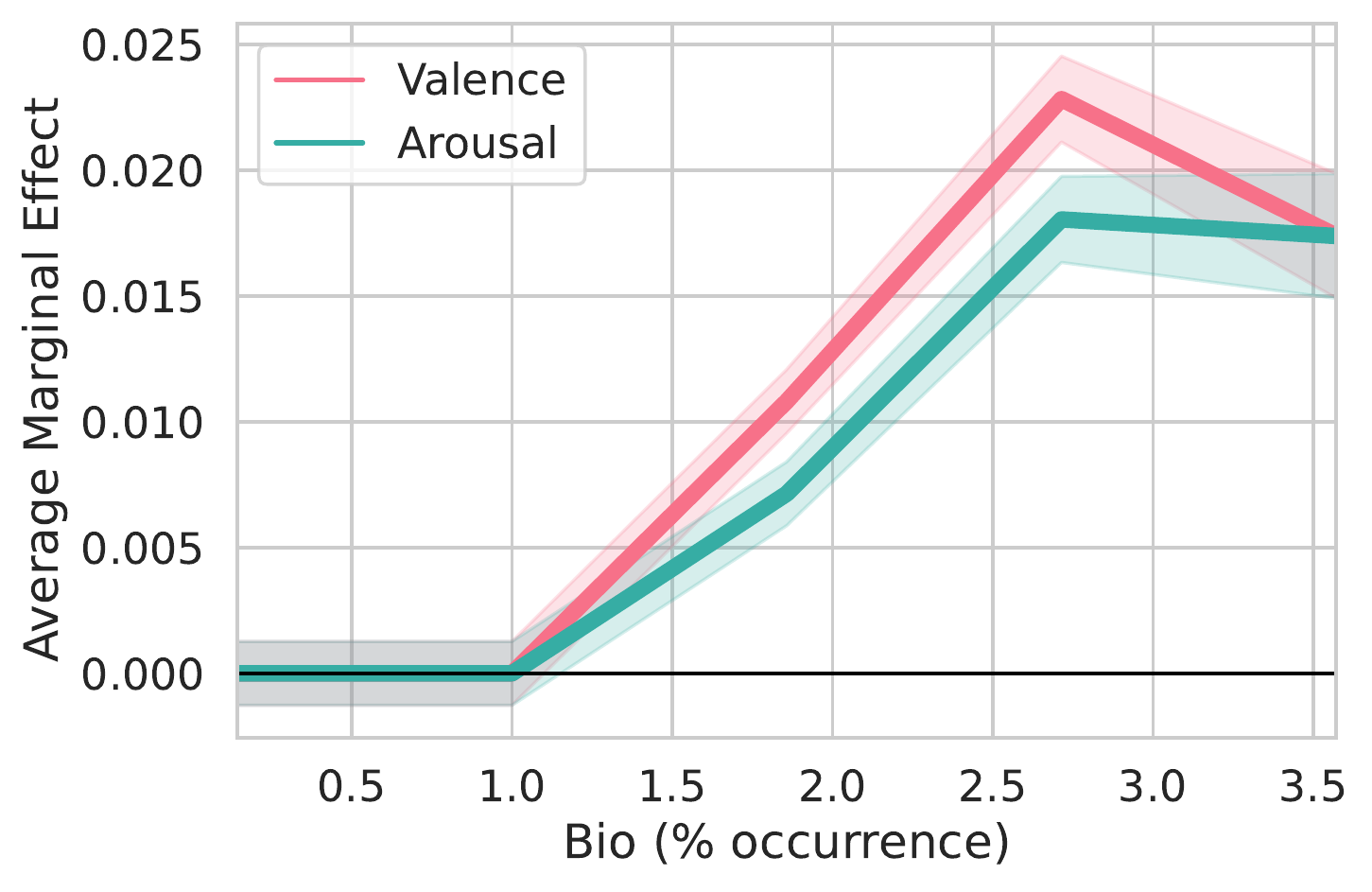}} & 
    {\includegraphics[width=0.30\textwidth]{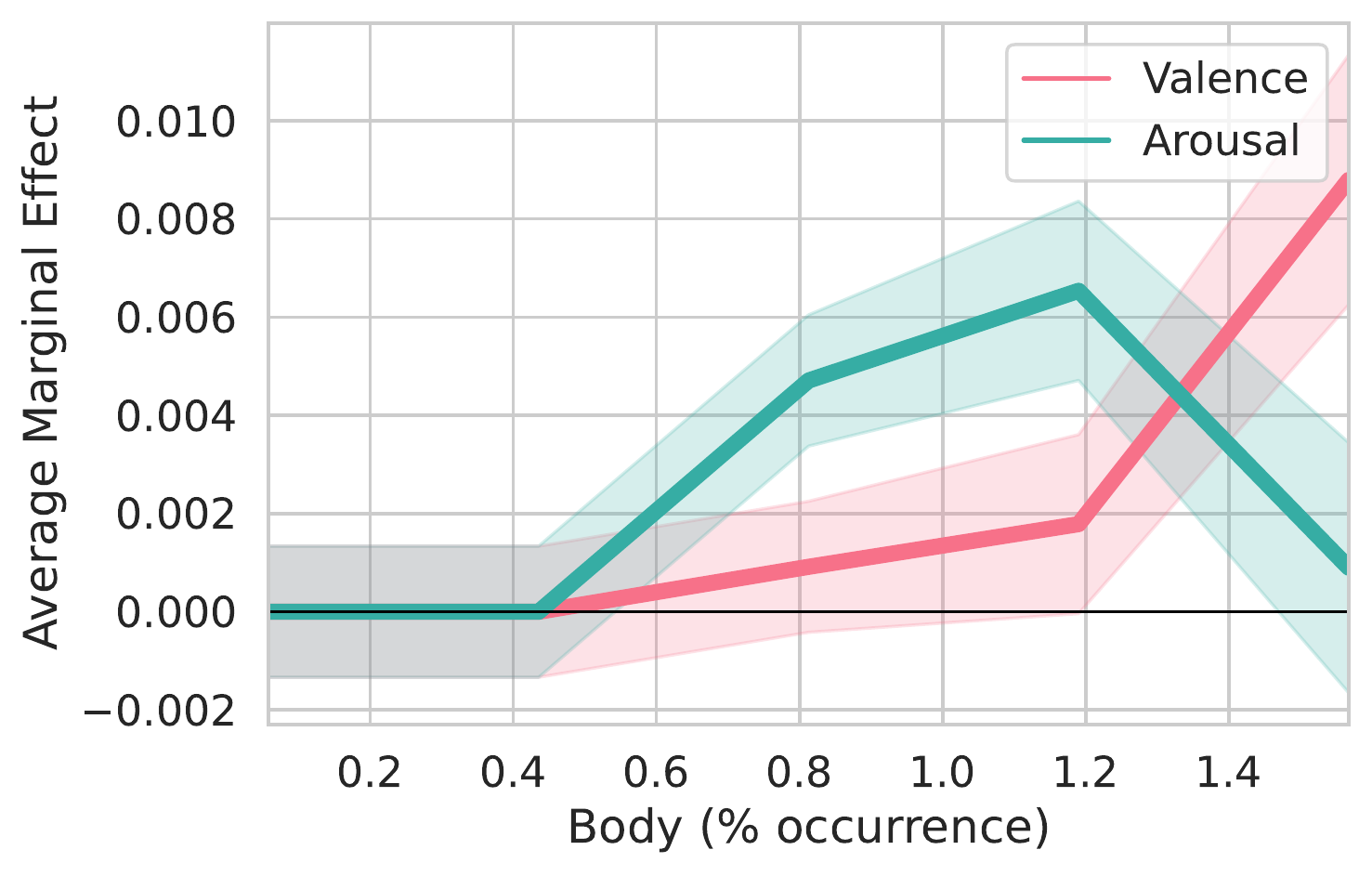}} \\ 
    {\includegraphics[width=0.30\textwidth]{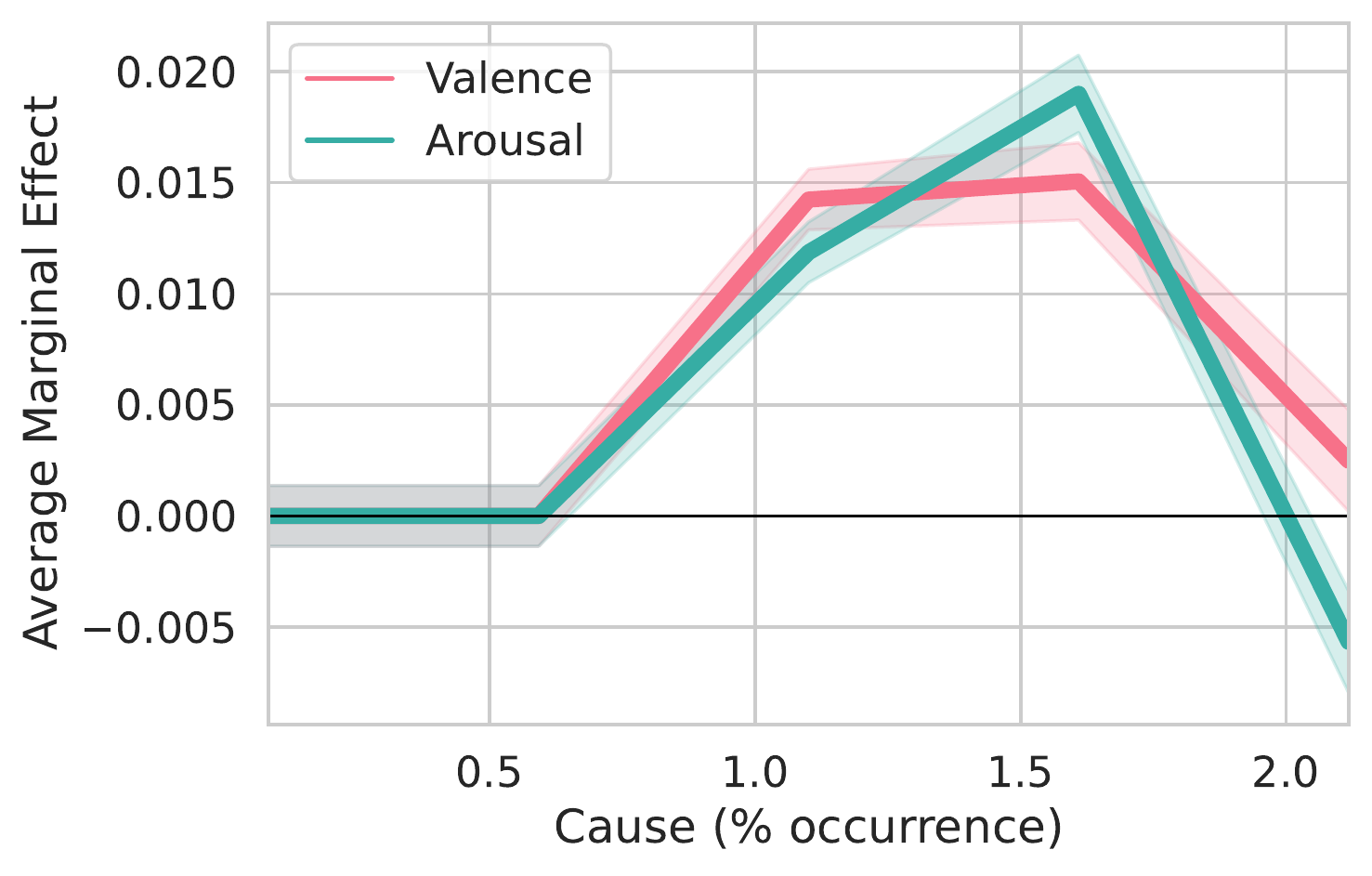}} & 
    {\includegraphics[width=0.30\textwidth]{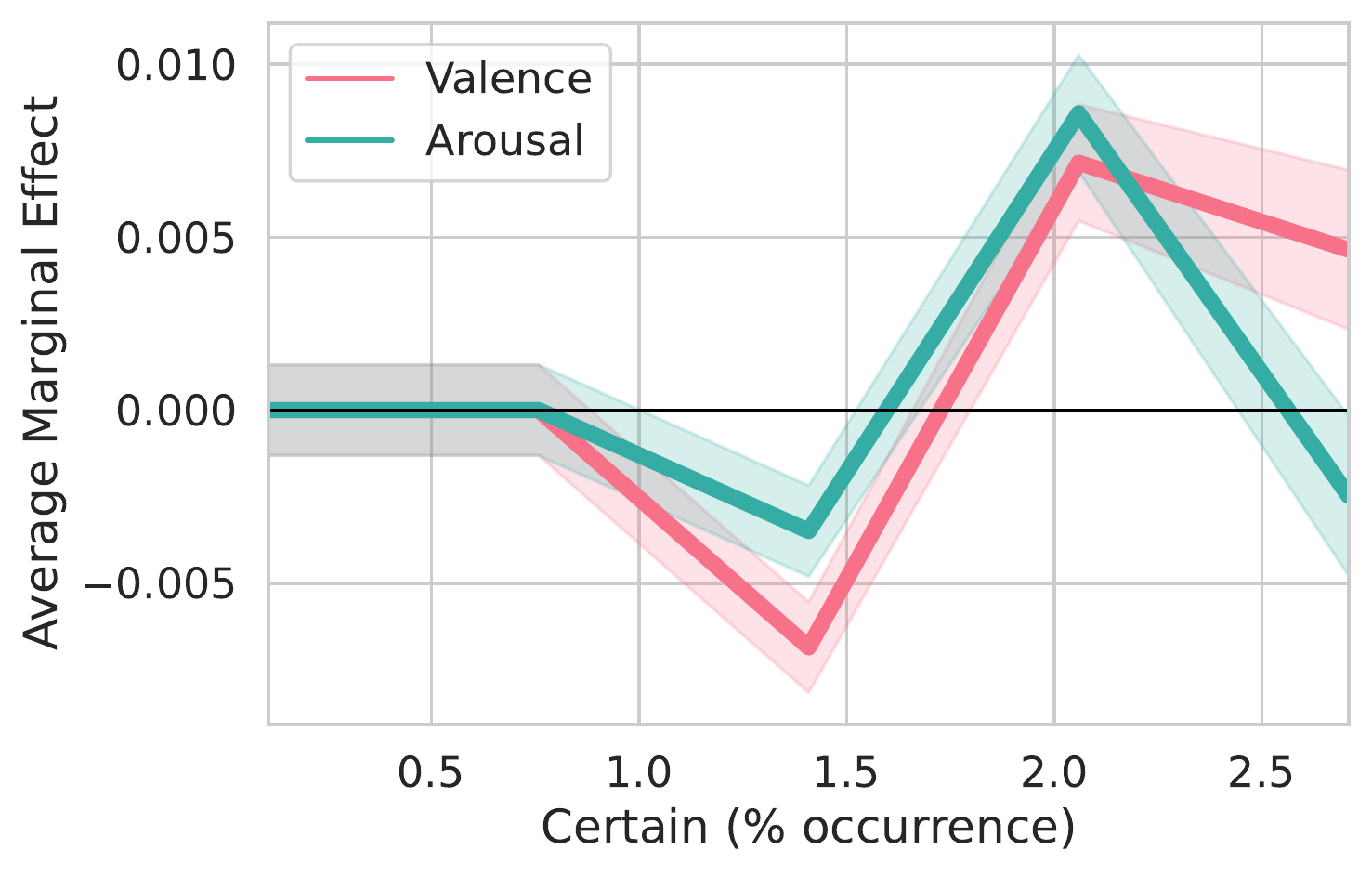}} & 
    {\includegraphics[width=0.30\textwidth]{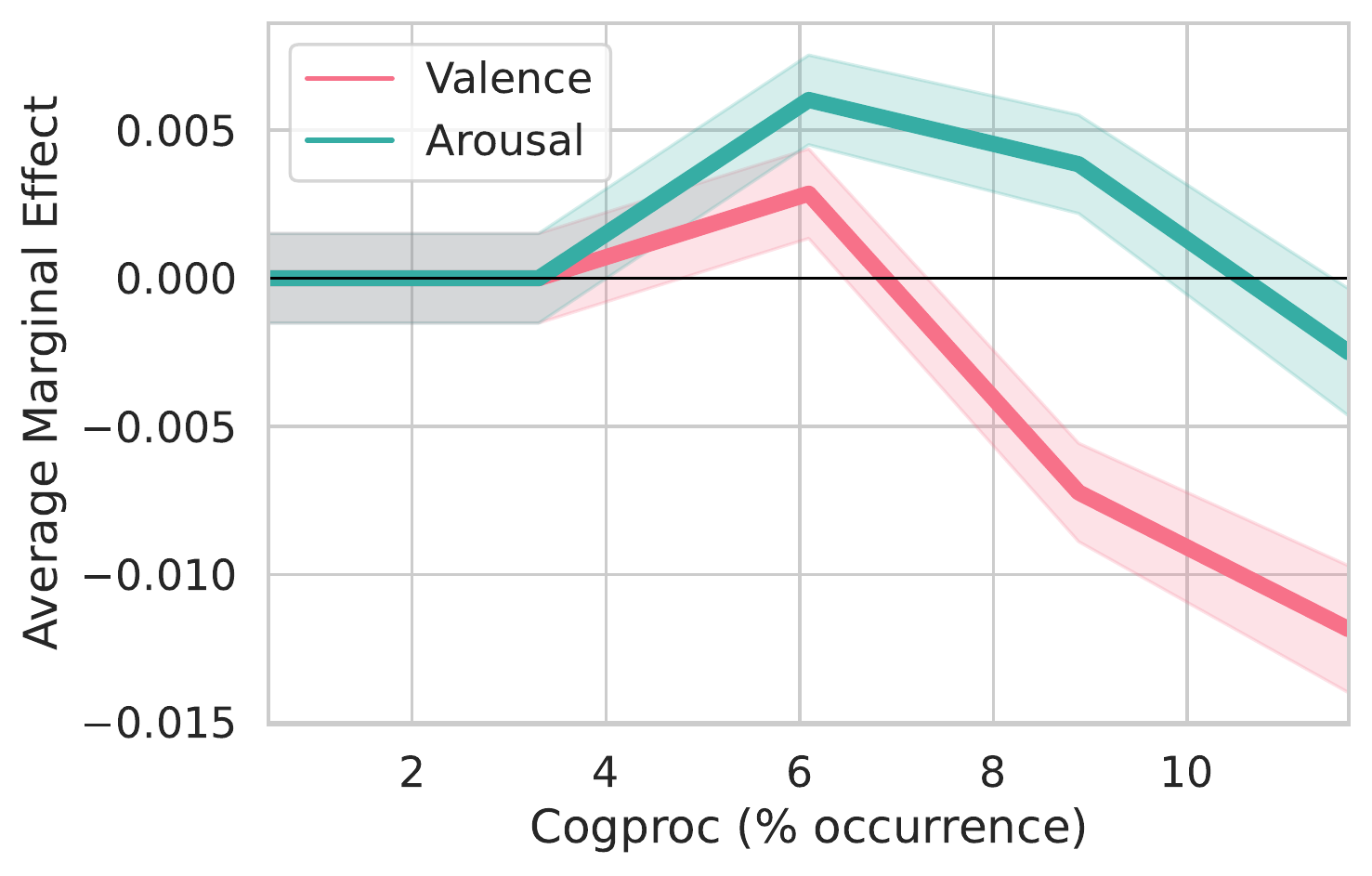}} \\ 
    {\includegraphics[width=0.30\textwidth]{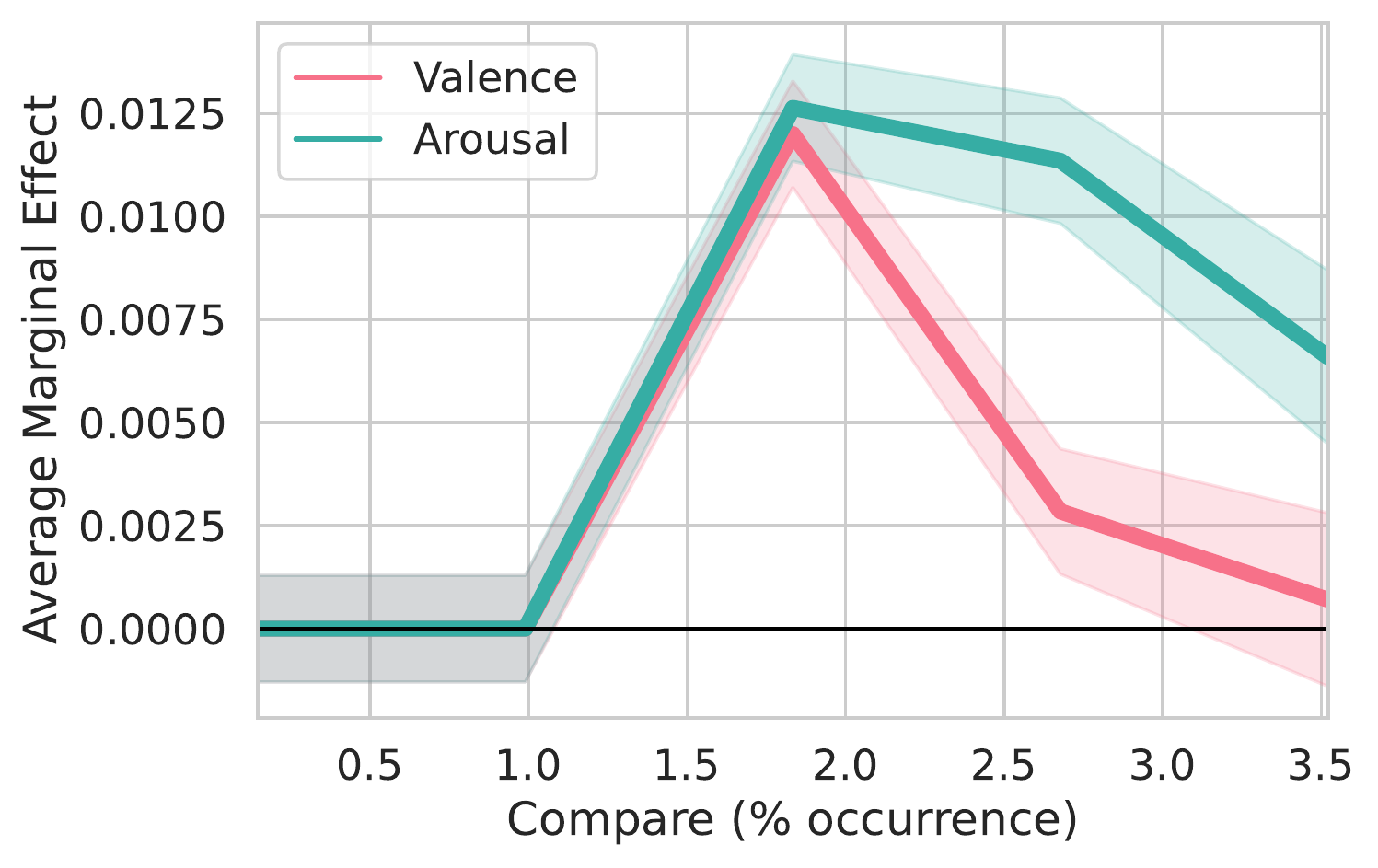}} & 
    {\includegraphics[width=0.30\textwidth]{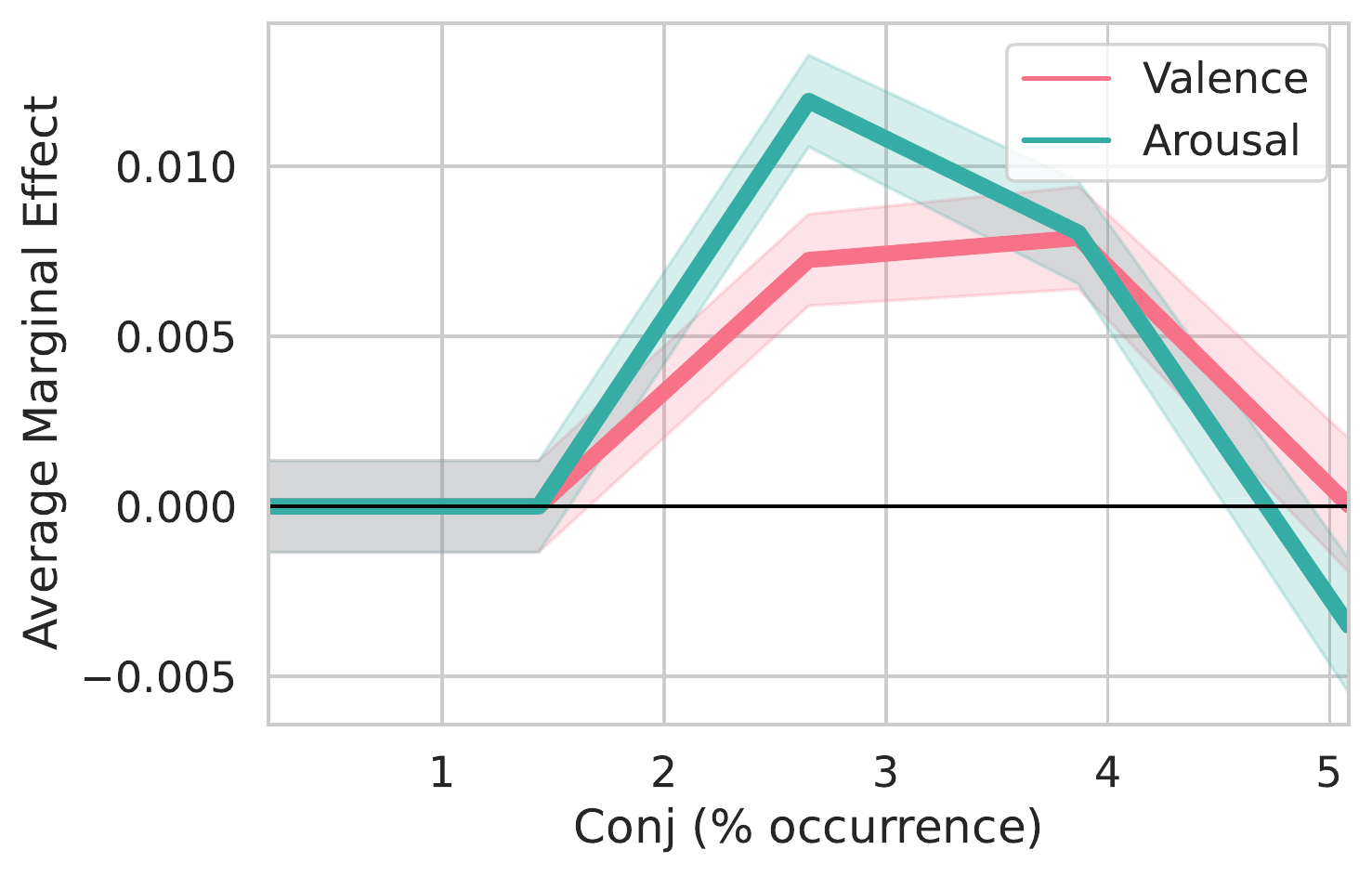}} & 
    {\includegraphics[width=0.30\textwidth]{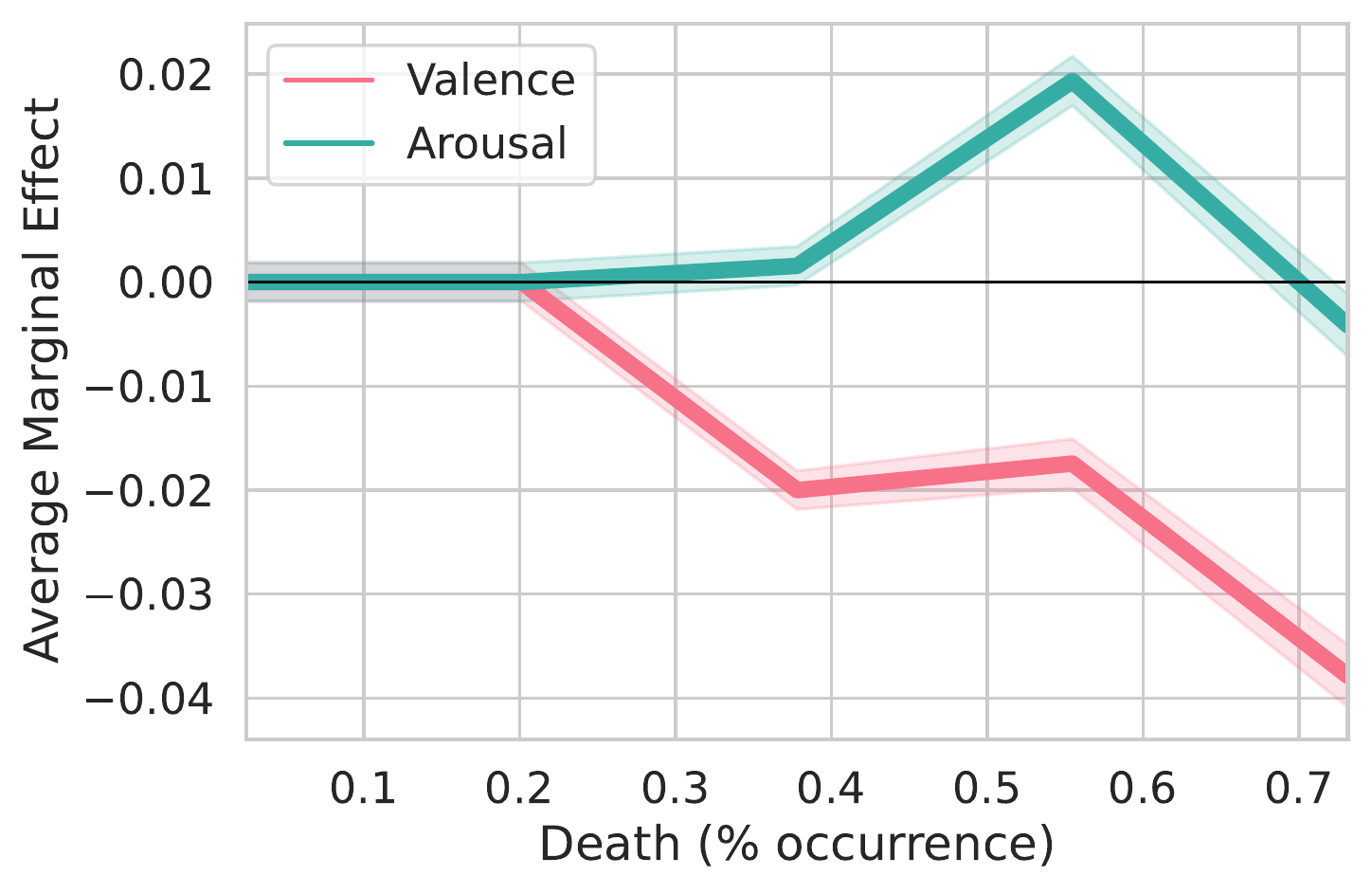}} \\ 
    {\includegraphics[width=0.30\textwidth]{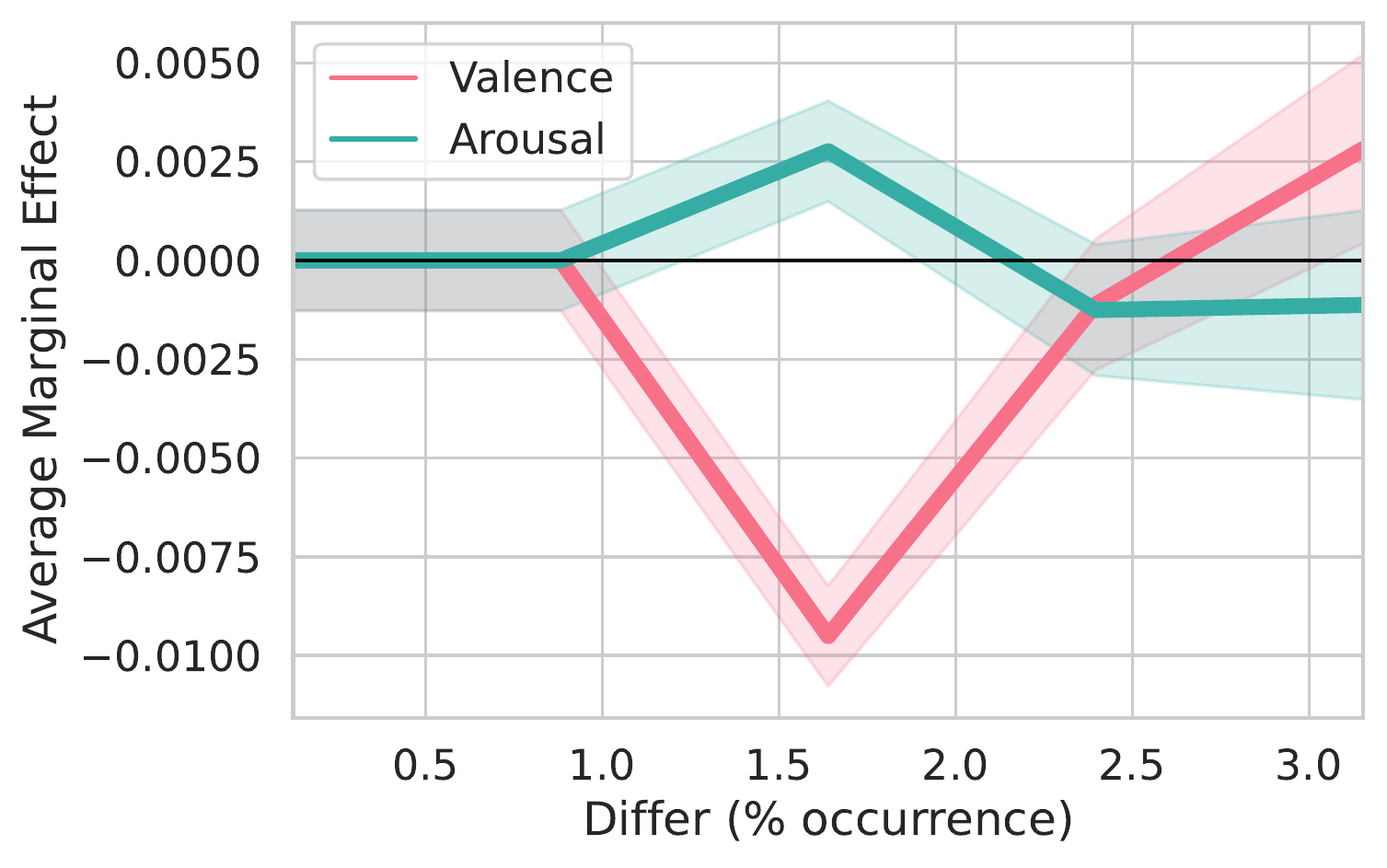}} & 
    {\includegraphics[width=0.30\textwidth]{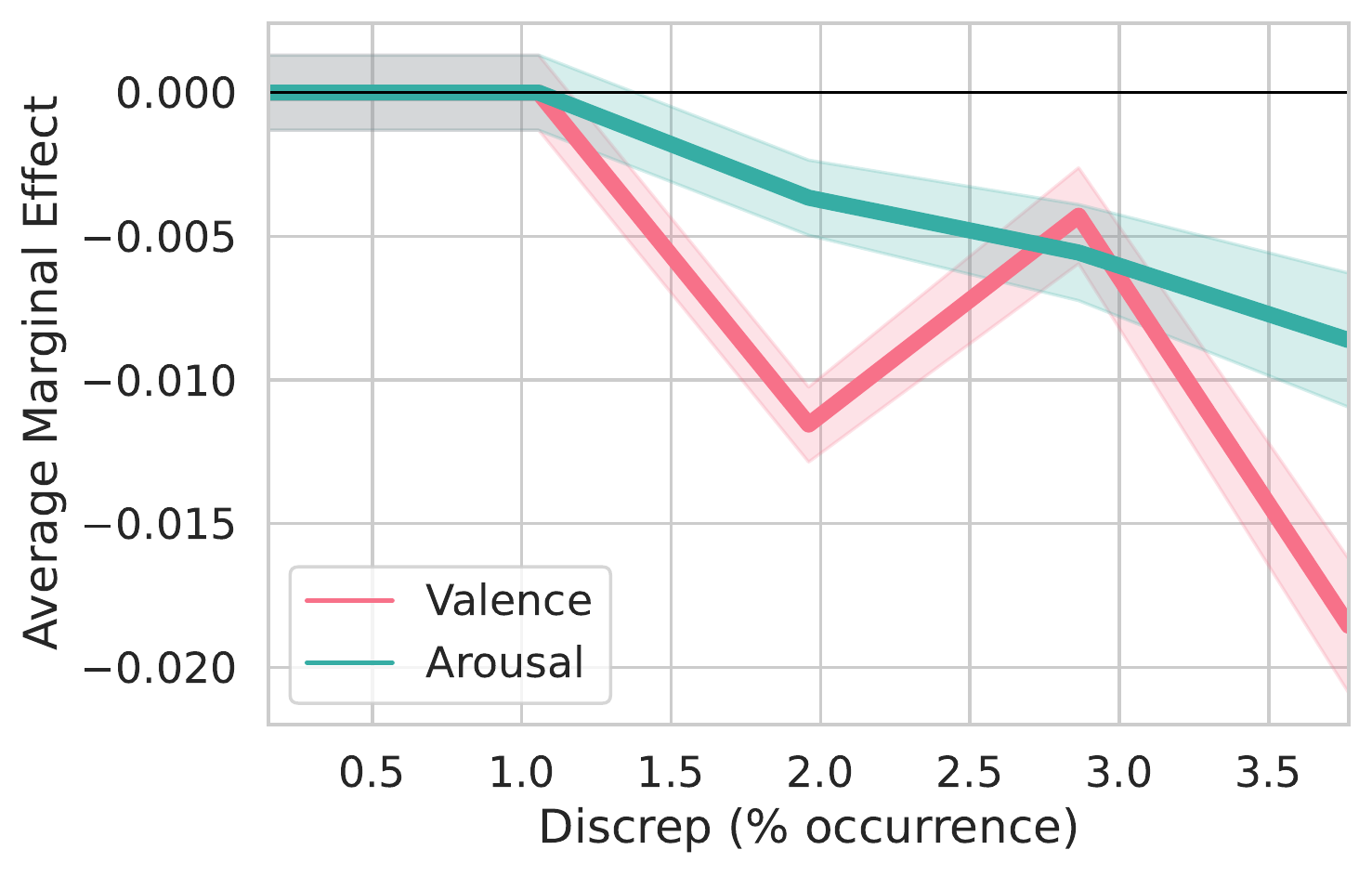}} & 
    {\includegraphics[width=0.30\textwidth]{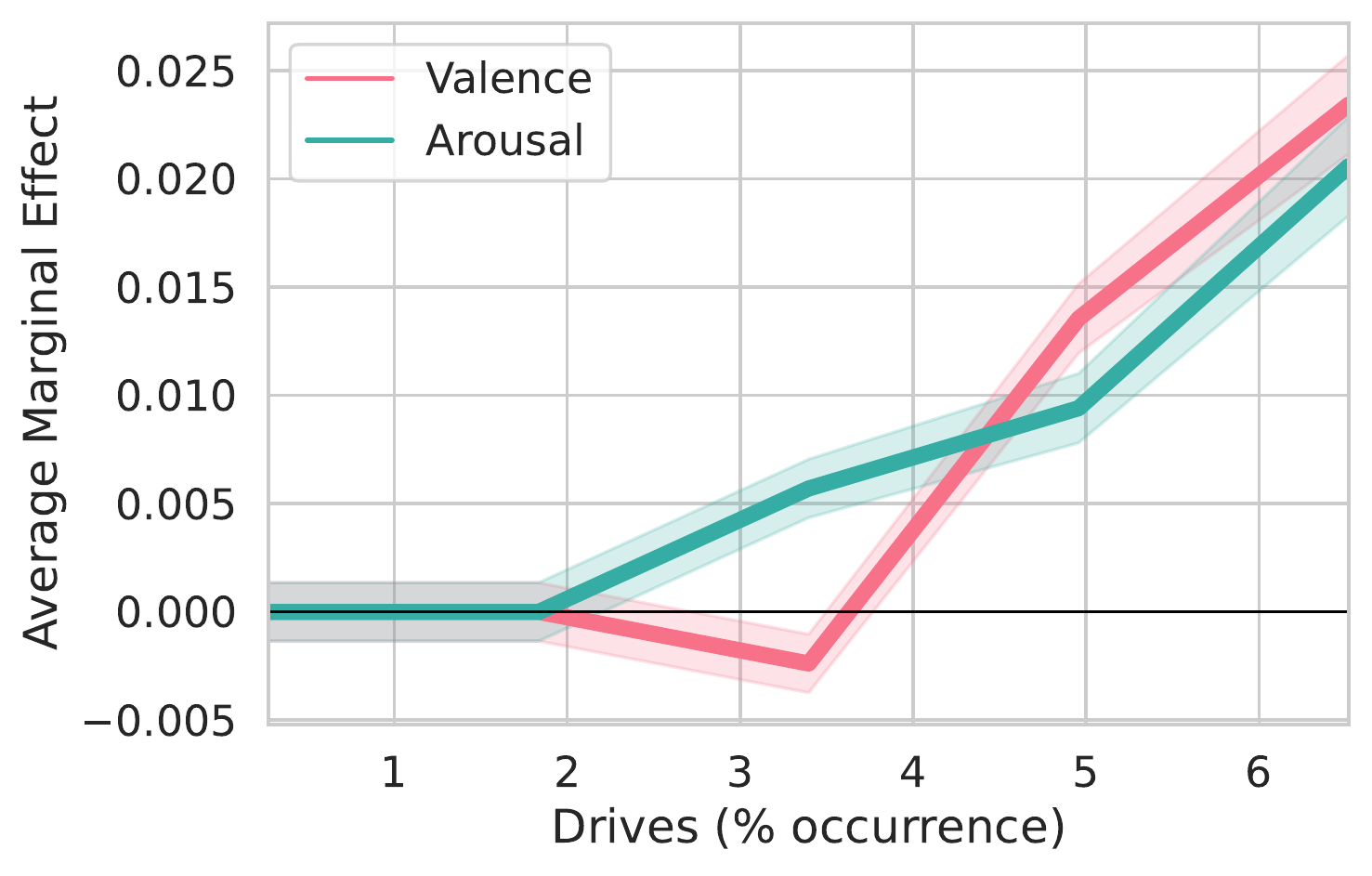}} \\ 
    {\includegraphics[width=0.30\textwidth]{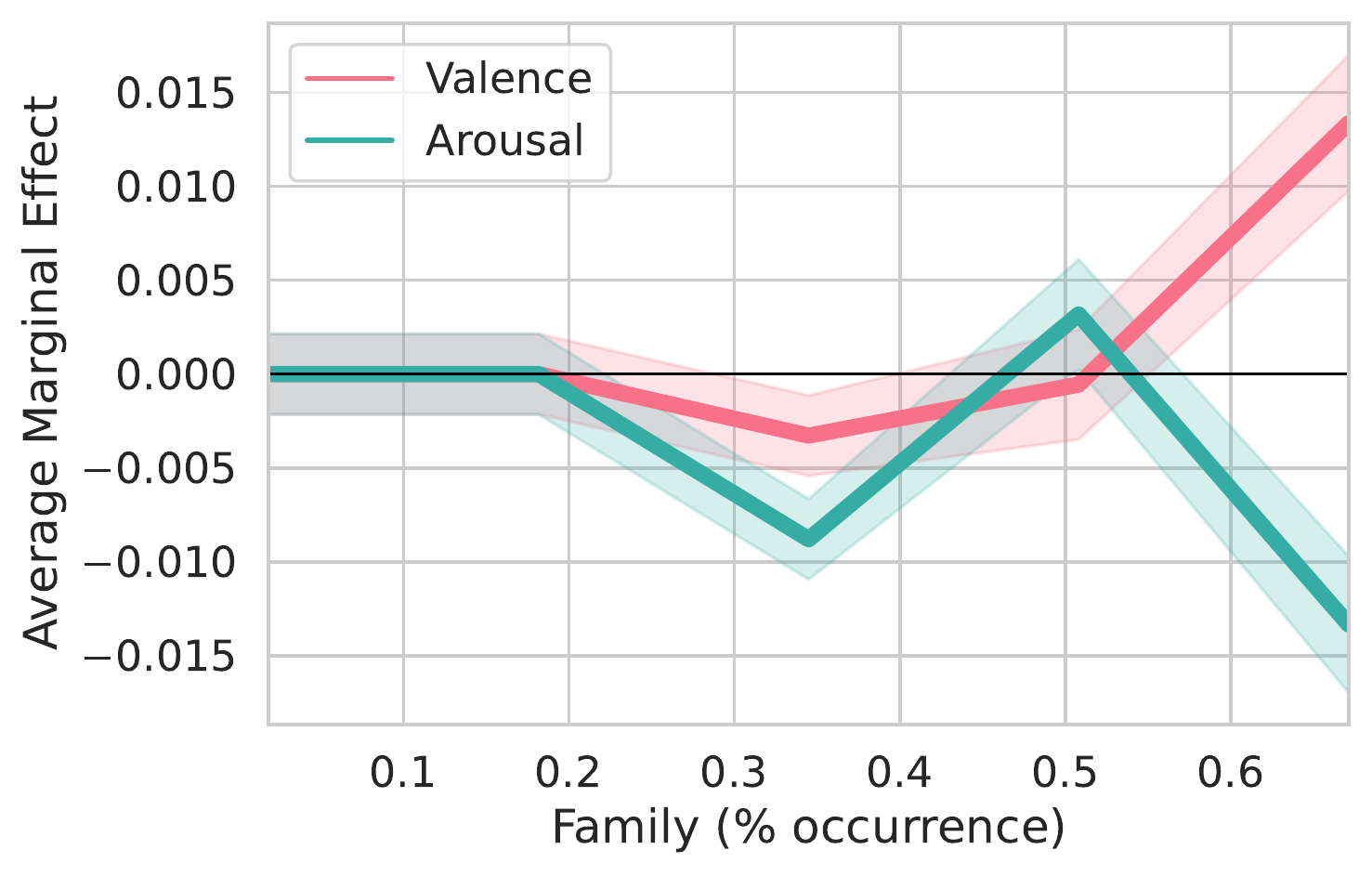}} & 
    {\includegraphics[width=0.30\textwidth]{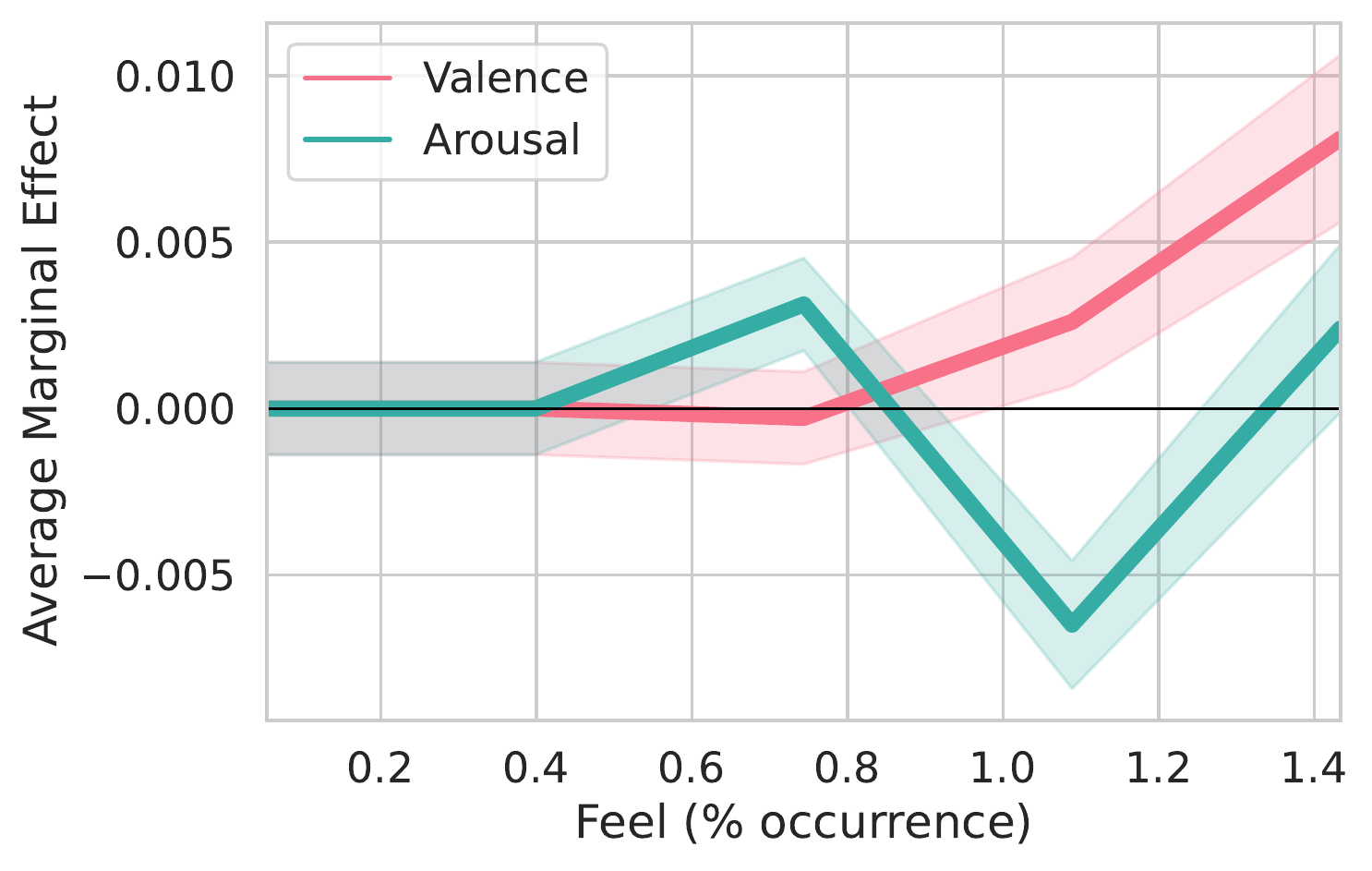}} & 
    {\includegraphics[width=0.30\textwidth]{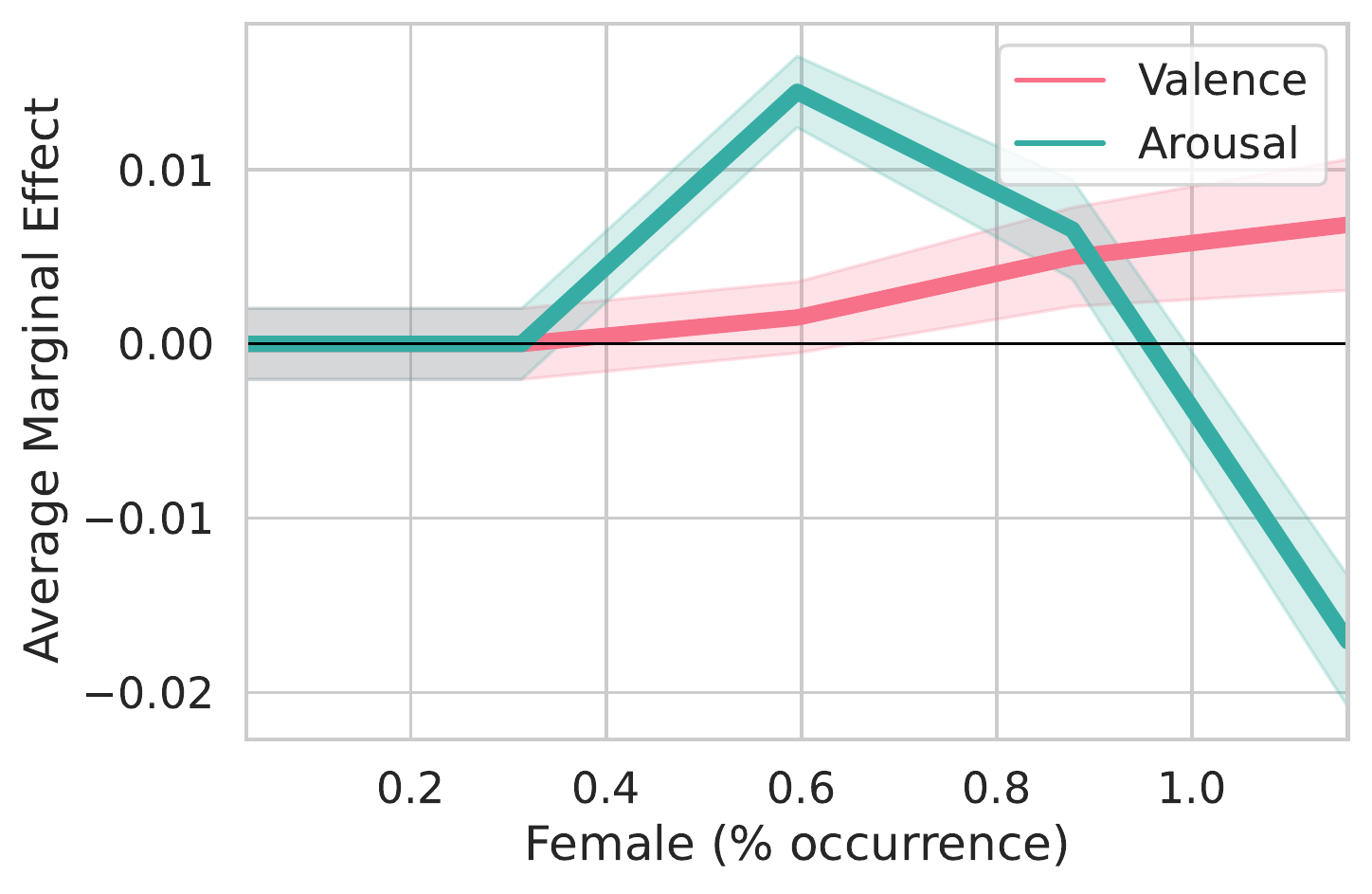}} \\ 
    \end{tabular}
    \caption{
    Average marginal effects of LIWC psycholinguistic lexical category \textbf{lyrical features} on listener affective responses, controlling for musical features and listener demographics. With the intent to reduce noise at the extremities, x-axis limits are capped at their 95\% quantile values.
    Arranged in alphabetical order, standard errors are shown;
    \textcolor{red}{valence} in \textcolor{red}{red}, \textcolor{blue}{arousal} in \textcolor{blue}{blue} (Part 1/4).
    }
    \label{fig:lyricfeatures_expanded_1}
\end{figure*}

\begin{figure*}[!t]
    \centering
    \begin{tabular}{ccc}
    {\includegraphics[width=0.30\textwidth]{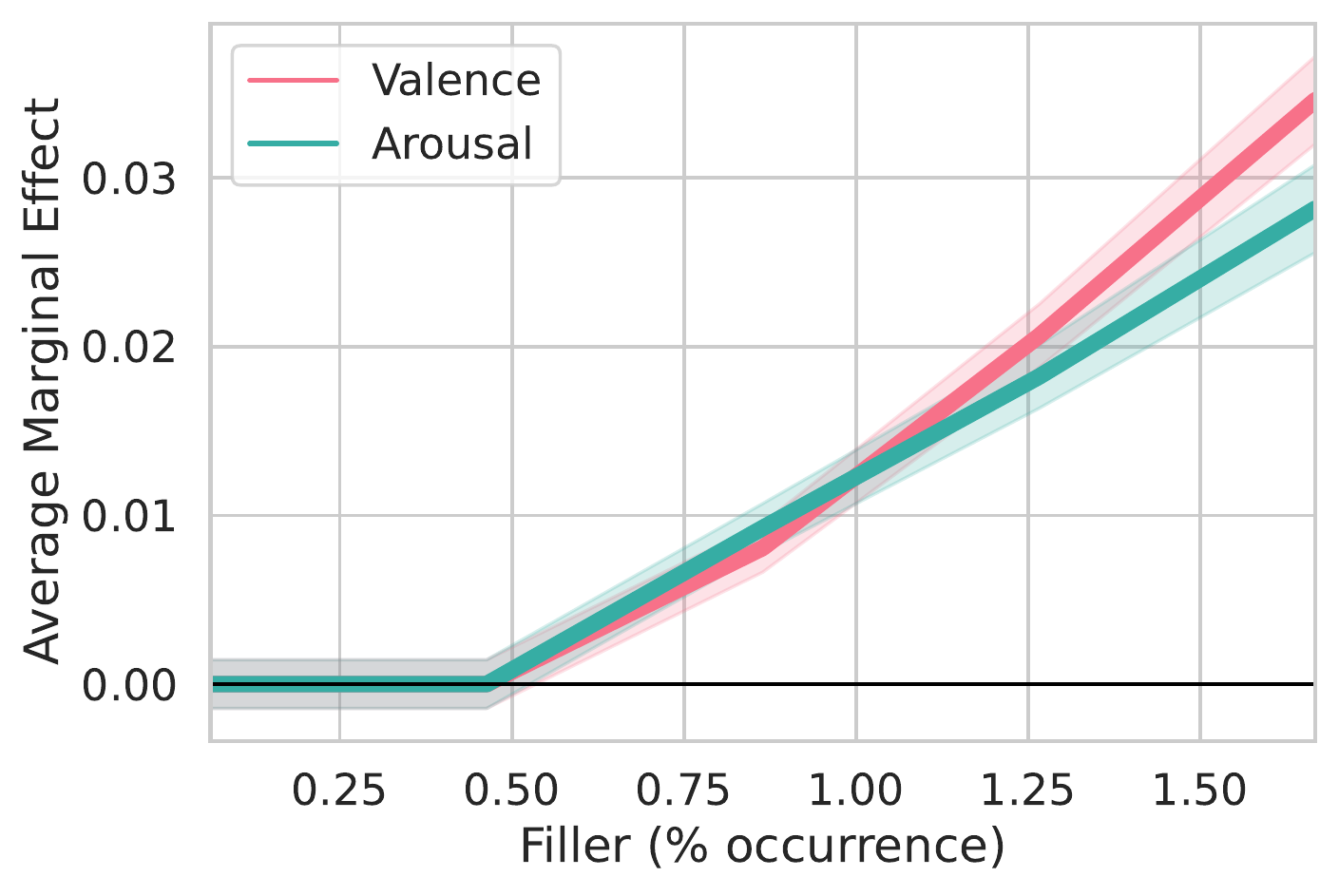}} &
    {\includegraphics[width=0.30\textwidth]{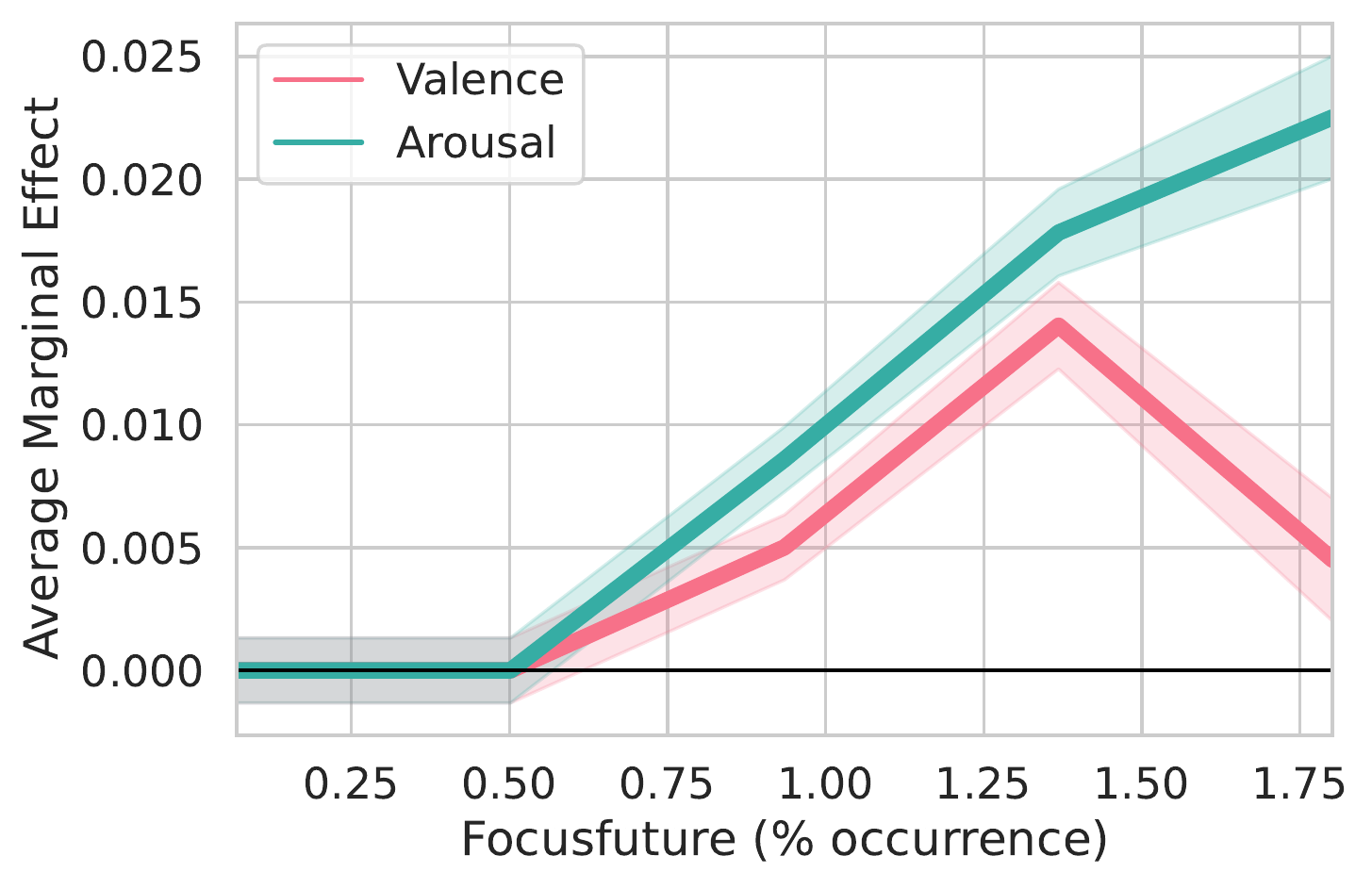}} & 
    {\includegraphics[width=0.30\textwidth]{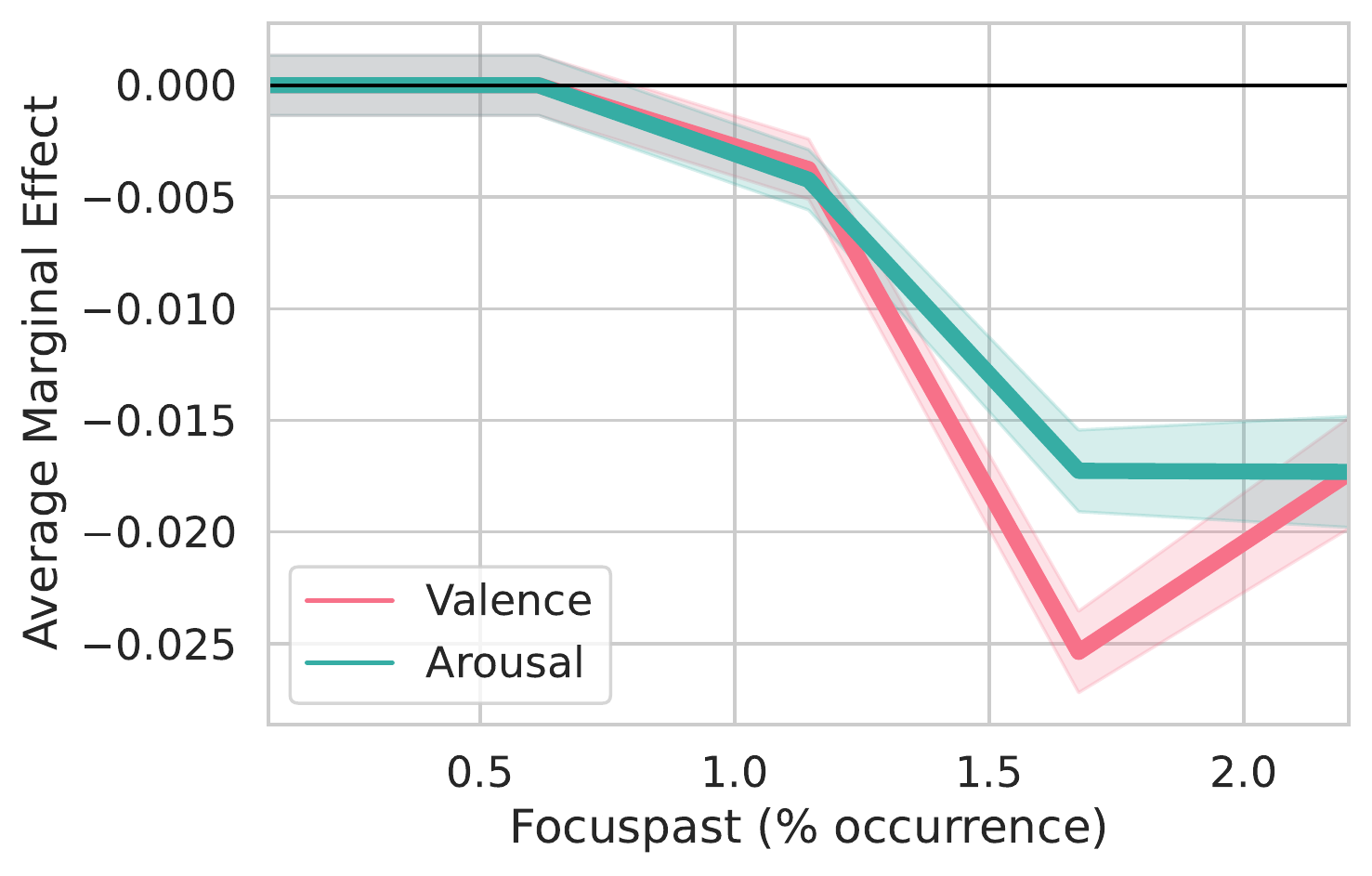}} \\
    {\includegraphics[width=0.30\textwidth]{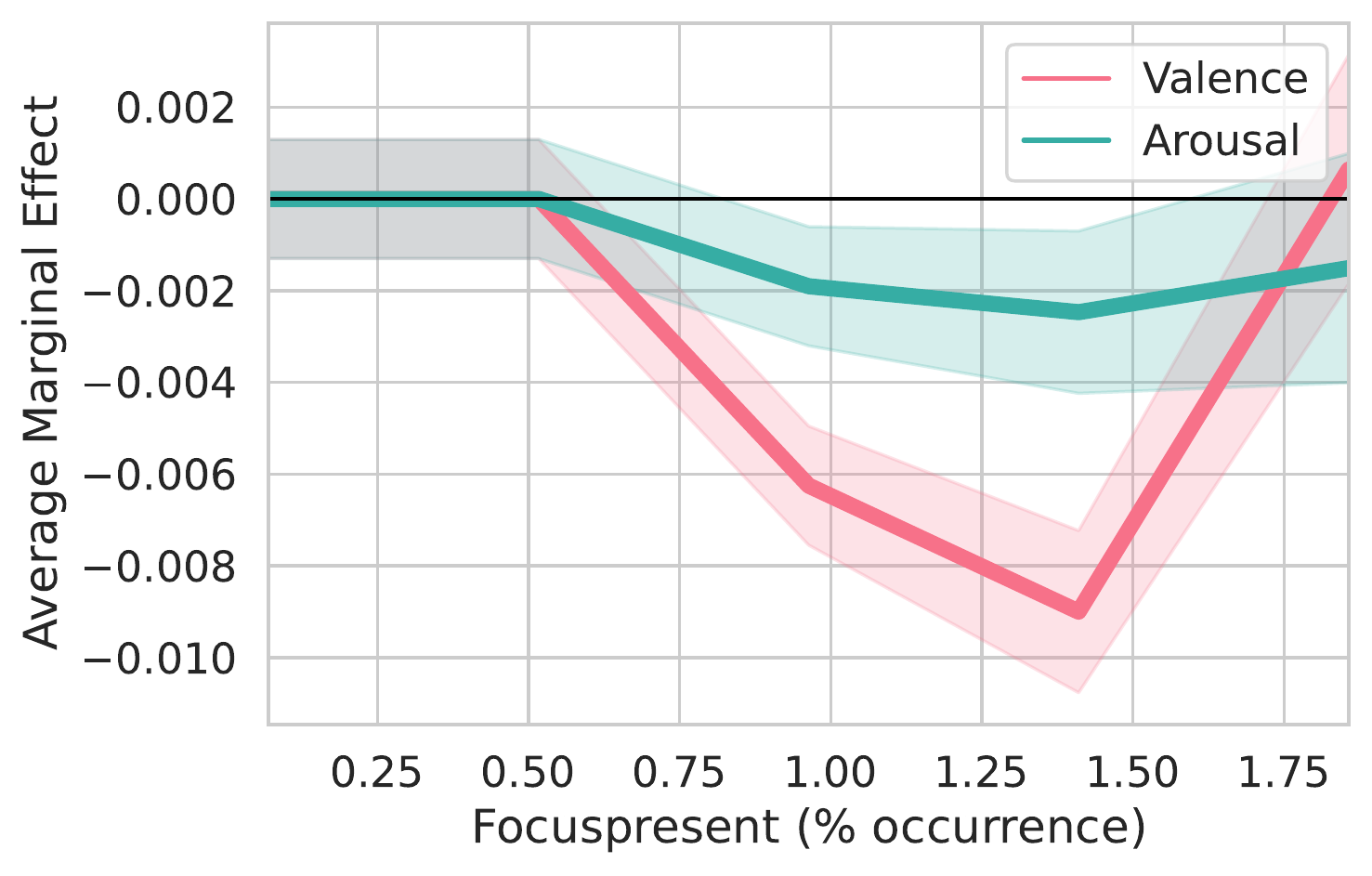}} & 
    {\includegraphics[width=0.30\textwidth]{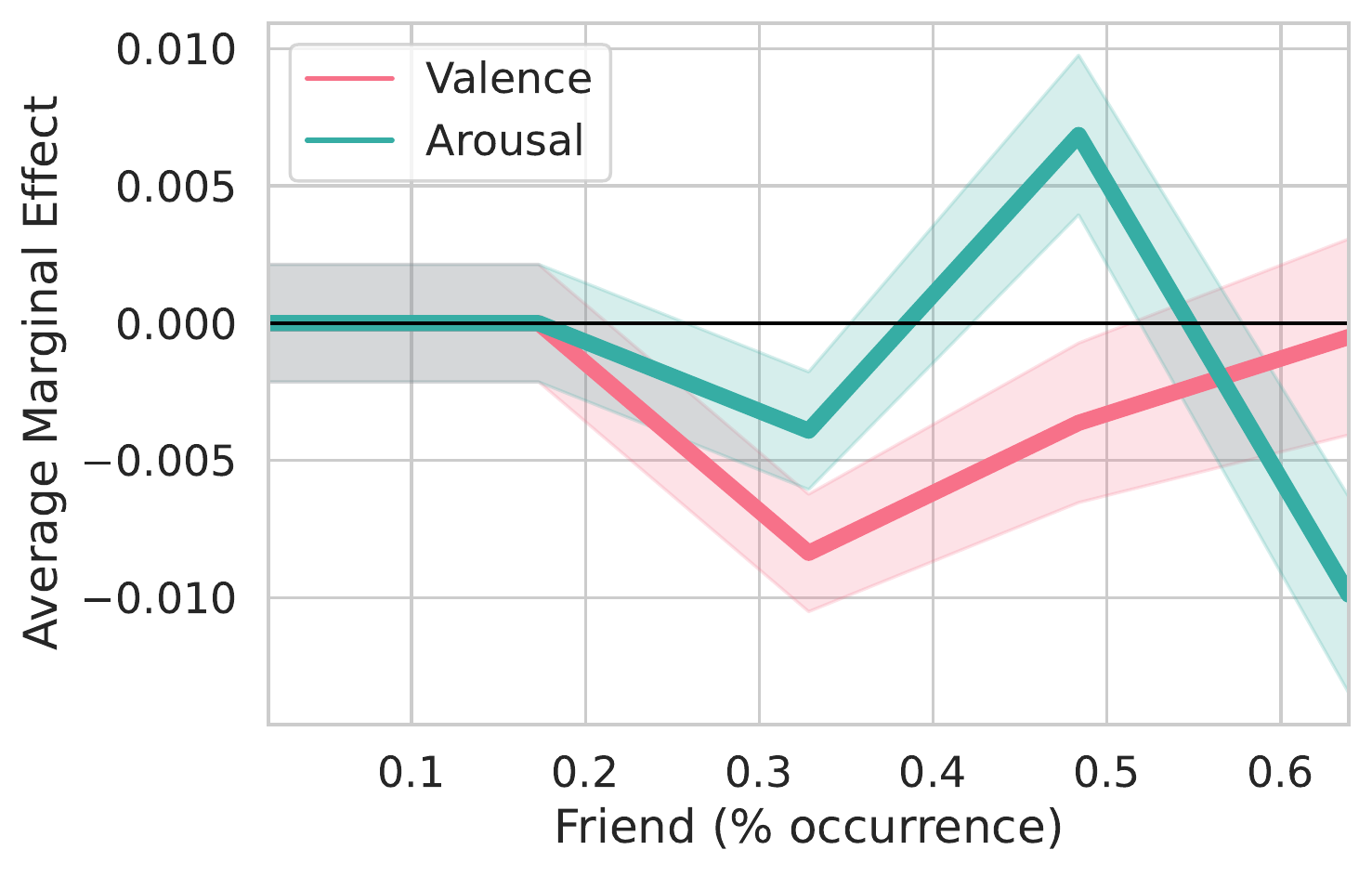}} & 
    {\includegraphics[width=0.30\textwidth]{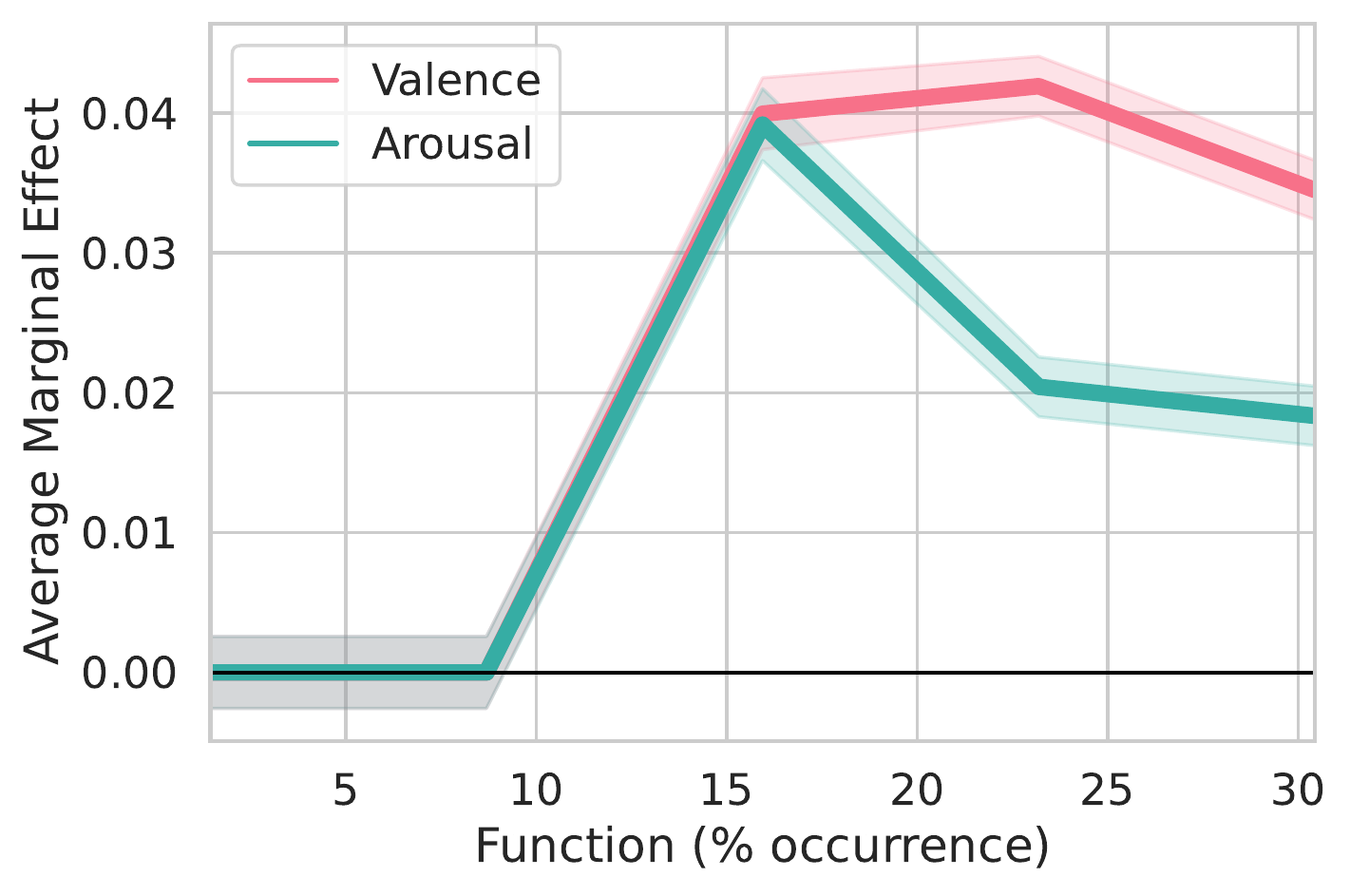}} \\ 
    {\includegraphics[width=0.30\textwidth]{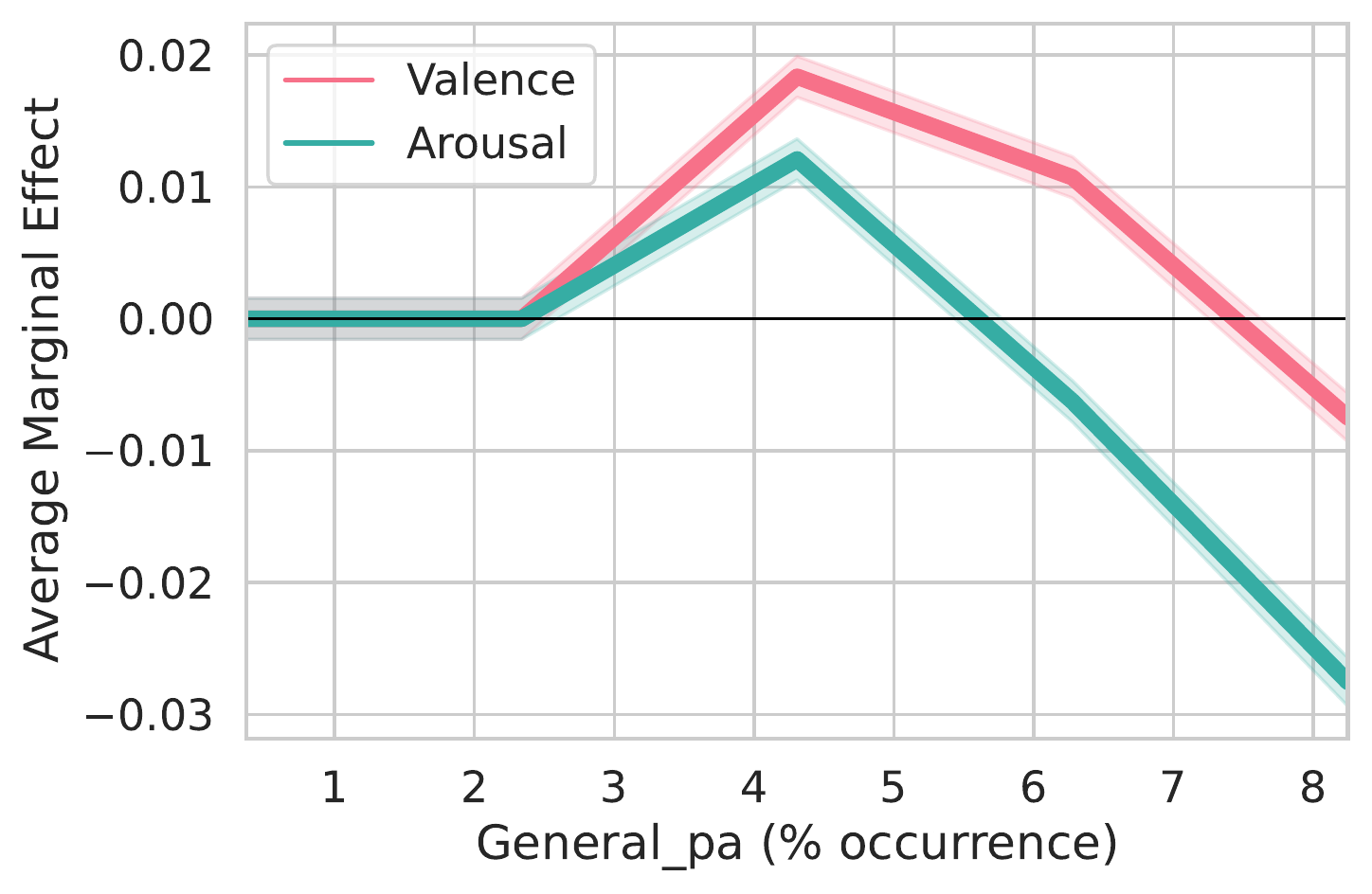}} & 
    {\includegraphics[width=0.30\textwidth]{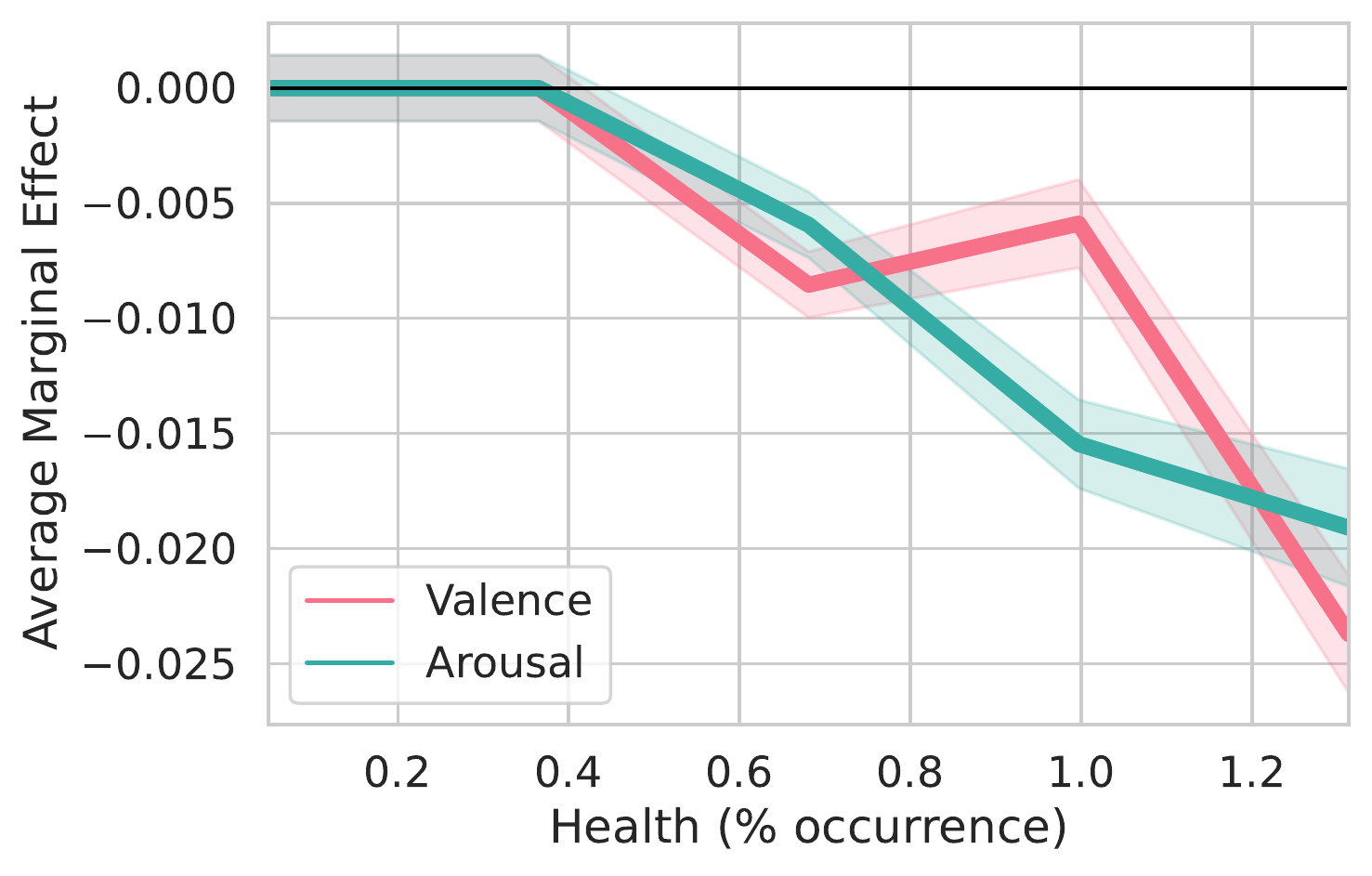}} & 
    {\includegraphics[width=0.30\textwidth]{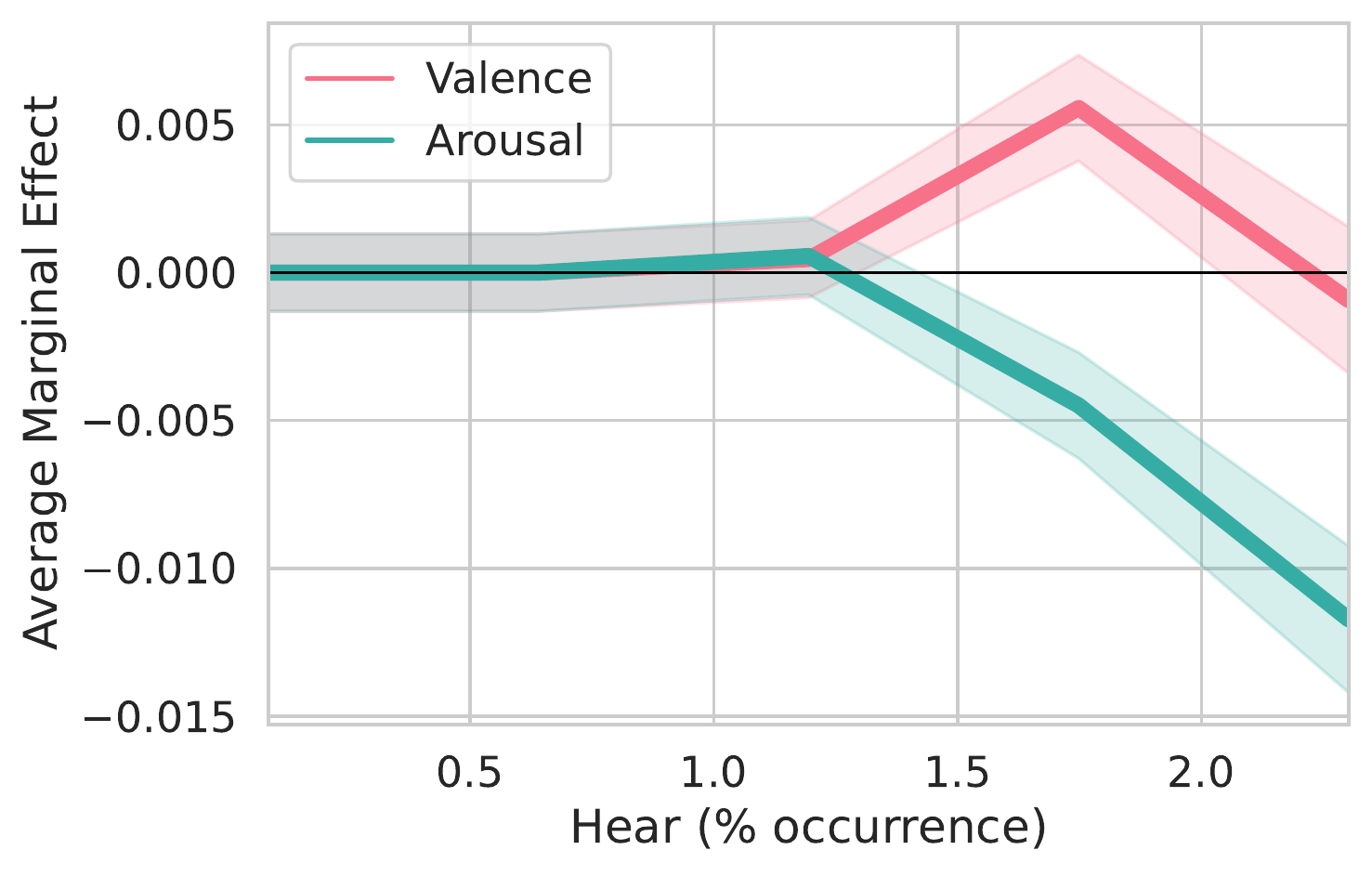}} \\ 
    {\includegraphics[width=0.30\textwidth]{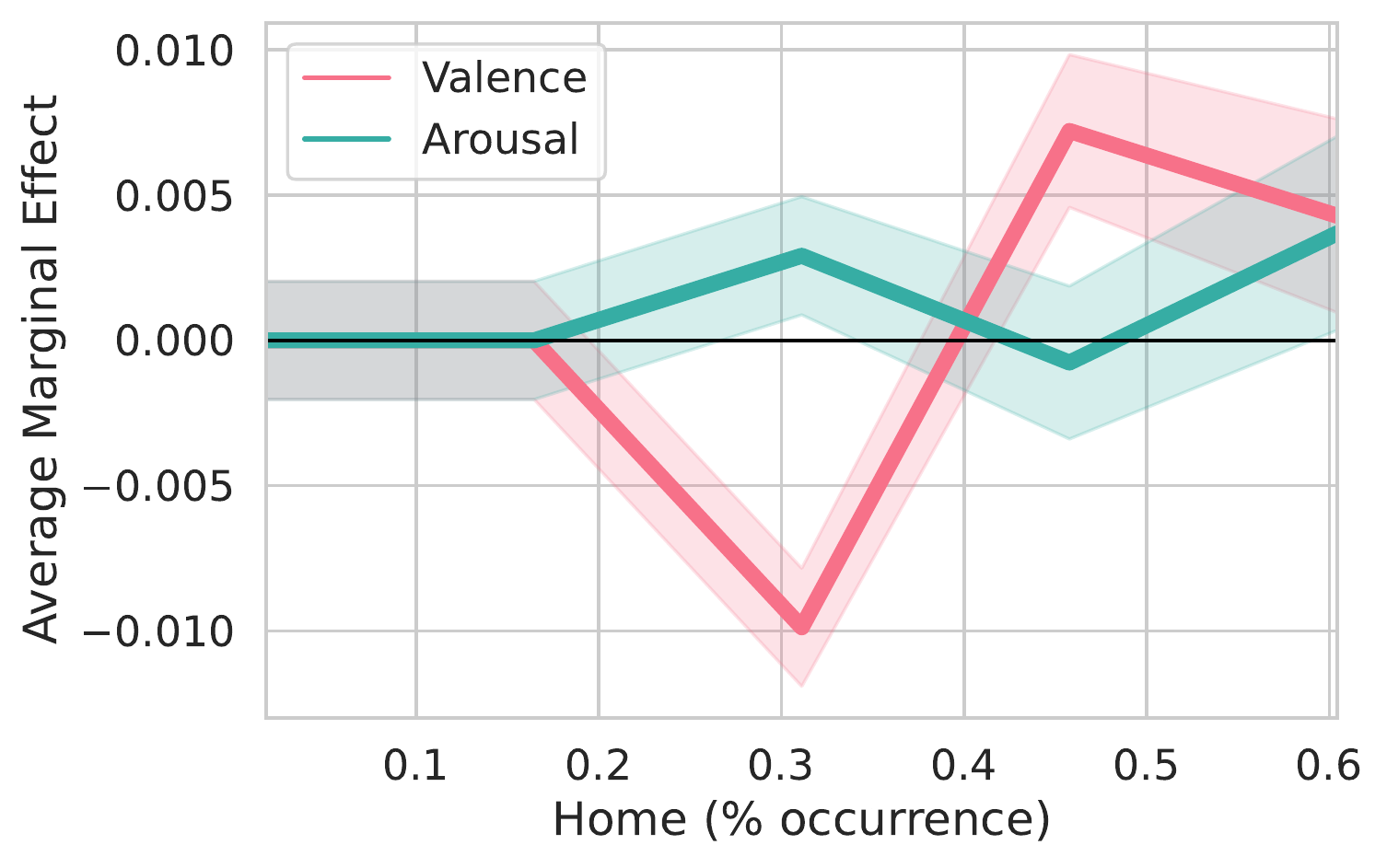}} & 
    {\includegraphics[width=0.30\textwidth]{plots/AME/lyric_i.pdf}} & 
    {\includegraphics[width=0.30\textwidth]{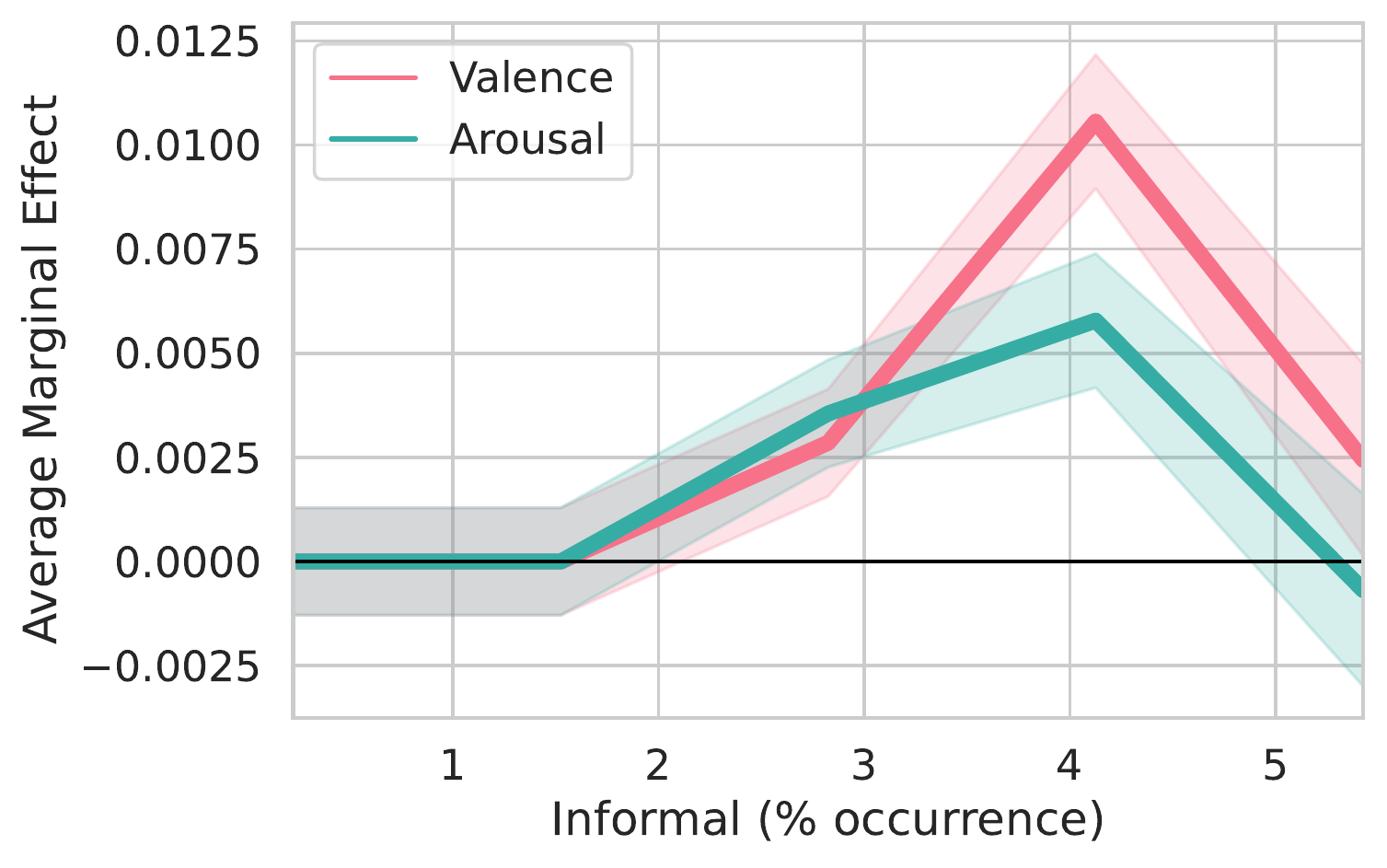}} \\ 
    {\includegraphics[width=0.30\textwidth]{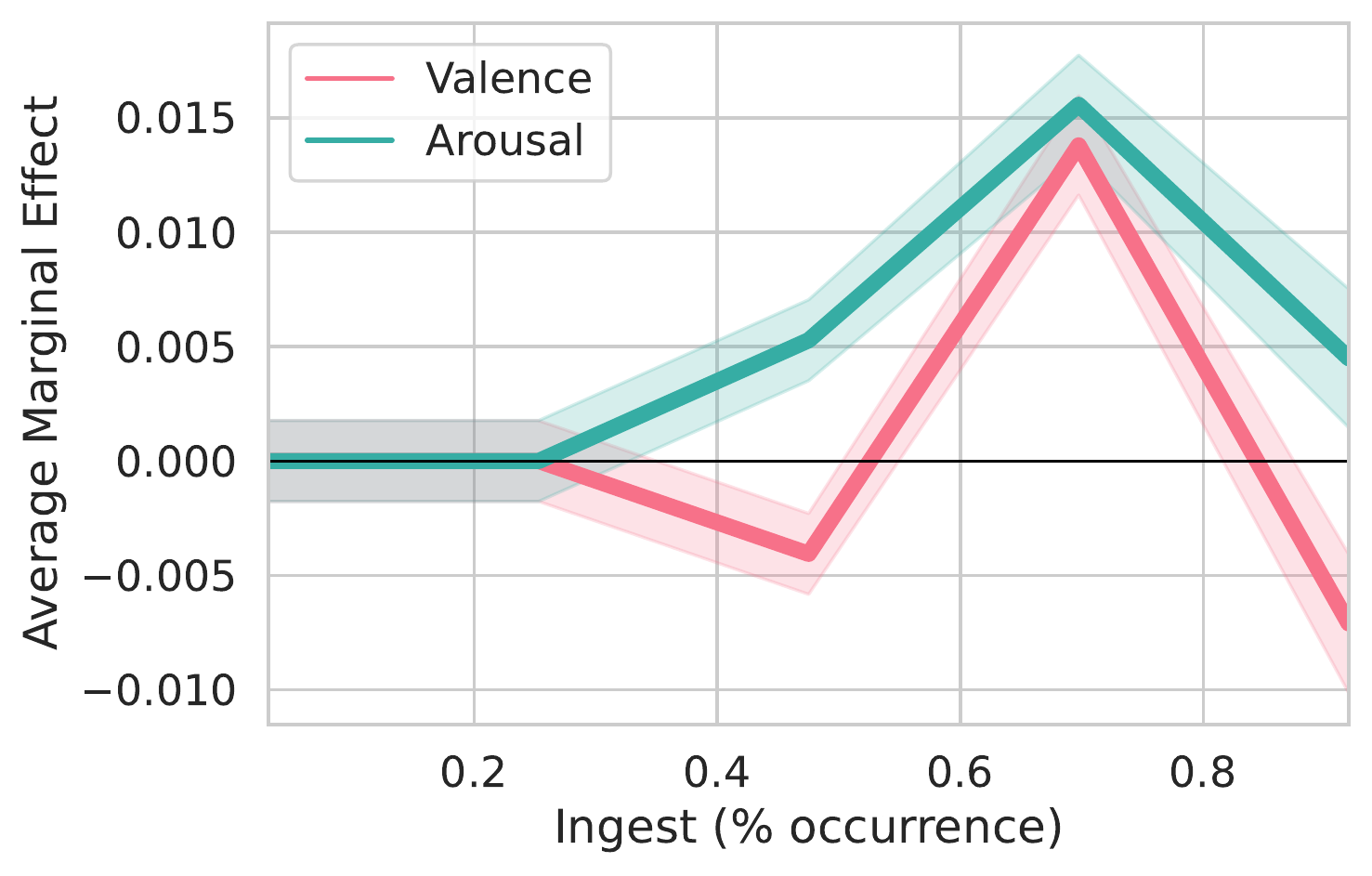}} & 
    {\includegraphics[width=0.30\textwidth]{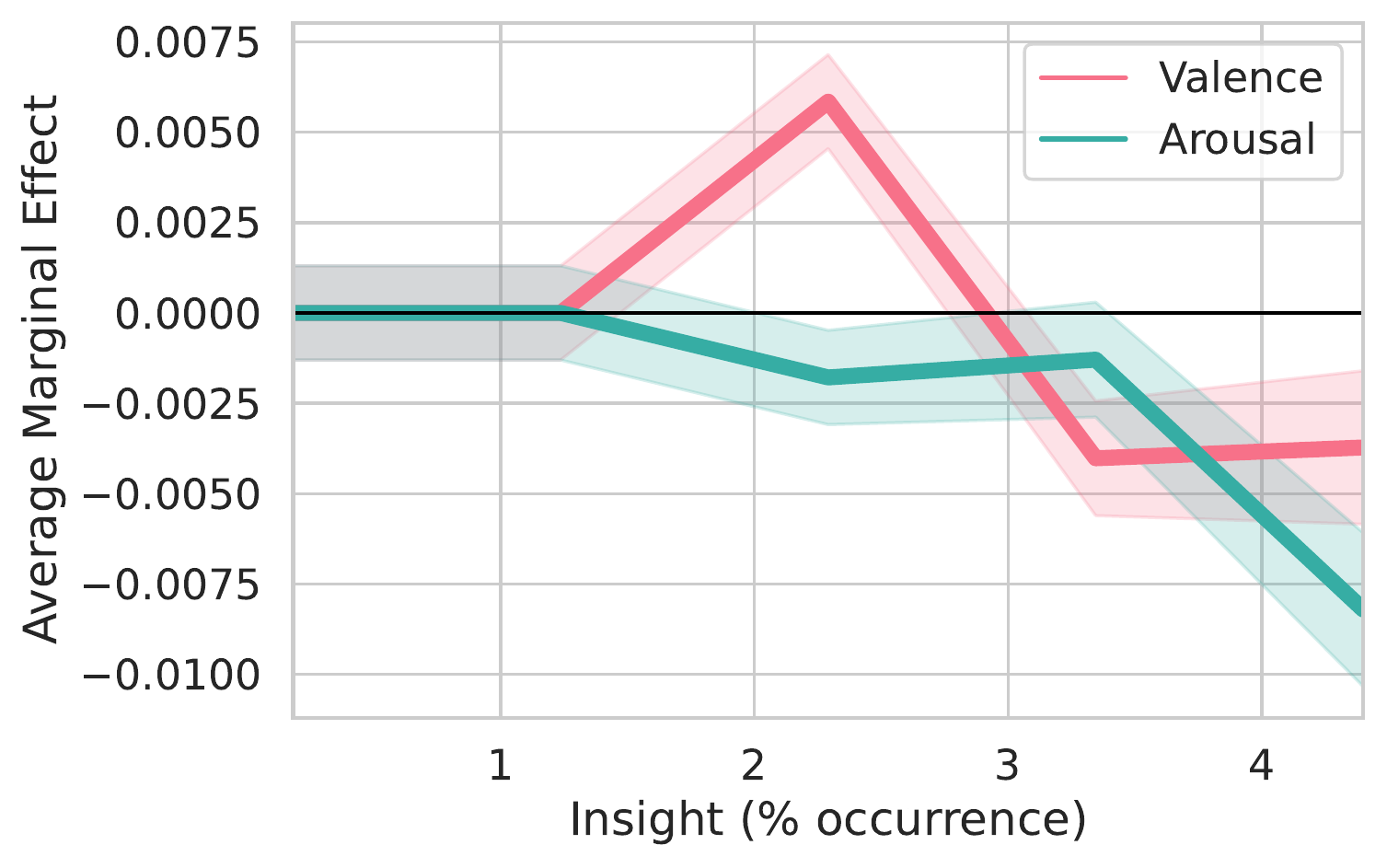}} & 
    {\includegraphics[width=0.30\textwidth]{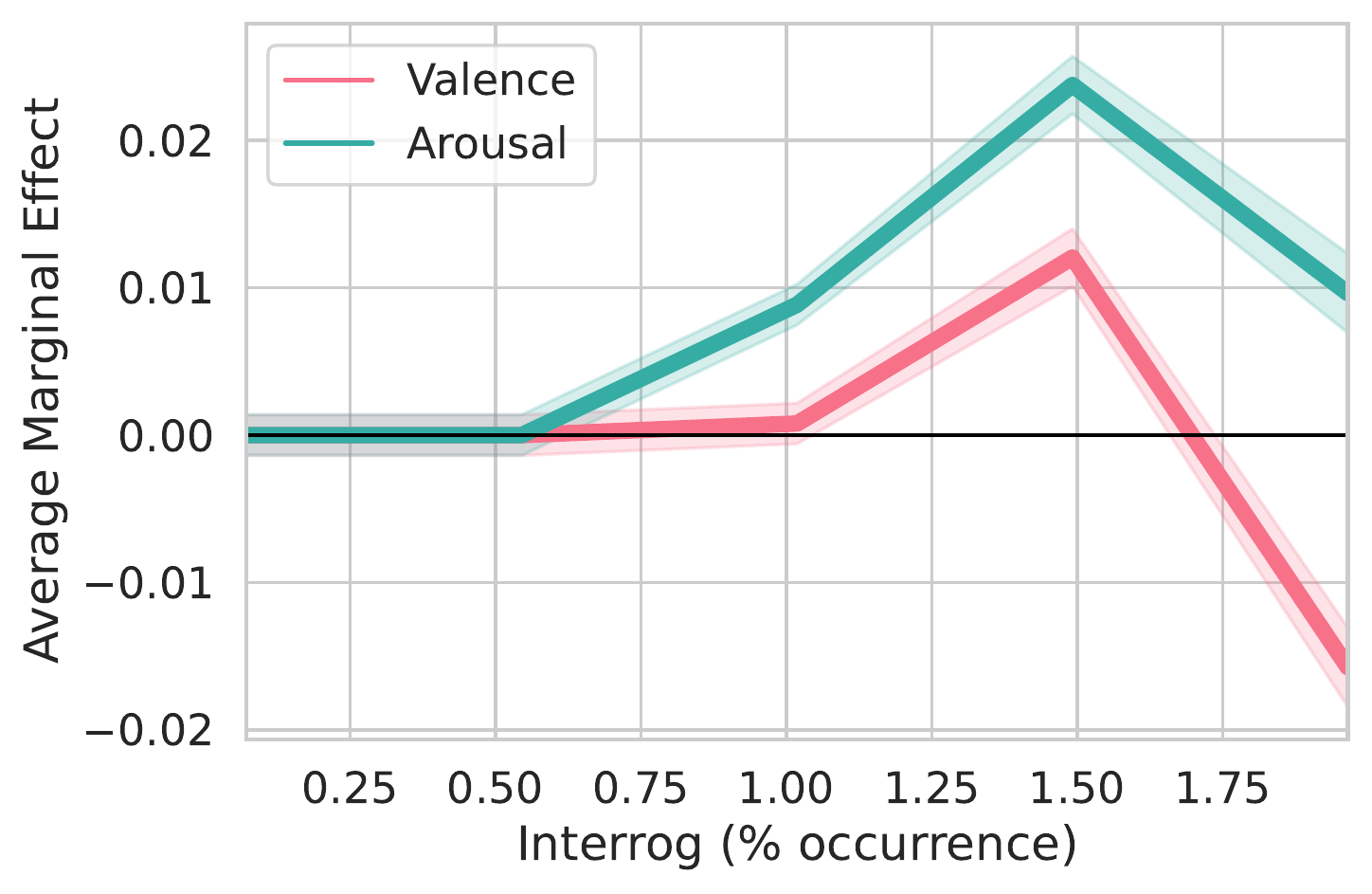}} \\ 
    {\includegraphics[width=0.30\textwidth]{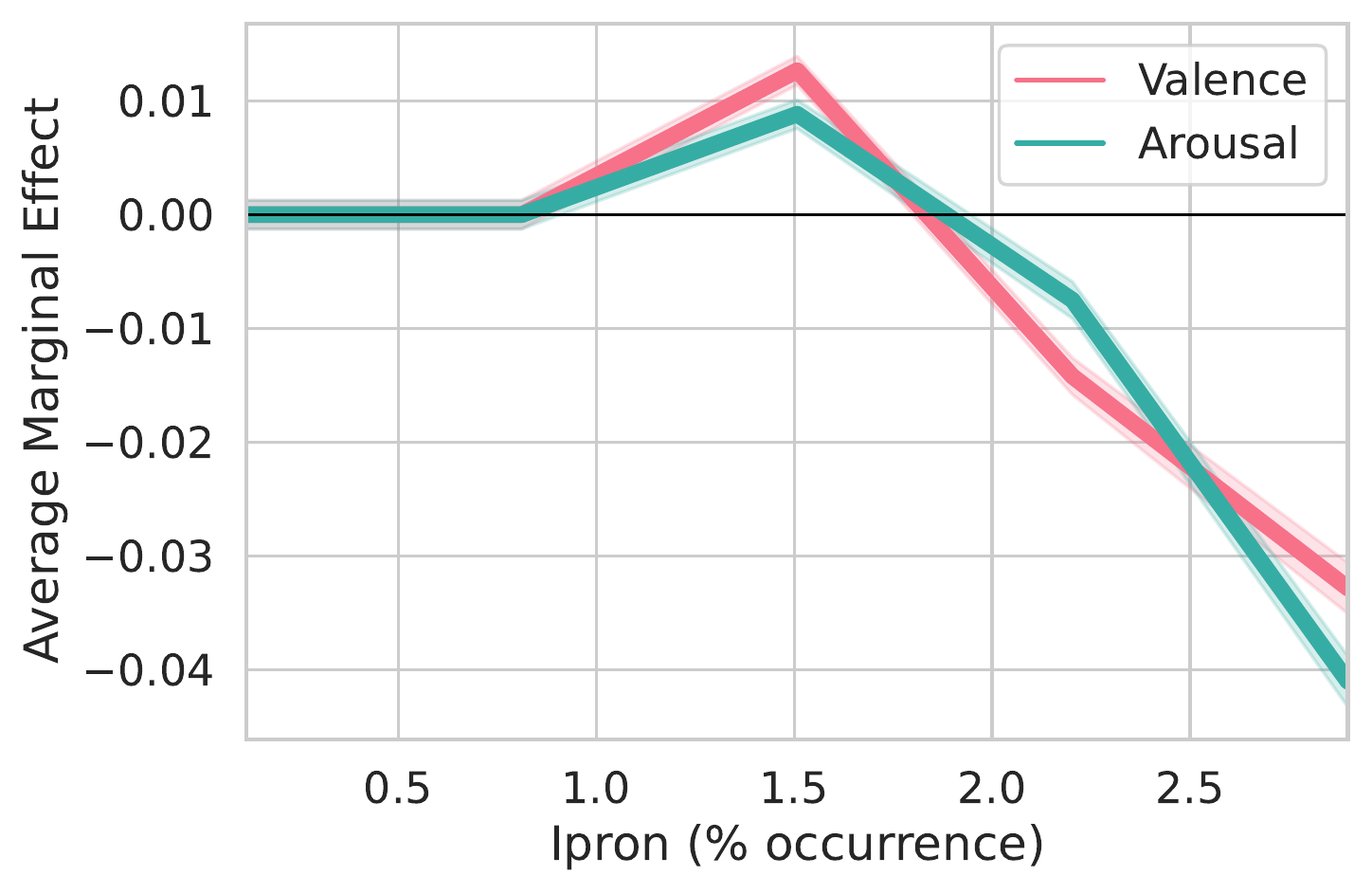}} & 
    {\includegraphics[width=0.30\textwidth]{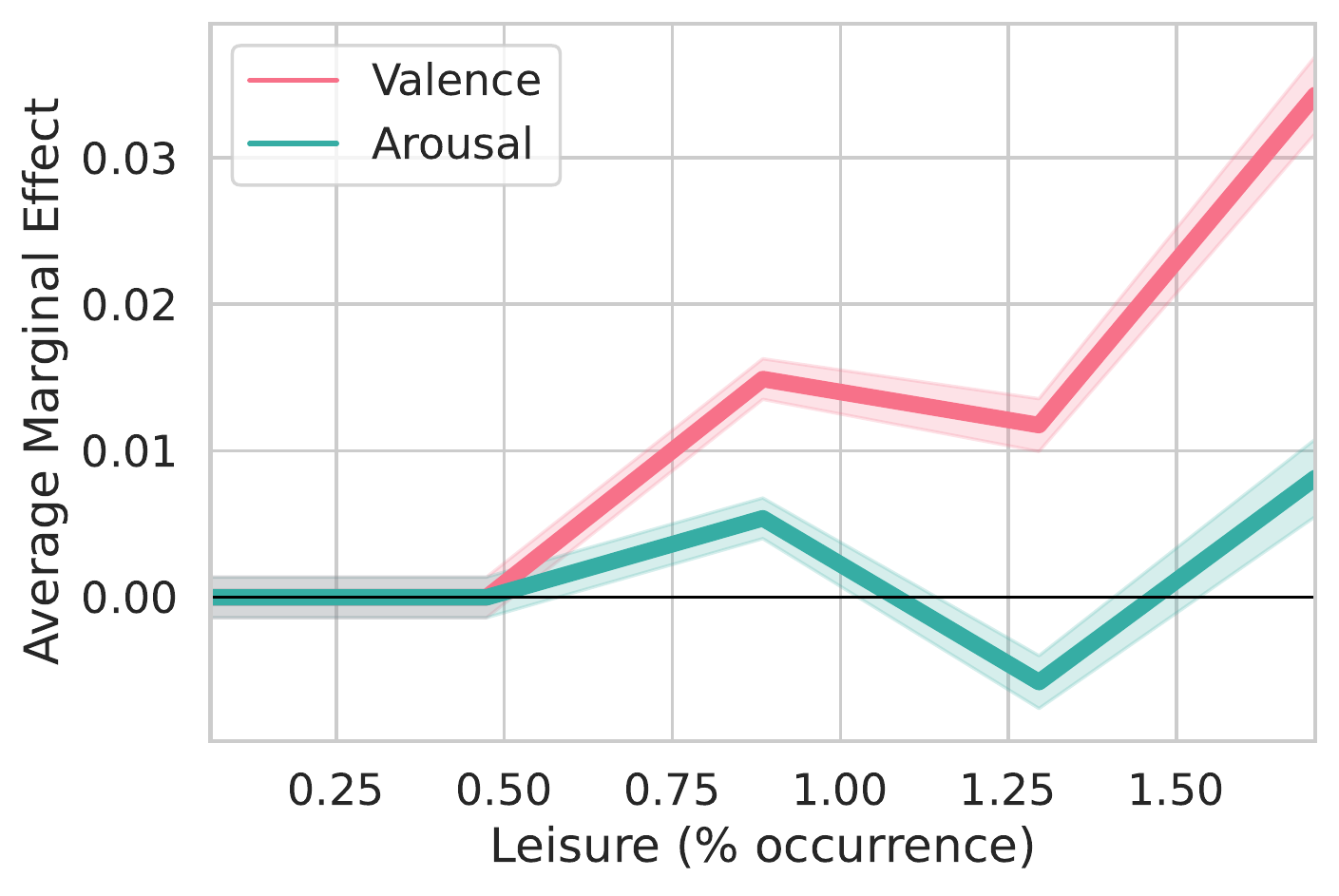}} & 
    {\includegraphics[width=0.30\textwidth]{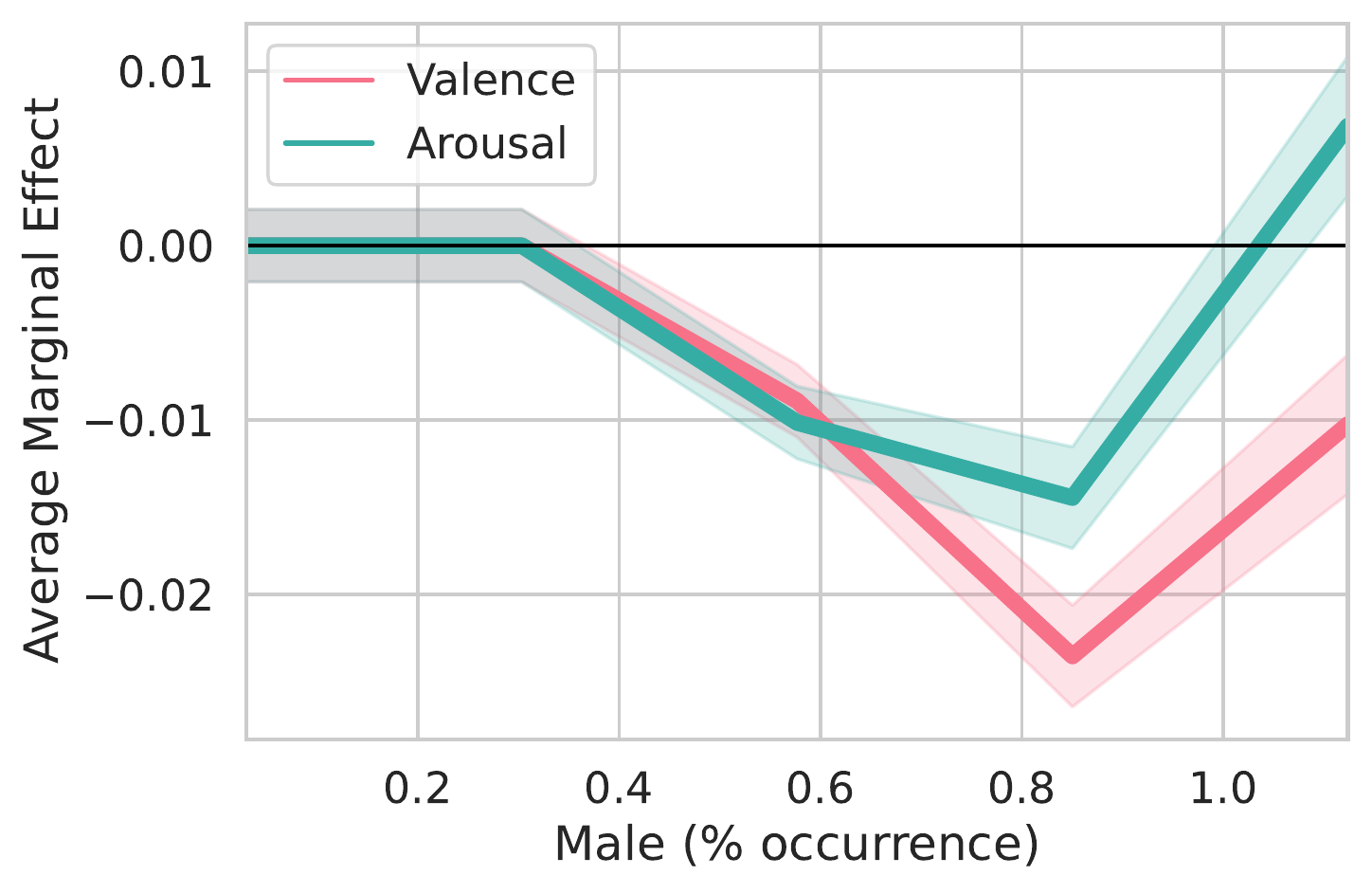}} \\ 
    {\includegraphics[width=0.30\textwidth]{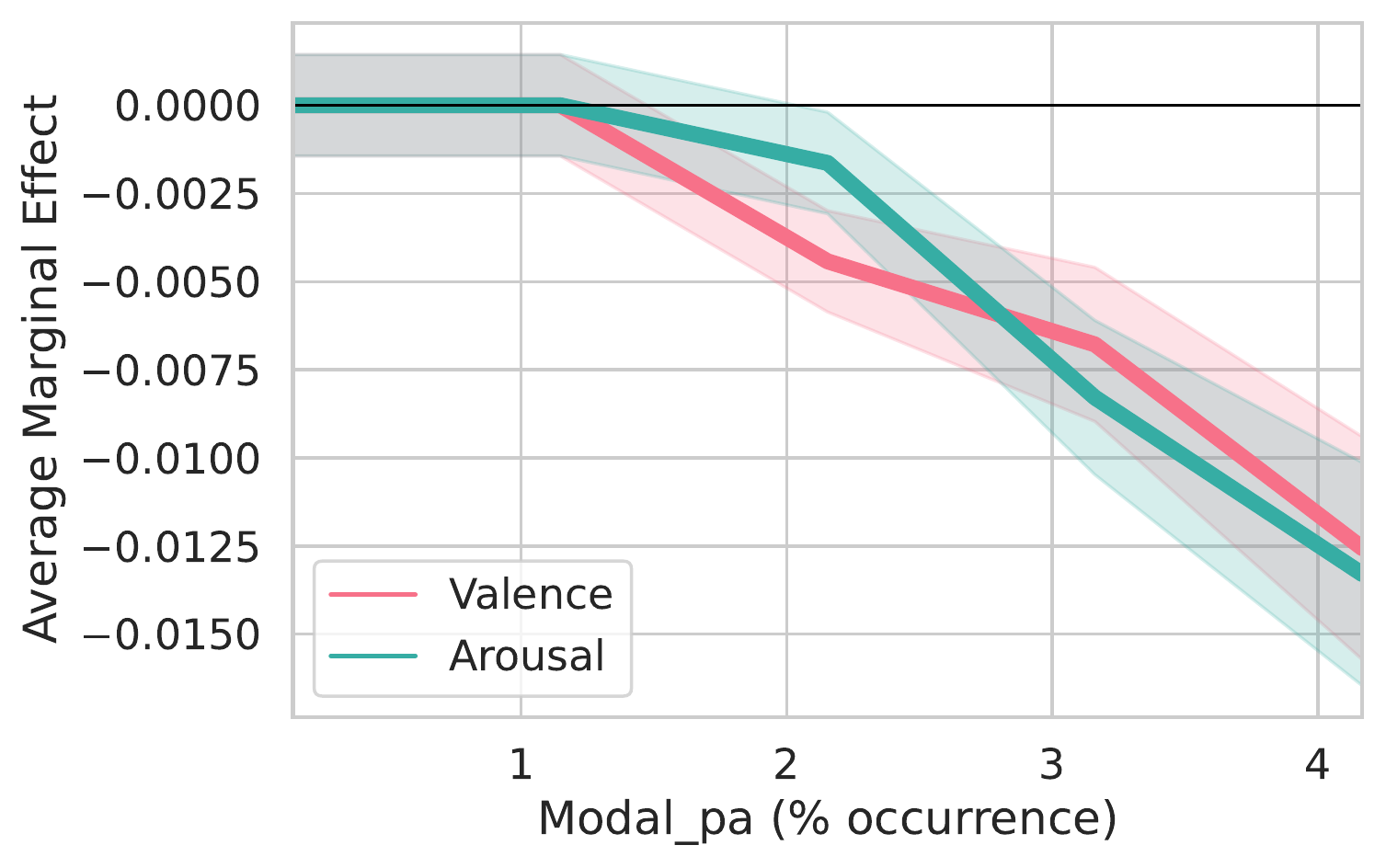}} & 
    {\includegraphics[width=0.30\textwidth]{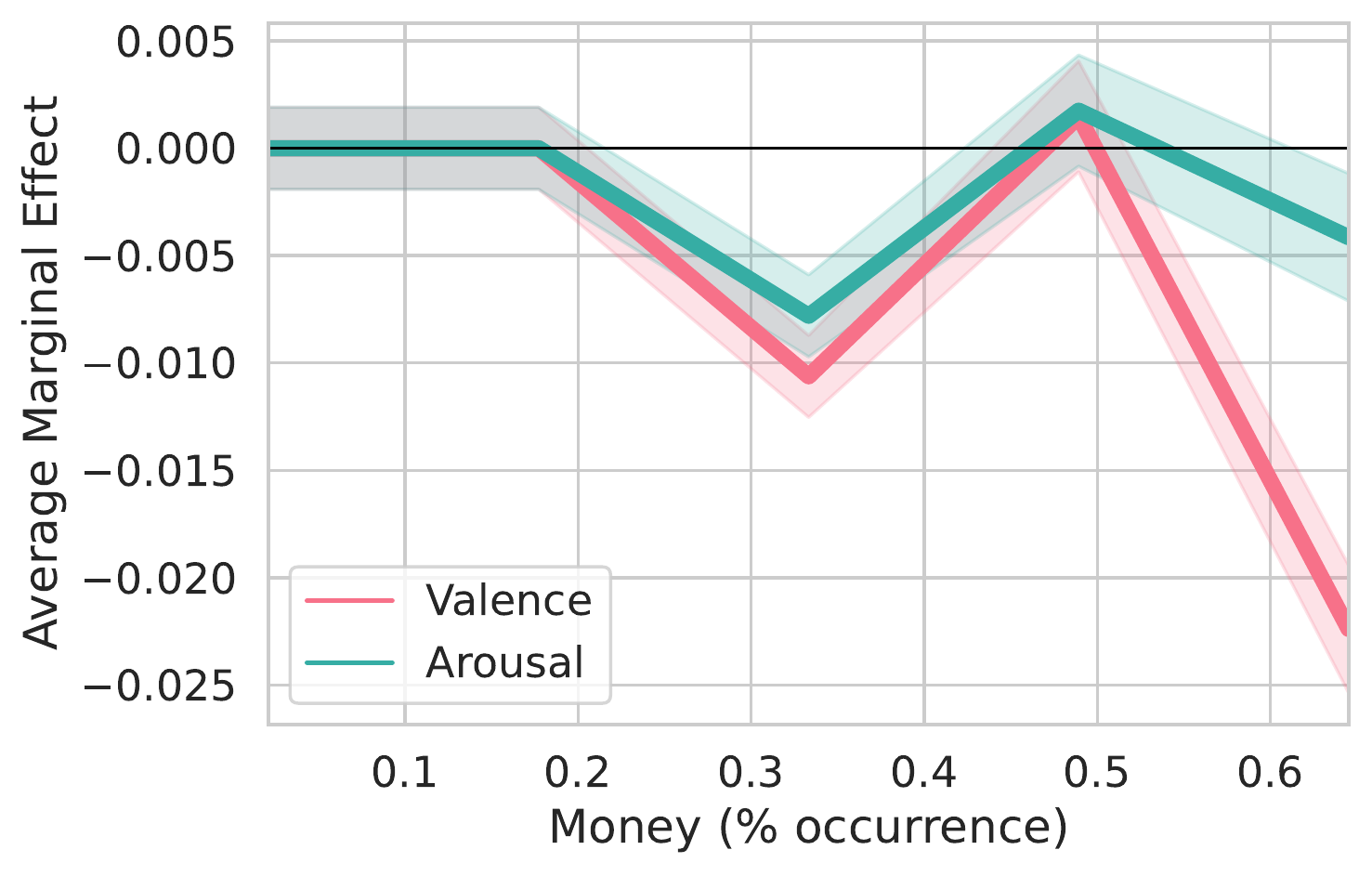}} & 
    {\includegraphics[width=0.30\textwidth]{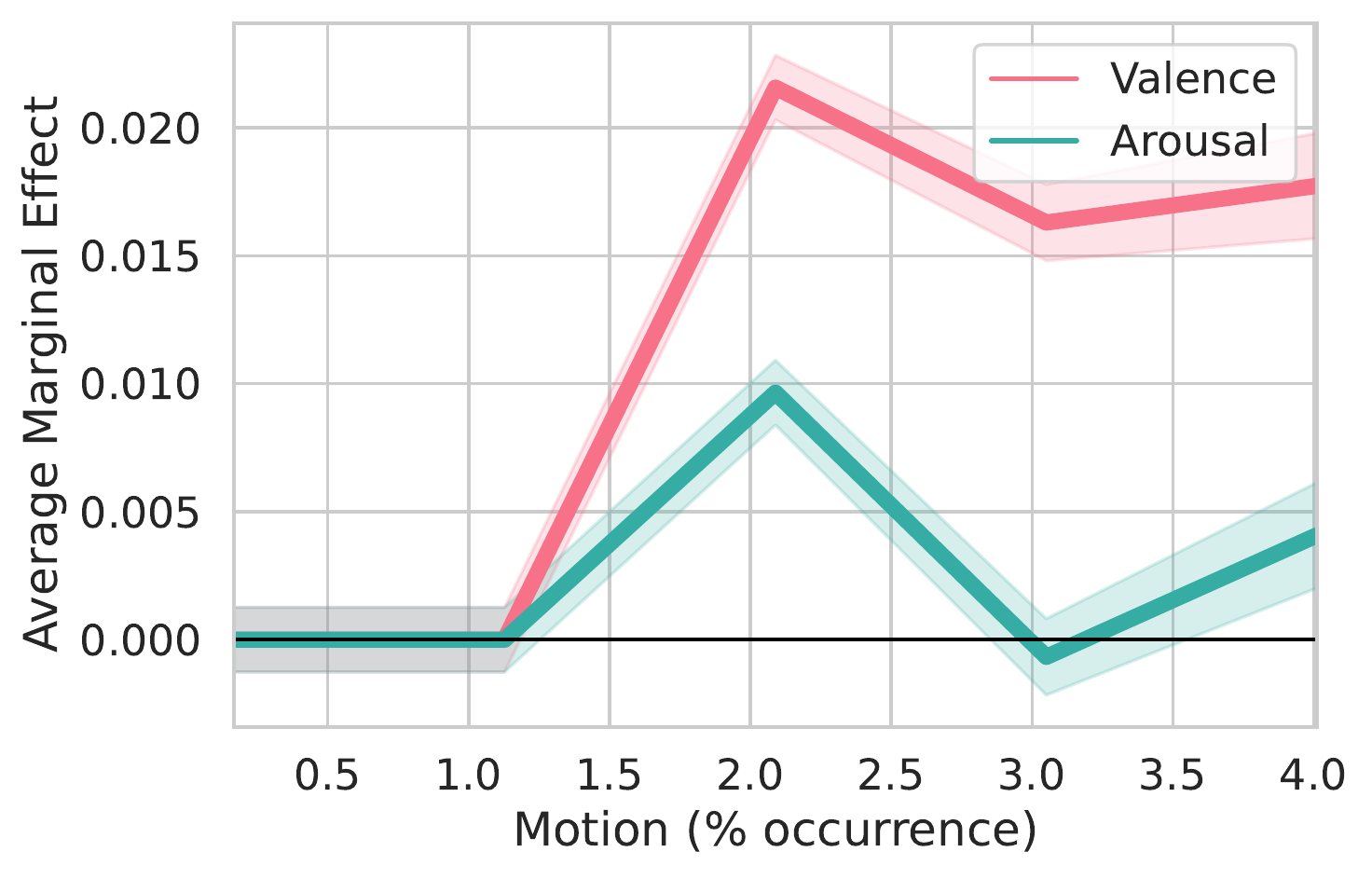}} \\ 
    \end{tabular}
    \caption{
    Average marginal effects of LIWC psycholinguistic lexical category \textbf{lyrical features} on listener affective responses, controlling for musical features and listener demographics. With the intent to reduce noise at the extremities, x-axis limits are capped at their 95\% quantile values.
    Arranged in alphabetical order, standard errors are shown;
    \textcolor{red}{valence} in \textcolor{red}{red}, \textcolor{blue}{arousal} in \textcolor{blue}{blue} (Part 2/4).
    }
    \label{fig:lyricfeatures_expanded_2}
\end{figure*}

\begin{figure*}[!t]
    \centering
    \begin{tabular}{ccc}
    {\includegraphics[width=0.30\textwidth]{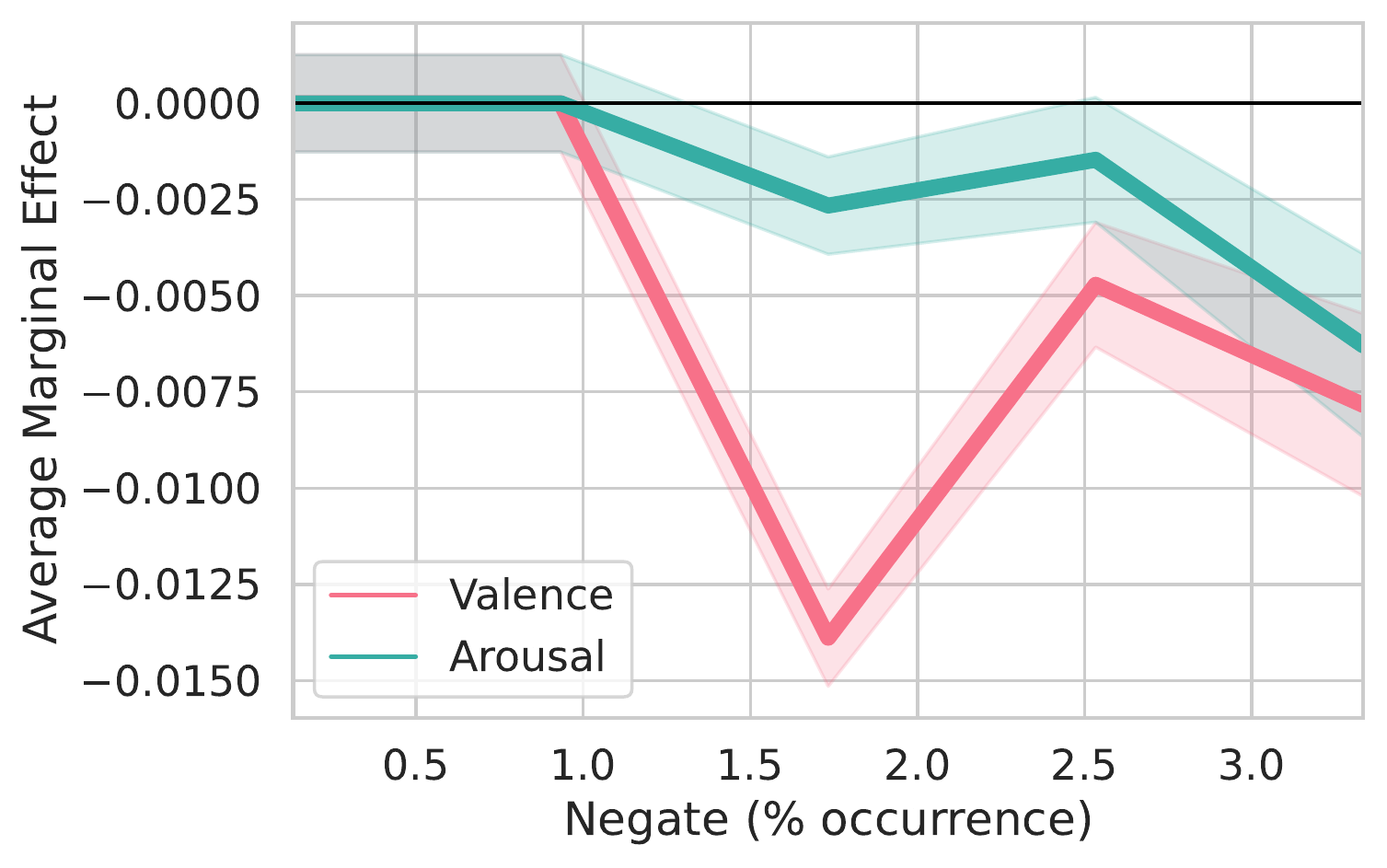}} &
    {\includegraphics[width=0.30\textwidth]{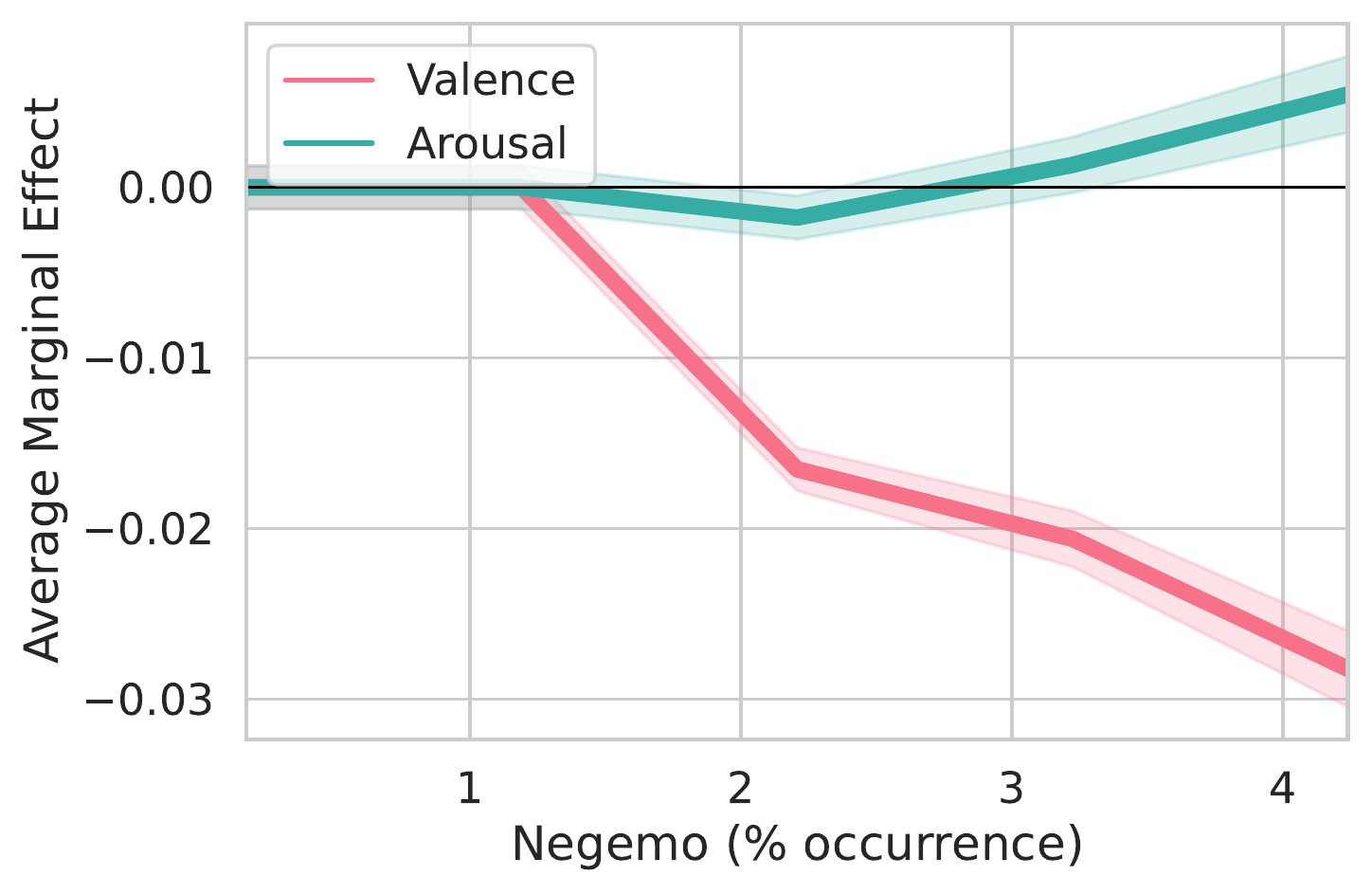}} & 
    {\includegraphics[width=0.30\textwidth]{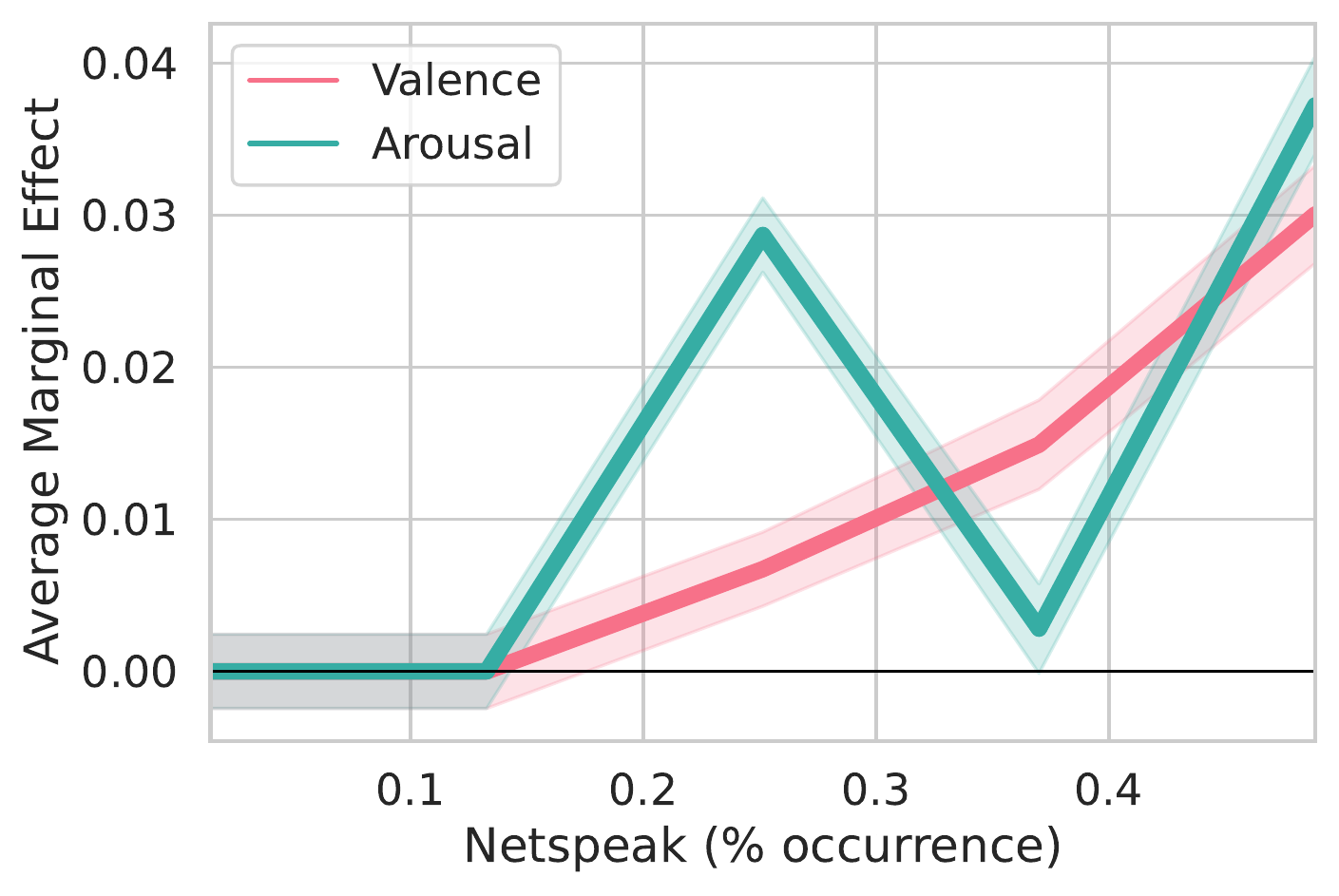}} \\
    {\includegraphics[width=0.30\textwidth]{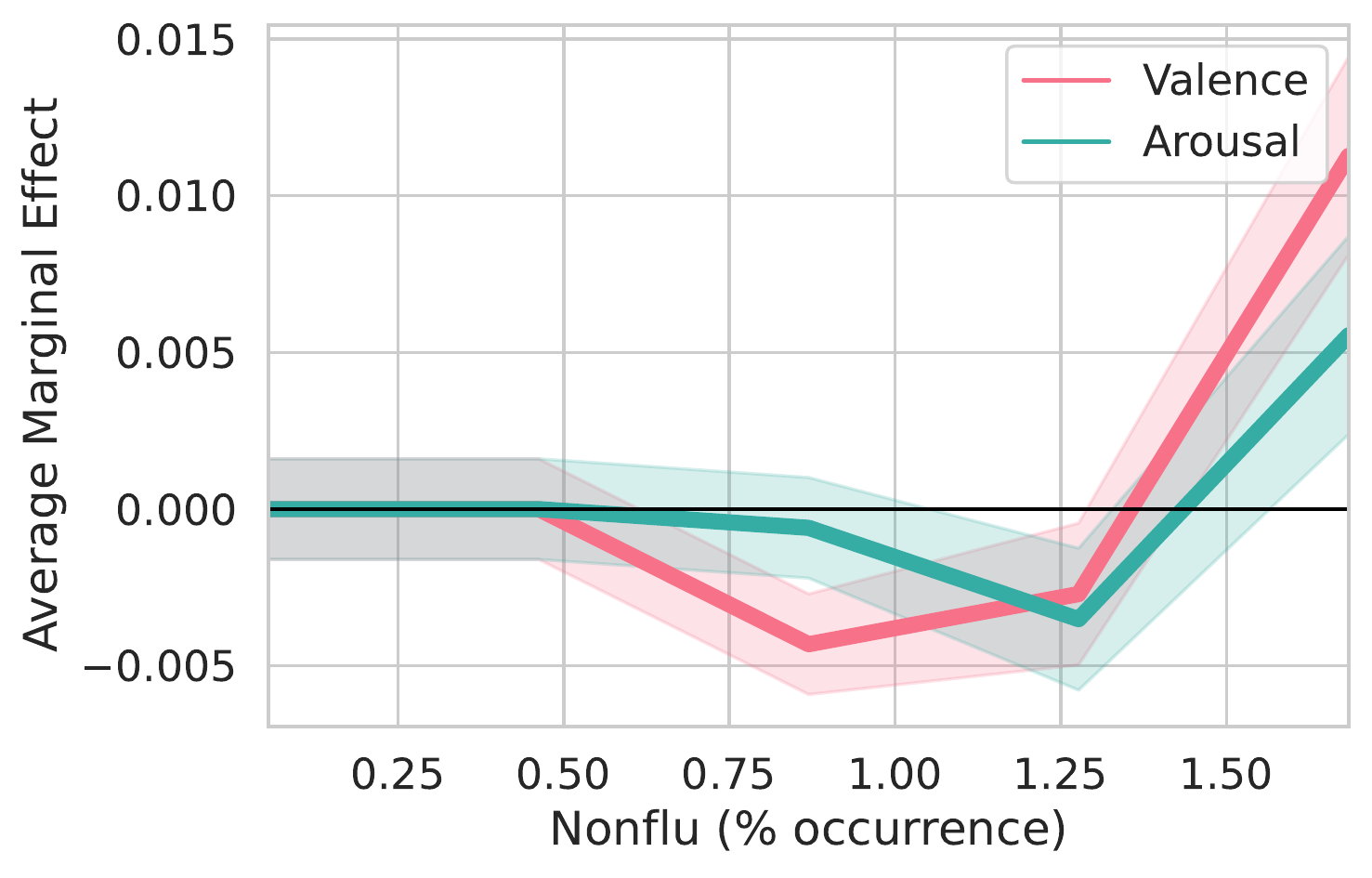}} & 
    {\includegraphics[width=0.30\textwidth]{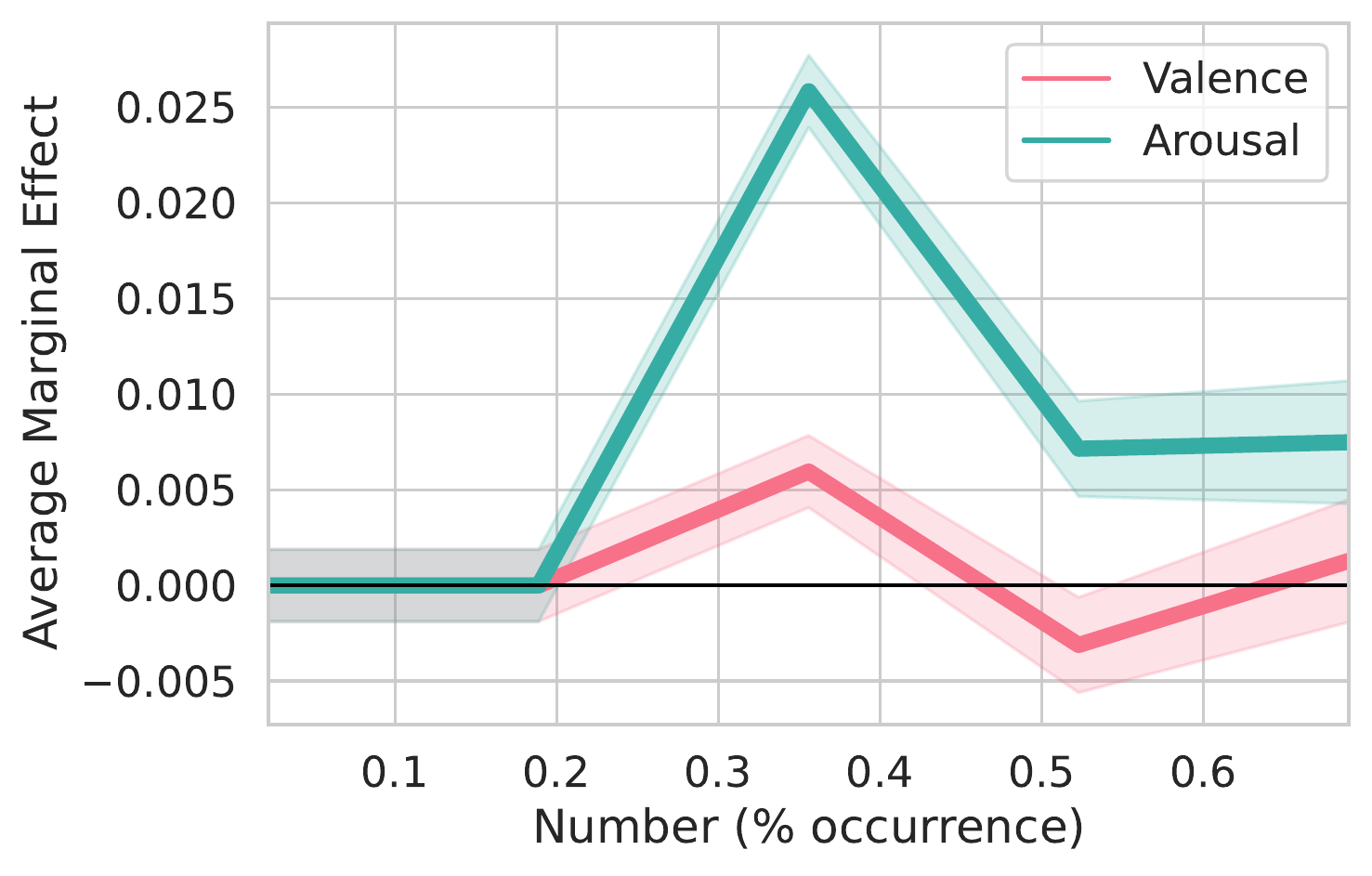}} & 
    {\includegraphics[width=0.30\textwidth]{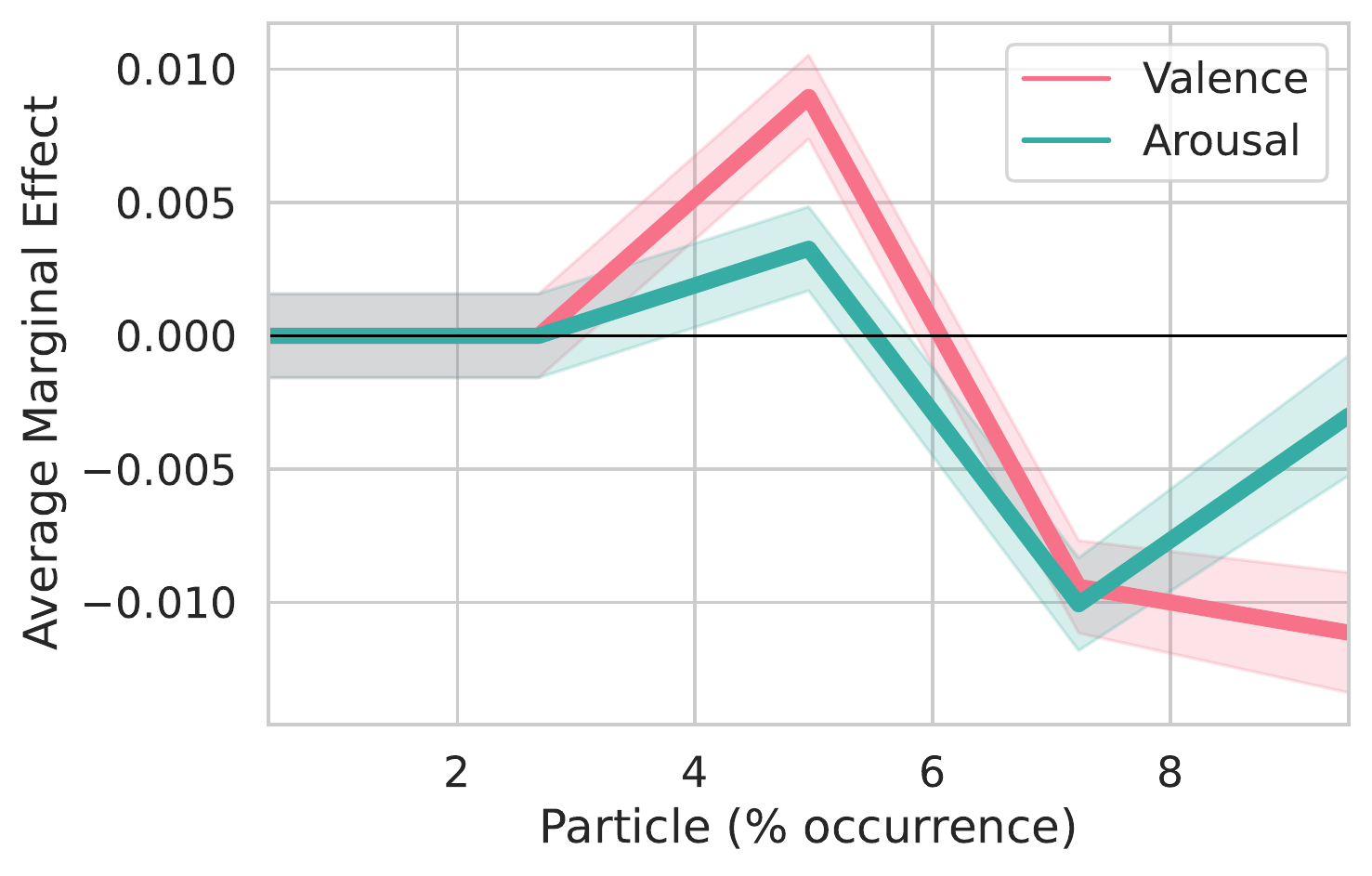}} \\ 
    {\includegraphics[width=0.30\textwidth]{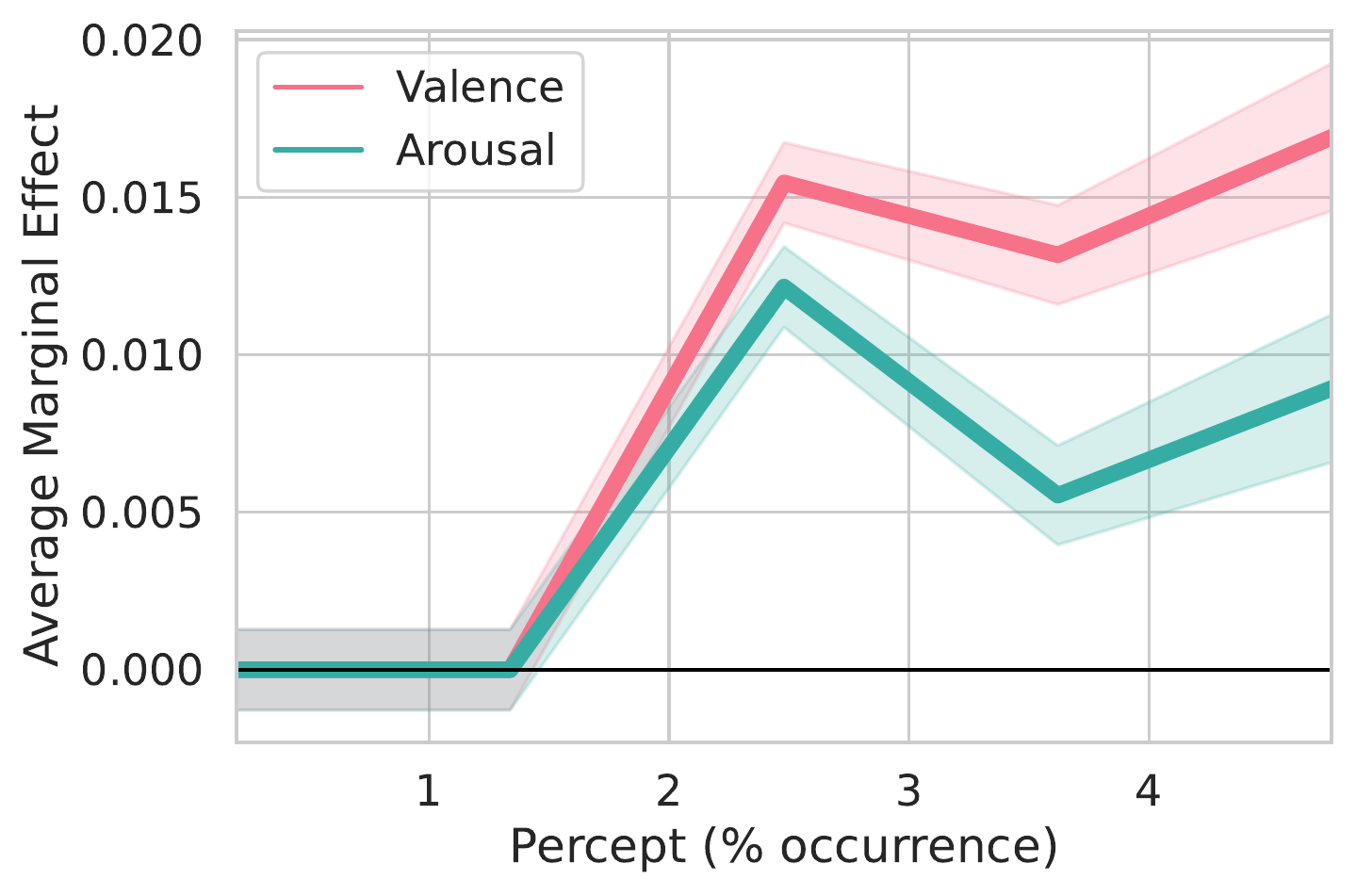}} & 
    {\includegraphics[width=0.30\textwidth]{plots/AME/lyric_posemo.pdf}} & 
    {\includegraphics[width=0.30\textwidth]{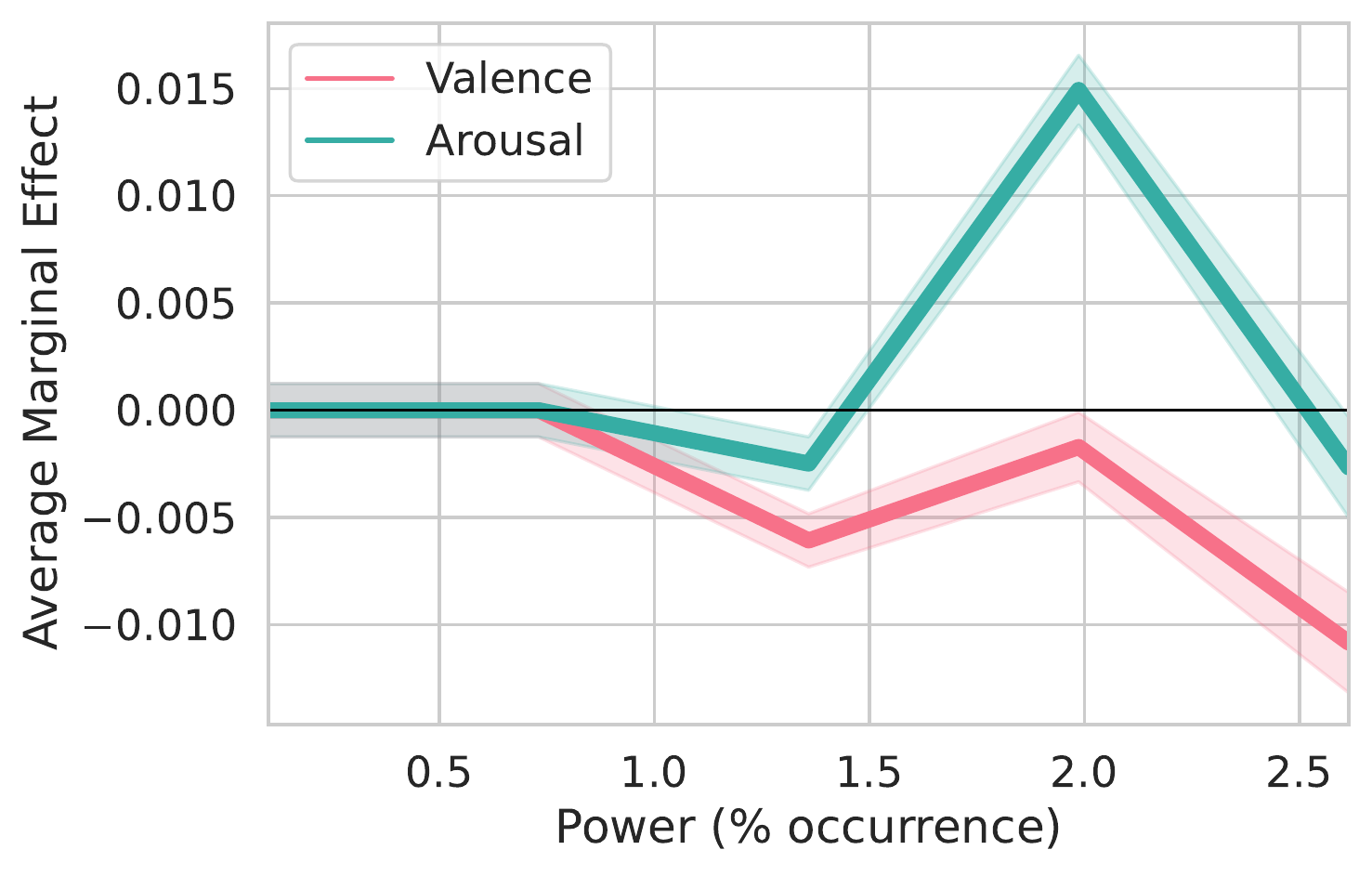}} \\
    {\includegraphics[width=0.30\textwidth]{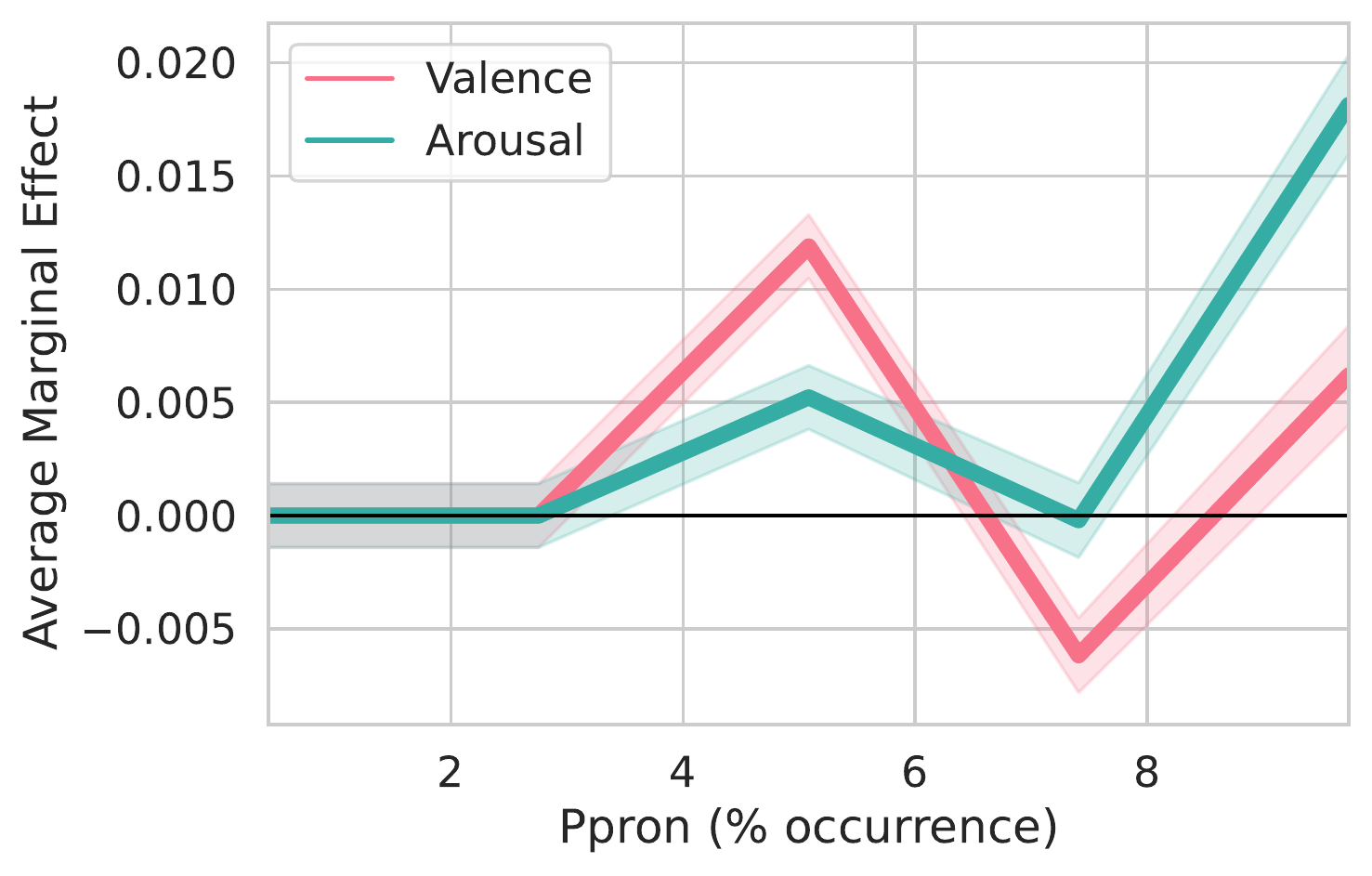}} & 
    {\includegraphics[width=0.30\textwidth]{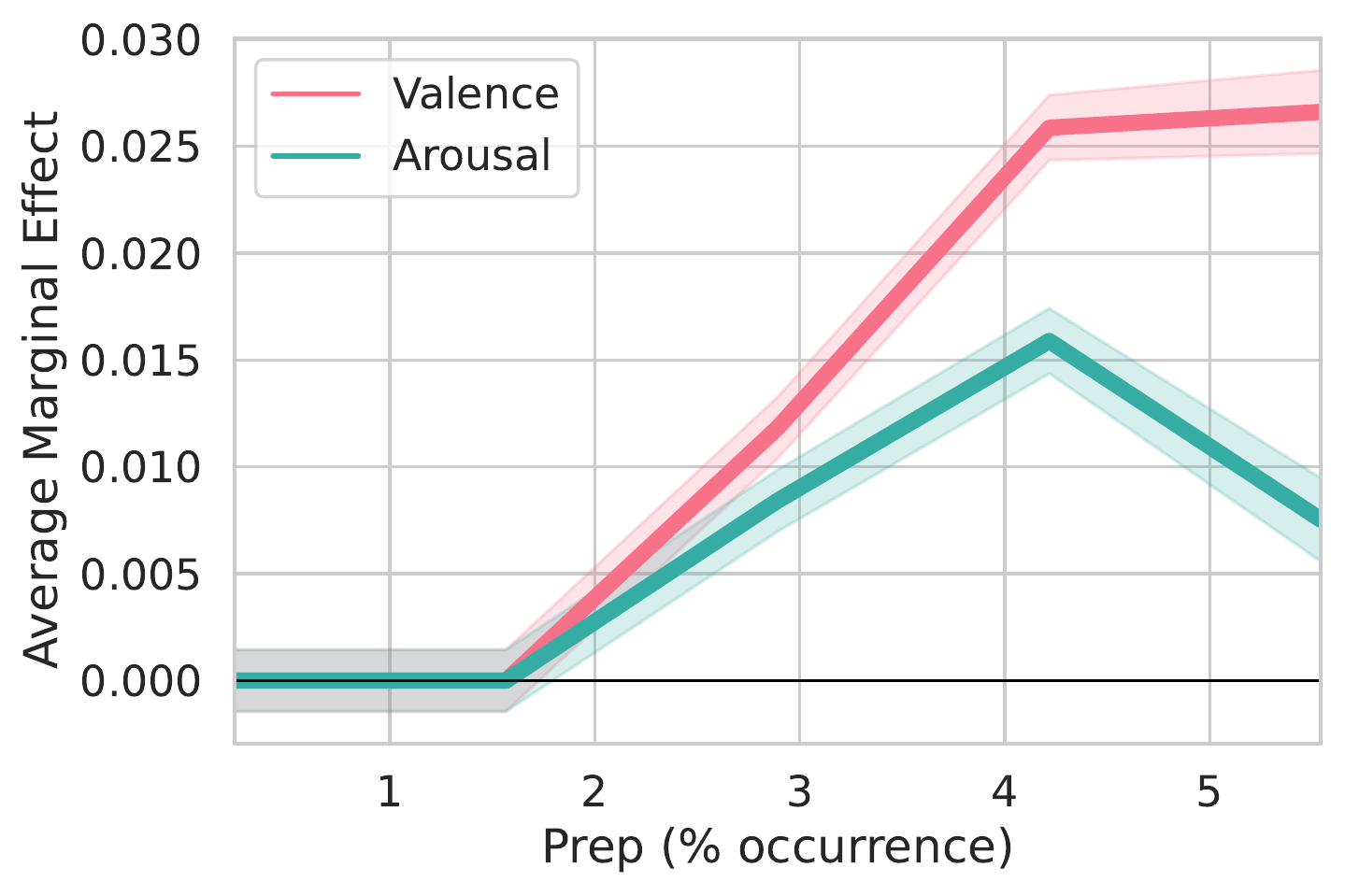}} &
    {\includegraphics[width=0.30\textwidth]{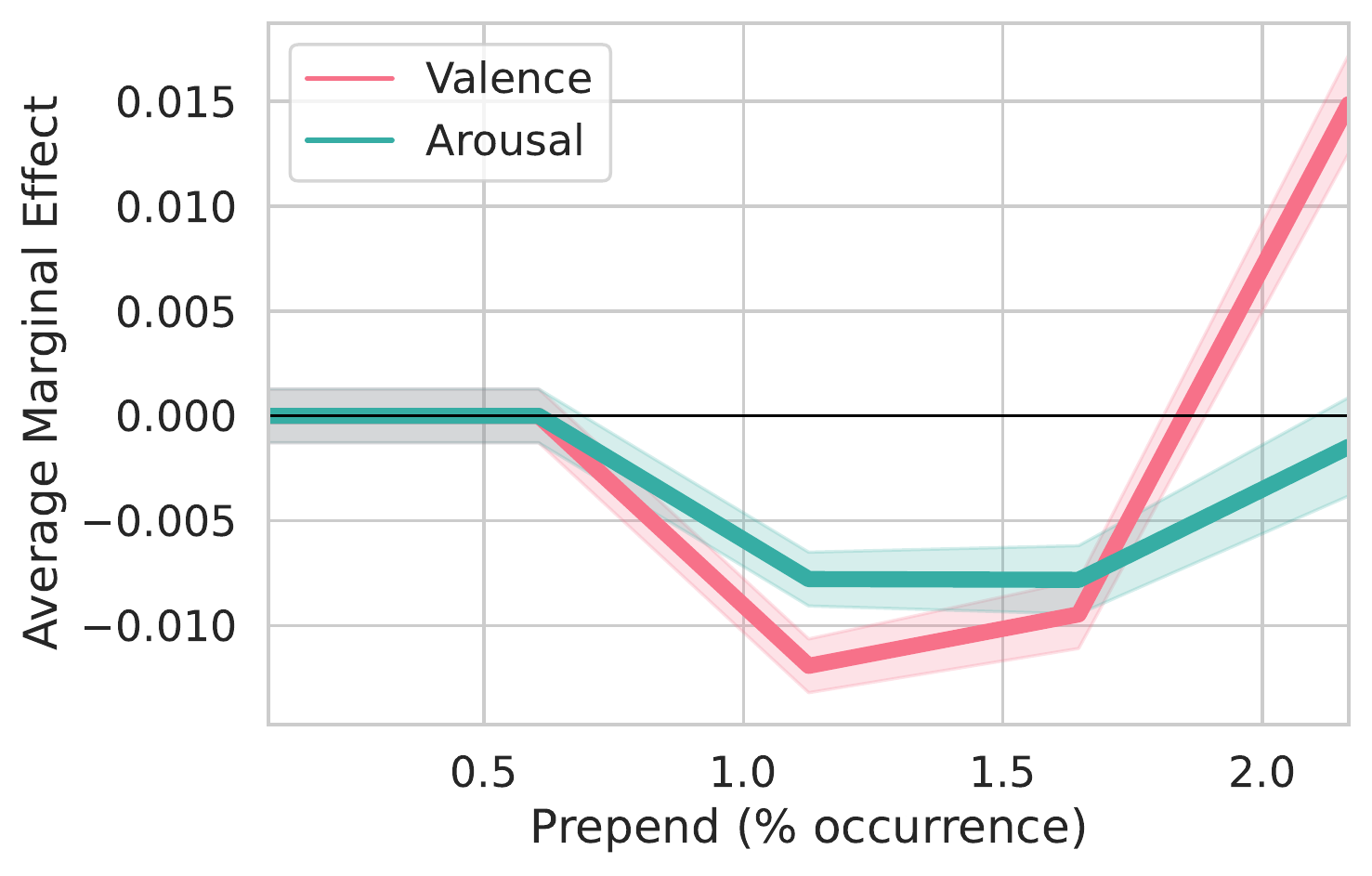}} \\
    {\includegraphics[width=0.30\textwidth]{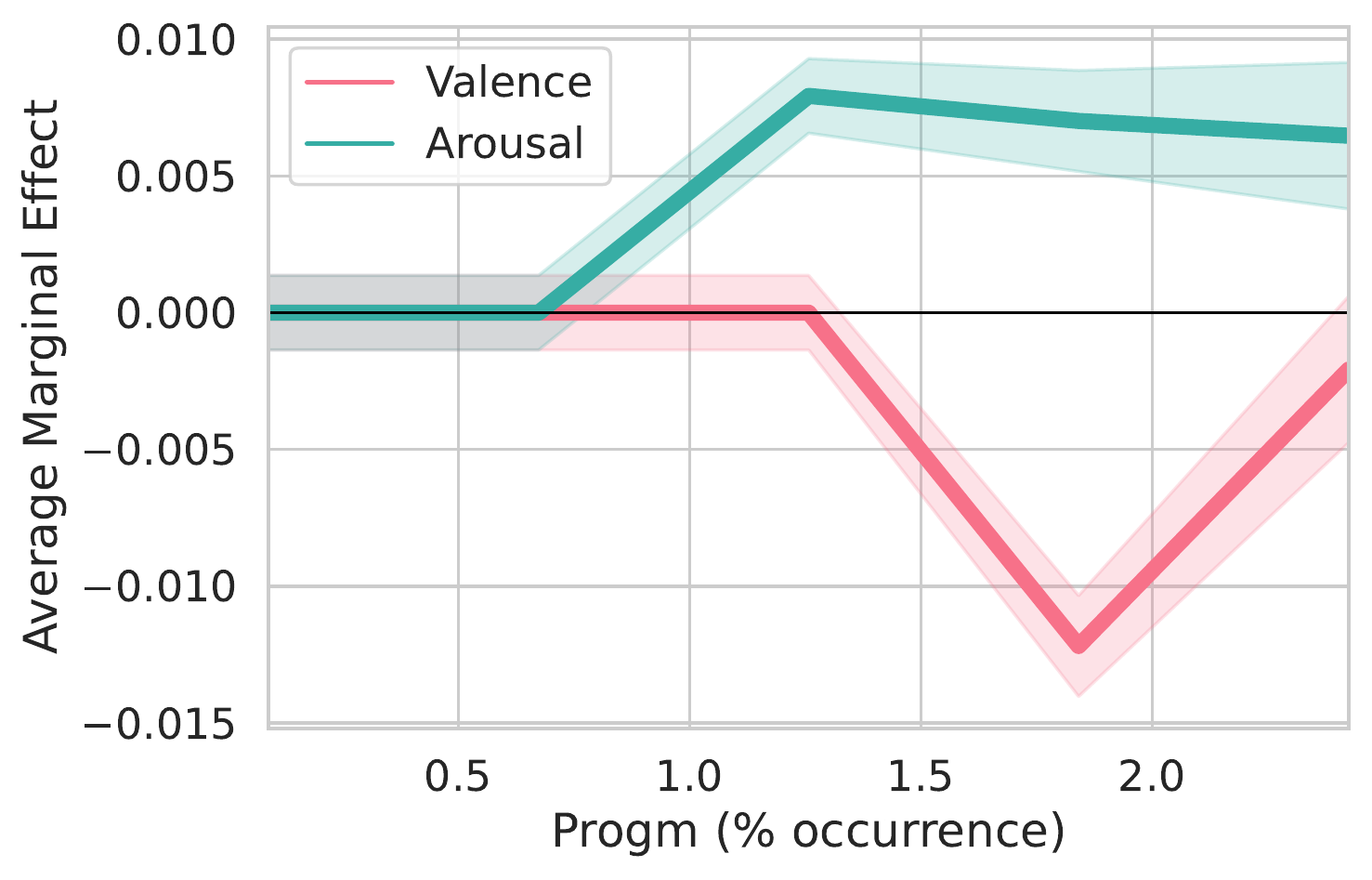}} & 
    {\includegraphics[width=0.30\textwidth]{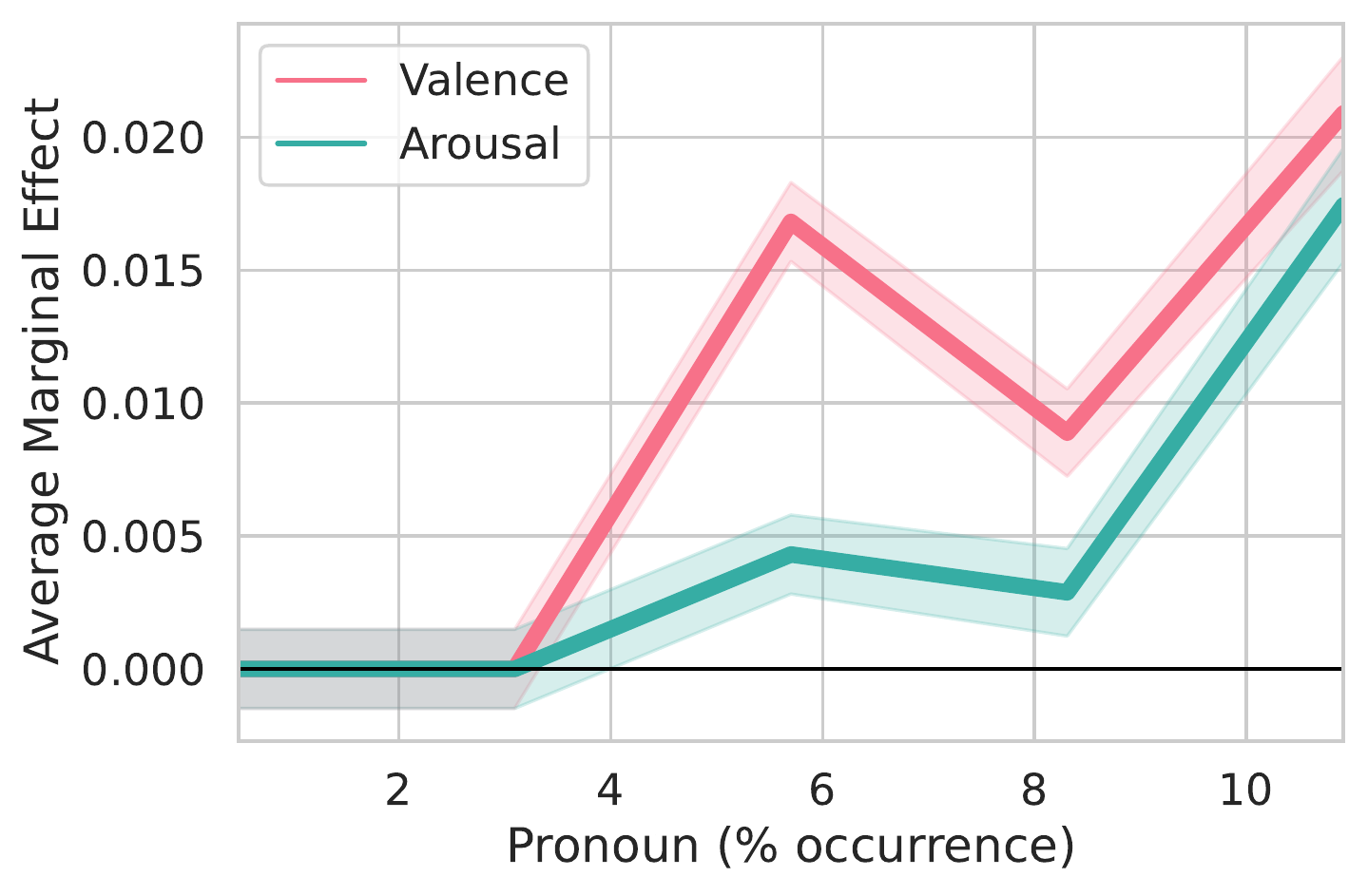}} &
    {\includegraphics[width=0.30\textwidth]{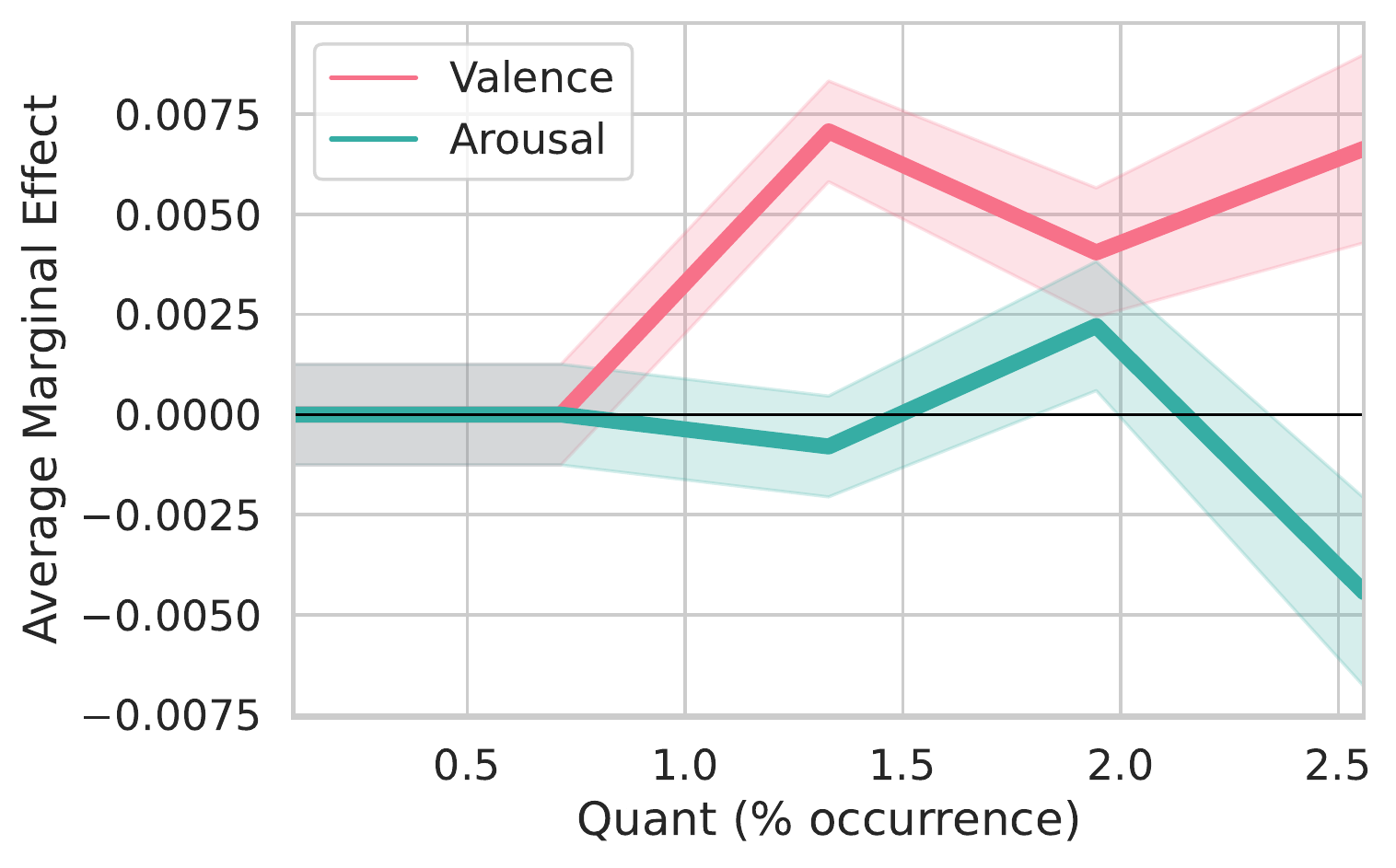}} \\
    {\includegraphics[width=0.30\textwidth]{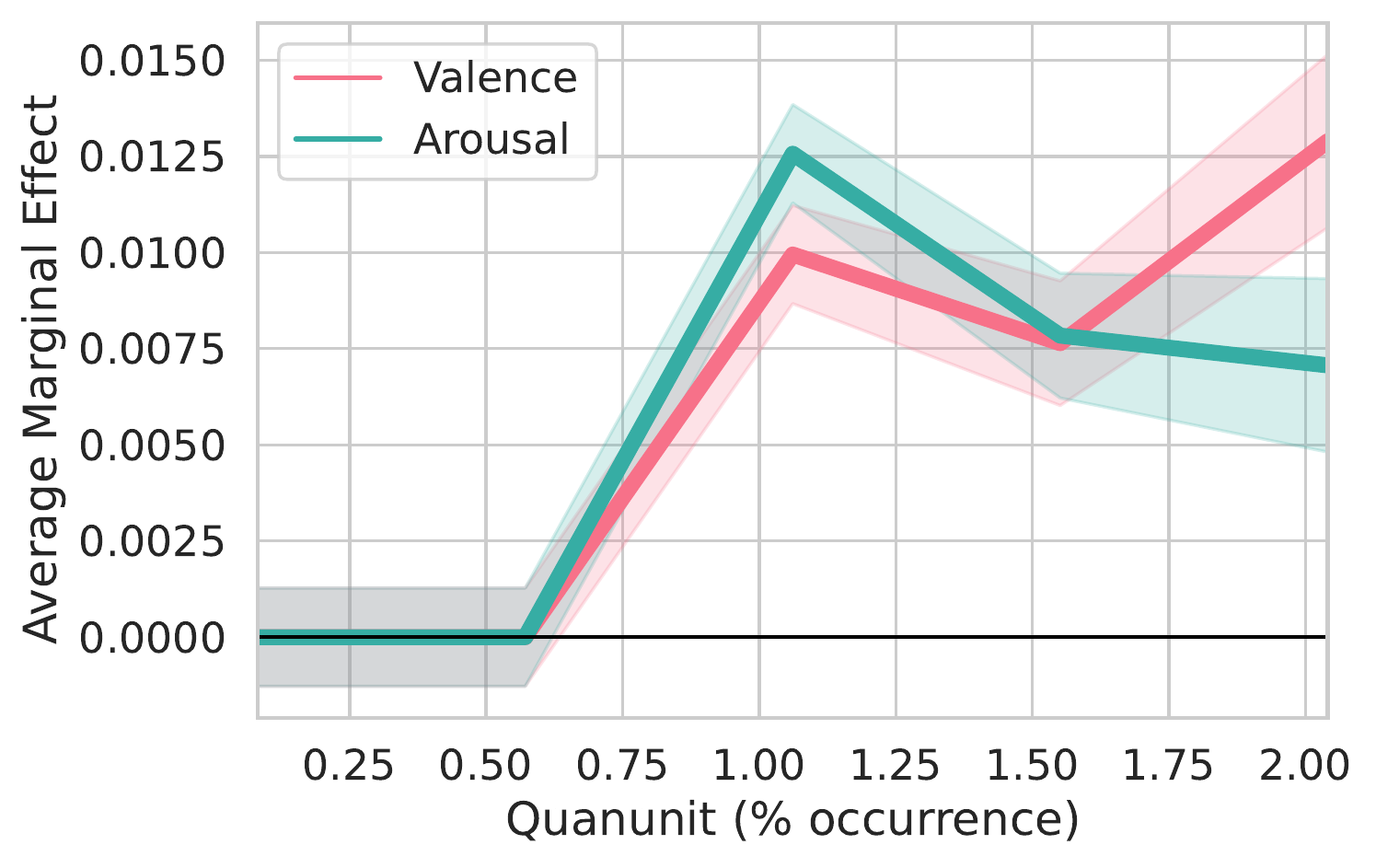}} & 
    {\includegraphics[width=0.30\textwidth]{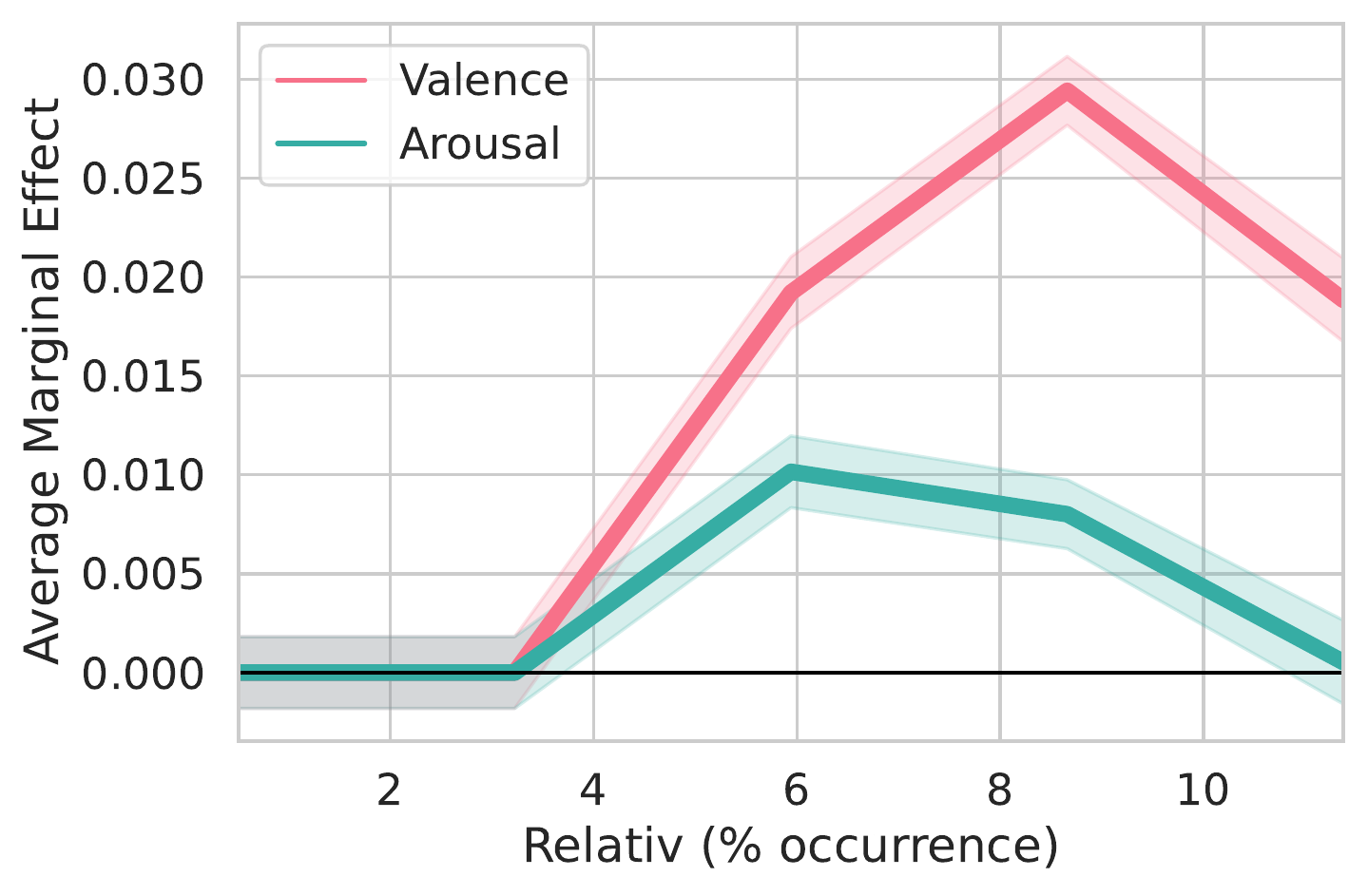}} & 
    {\includegraphics[width=0.30\textwidth]{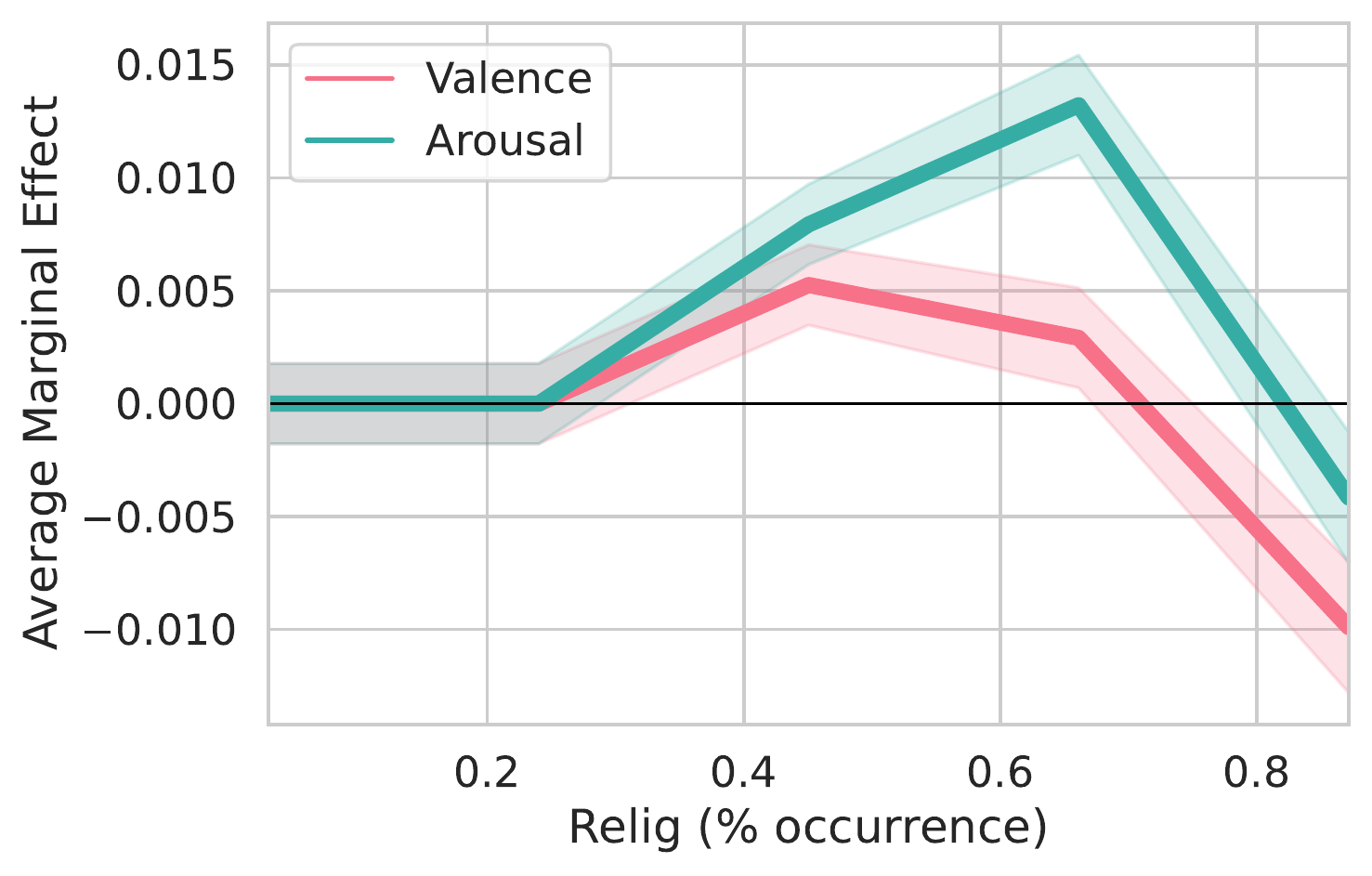}} \\
    {\includegraphics[width=0.30\textwidth]{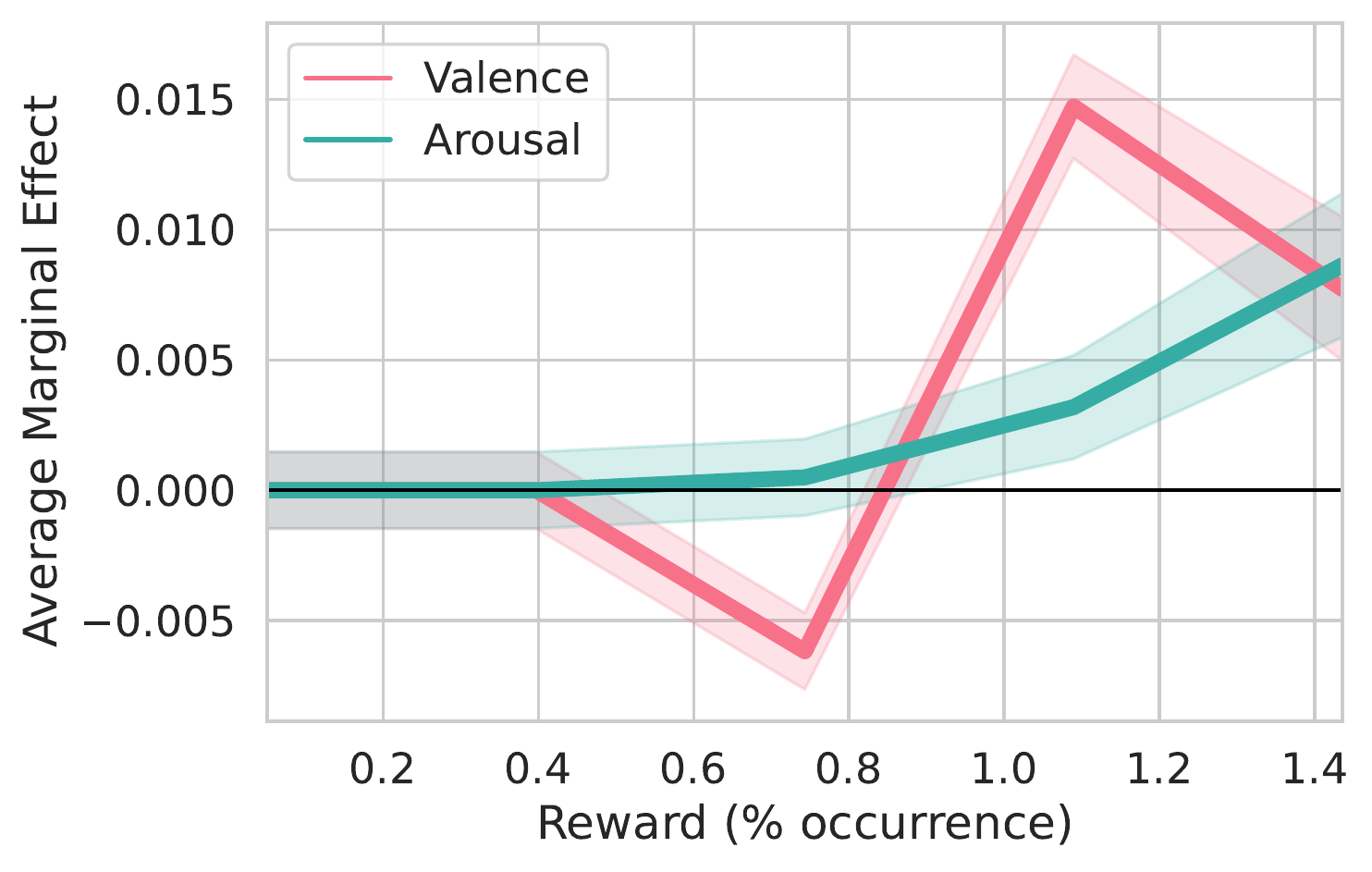}} & 
    {\includegraphics[width=0.30\textwidth]{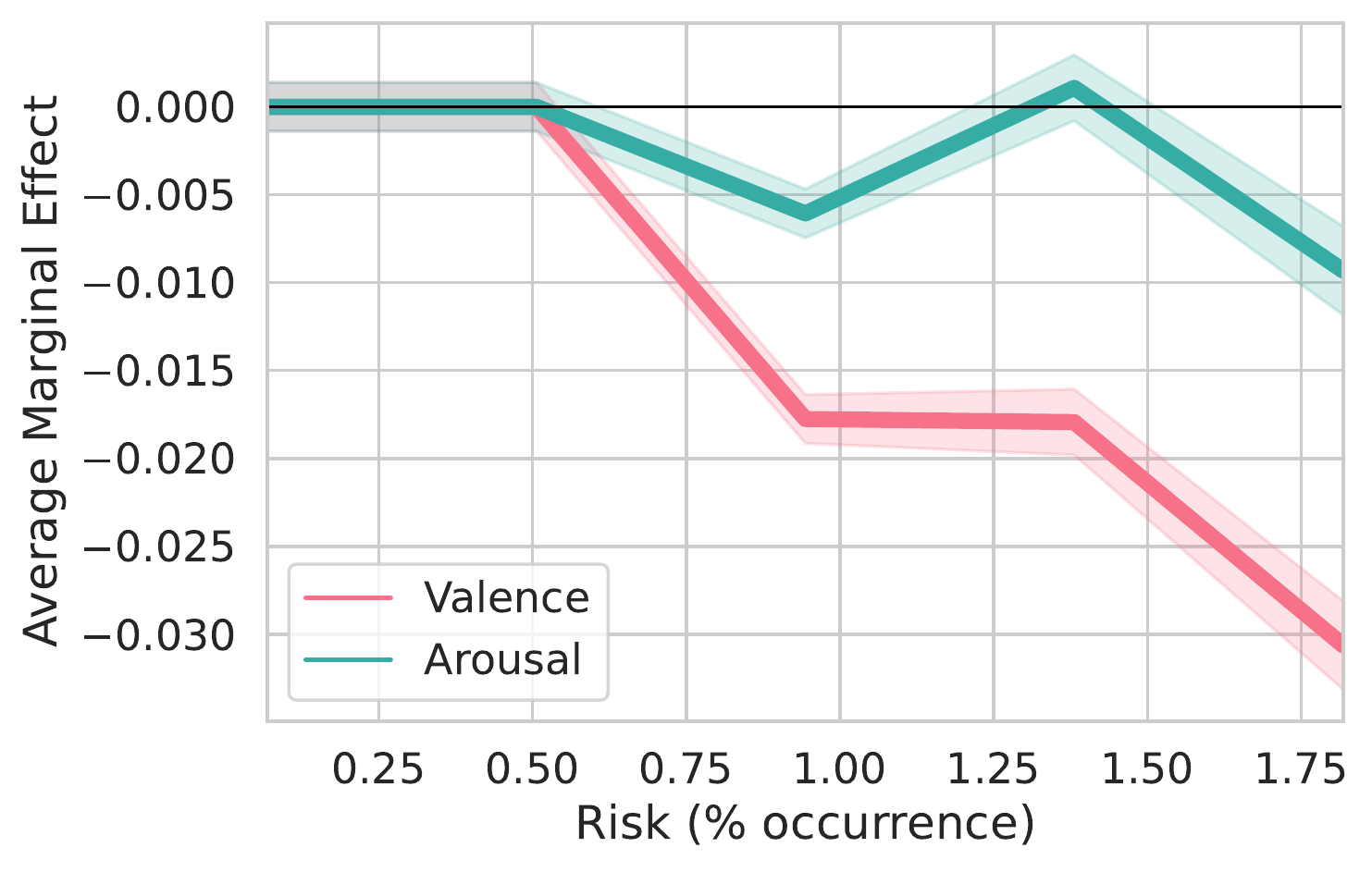}} & 
    {\includegraphics[width=0.30\textwidth]{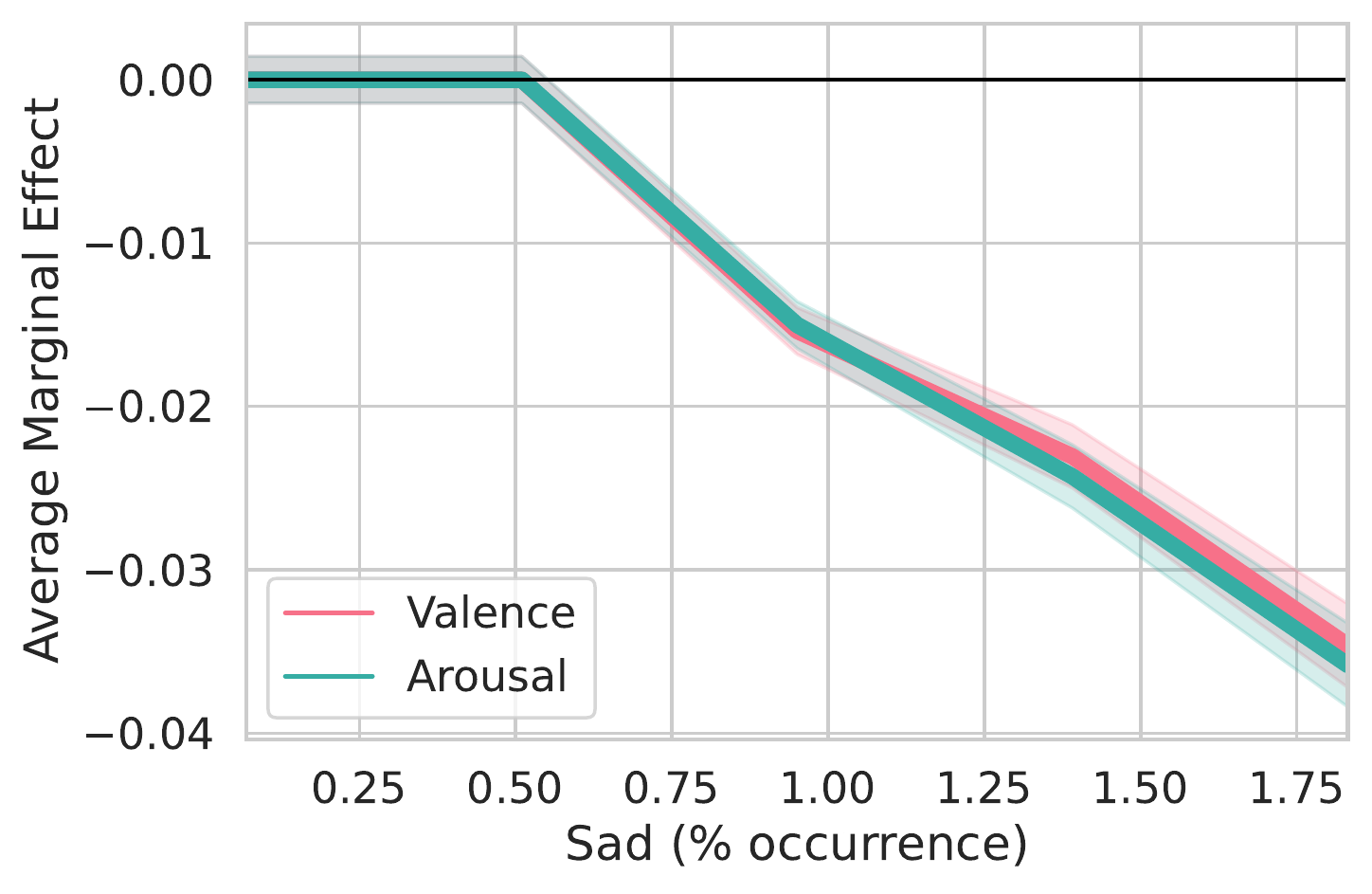}} \\
    \end{tabular}
    \caption{
    Average marginal effects of LIWC psycholinguistic lexical category \textbf{lyrical features} on listener affective responses, controlling for musical features and listener demographics. With the intent to reduce noise at the extremities, x-axis limits are capped at their 95\% quantile values.
    Arranged in alphabetical order, standard errors are shown;
    \textcolor{red}{valence} in \textcolor{red}{red}, \textcolor{blue}{arousal} in \textcolor{blue}{blue} (Part 3/4).
    }
    \label{fig:lyricfeatures_expanded_3}
\end{figure*}

\begin{figure*}[!t]
    \centering
    \begin{tabular}{ccc}
    {\includegraphics[width=0.30\textwidth]{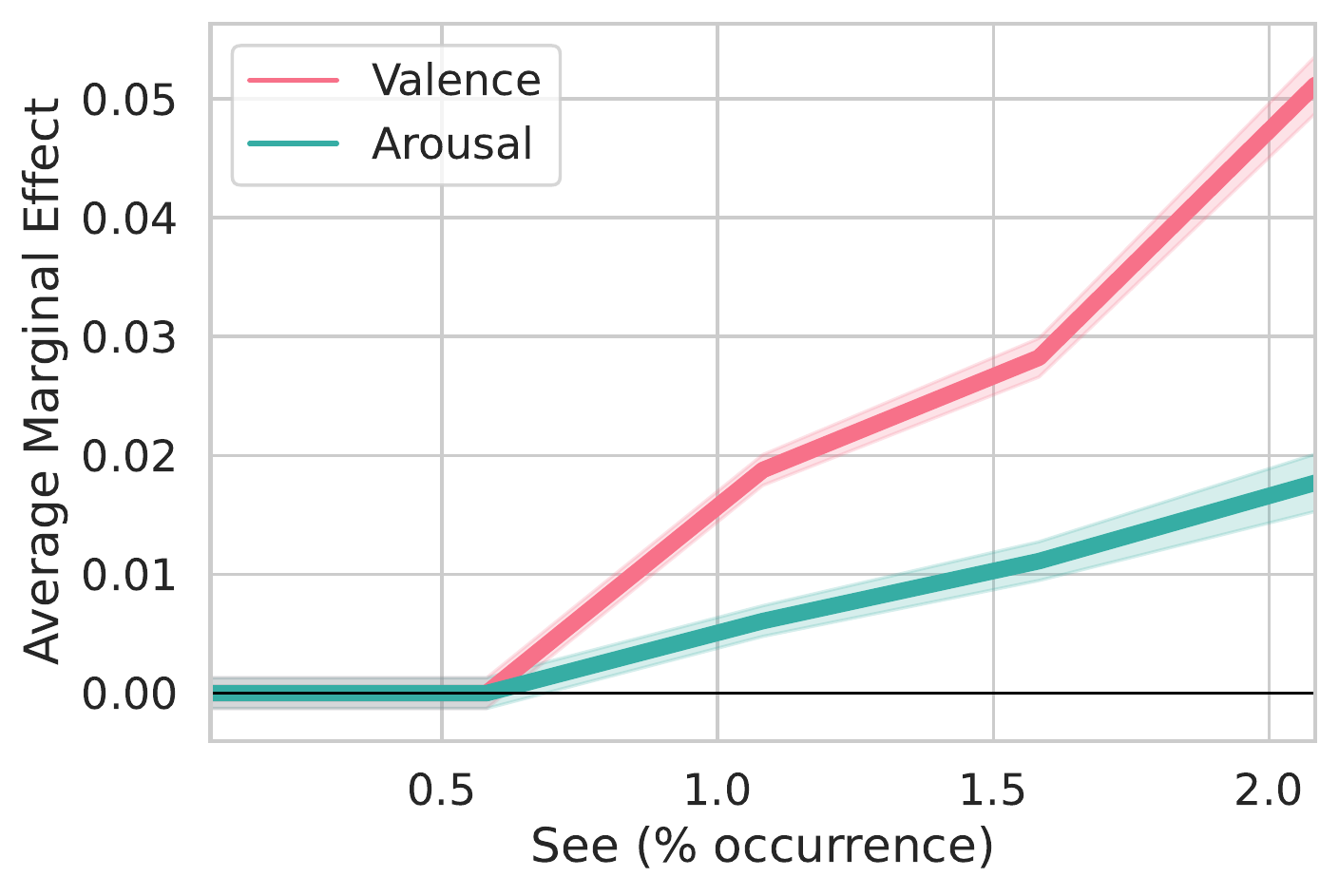}} & 
    {\includegraphics[width=0.30\textwidth]{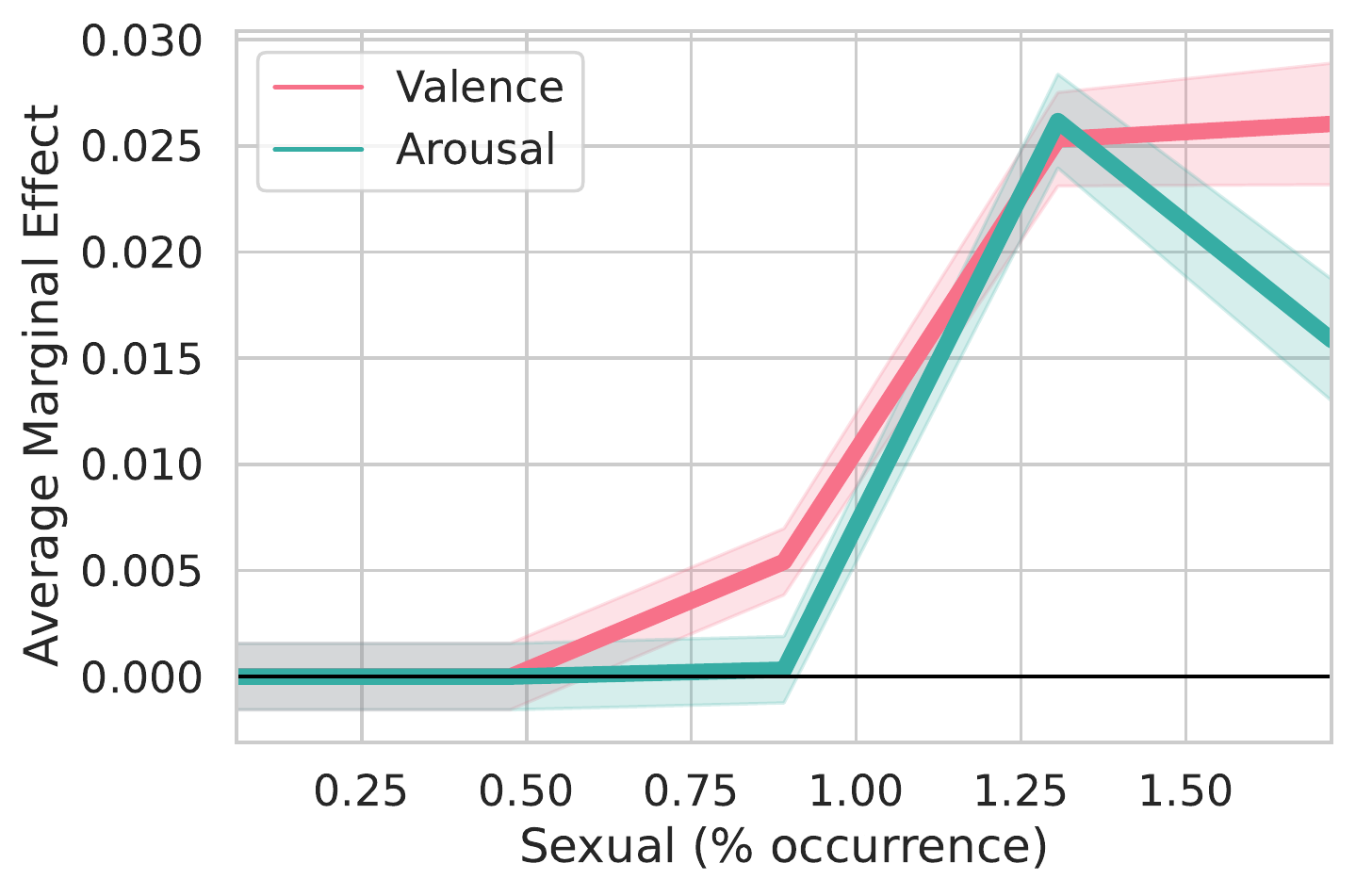}} &
    {\includegraphics[width=0.30\textwidth]{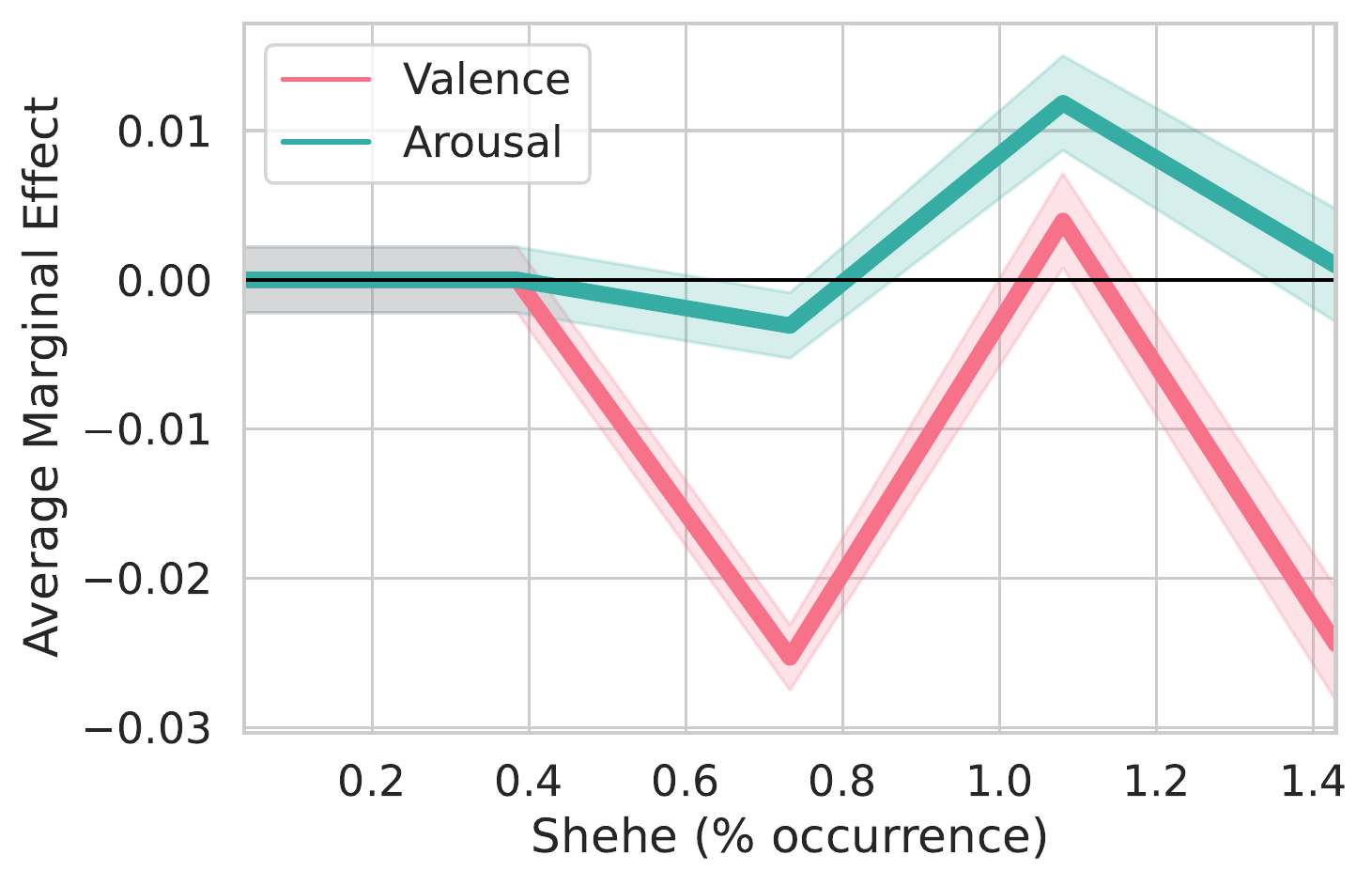}} \\
    {\includegraphics[width=0.30\textwidth]{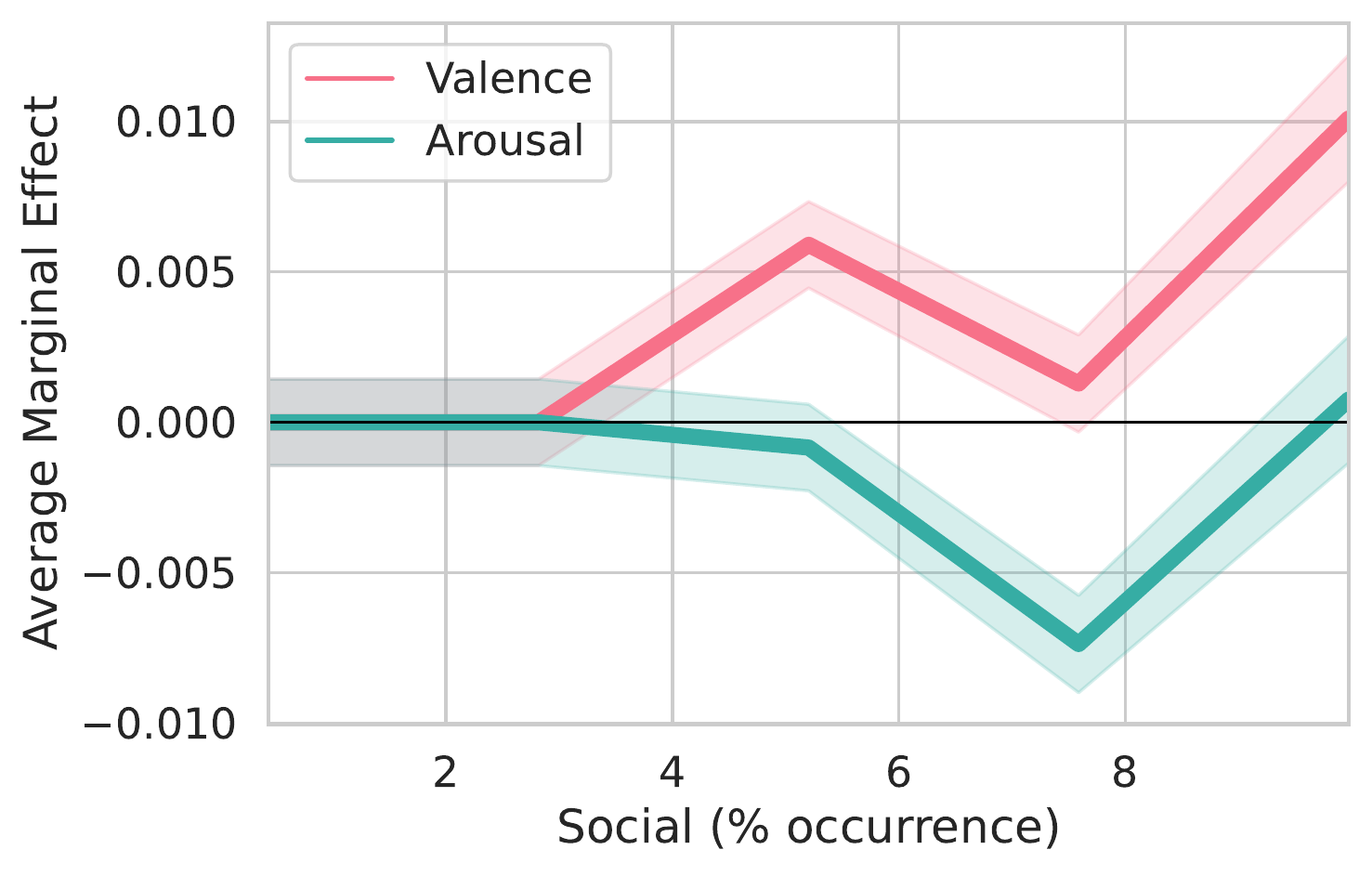}} & 
    {\includegraphics[width=0.30\textwidth]{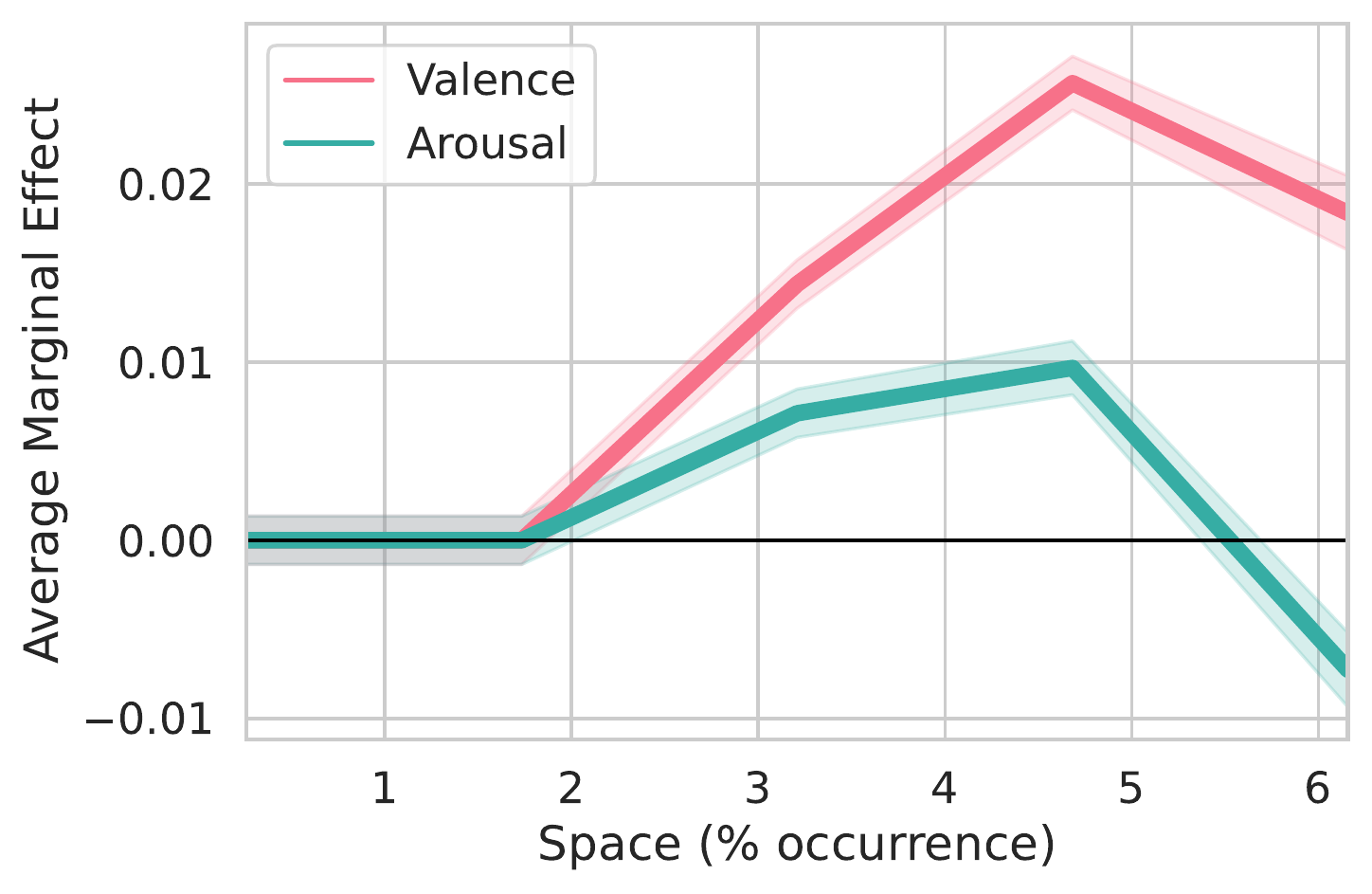}} & 
    {\includegraphics[width=0.30\textwidth]{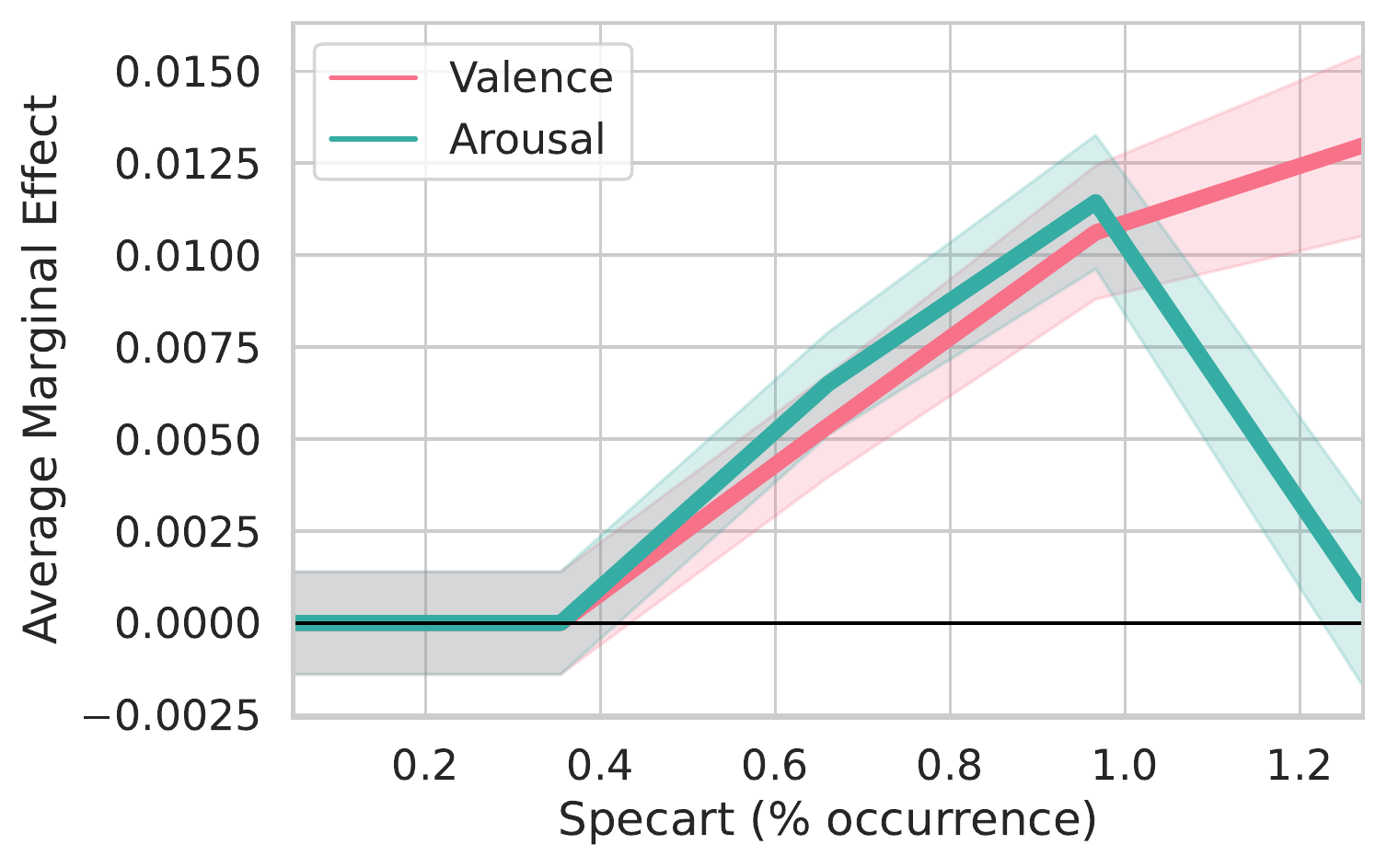}} \\ 
    {\includegraphics[width=0.30\textwidth]{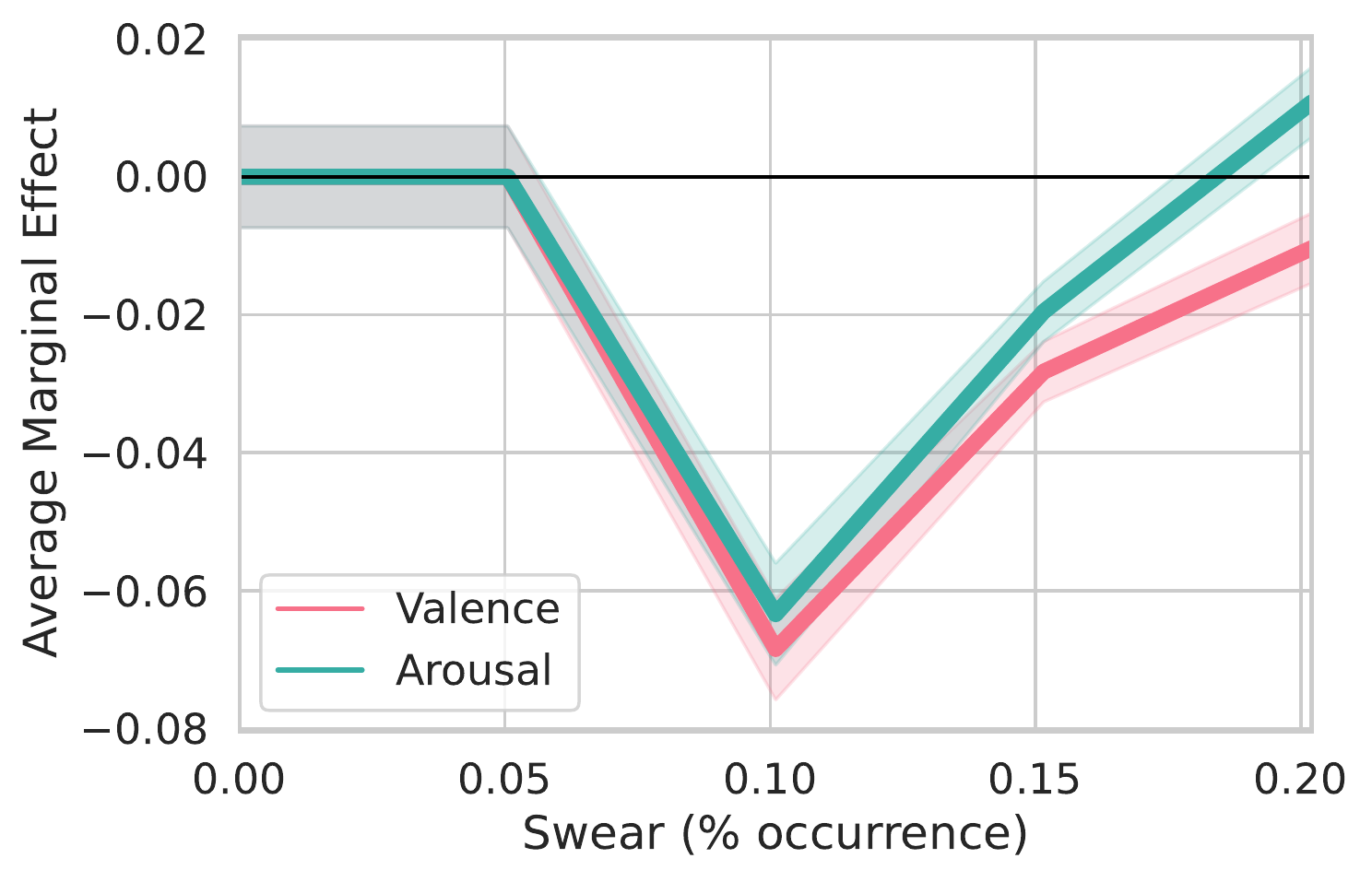}} &
    {\includegraphics[width=0.30\textwidth]{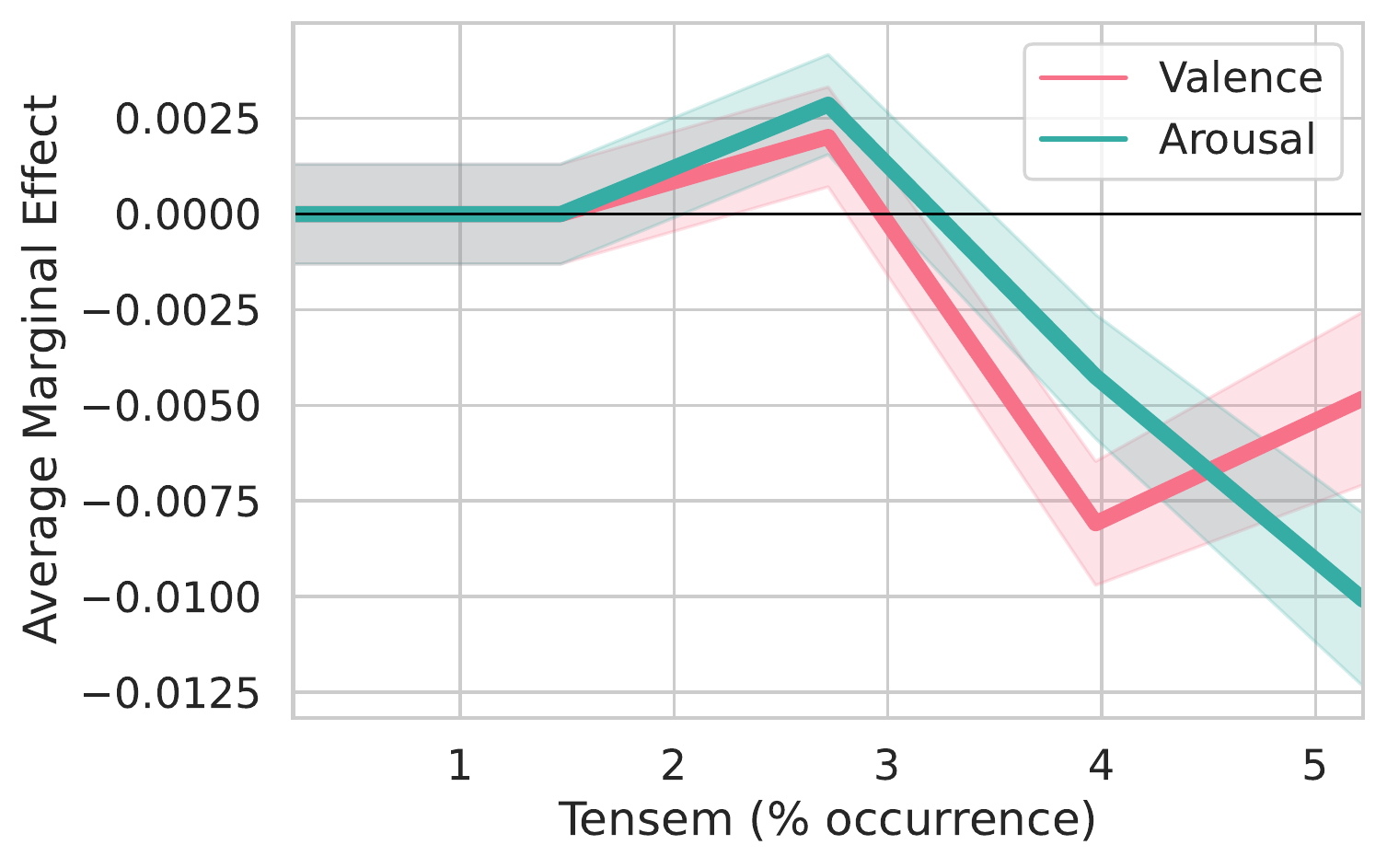}} &
    {\includegraphics[width=0.30\textwidth]{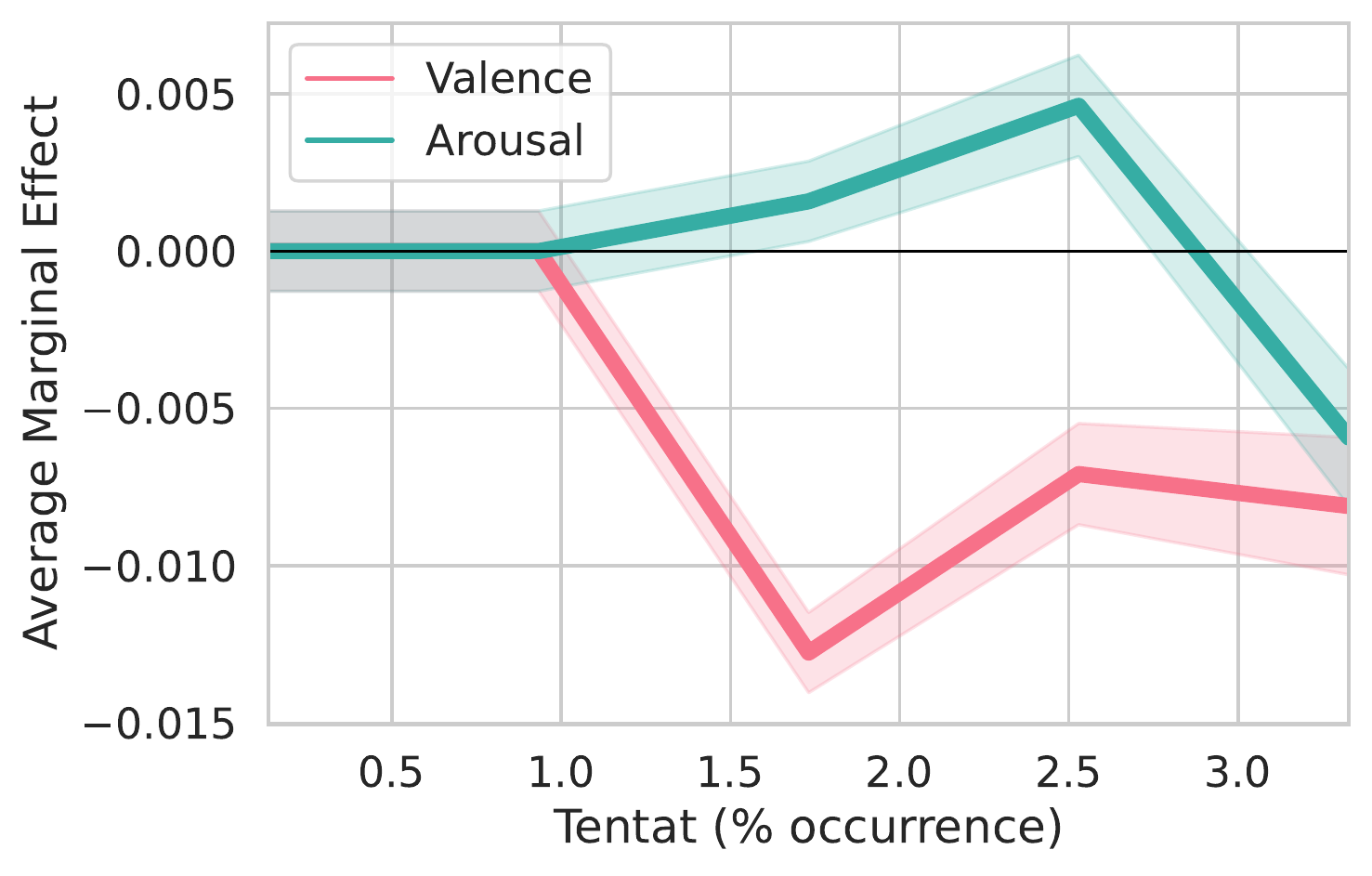}} \\ 
    {\includegraphics[width=0.30\textwidth]{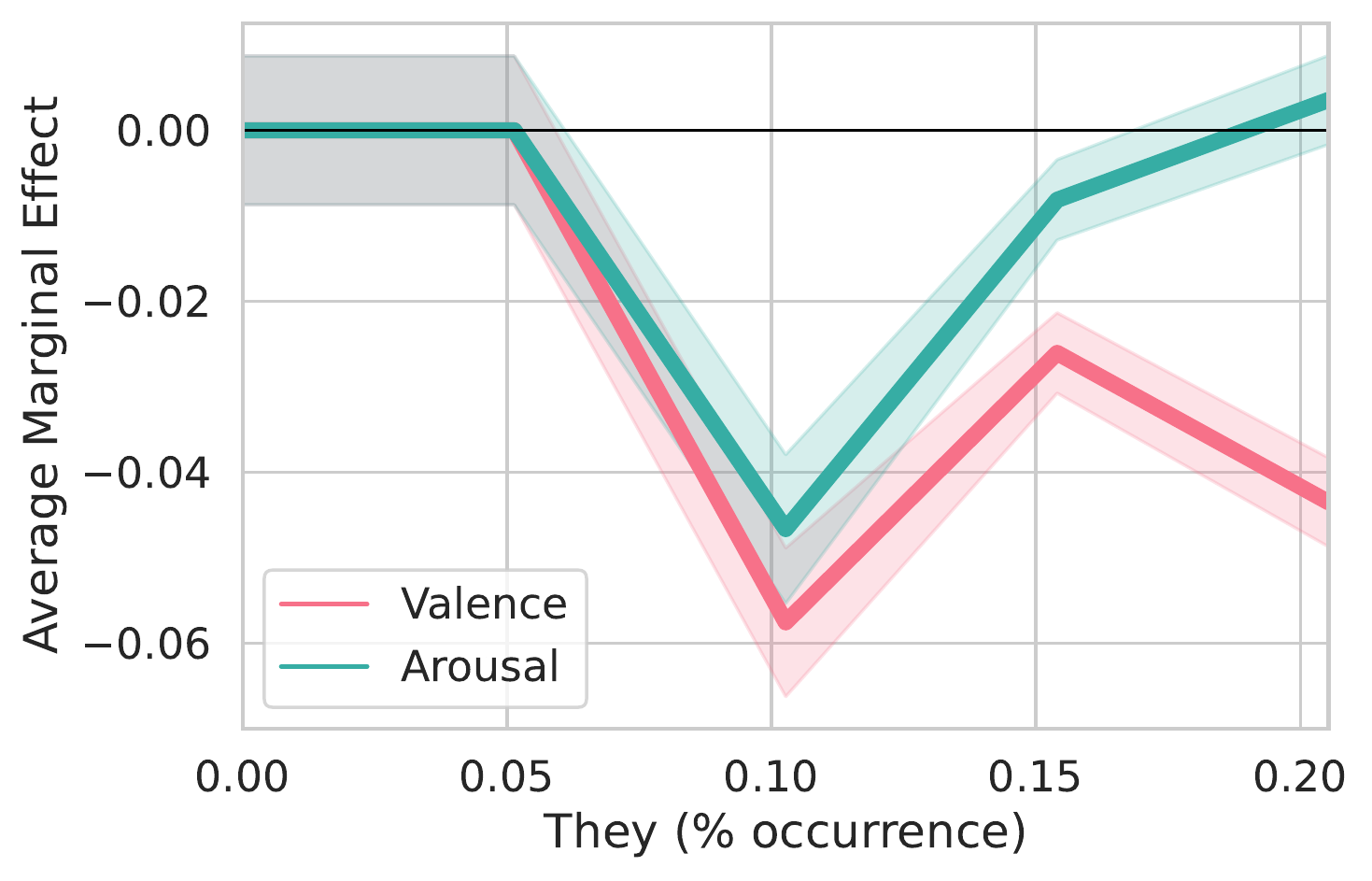}} & 
    {\includegraphics[width=0.30\textwidth]{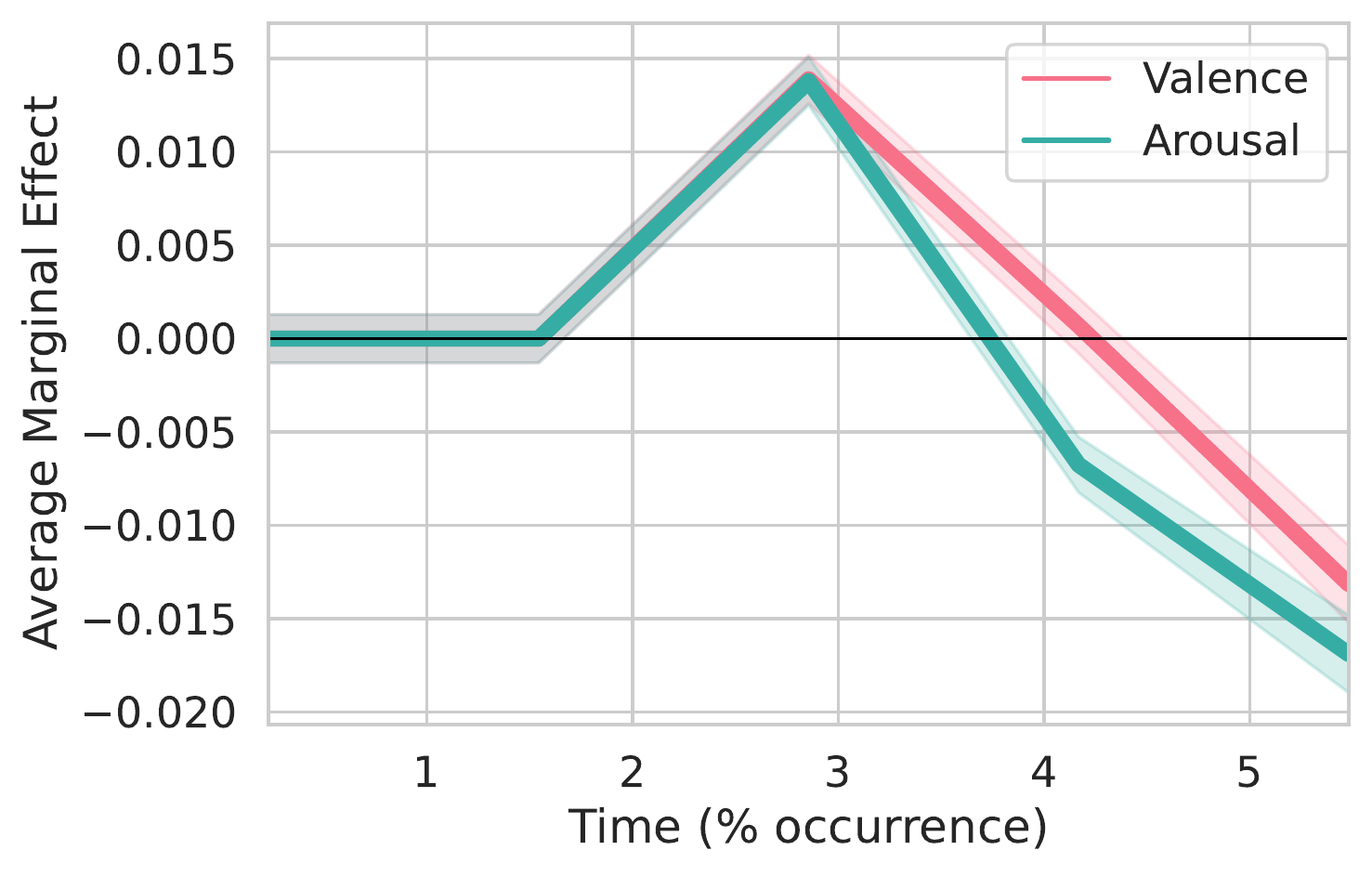}} &
    {\includegraphics[width=0.30\textwidth]{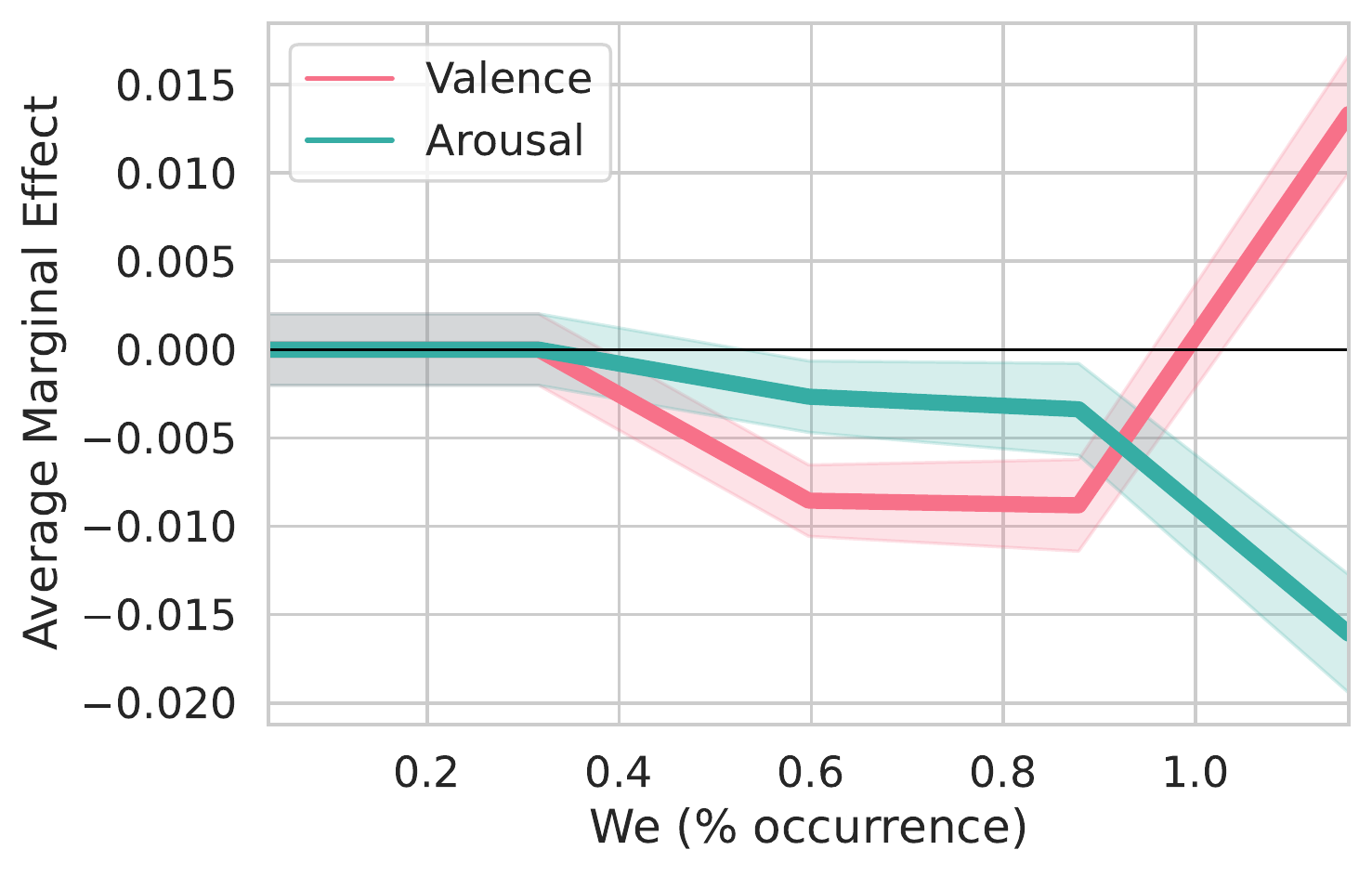}} \\
    {\includegraphics[width=0.30\textwidth]{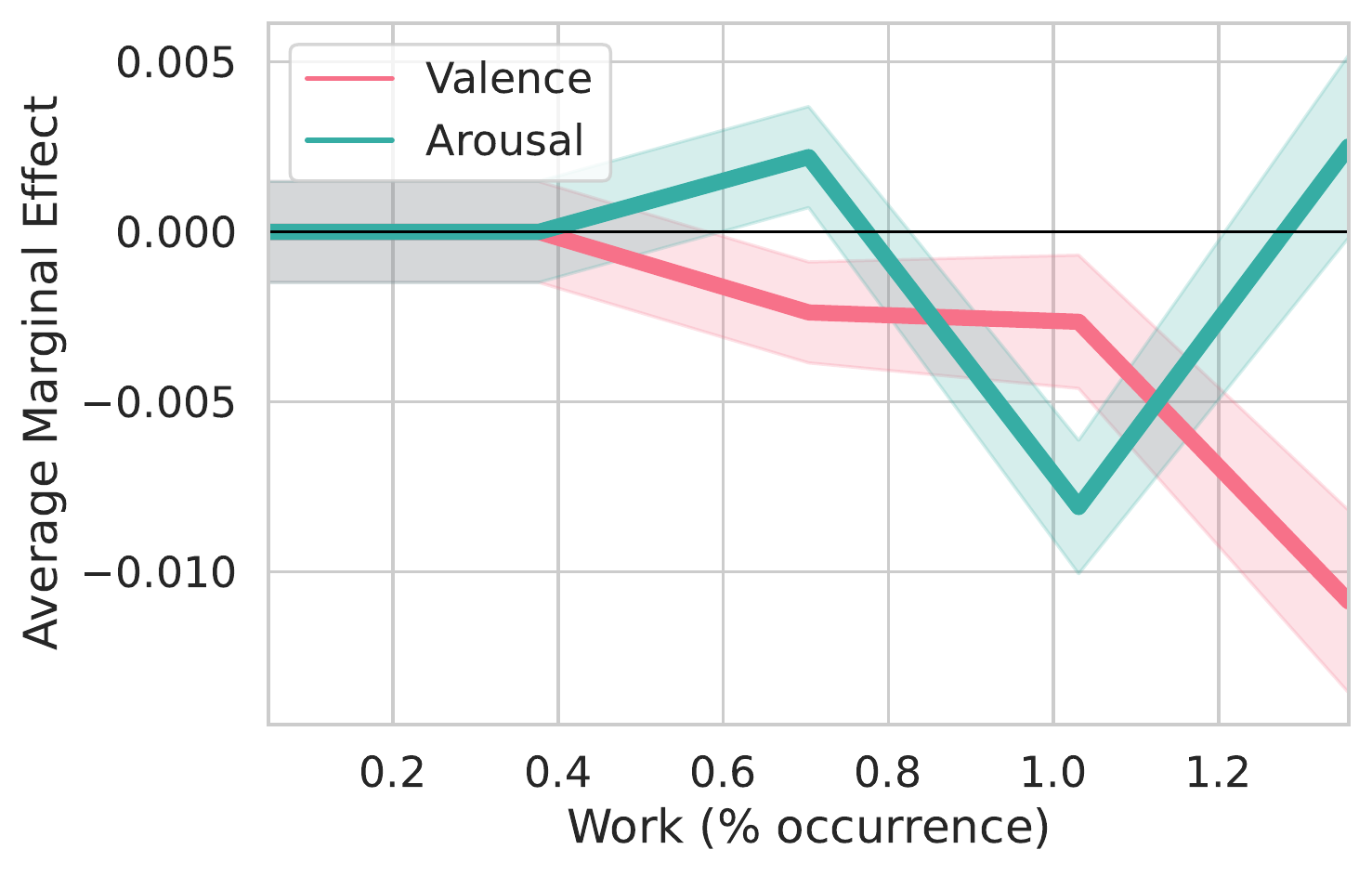}} & 
    {\includegraphics[width=0.30\textwidth]{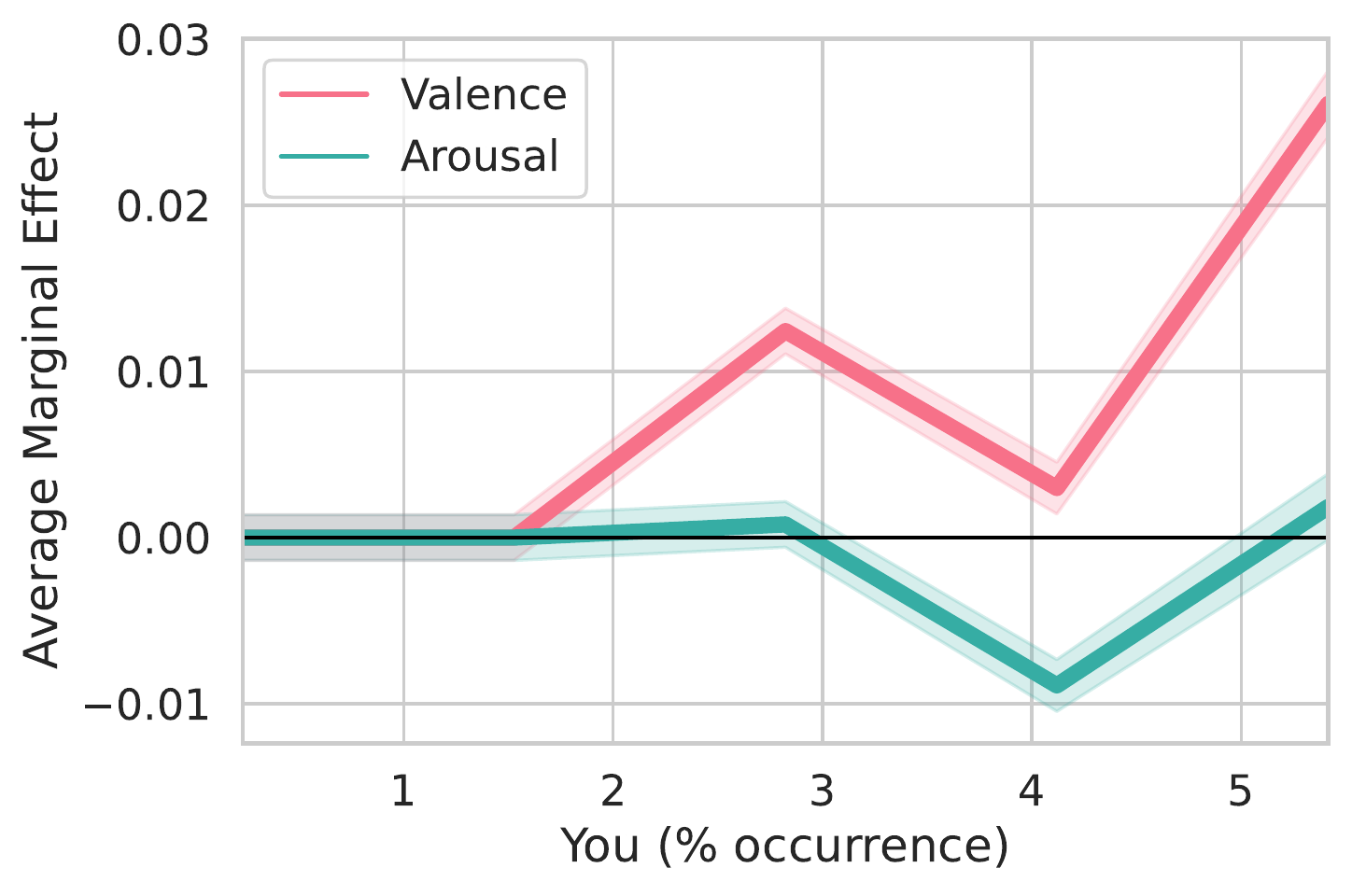}} & 
    \end{tabular}
    \caption{
    Average marginal effects of LIWC psycholinguistic lexical category \textbf{lyrical features} on listener affective responses, controlling for musical features and listener demographics. With the intent to reduce noise at the extremities, x-axis limits are capped at their 95\% quantile values.
    Arranged in alphabetical order, standard errors are shown;
    \textcolor{red}{valence} in \textcolor{red}{red}, \textcolor{blue}{arousal} in \textcolor{blue}{blue} (Part 4/4).
    }
    \label{fig:lyricfeatures_expanded_4}
\end{figure*}

\begin{figure*}[!t]
    \centering
    \raisebox{-0.5\height}{\includegraphics[width=0.23\textwidth]{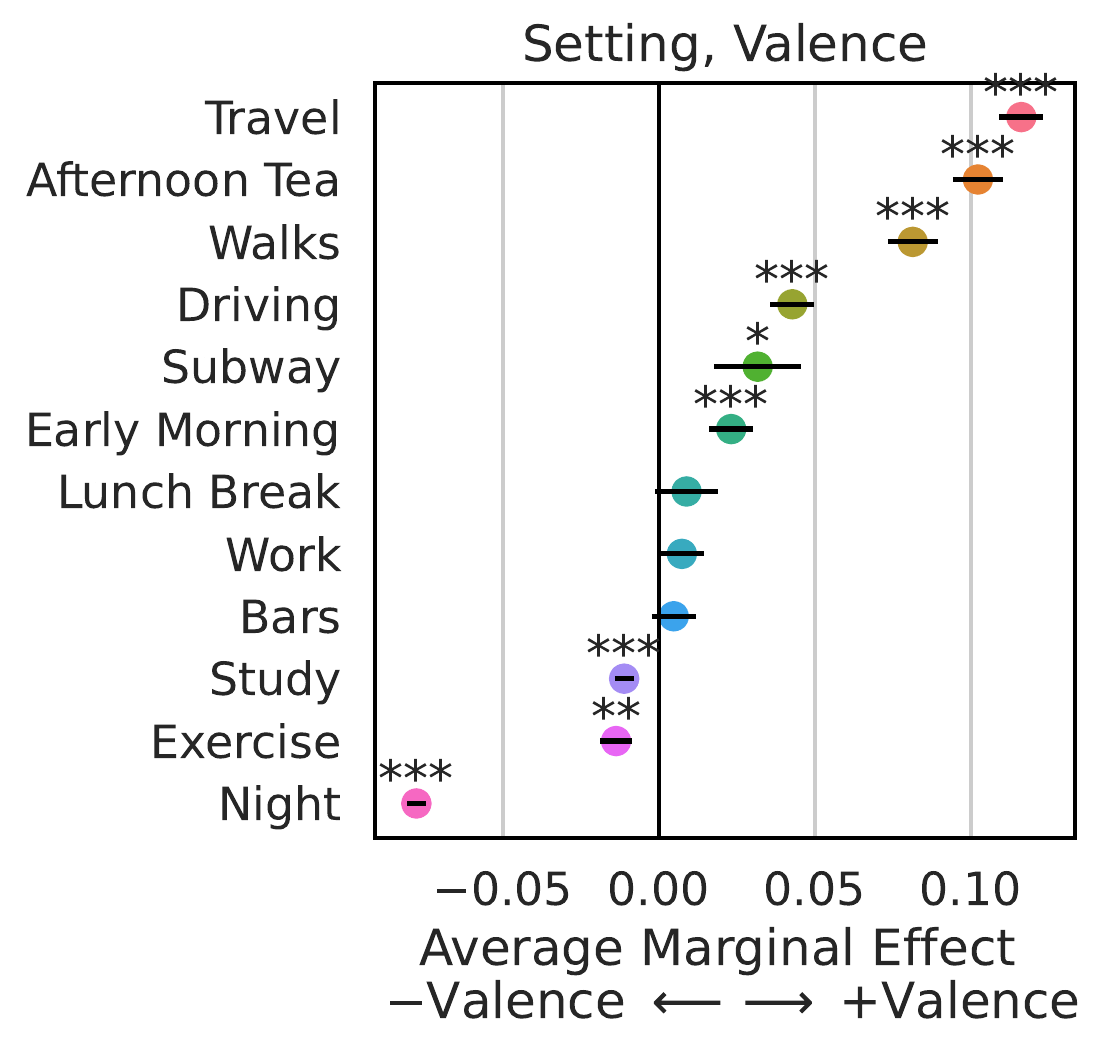}}
    \raisebox{-0.5\height}{\includegraphics[width=0.23\textwidth]{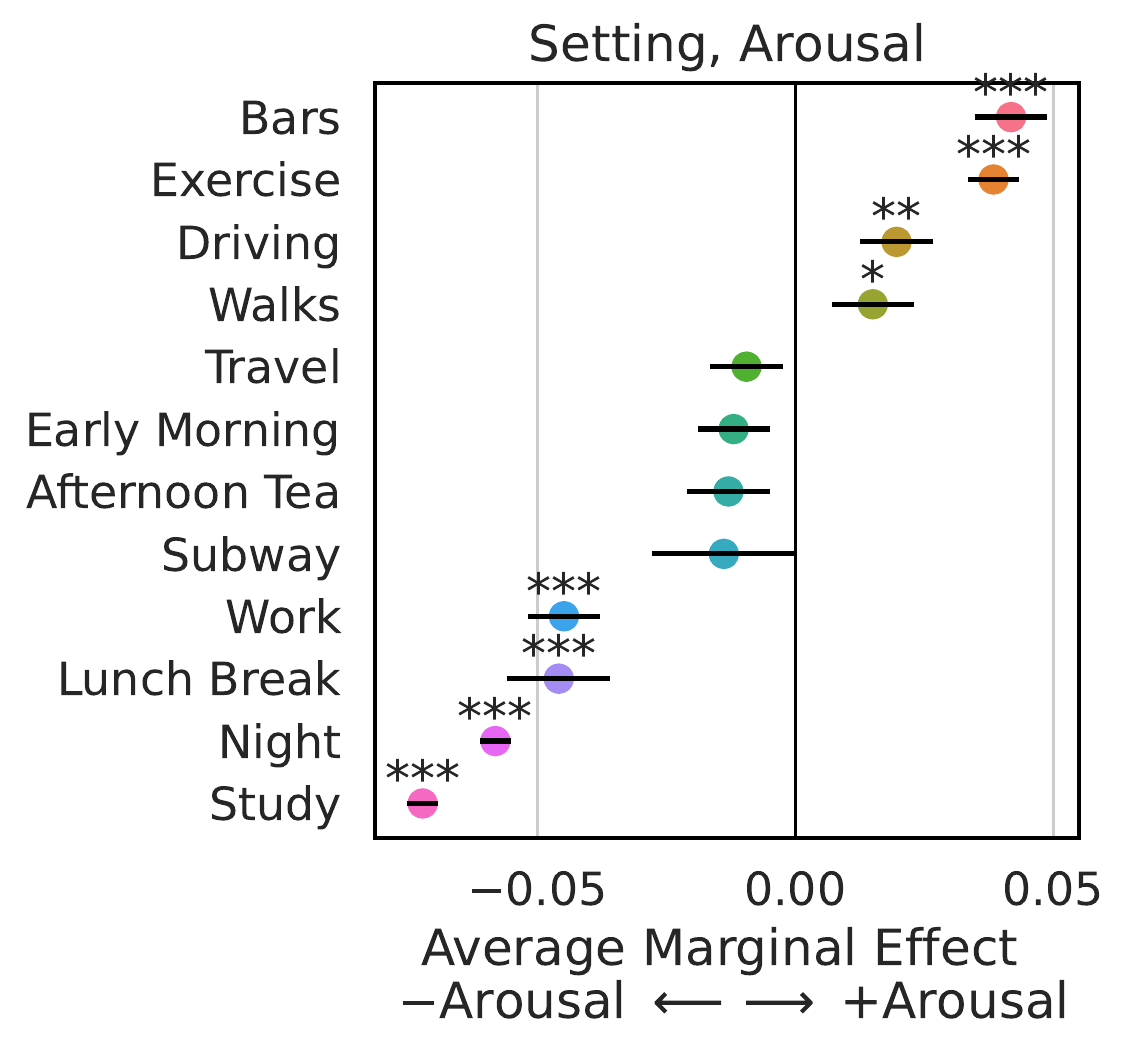}} 
    \caption{Average marginal effects of listening contexts in 
    \textbf{setting-tagged} playlists on listener affective responses, controlling for songs and user demographic variables; standard errors are shown.} 
    \label{fig:setting_overall_setting}
\end{figure*}

\begin{figure*}[!t]
    \centering
    \raisebox{-0.5\height}{\includegraphics[width=0.23\textwidth]{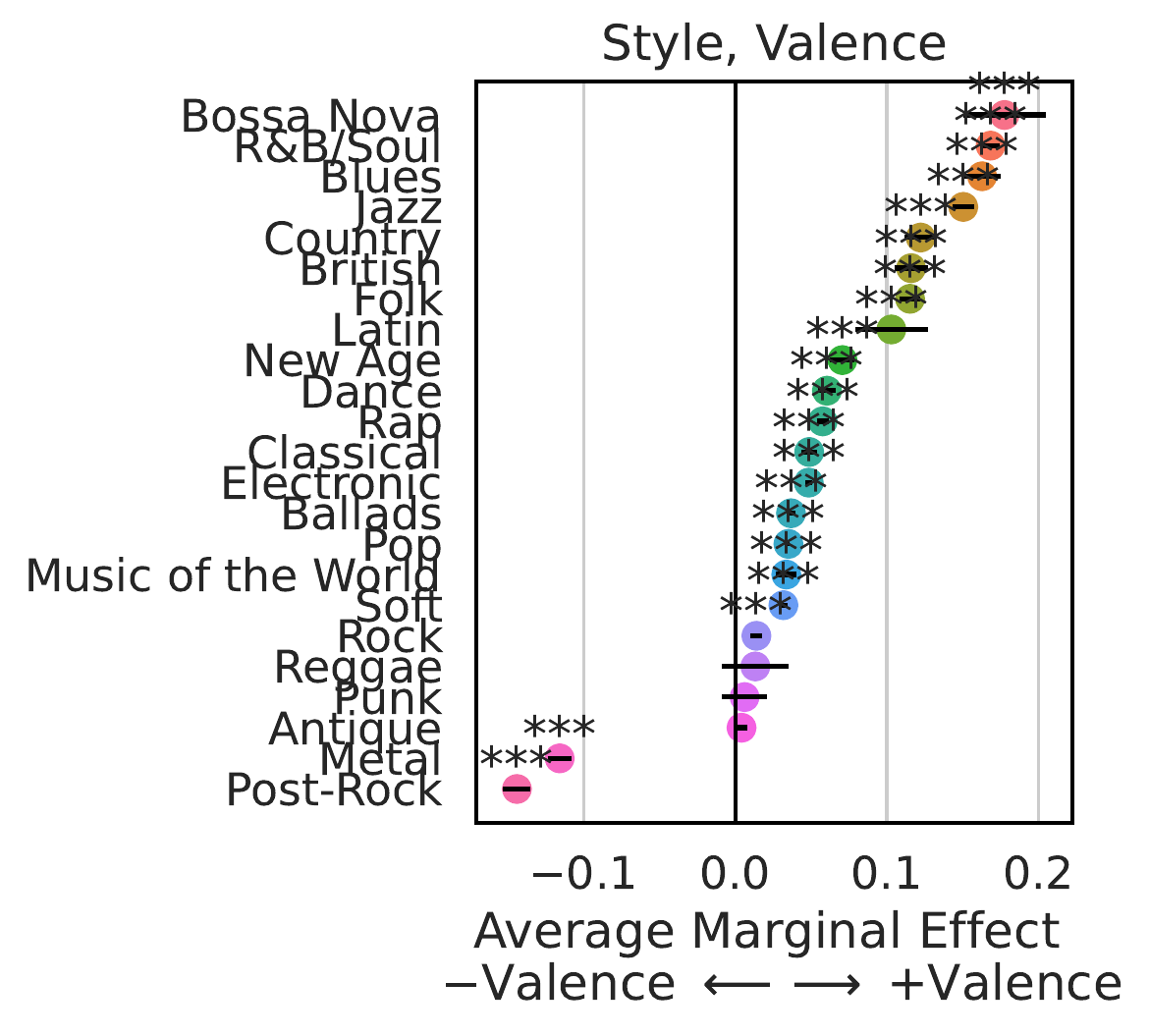}}
    \raisebox{-0.5\height}{\includegraphics[width=0.23\textwidth]{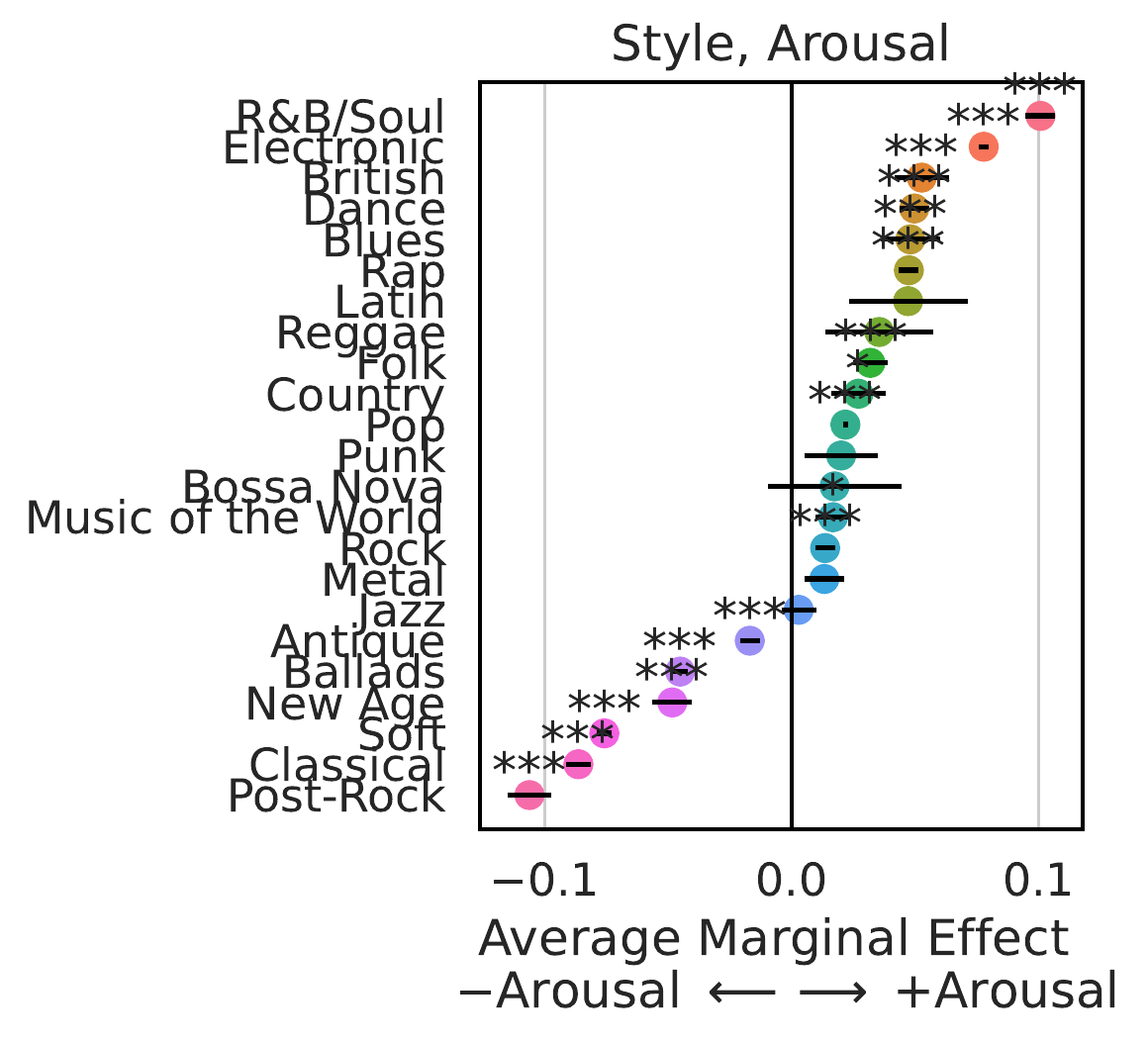}} 
    \caption{Average marginal effects of listening contexts in 
    \textbf{style-tagged} playlists on listener affective responses, controlling for songs and user demographic variables; standard errors are shown.} 
    \label{fig:setting_overall_style}
\end{figure*}

\begin{figure*}[!t]
    \centering
    \raisebox{-0.5\height}{\includegraphics[width=0.23\textwidth]{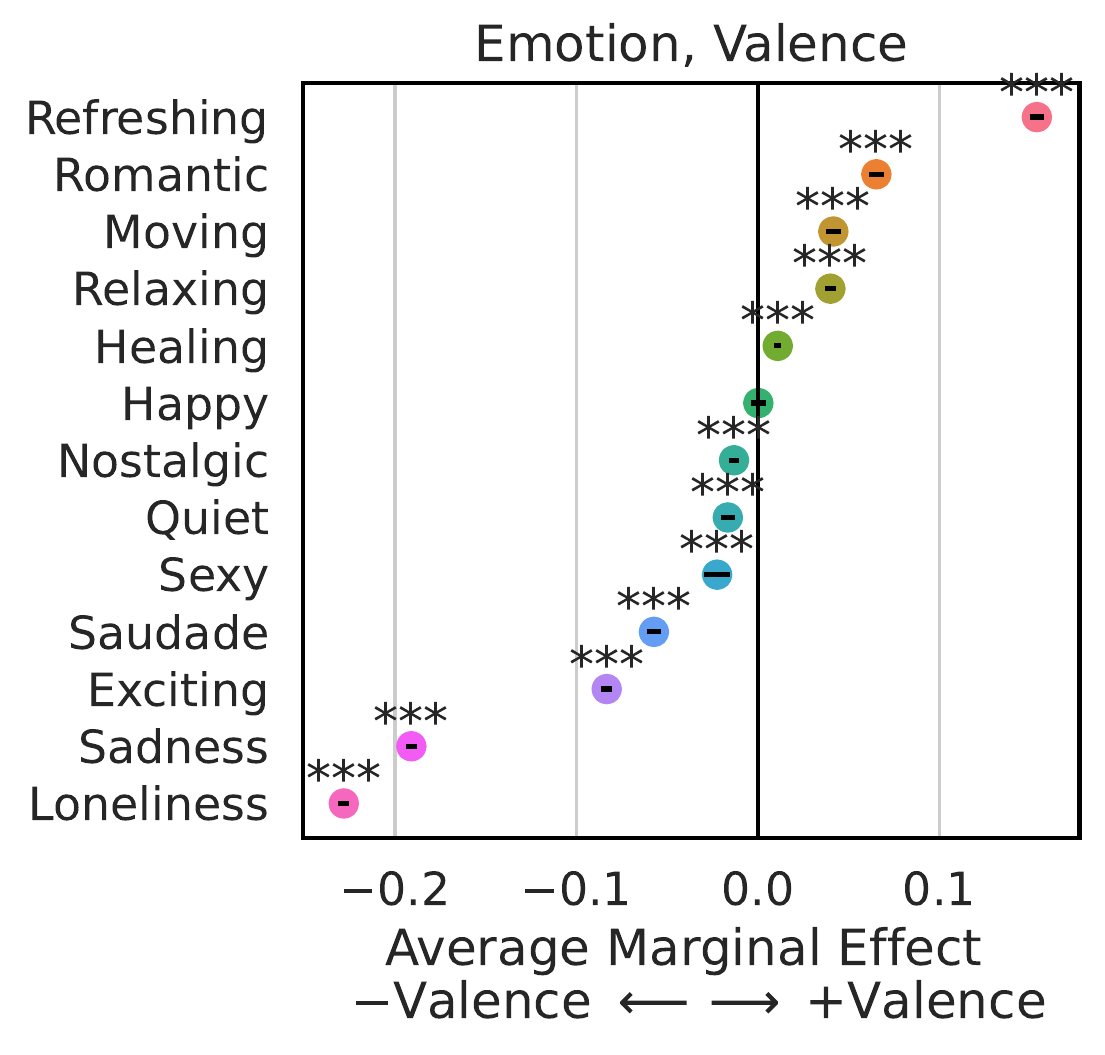}}
    \raisebox{-0.5\height}{\includegraphics[width=0.23\textwidth]{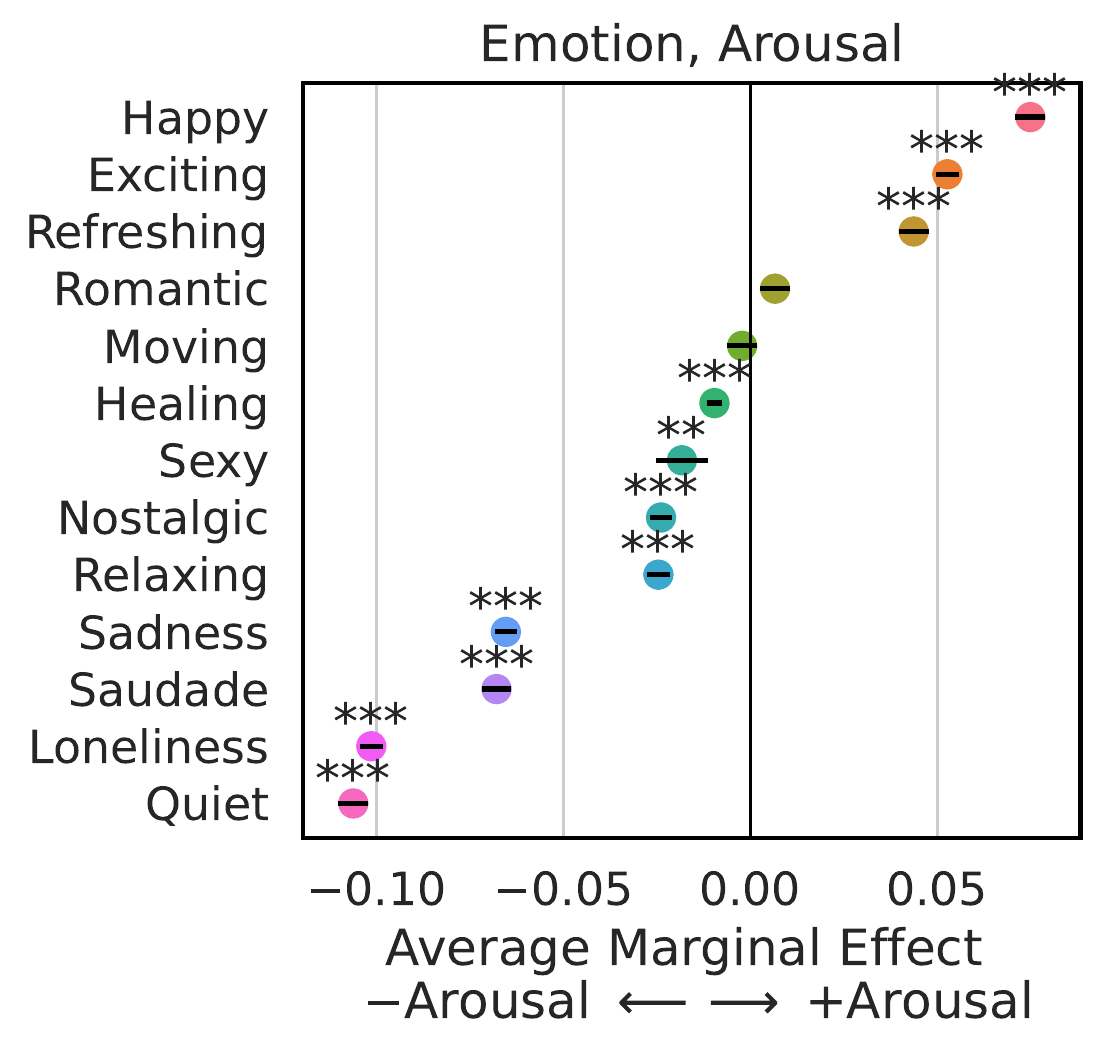}} 
    \caption{Average marginal effects of listening contexts in 
    \textbf{emotion-tagged} playlists on listener affective responses, controlling for songs and user demographic variables; standard errors are shown.} 
    \label{fig:setting_overall_emotion}
\end{figure*}

\begin{figure*}[!t]
    \centering
    \raisebox{-0.5\height}{\includegraphics[width=0.23\textwidth]{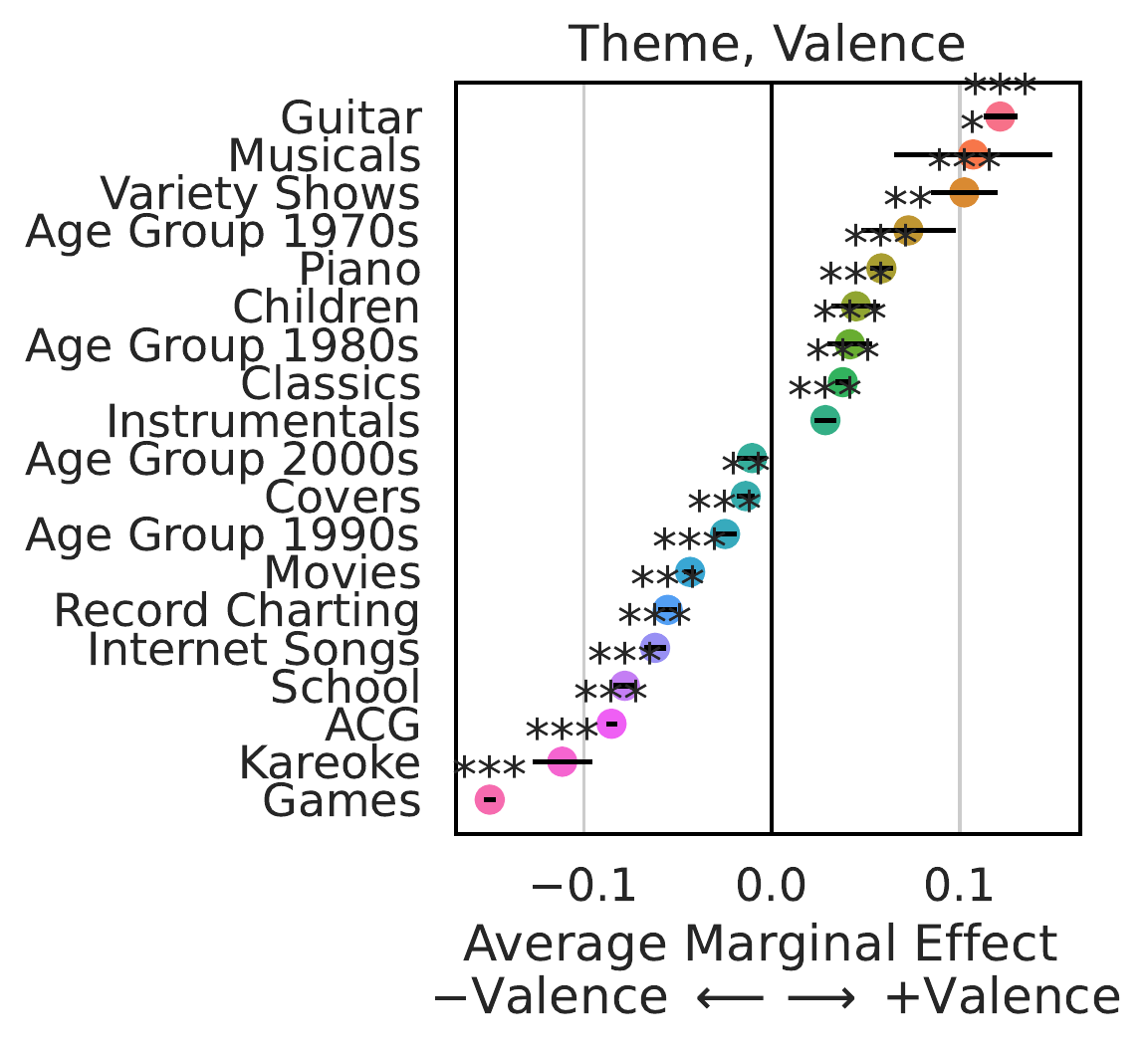}}
    \raisebox{-0.5\height}{\includegraphics[width=0.23\textwidth]{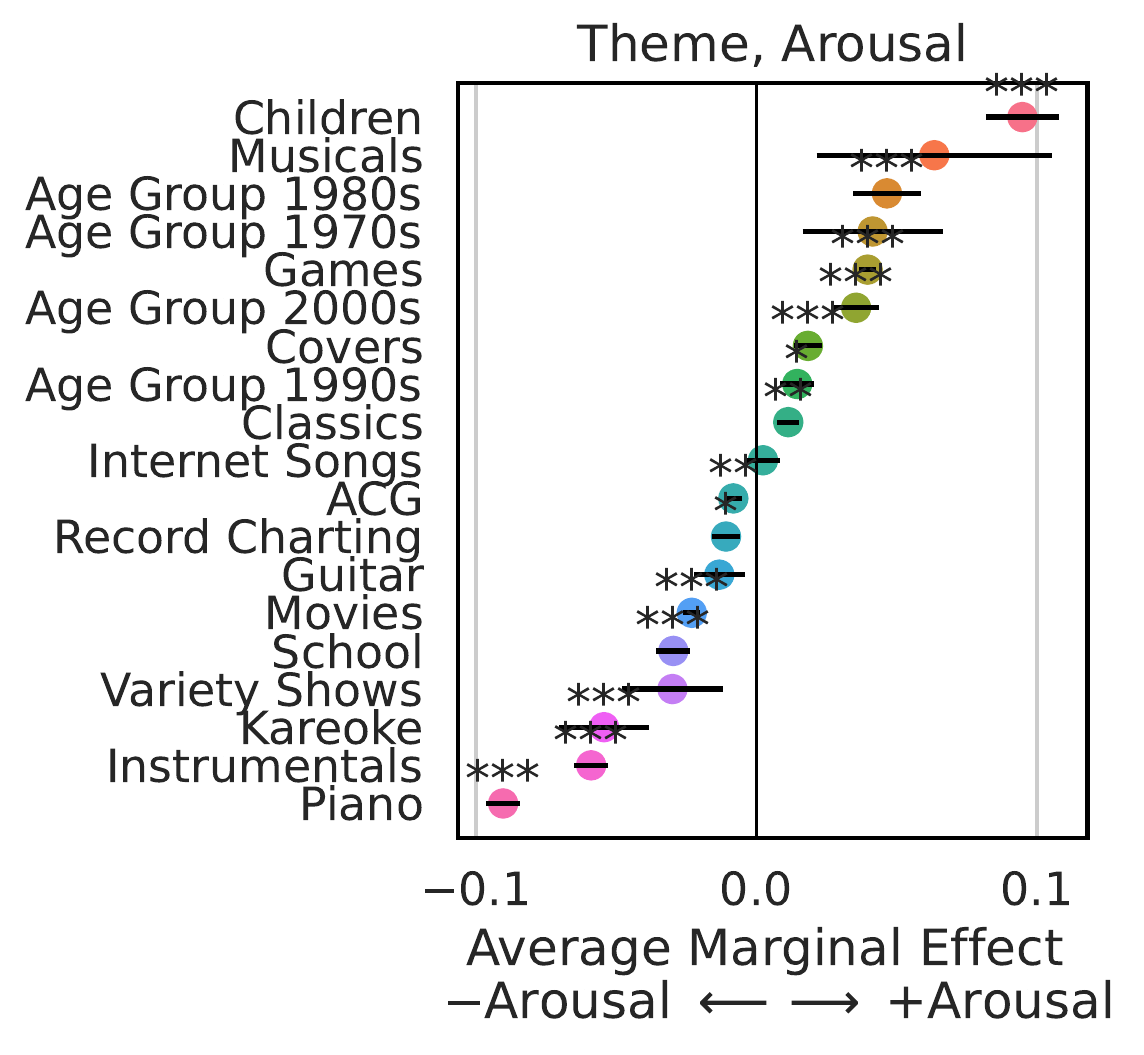}} 
    \caption{Average marginal effects of listening contexts in 
    \textbf{theme-tagged} playlists on listener affective responses, controlling for songs and user demographic variables; standard errors are shown.} 
    \label{fig:setting_overall_theme}
\end{figure*}

\begin{figure*}[!t]
    \centering
    \raisebox{-0.5\height}{\includegraphics[width=0.23\textwidth]{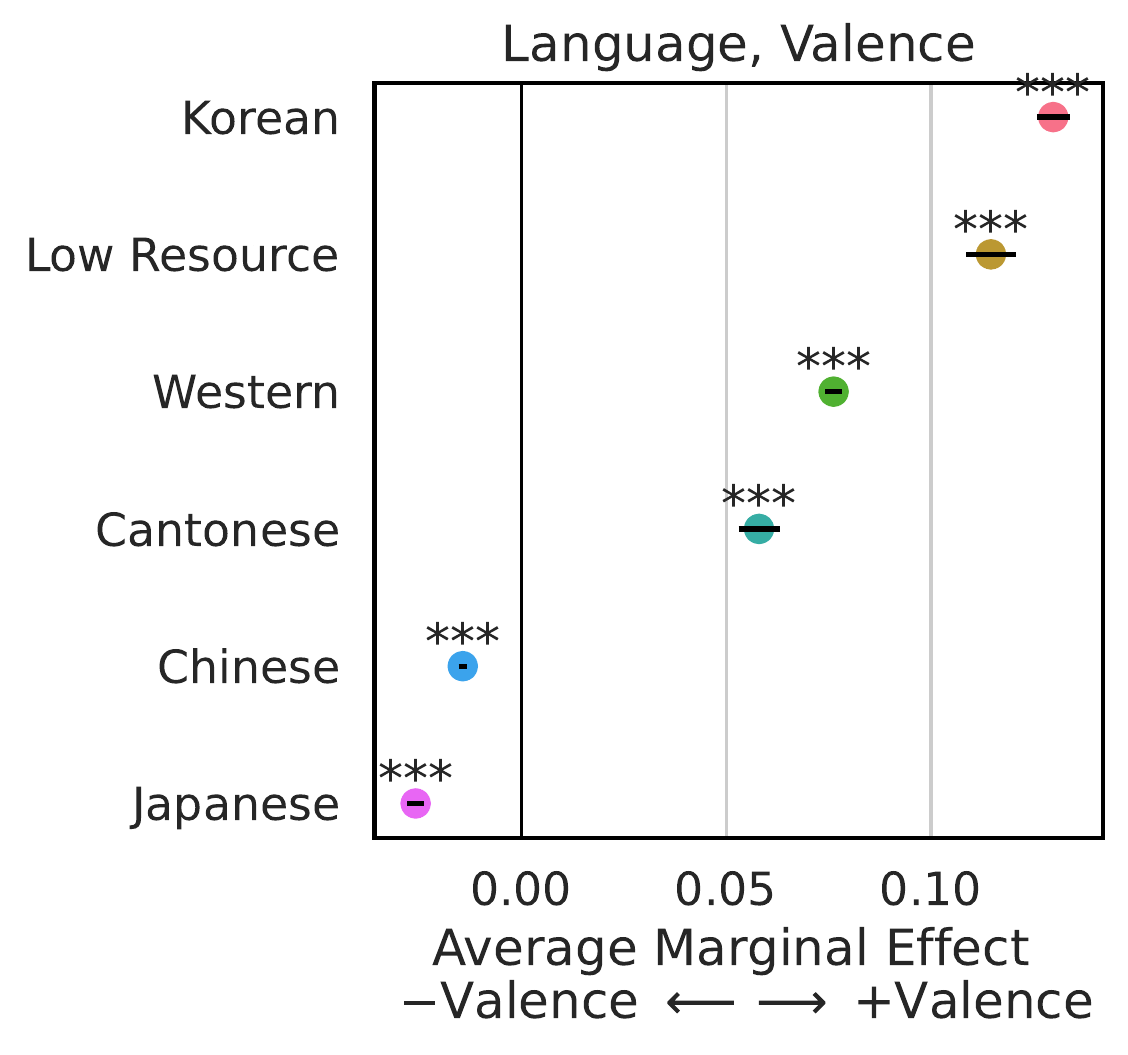}}
    \raisebox{-0.5\height}{\includegraphics[width=0.23\textwidth]{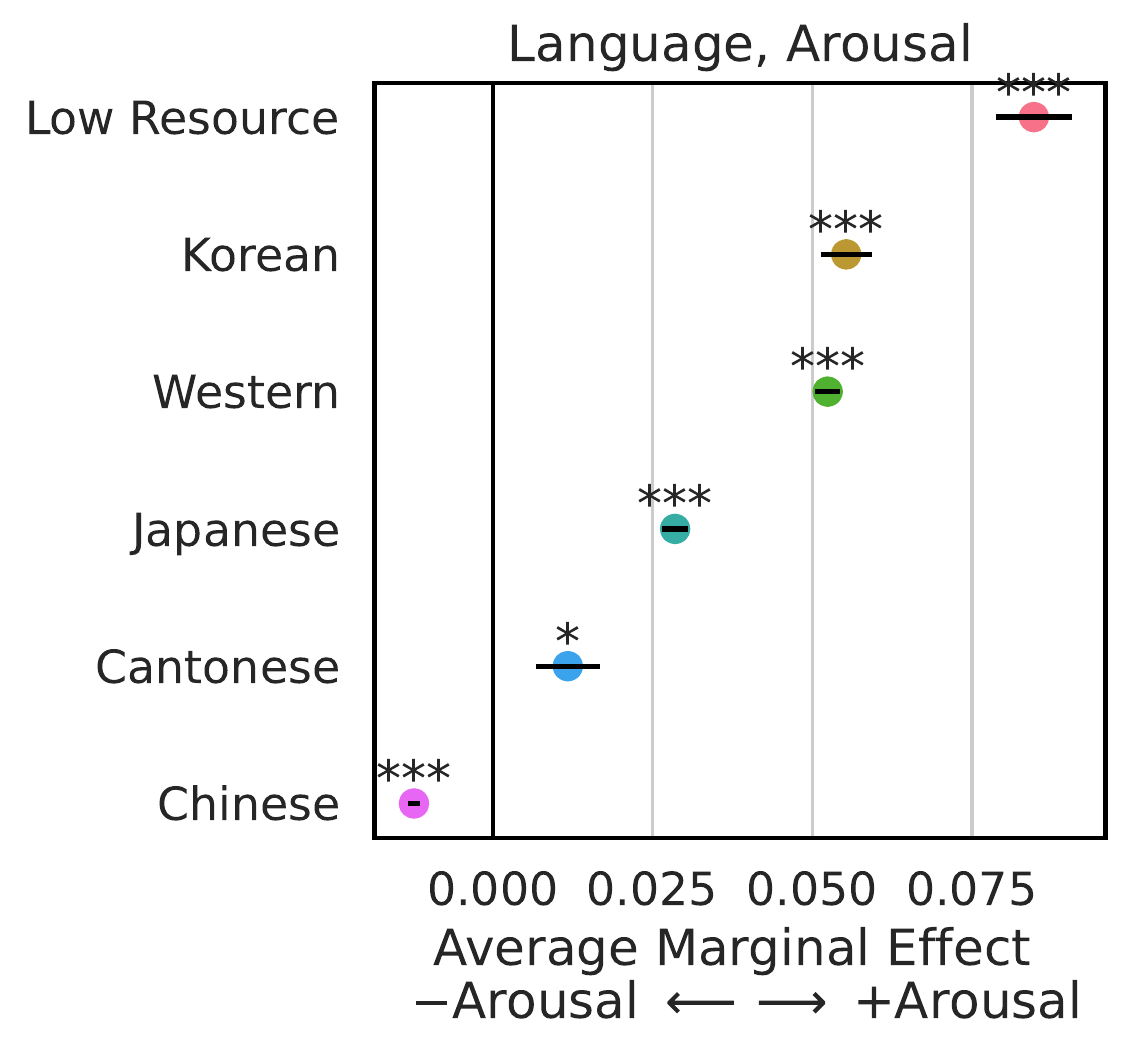}} 
    \caption{Average marginal effects of listening contexts in 
    \textbf{language-tagged} playlists on listener affective responses, controlling for songs and user demographic variables; standard errors are shown.} 
    \label{fig:setting_overall_language}
\end{figure*}

\begin{figure*}[!t]
    \centering
    \begin{tabular}{ccc}
    {\includegraphics[width=0.30\textwidth]{plots/ATE/women_men_mp3_features_tempo.pdf}} & 
    {\includegraphics[width=0.30\textwidth]{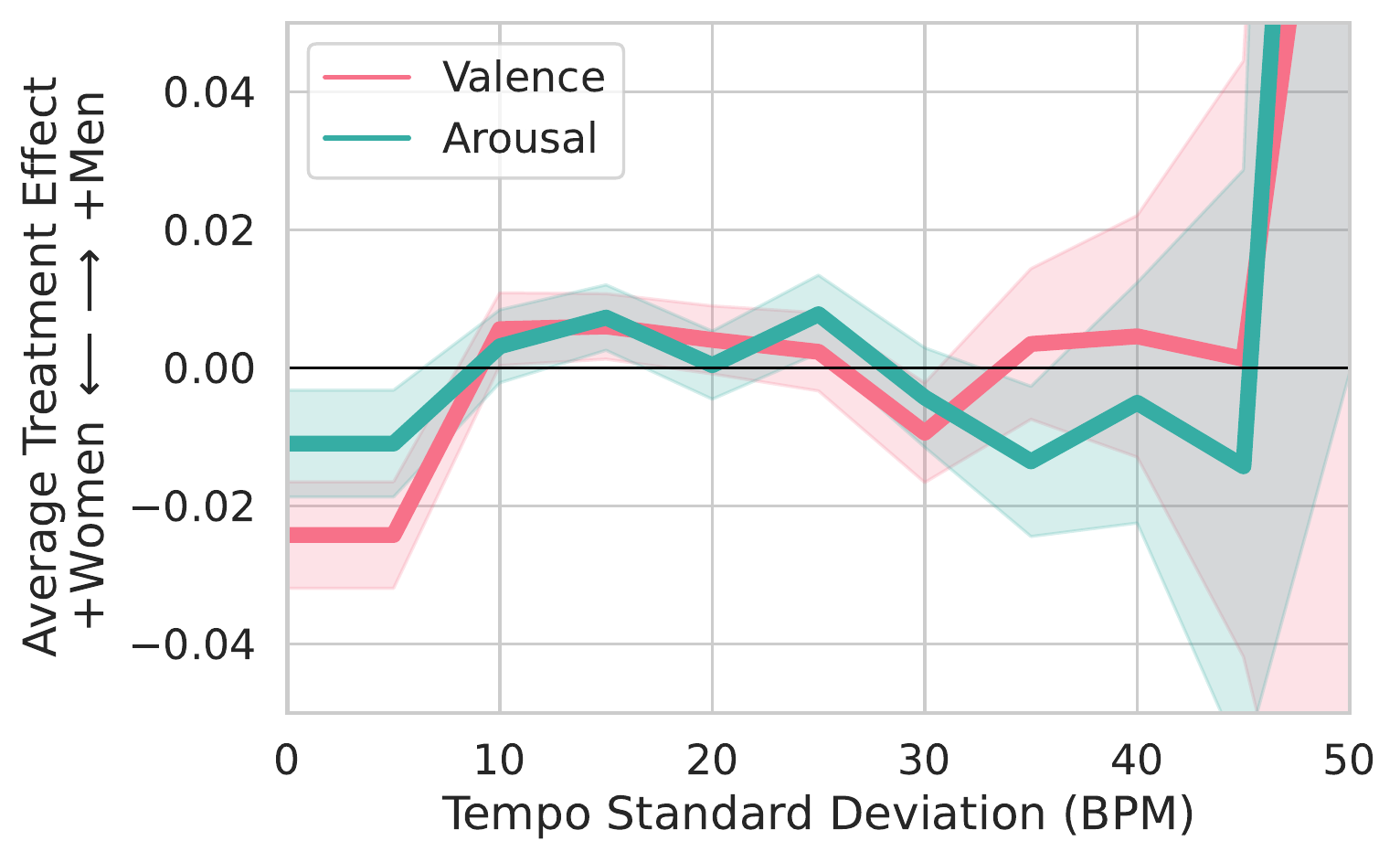}} &
    {\includegraphics[width=0.30\textwidth]{plots/ATE/women_men_mp3_features_loudness.pdf}} \\ 
    {\includegraphics[width=0.30\textwidth]{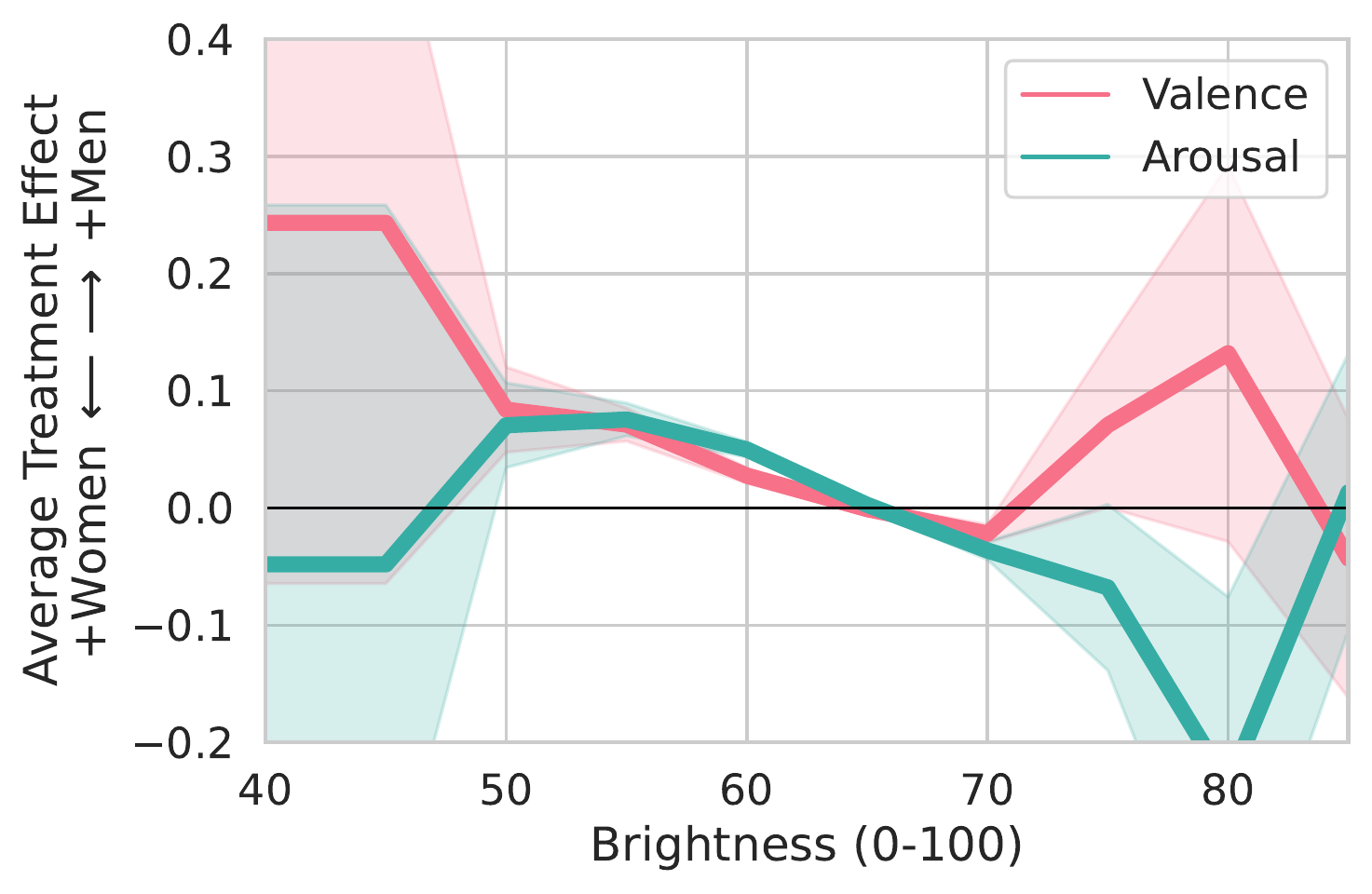}} &
    {\includegraphics[width=0.30\textwidth]{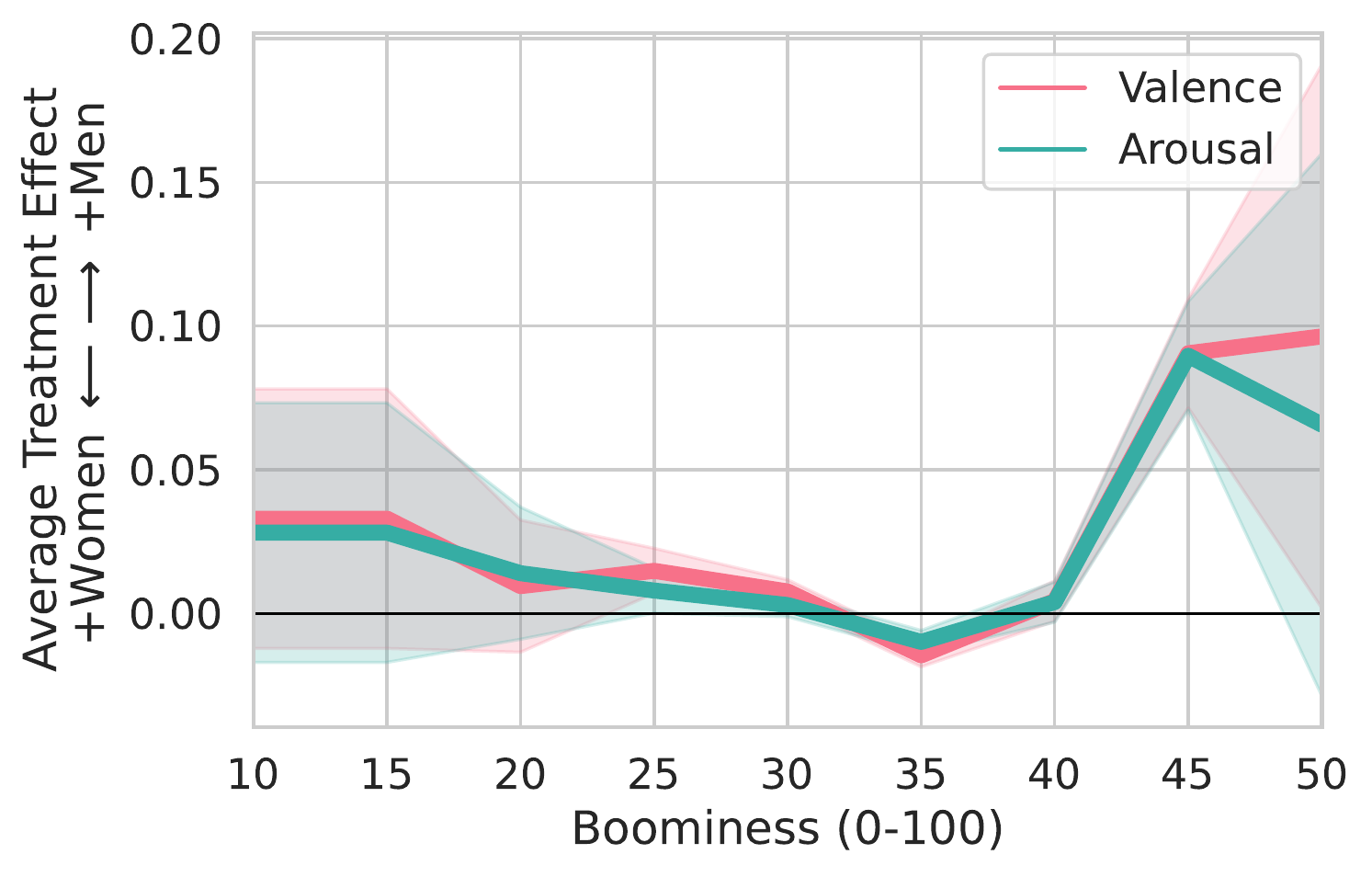}} &
    {\includegraphics[width=0.30\textwidth]{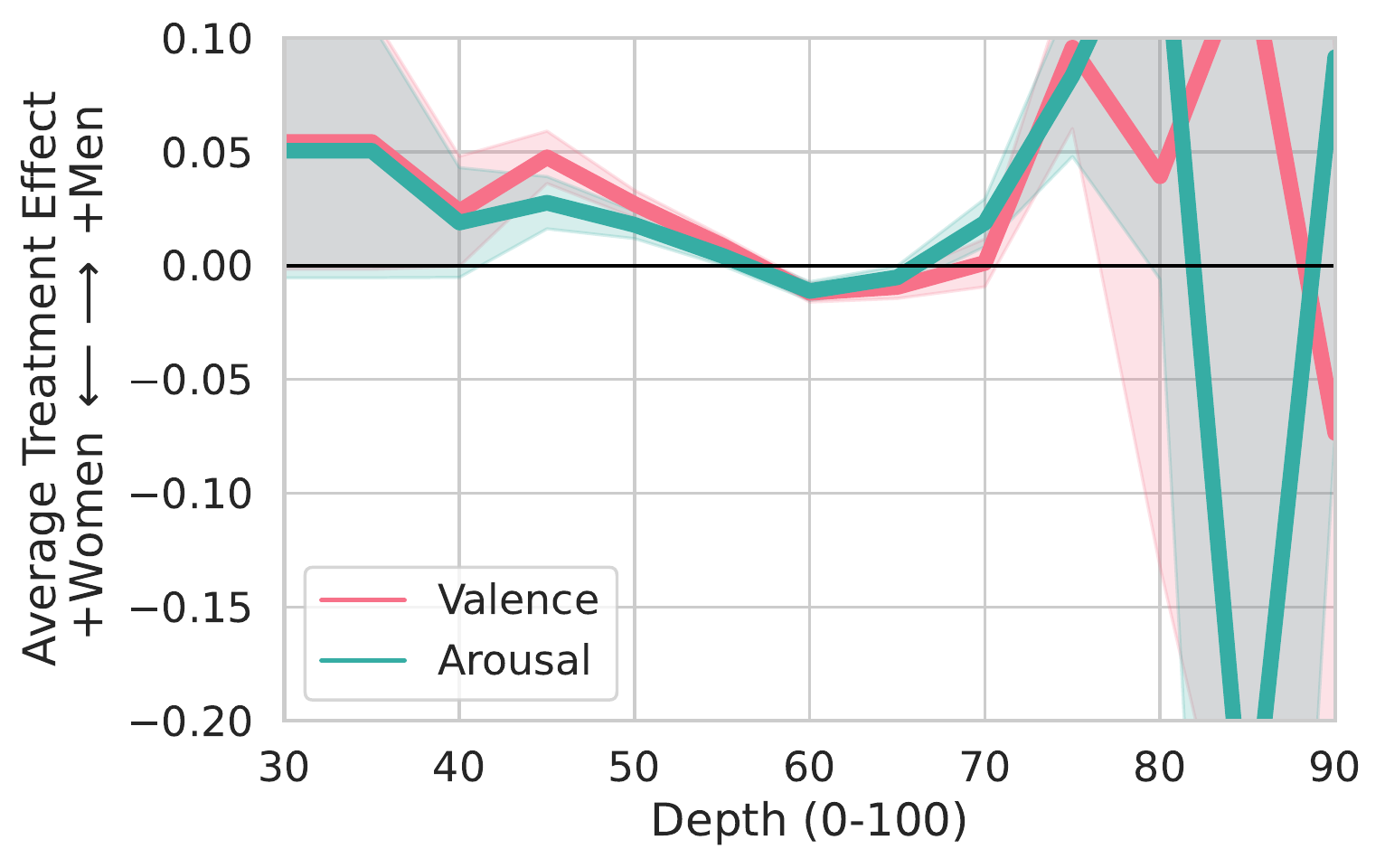}} \\ 
    {\includegraphics[width=0.30\textwidth]{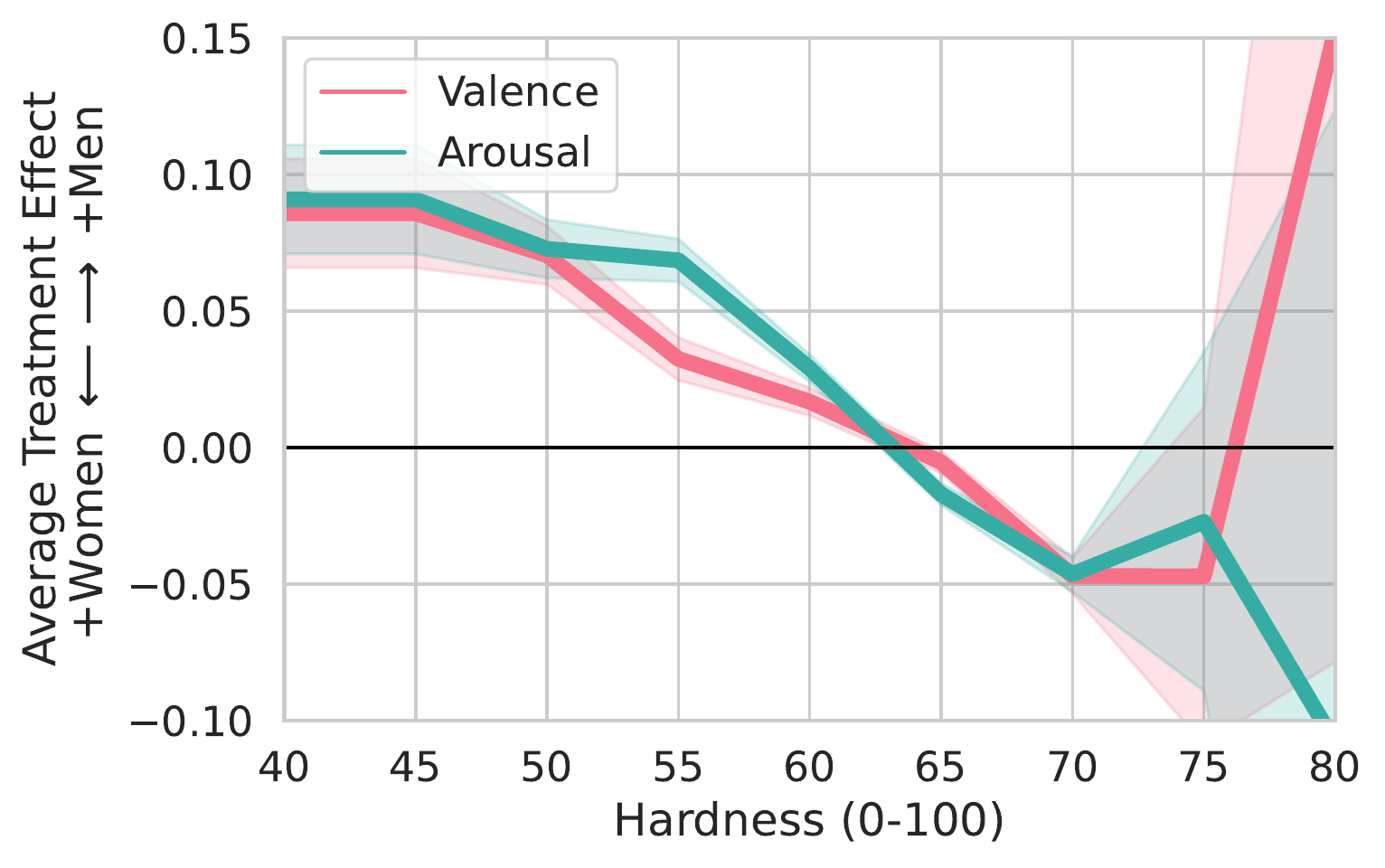}} & 
    {\includegraphics[width=0.30\textwidth]{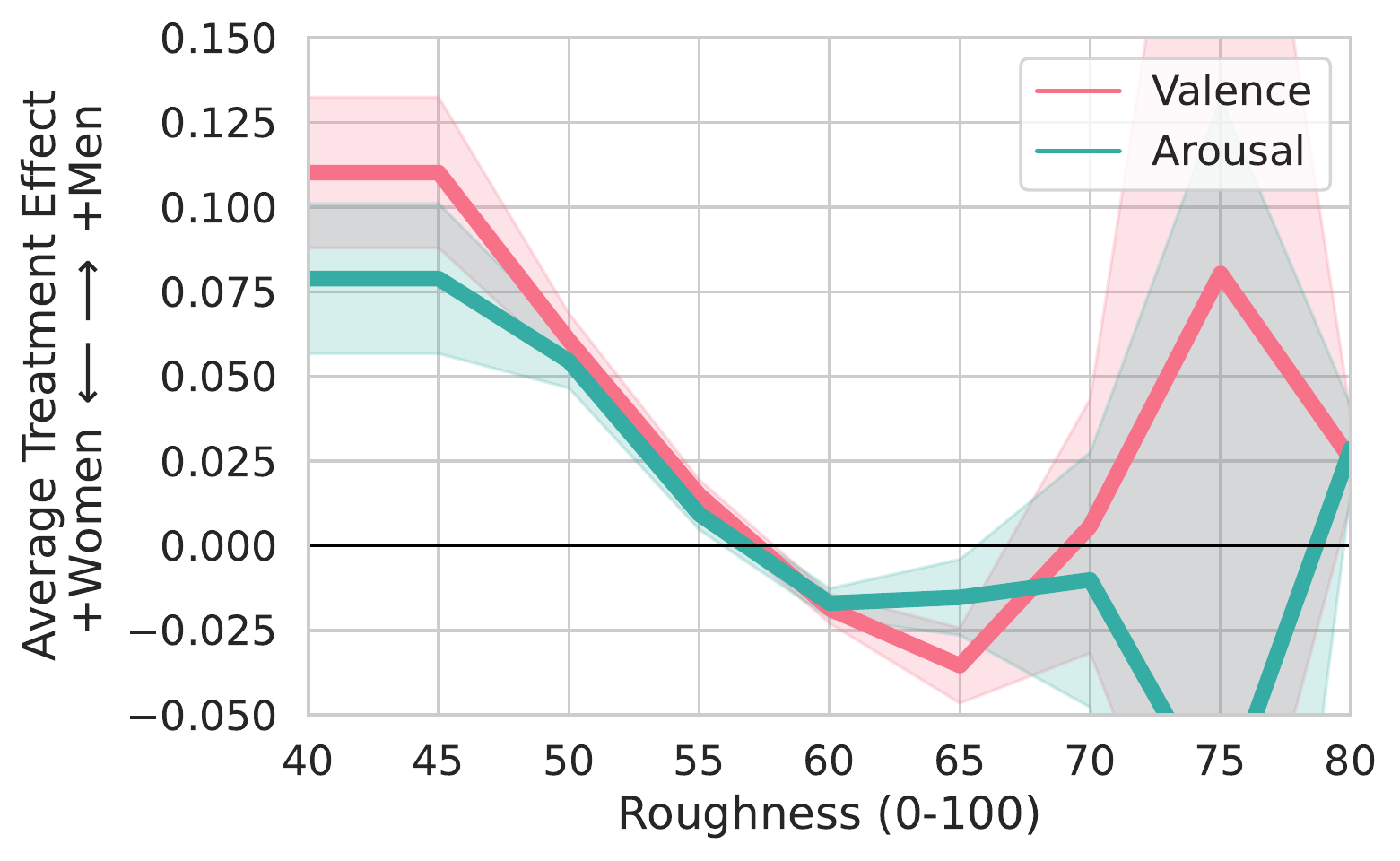}} &
    {\includegraphics[width=0.30\textwidth]{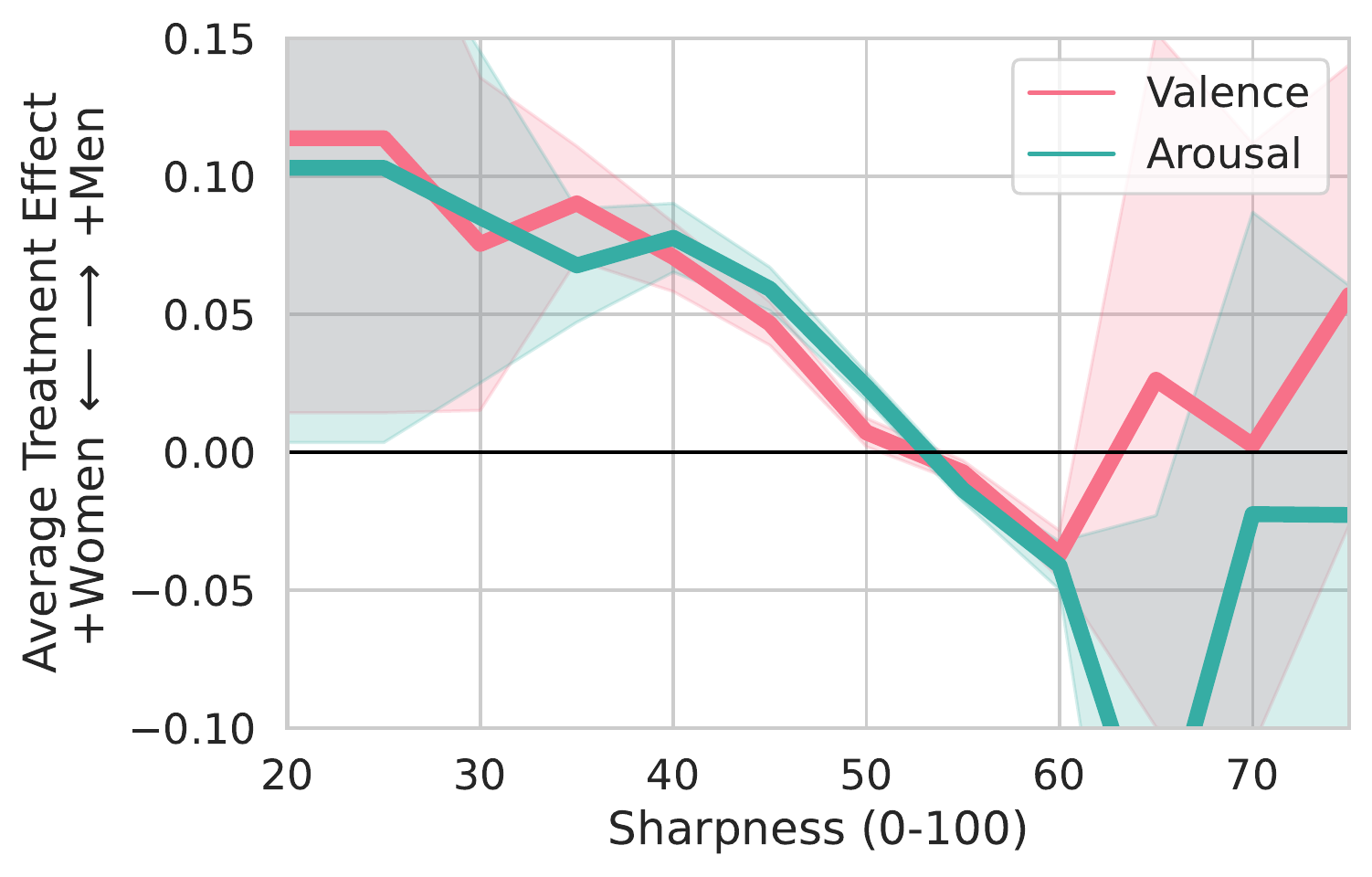}} \\
    {\includegraphics[width=0.30\textwidth]{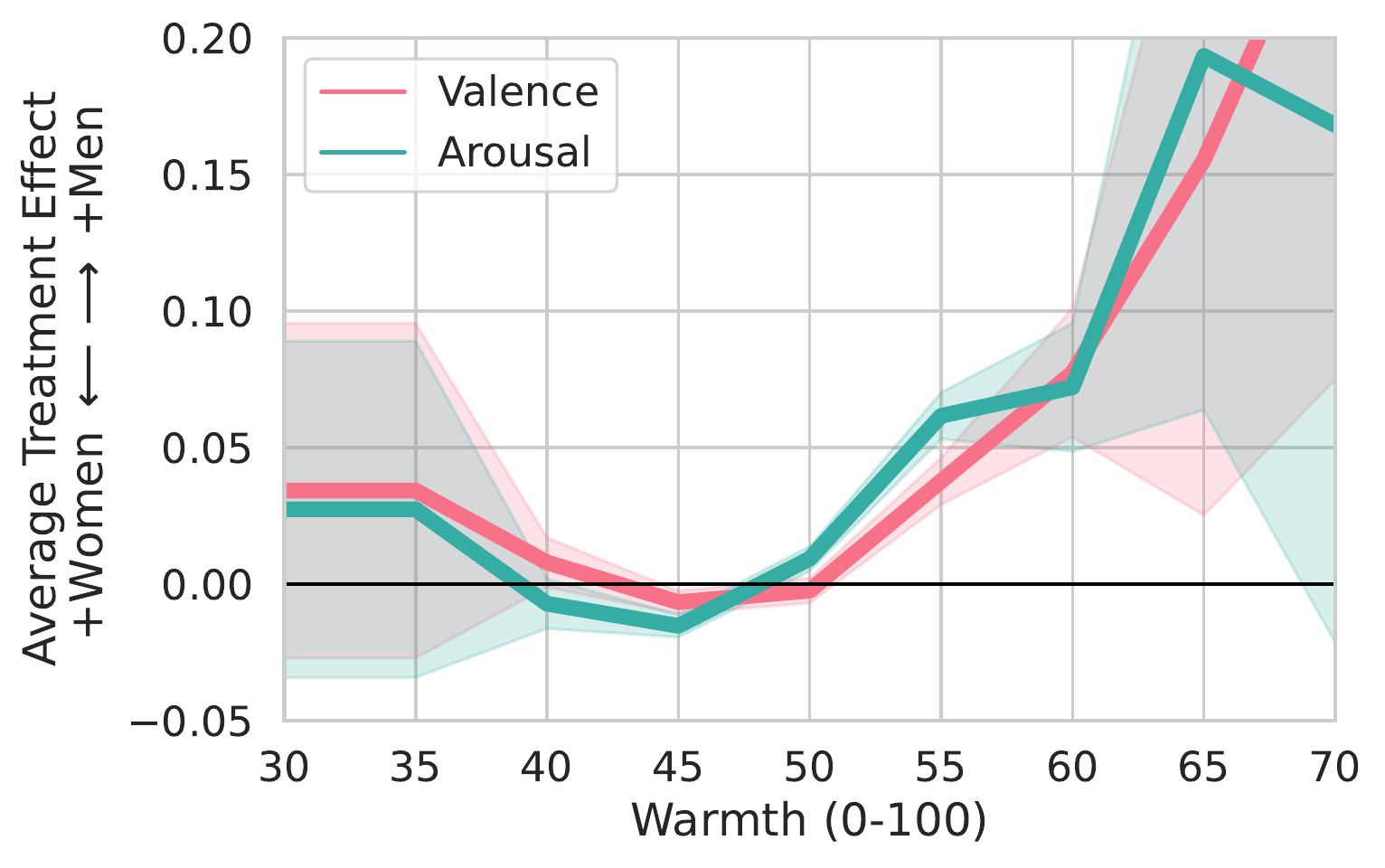}} & 
    {\includegraphics[width=0.30\textwidth]{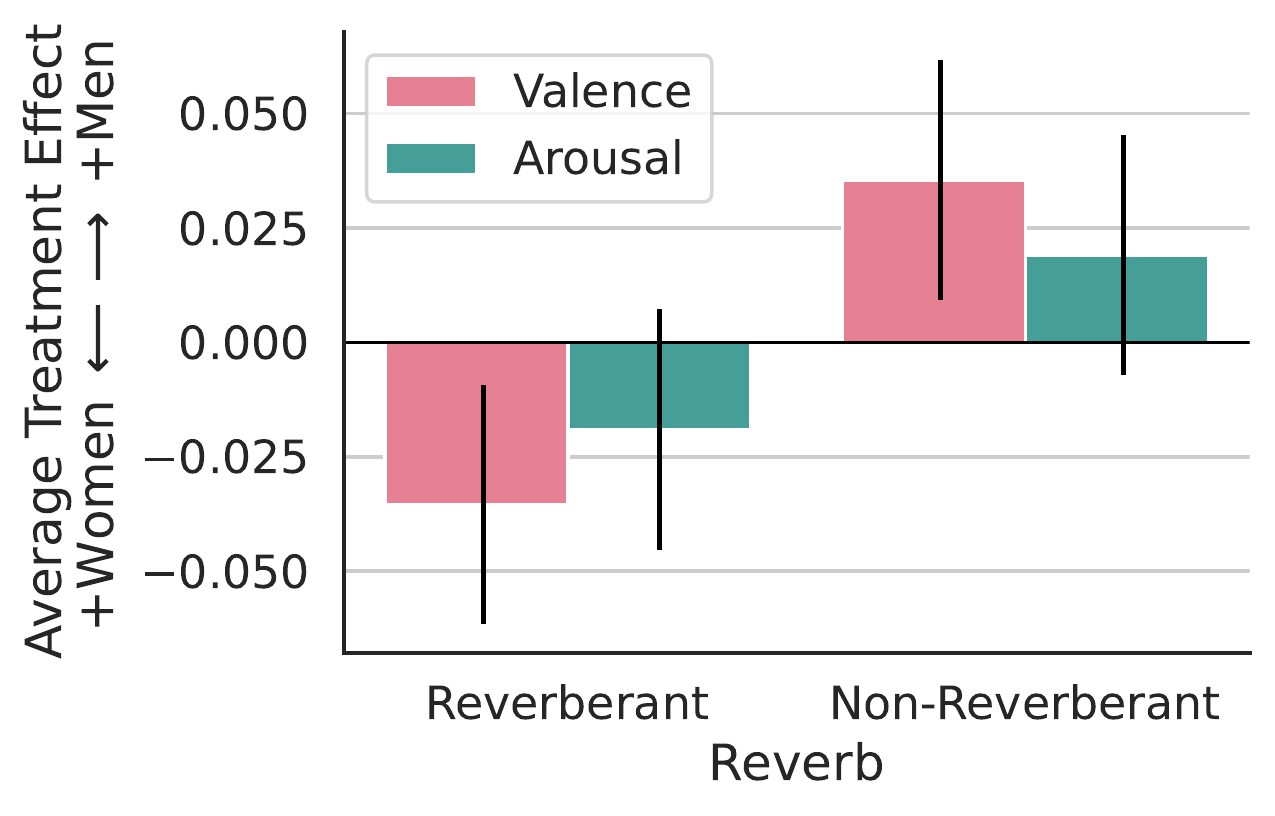}} & 
    {\includegraphics[width=0.30\textwidth]{plots/ATE/women_men_mp3_features_mode.pdf}}
    \end{tabular}
    \caption{
    Average treatment effects of listener \textbf{gender} on response valence and arousal relative to \textbf{musical} features. A \textit{positive} ATE here indicates a larger percent increase in valence or arousal for \textit{men}, and a \textit{negative} ATE here indicates a larger percent increase in valence or arousal for \textit{women}. Standard errors are shown; \textcolor{red}{valence} in \textcolor{red}{red}, \textcolor{blue}{arousal} in \textcolor{blue}{blue}. 
    }
    \label{fig:demographics_expanded_women_men}
\end{figure*}

\begin{figure*}[!t]
    \centering
    \begin{tabular}{ccc}
    {\includegraphics[width=0.30\textwidth]{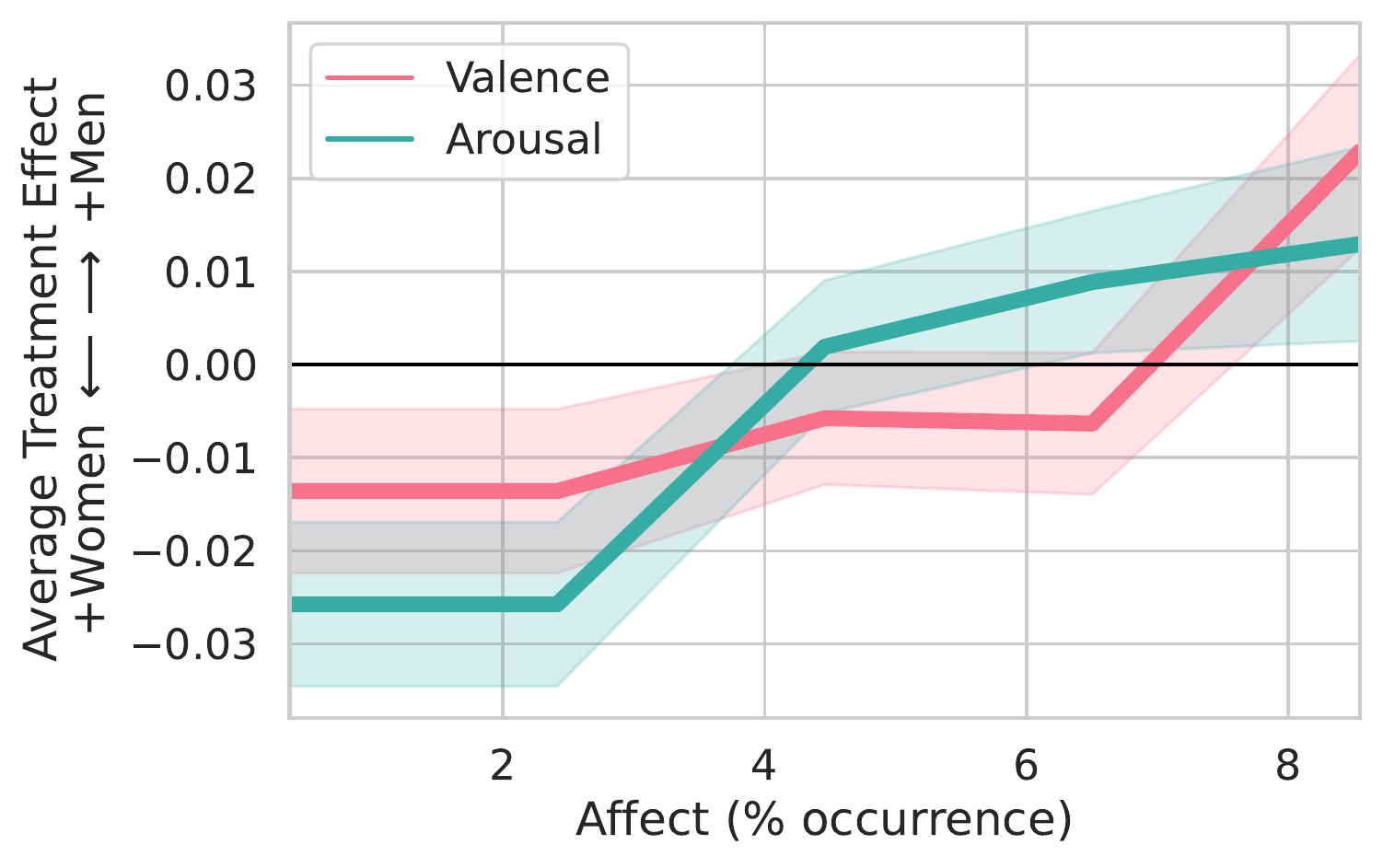}} & 
    {\includegraphics[width=0.30\textwidth]{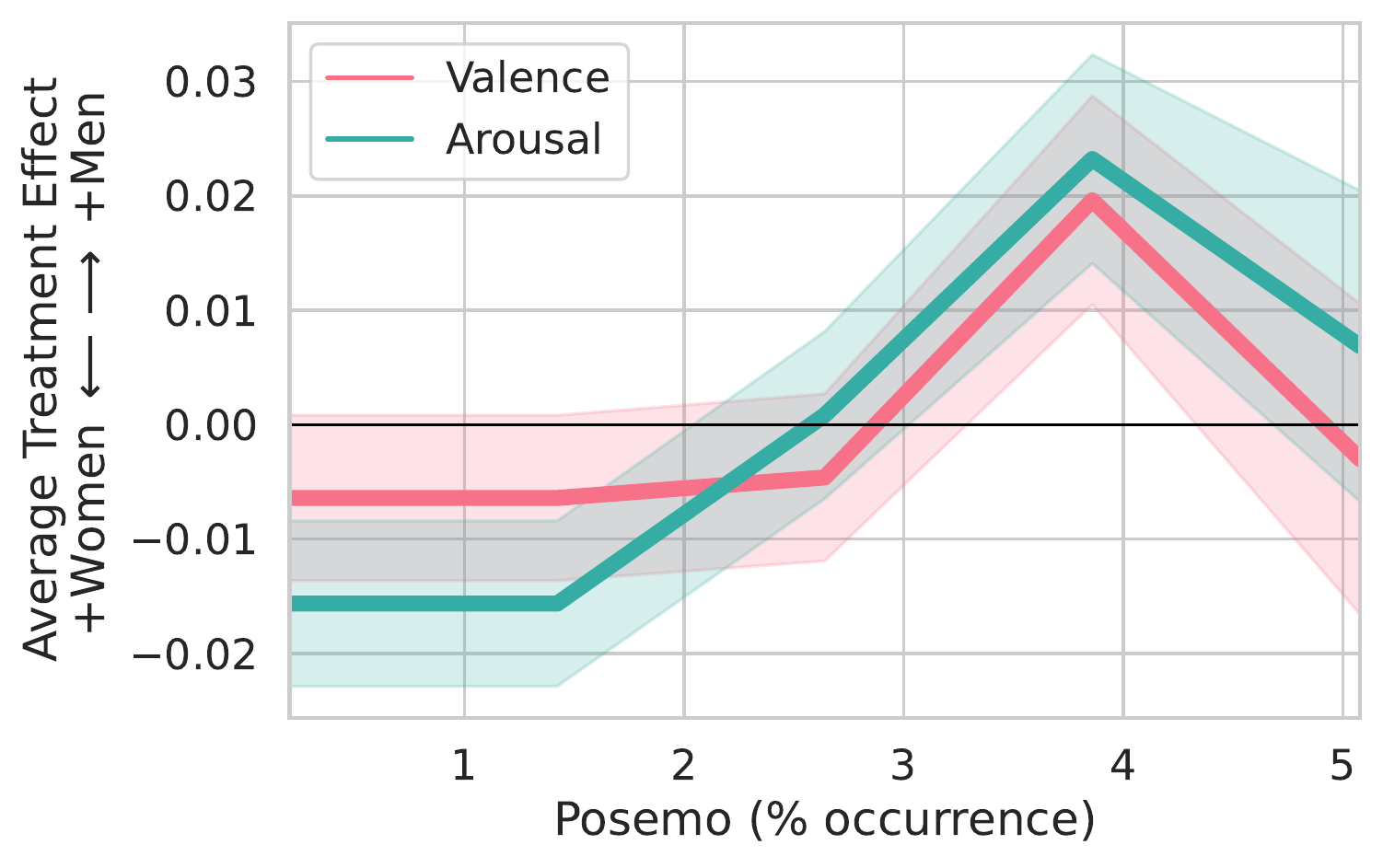}} &
    {\includegraphics[width=0.30\textwidth]{plots/ATE/women_men_lyric_negemo.pdf}} \\ 
    \end{tabular}
    \caption{
    Average treatment effects of listener \textbf{gender} on response valence and arousal relative to \textbf{lyrical} features on LIWC affective processes. Observations show that men are more positively affected by greater \textit{posemo} use, while women are more negatively affected by greater \textit{negemo} use. With the intent to reduce noise at the extremities, x-axis limits are capped at their 95\% quantile values. Standard errors are shown; \textcolor{red}{valence} in \textcolor{red}{red}, \textcolor{blue}{arousal} in \textcolor{blue}{blue}. 
    }
    \label{fig:demographics_expanded_women_men_liwc}
\end{figure*}

\begin{figure*}[!t]
    \centering
    \begin{tabular}{ccc}
    {\includegraphics[width=0.30\textwidth]{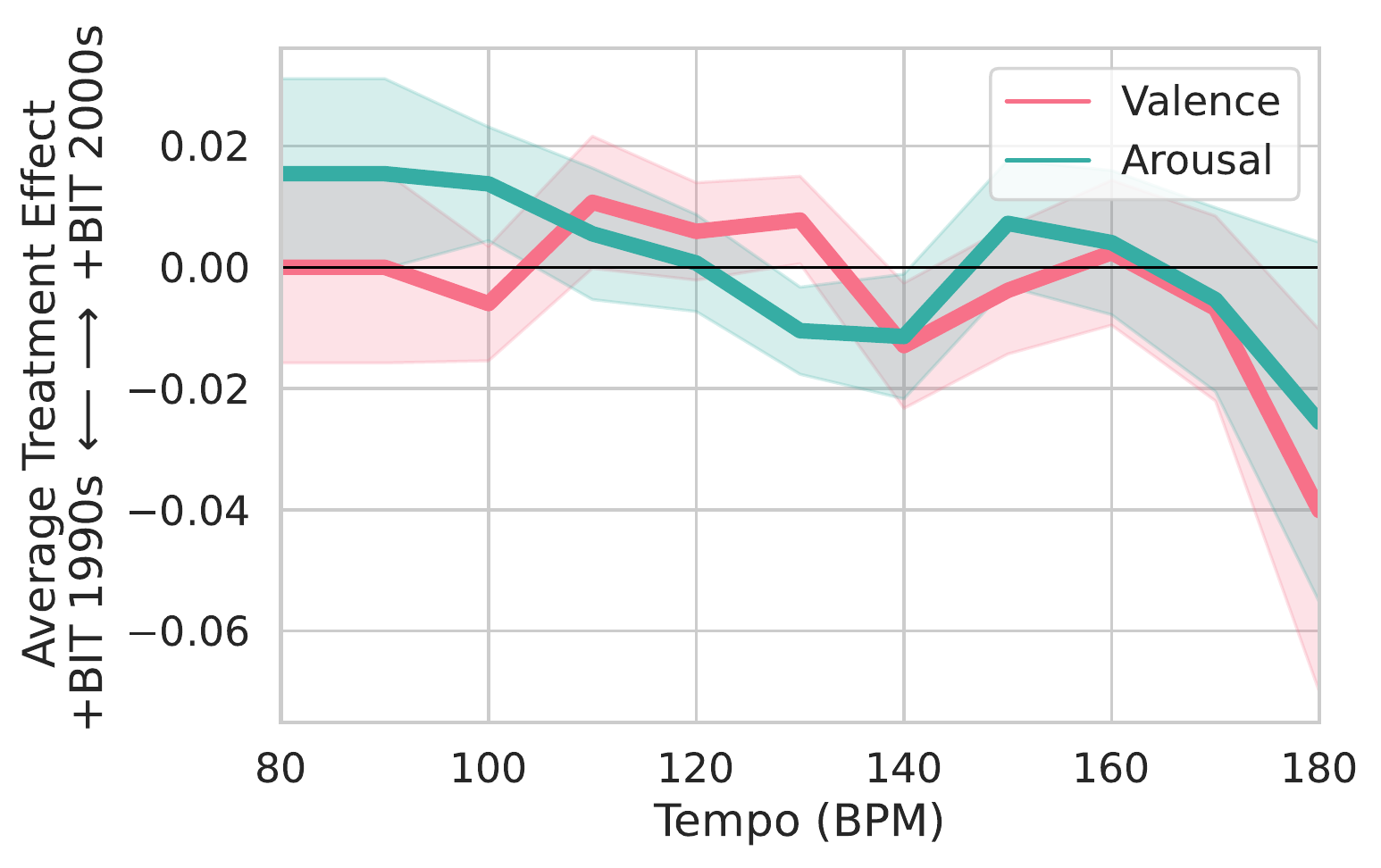}} & 
    {\includegraphics[width=0.30\textwidth]{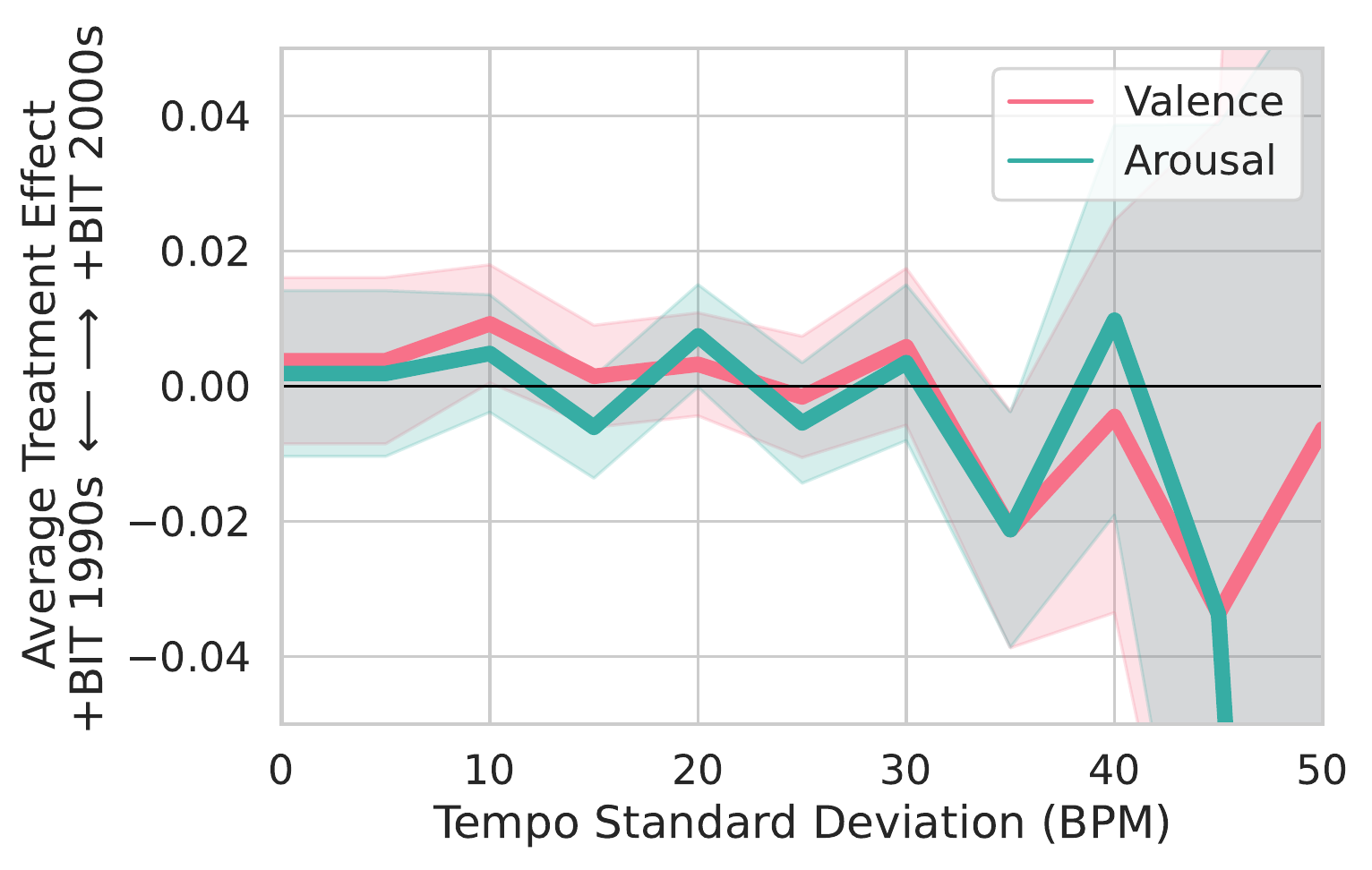}} &
    {\includegraphics[width=0.30\textwidth]{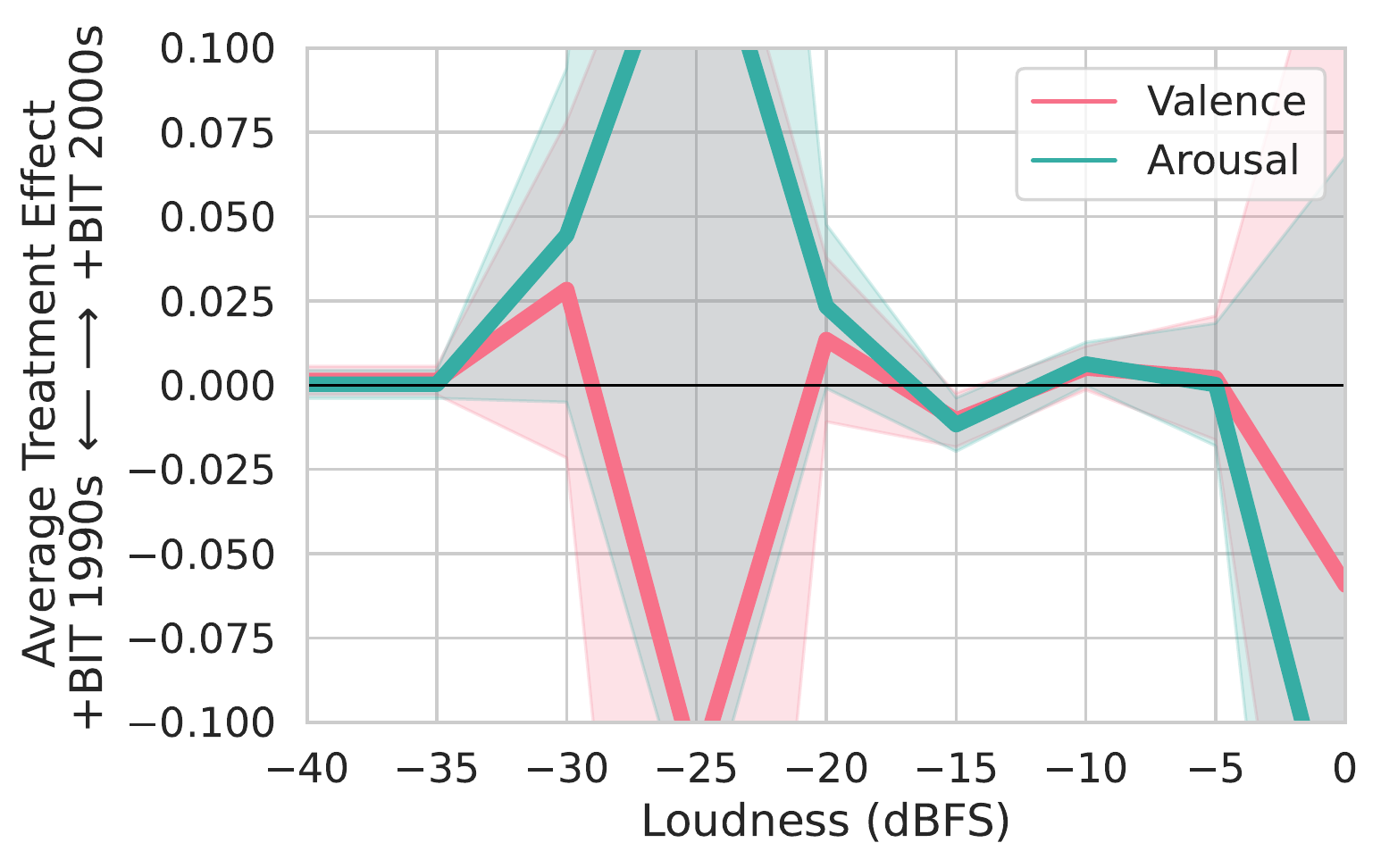}} \\ 
    {\includegraphics[width=0.30\textwidth]{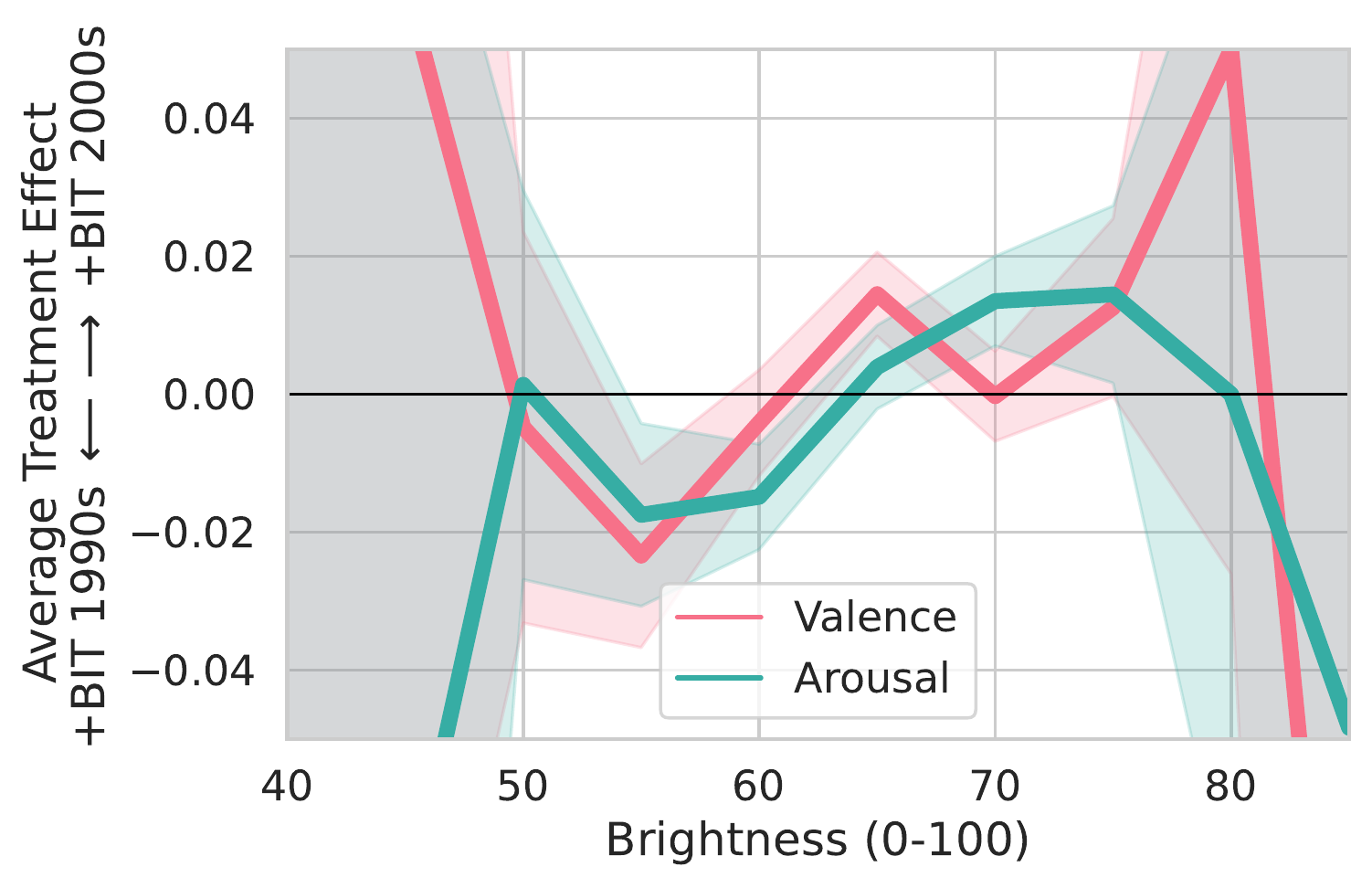}} &
    {\includegraphics[width=0.30\textwidth]{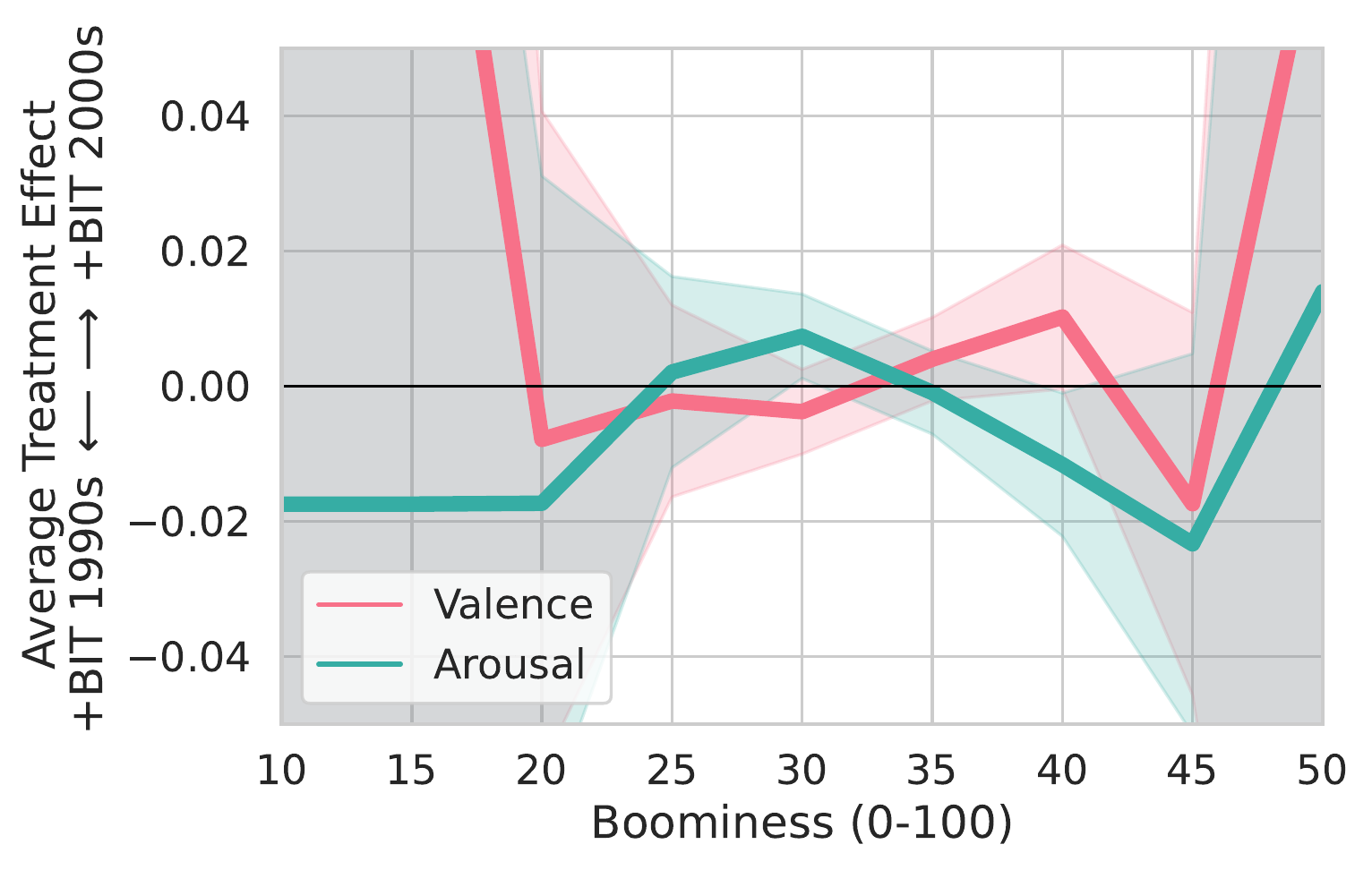}} &
    {\includegraphics[width=0.30\textwidth]{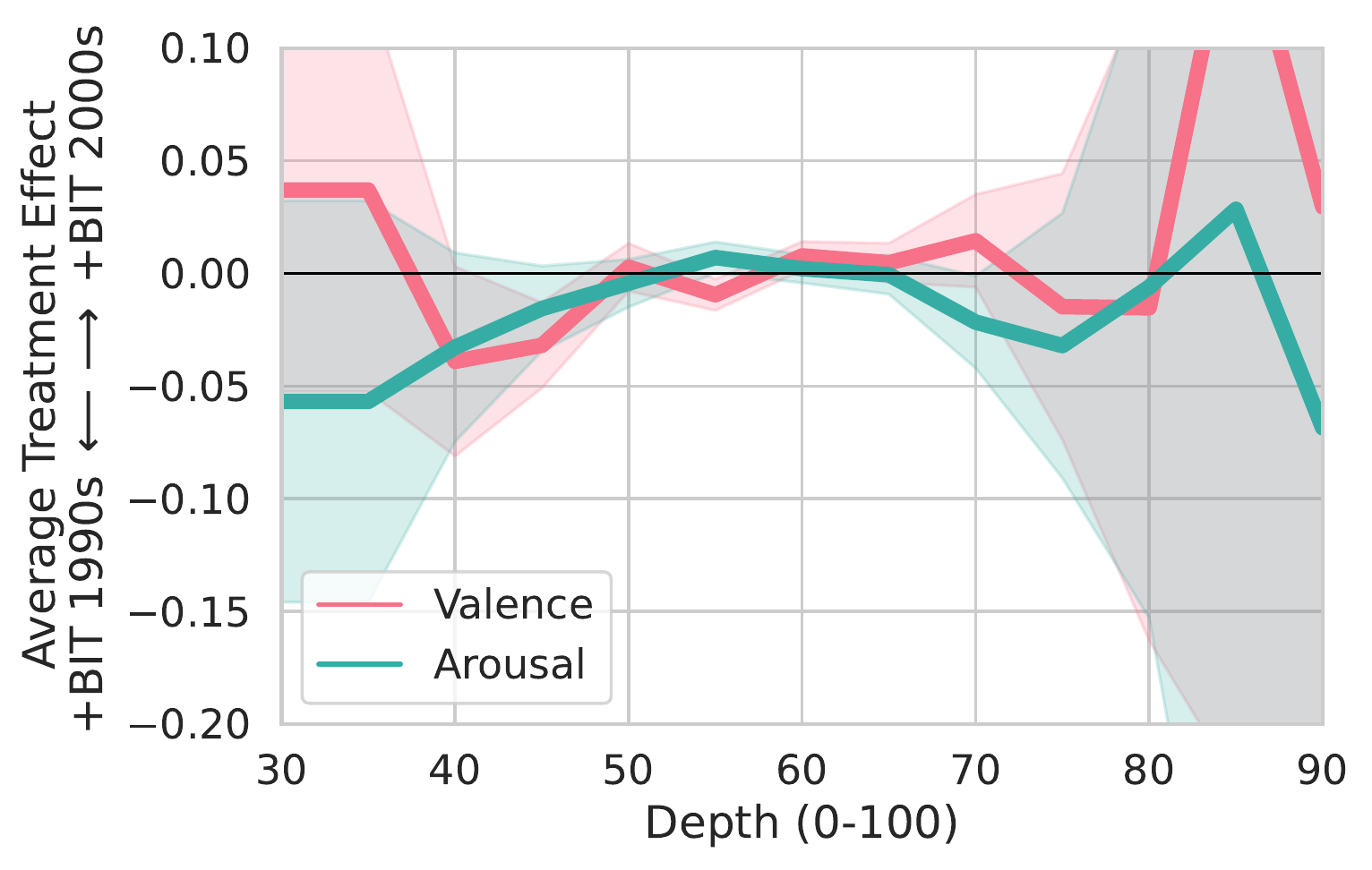}} \\ 
    {\includegraphics[width=0.30\textwidth]{plots/ATE/1990_2000_mp3_features_timbre_hardness.pdf}} & 
    {\includegraphics[width=0.30\textwidth]{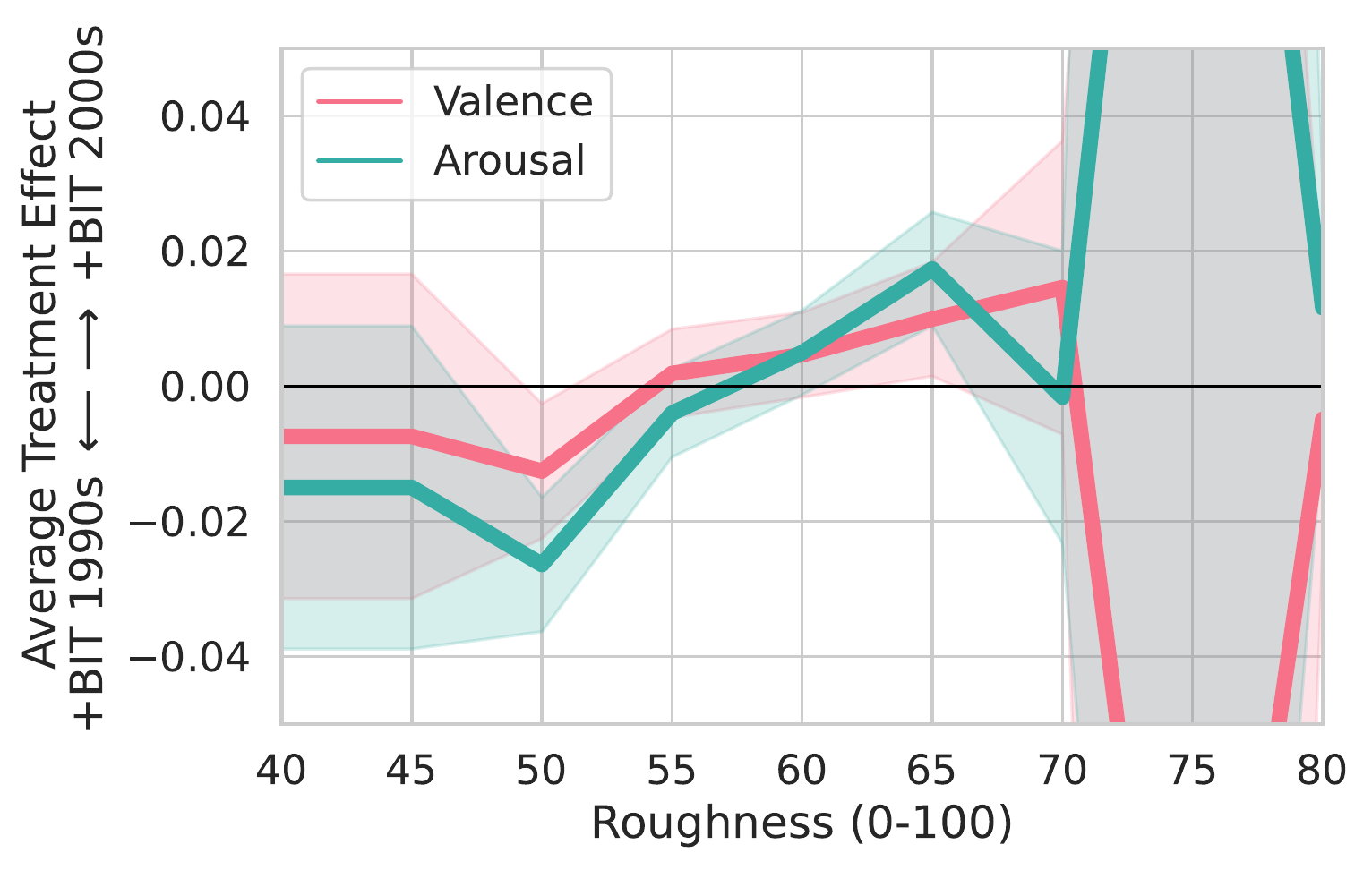}} &
    {\includegraphics[width=0.30\textwidth]{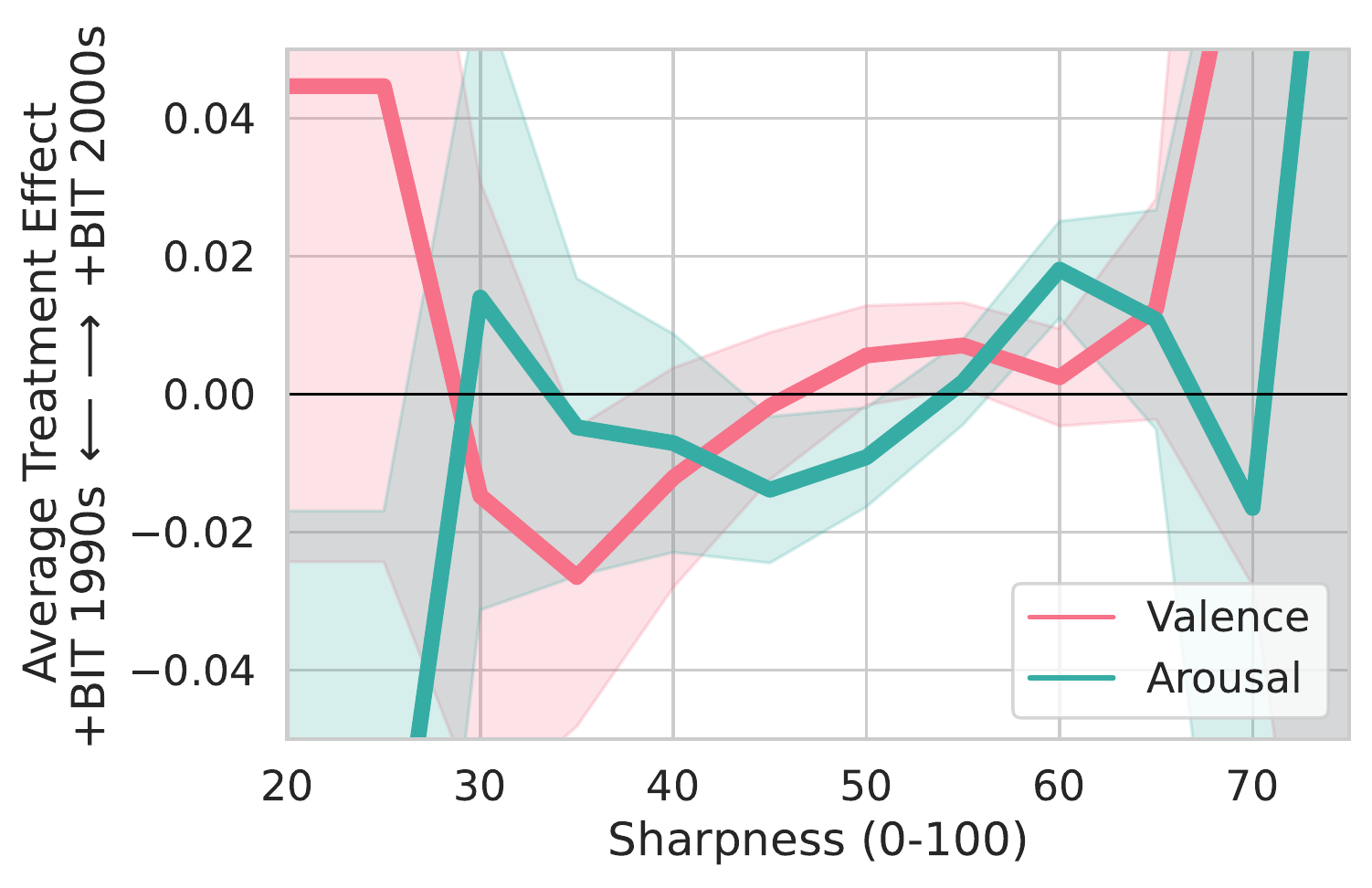}} \\
    {\includegraphics[width=0.30\textwidth]{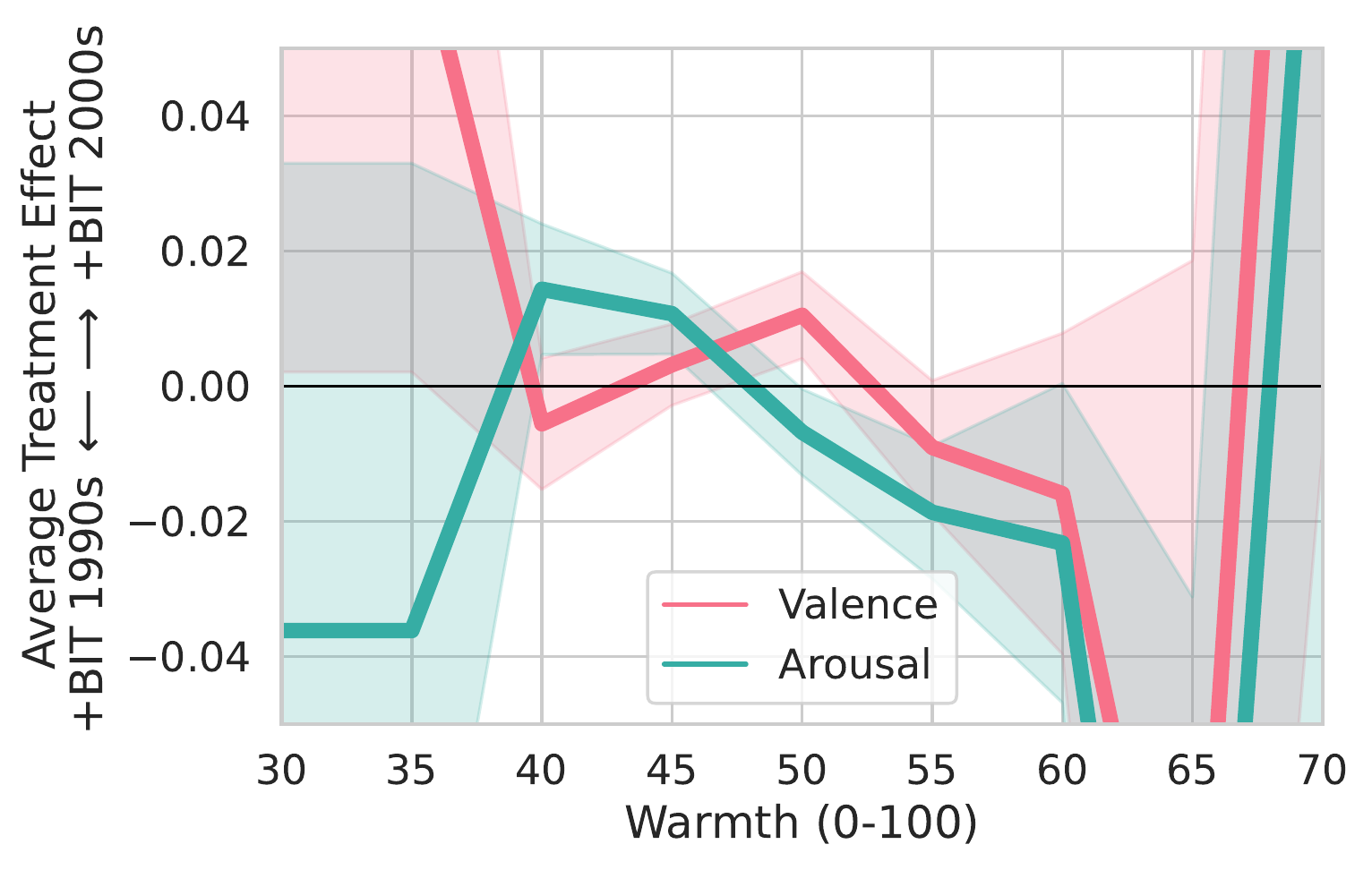}} & 
    {\includegraphics[width=0.30\textwidth]{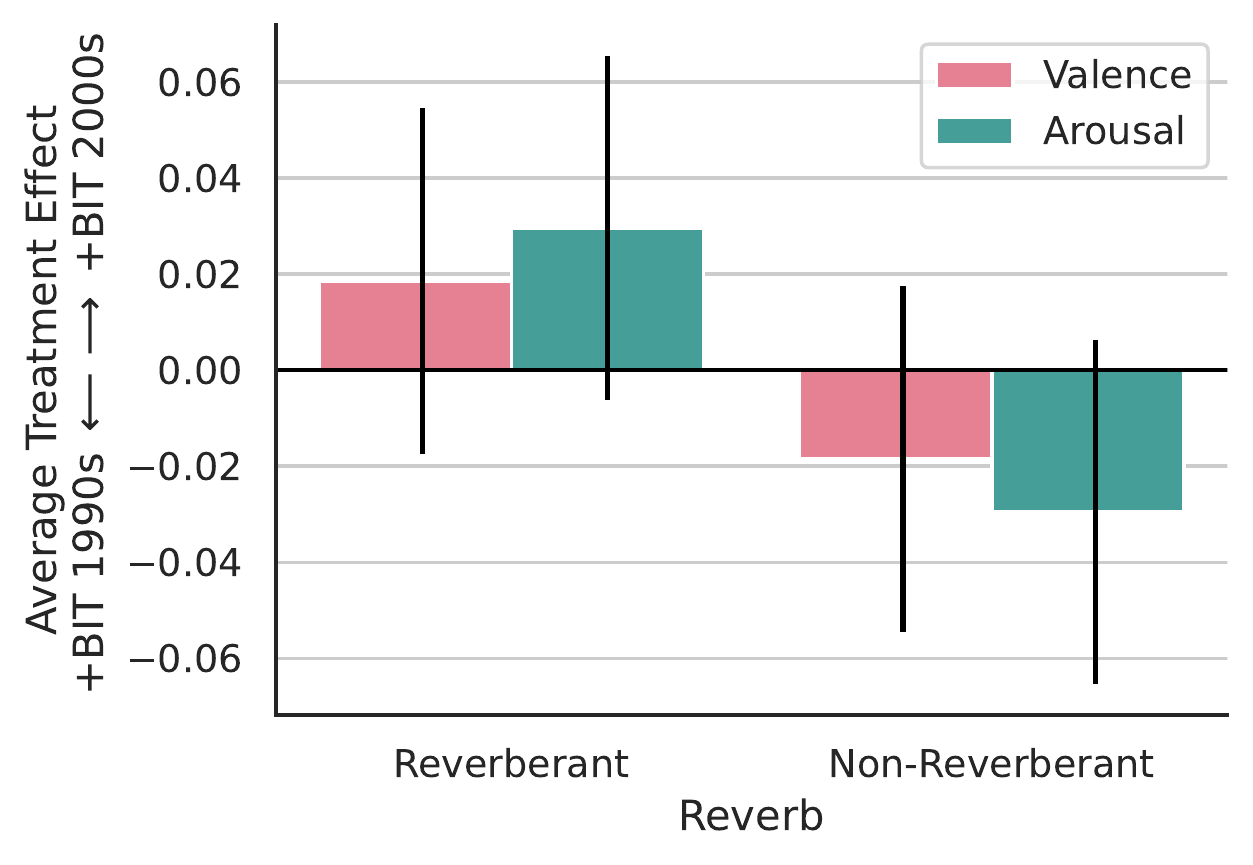}} & 
    {\includegraphics[width=0.30\textwidth]{plots/ATE/1990_2000_mp3_features_mode.pdf}} 
    \end{tabular}
    \caption{
    Average treatment effects of listener \textbf{age} on response valence and arousal relative to \textbf{musical} features. A \textit{positive} ATE here indicates a larger percent increase in valence or arousal for those born in the (b.i.t.) \textit{2000s}, and a \textit{negative} ATE here indicates a larger percent increase in valence or arousal for those b.i.t. \textit{1990s}. Standard errors are shown; \textcolor{red}{valence} in \textcolor{red}{red}, \textcolor{blue}{arousal} in \textcolor{blue}{blue}. 
    }
    \label{fig:demographics_expanded_1990_2000}
\end{figure*}

Expanded results for variations in affective responses are shown in Figures \ref{fig:musicfeatures_expanded} and \ref{fig:musicfeatures_key} for musical features, Figures \ref{fig:lyricfeatures_expanded_1}-\ref{fig:lyricfeatures_expanded_4} for LIWC lyrical features, Figures \ref{fig:setting_overall_setting}-\ref{fig:setting_overall_language} for settings and other playlist tags, and Figures \ref{fig:demographics_expanded_women_men}-\ref{fig:demographics_expanded_1990_2000} for demographic effects of gender and age.

\section{Expanded Results for Disclosures of Mental Health Disorders}
\label{sec:mentalhealthdisclosures}

\myparagraph{Regular Expressions.} We source mental health disorders from the DSM-5-TR \cite{american2013diagnostic} and construct regular expressions to identify possible comments that self-disclose a diagnosis of a mental health disorder; specific regular expressions are shown below in Table \ref{tab:conditions}. These regular expressions return 2319 matched comments in total.

\myparagraph{Manual Filtering.} Two Chinese native speakers then manually screened out comments that lacked a clear statement of self-disclosure. These comments primarily consisted of those that (1) described other people's diagnoses, i.e. relatives, friends, or celebrities, (2) described recovery from a disorder, (3) were of speculative nature on the diagnosis, e.g. \chinese{``我觉得...''} (``I think...''), (4) described what a diagnosis was but not necessarily that the listener themselves was diagnosed, and (5) only used diagnosis terms to exaggerate sentiment. Examples of positive and false positive comments are shown below in Table \ref{tab:commentexamples}. Following manual annotation, 46.9\% of regular expression-matched comments were eliminated, leaving 1231 comments deemed as self-disclosures.

\begin{table}[H]
    \centering
    \begin{tabular}{c} 
    \toprule
    
    \chinese{精神分裂, 精分, 人格分裂, 妄想, 思觉失调,} \\
    \chinese{强迫症, 创伤后应激, 情感障碍, 情绪障碍,} \\
    \chinese{情绪失调, 躁狂, 狂躁, 躁郁, 双向, 双相, } \\
    \chinese{抑郁, 忧郁, 重郁, 轻郁, 焦虑, 社交焦虑,} \\
    \chinese{社焦, 社交恐惧, 人群恐惧, 社恐, 余恐,} \\
    \chinese{恐惧, 恐慌, 广场恐惧, 分离焦虑, 缄默,} \\
    \chinese{人格障碍, 躯体变形, 体象障碍, 适应障碍,} \\
    \chinese{多重人格, 人格解体, 现实解体, 创伤性失忆,} \\
    \chinese{解离性身份障碍, 躯体症状, 歇斯底里,} \\
    \chinese{转换症, 转换障碍, 做作性障碍, 装病候群,} \\
    \chinese{代理性孟乔森, 进食障碍, 摄食障碍,} \\
    \chinese{神经性饮食失调, 反刍, 厌食, 贪食,} \\
    \chinese{暴食, 暴饮暴食, 异食, 失眠, 睡眠障碍,} \\
    \chinese{嗜睡, 睡眠相位后移, 快动眼睡眠, 睡瘫,} \\
    \chinese{睡眠瘫痪, 梦魇症, 易怒症, 暴怒症,} \\
    \chinese{行为障碍, 品行障碍, 偏执, 边缘性人格,} \\
    \chinese{边缘性人格, 边缘型人格, 边缘性格,} \\
    \chinese{做作.*人格, 自恋.*人格, 回避.*人格, } \\
    \chinese{依赖.*人格, 精神病, 心理疾病} \\
    
    \bottomrule
    \end{tabular}
    \caption{
    Mental health disorder condition name strings; these are prefixed/suffixed with the strings for ``diagnosed`` (``\chinese{确诊.*}'') and ``diagnosis'' (``\chinese{诊断.*}''), i.e. \chinese{``确诊抑郁''} for ``diagnosed with depression'', to act as initial regular expression filters for users who self-disclose a diagnosis of a mental health disorder. 
    }
    \label{tab:conditions}
\end{table}

\begin{table}[H]
    \centering
    \begin{tabular}{l} 
    \toprule
    Positives \\
    \midrule
    \chinese{今天，我被确诊为抑郁症了。} \\
    \chinese{去了大城市诊断的双相情感障碍} \\
    \chinese{524双子，确诊精分。} \\
    \chinese{诊断人格分裂。即使这样也要活下去啊！} \\
    \chinese{2017.12.25确诊焦虑症中度抑郁我很痛苦} \\
    \midrule
    False Positives \\
    \midrule
    \chinese{医院等待医生上班确诊我是否患有抑郁症} \\
    \chinese{整的我有点精神分裂症自我诊断上瘾了} \\
    \chinese{确诊中度抑郁症，现在已经走出来啦} \\
    \chinese{我已经几个朋友确诊抑郁了。} \\
    \chinese{我认知是抑郁症心境低落与其处境不相符} \\
    \bottomrule
    \end{tabular}
    \caption{
    Examples of positive and false positive self-disclosure statements of mental health disorders encountered in our manual labeling of comments matched with regular expressions. Partial comments are shown.
    }
    \label{tab:commentexamples}
\end{table}

\myparagraph{Disorder Matches.} In total, 1133 users made self-disclosure statements. A breakdown of users by disorder class is shown in Table \ref{tab:disordercounts}; note that the total users per class across all classes exceeds 1133 due to comorbidities. 

\begin{table}[H]
    \centering
    \begin{tabular}{ll} 
    \toprule
    Disclosed Disorder Class & Matched Users \\
    \midrule
    Depressive & 920 \\
    Anxiety & 225 \\
    Bipolar and Related & 201 \\
    Schizophrenia Spectrum & 108 \\ and Other Psychotic & \\
    Sleep-Wake & 35 \\
    Personality & 18 \\
    Feeding and Eating & 11 \\
    Obsessive-Compulsive & 5 \\ and Related & \\
    Somatic Symptom & 4 \\ and Related & \\
    Dissociative & 1 \\
    Trauma- & 1 \\ and Stressor-Related & \\
    \bottomrule
    \end{tabular}
    \caption{
    The number of users who self-disclose a mental health disorder, stratified over broad disorder classes.
    }
    \label{tab:disordercounts}
\end{table}

\myparagraph{Diurnal User Activity.} Stratifying user activity across hours and measuring the relative comments made per stratum, we observe that disclosers show greater platform activity in the AMs (1-5 AM) and around 11AM-5PM compared to the set of all users. Shown in Figure \ref{fig:diurnal} below, these observations are consistent with insomnia-aligned diurnal user activity, prevalent in individuals diagnosed with affective disorders \cite{taylor2005epidemiology, harvey2008sleep}. Note here that due to platform data limitations, while comment \textit{dates} are available for all comments on the platform, only those made in the past year had \textit{times} recorded and, as a result, are what we use in our analysis here on diurnal user activity. Thus, it is important to interpret these in the context of the COVID-19 pandemic, which has caused an increase in the prevalence of anxiety and depression worldwide \cite{bareeqa2021prevalence}.

\begin{figure}[H]
    \centering
    \begin{tabular}{ccc}
    {\includegraphics[width=0.45\textwidth]{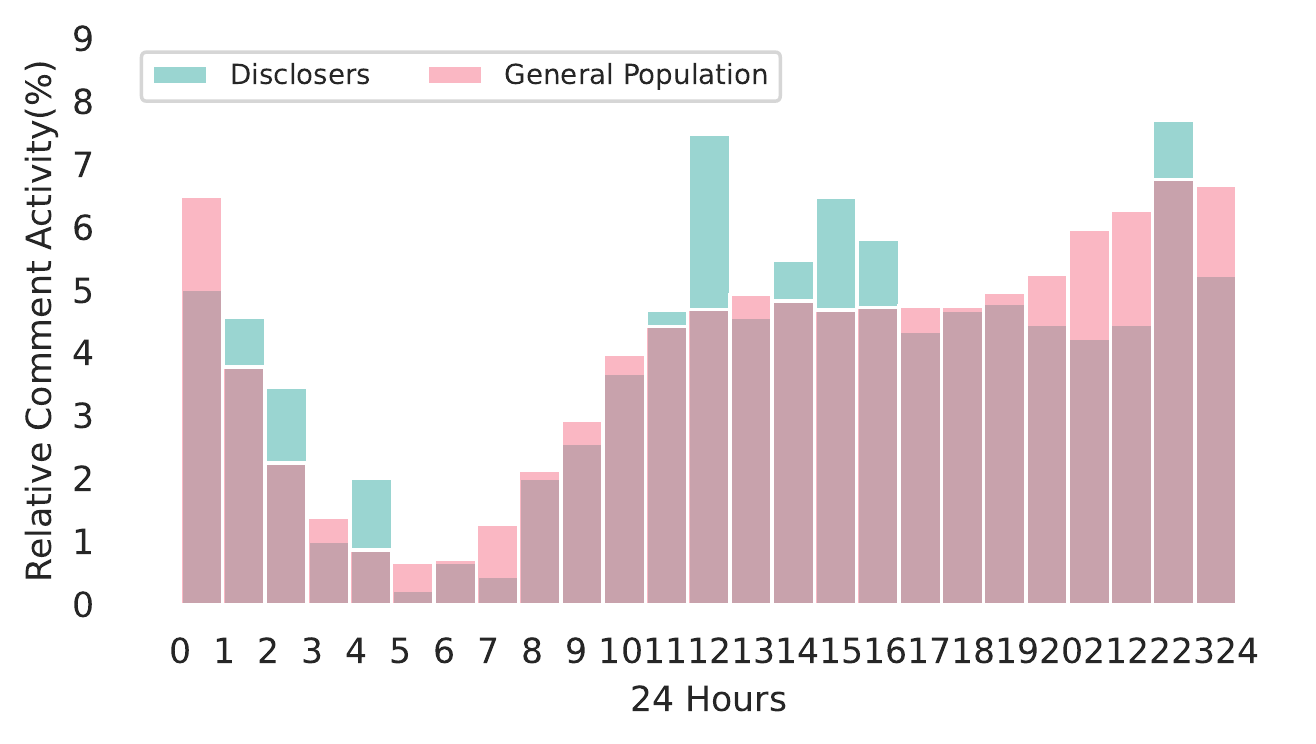}}
    \end{tabular}
    \caption{
    Diurnal commenting activity between \textcolor{blue}{disclosers} and the set of \textcolor{red}{all users}.
    }
    \label{fig:diurnal}
\end{figure}

\begin{figure*}
    \centering
    \begin{tabular}{cc}
    {\includegraphics[height=4cm]{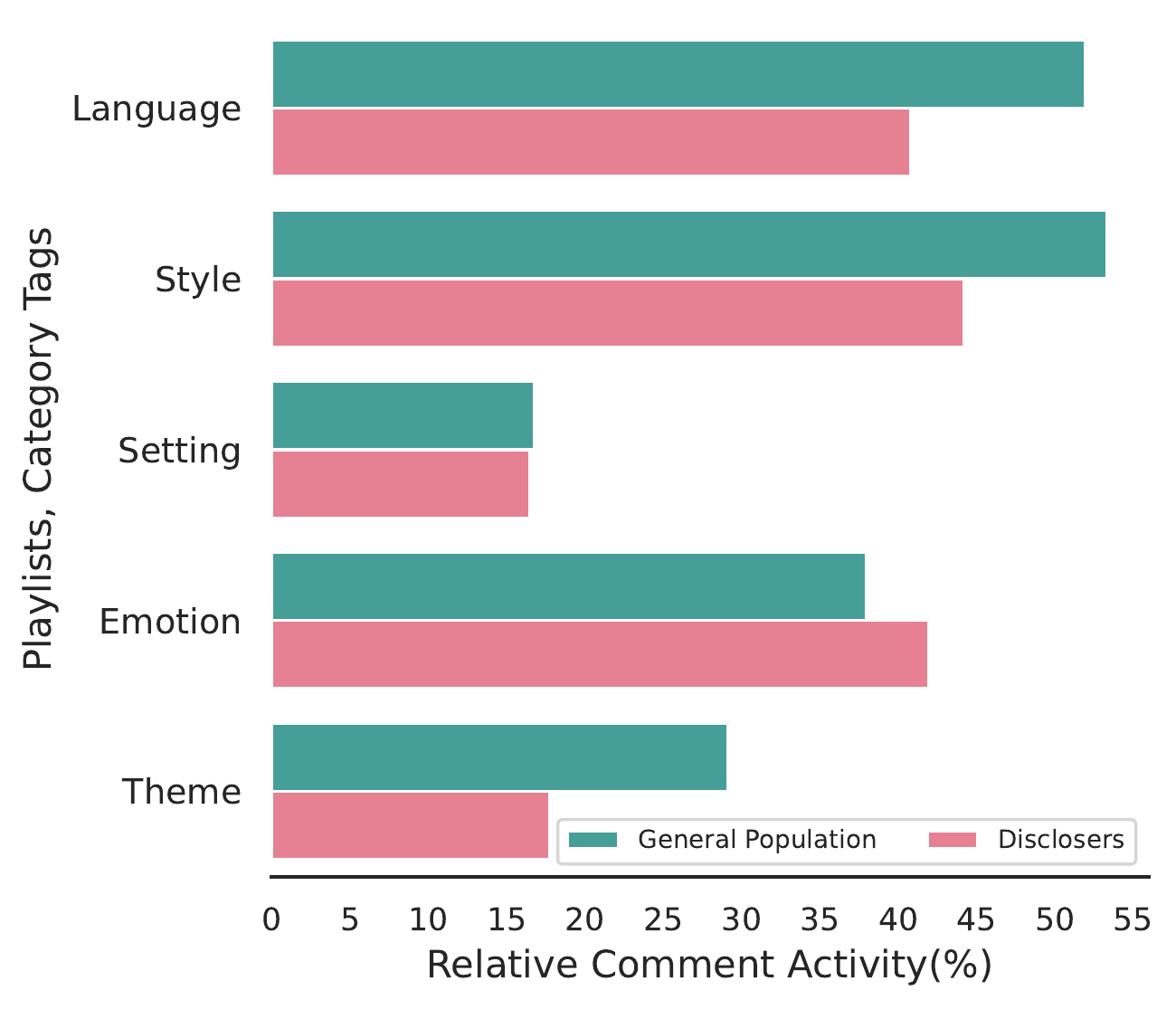}} &
    {\includegraphics[height=4cm]{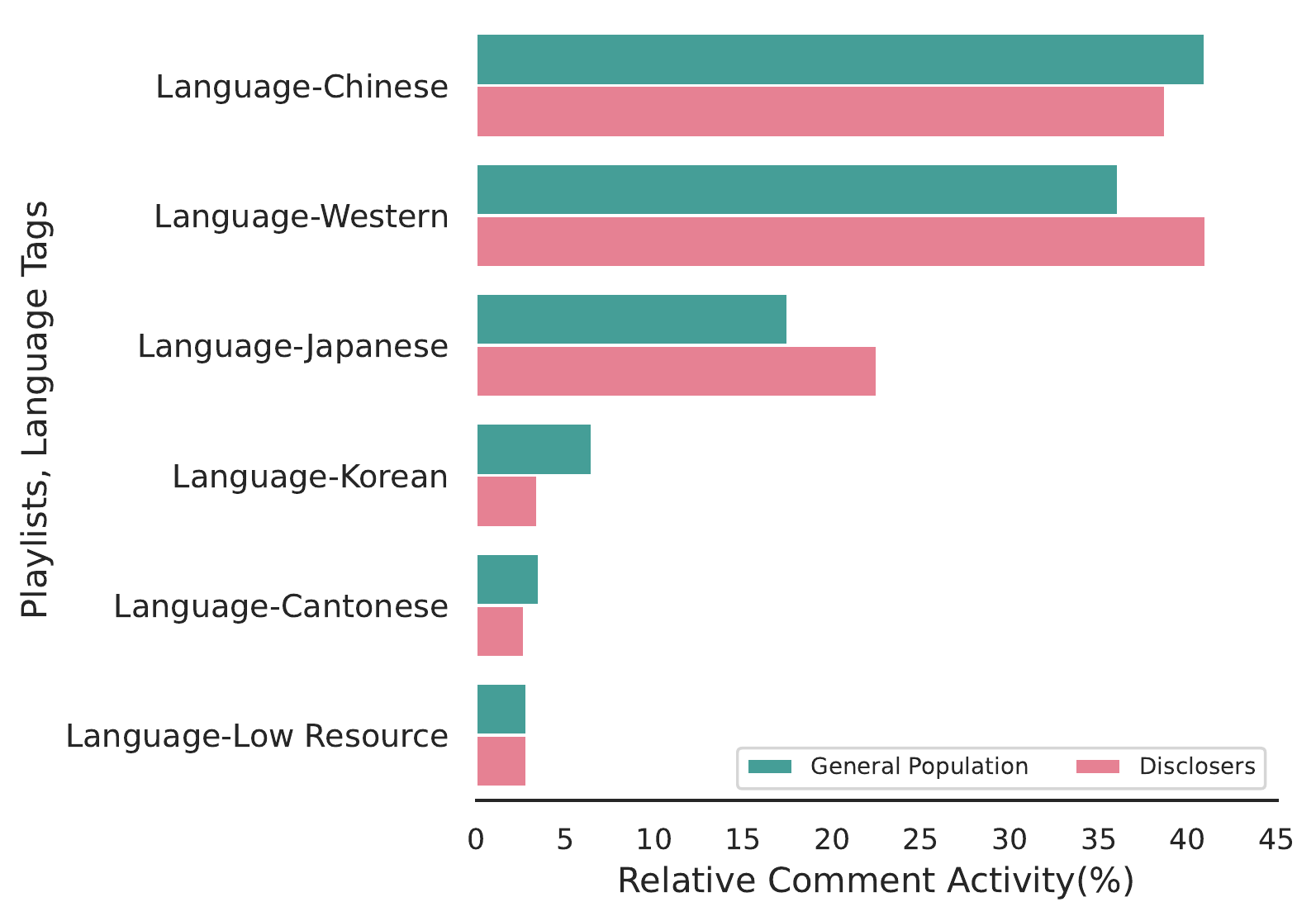}} \\
    {\includegraphics[height=5cm]{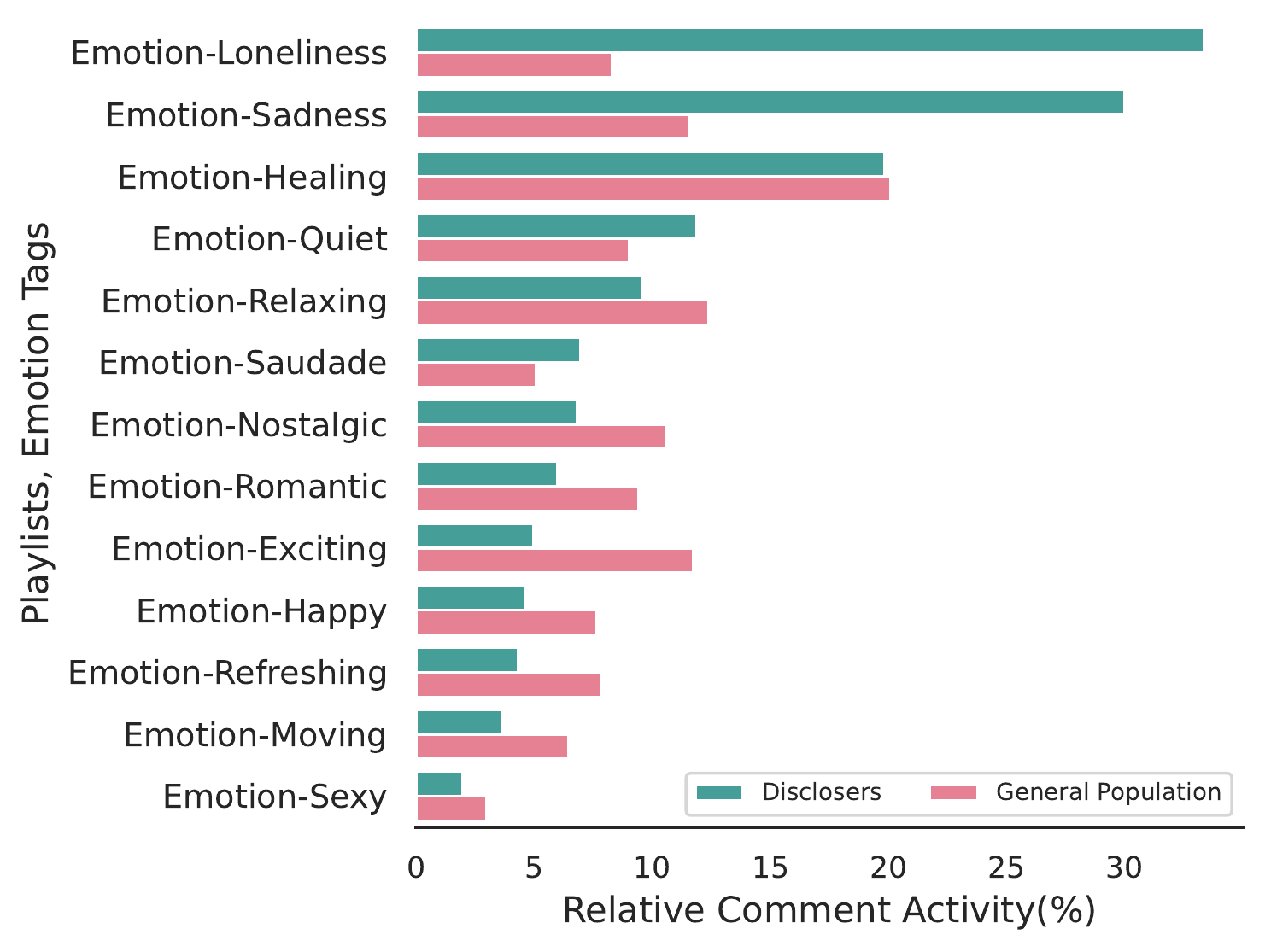}} &
    {\includegraphics[height=5cm]{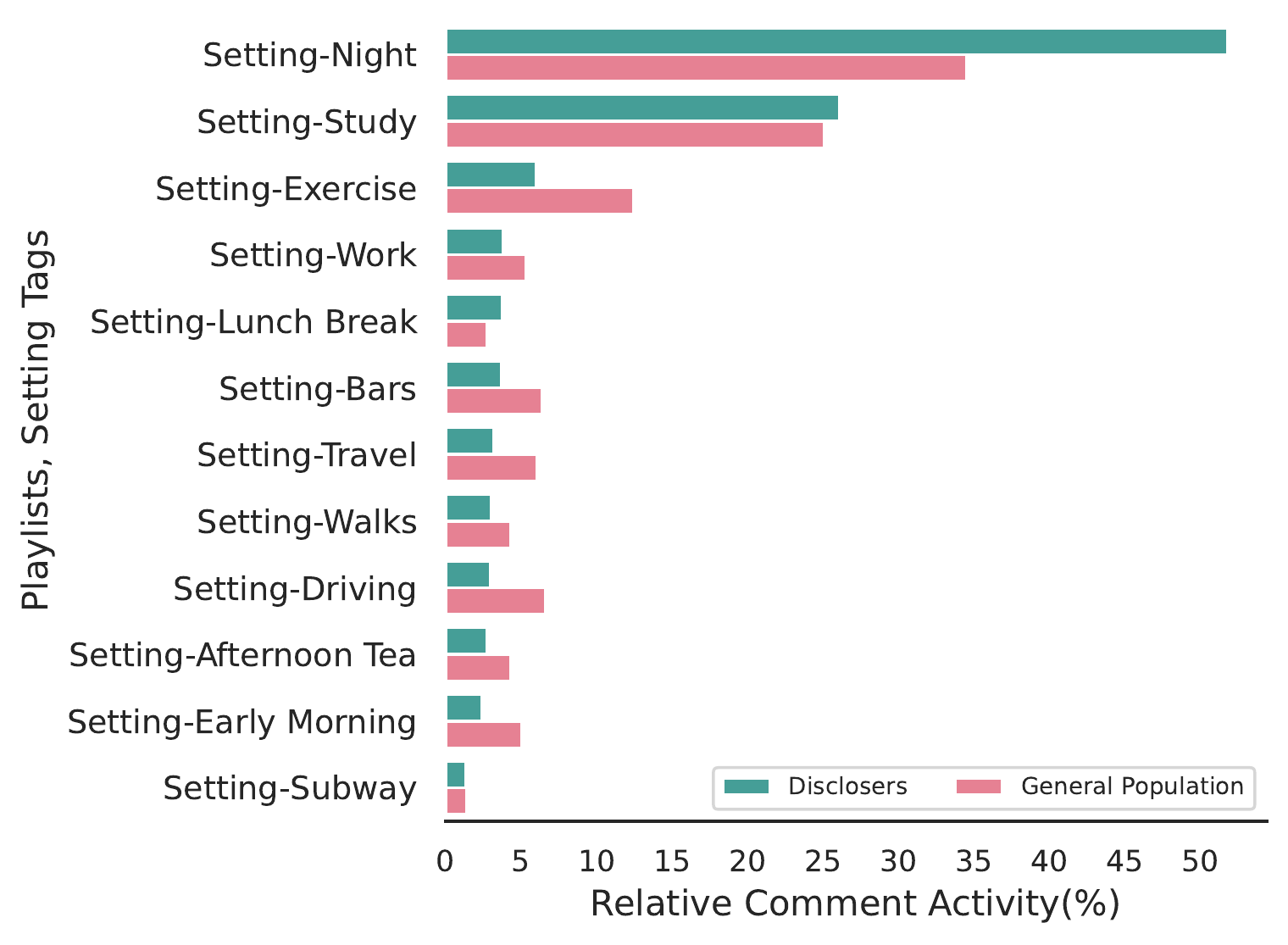}} \\
    {\includegraphics[height=8cm]{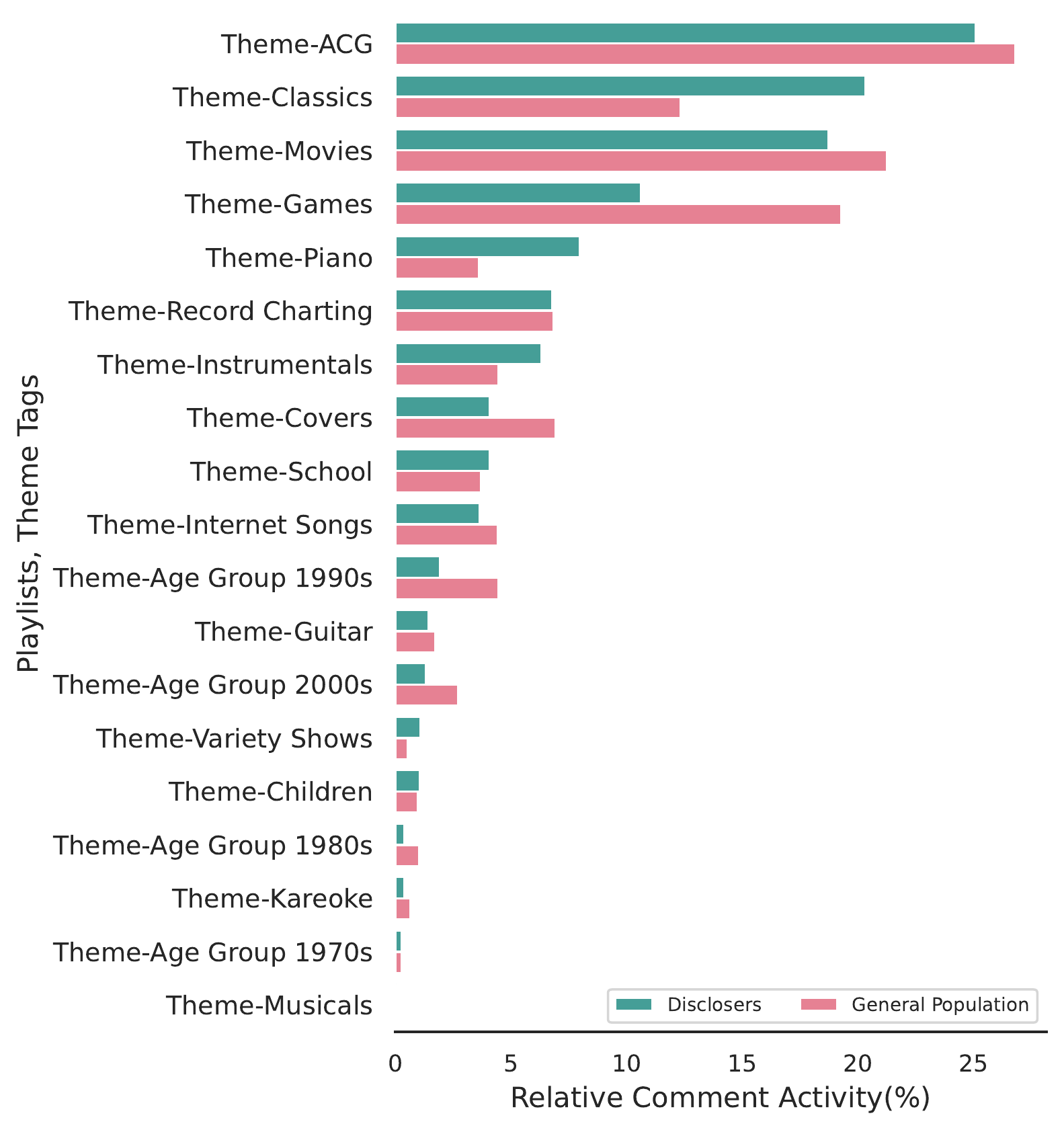}} &
    {\includegraphics[height=8cm]{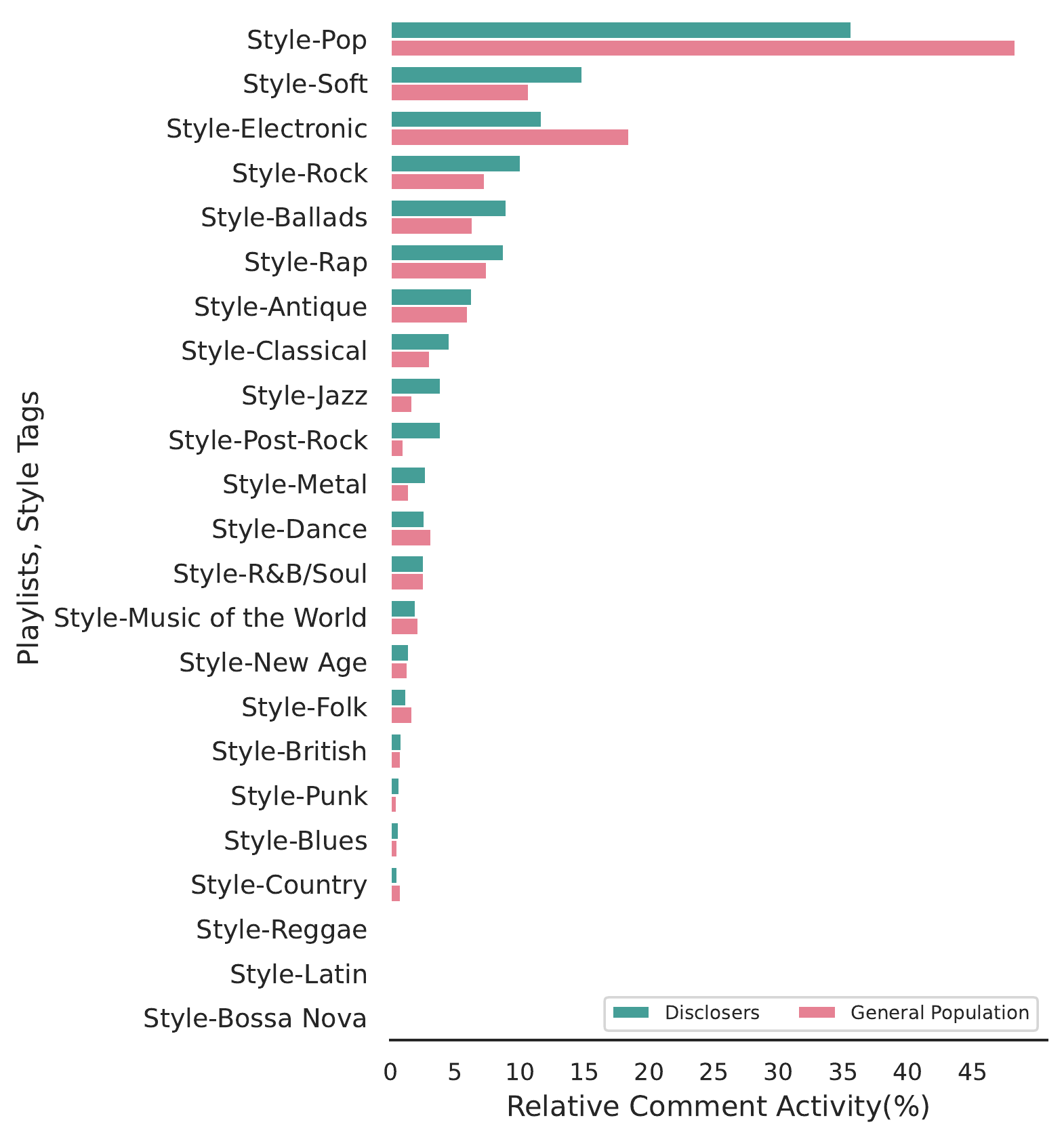}} \\
    \end{tabular}
    \caption{
    Relative tagged playlist commenting activity between disclosers and the set of all users. A breakdown of engagement with the five broad tag categories is shown on the \textbf{top left}, while other figures show each category's relative tag engagements. Note that as each playlist may have up to three unique tags, relative tag percentages do not add up to 100\%.
    }
    \label{fig:playlistengagement}
\end{figure*}

\myparagraph{Playlist Engagement.} Relative tagged playlist engagements are shown in Figure \ref{fig:playlistengagement} for disclosers and the set of all users; these are expanded figures as noted in Section \ref{sec:mentalhealthmain} of the main paper. Notably, disclosers show greater engagement with emotion tagged playlists; within emotional and setting tags, disclosers show overwhelmingly greater engagement with tagged playlists of sadder nature, i.e. \texttt{loneliness} (+302\%), \texttt{sadness} (+158\%), and \texttt{night} (+50.1\%), as well as decreased engagement with playlists of more active natures, e.g. \texttt{exercise} (-51.7\%). These observations mirror affective disorder activity trends \cite{cooney2013exercise} and suggest that people with affective disorders are more likely to use music reflective of negative emotions than positive emotions to manage feelings of sadness and depression \cite{stewart2019music}. 

\begin{figure*}[!t]
    \centering
    \begin{tabular}{ccc}
    {\includegraphics[width=0.4\textwidth]{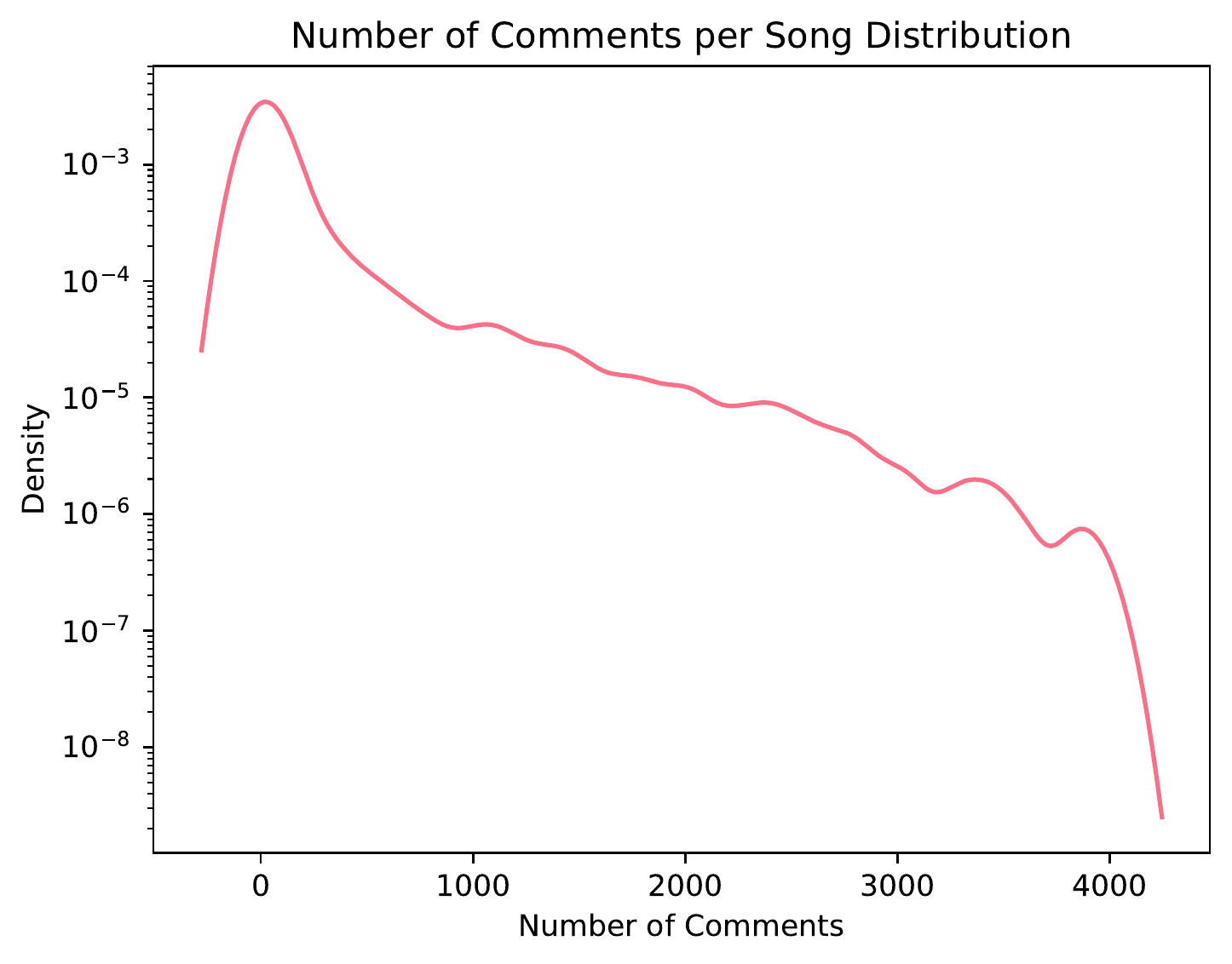}} & 
    {\includegraphics[width=0.4\textwidth]{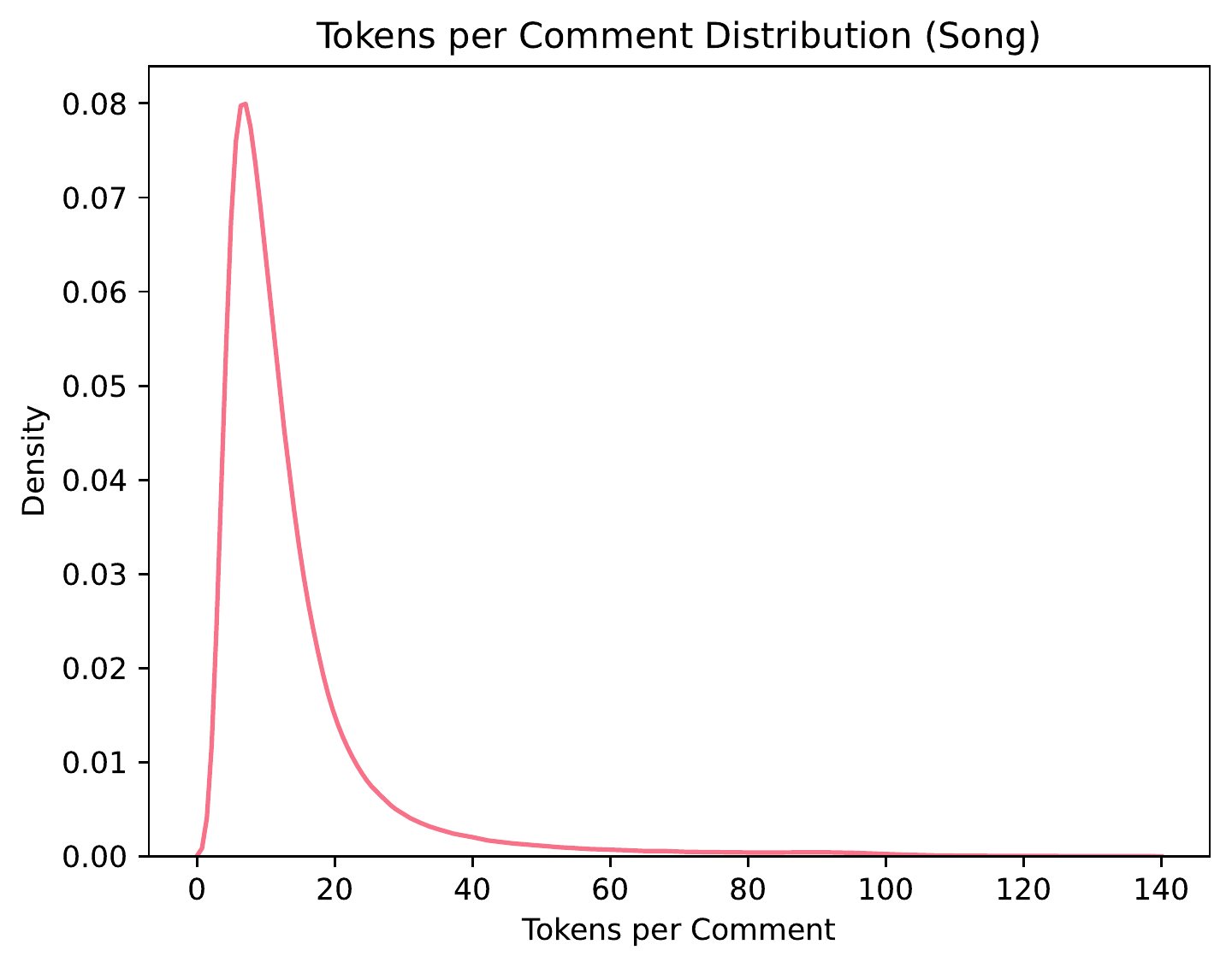}}
    \end{tabular}
    \caption{
    Comment (left) and comment token (right) distributions across all songs with at least one comment.
    }
    \label{fig:song-comments}
\end{figure*}

\end{document}